\documentclass[final, journal]{IEEEtran}
\ifCLASSINFOpdf
\else
\fi
\usepackage{amsmath,graphicx,bm,amsfonts,amssymb}
\usepackage{booktabs,subfigure,array}
\usepackage[numbers,sort&compress]{natbib}
\usepackage{multirow}
\usepackage{tabularx,booktabs,xcolor}
\newcolumntype{Y}{>{\centering\arraybackslash}X}
\usepackage{mathrsfs}
\usepackage{diagbox}

\usepackage{float}
\usepackage[ruled]{algorithm2e}

\newcommand{\diag}{\mathop{\rm diag}}

\makeatletter
\newcommand*{\rom}[1]{\expandafter\@slowromancap\romannumeral #1@}
\makeatother

\makeatletter
\def\ps@myheadings{%
    \let\@oddfoot\@empty\let\@evenfoot\@empty
    \def\@evenhead{\thepage\hfil\slshape\leftmark}%
    \def\@oddhead{{\slshape\rightmark}\hfil\thepage}%
    \let\@mkboth\@gobbletwo
    \let\sectionmark\@gobble
    \let\subsectionmark\@gobble
    }
  \if@titlepage
  \renewcommand\maketitle{\begin{titlepage}%
  \let\footnotesize\small
  \let\footnoterule\relax
  \let \footnote \thanks
  \null\vfil
  \vskip 60\p@
  \begin{center}%
    {\LARGE \@title \par}%
    \vskip 3em%
    {\large
     \lineskip .75em%
      \begin{tabular}[t]{c}%
        \@author
      \end{tabular}\par}%
      \vskip 1.5em%
    {\large \@date \par}
  \end{center}\par
  \@thanks
  \vfil\null
  \end{titlepage}%
  \setcounter{footnote}{0}%
}
\else
\renewcommand\maketitle{\par
  \begingroup
    \renewcommand\thefootnote{\@fnsymbol\c@footnote}%
    \def\@makefnmark{\rlap{\@textsuperscript{\normalfont\@thefnmark}}}%
    \long\def\@makefntext##1{\parindent 1em\noindent
            \hb@xt@1.8em{%
                \hss\@textsuperscript{\normalfont\@thefnmark}}##1}%
    \if@twocolumn
      \ifnum \col@number=\@ne
        \@maketitle
      \else
        \twocolumn[\@maketitle]%
      \fi
    \else
      \newpage
      \global\@topnum\z@   
      \@maketitle
    \fi
    \thispagestyle{plain}\@thanks
  \endgroup
  \setcounter{footnote}{0}%
}
\makeatother

\begin{document}

\title{Manifold-Inspired Single Image Interpolation}

\author{Lantao~Yu,~\IEEEmembership{Member,~IEEE}, Kuida~Liu,
        Michael~T.~Orchard,~\IEEEmembership{Fellow,~IEEE}
\thanks{Lantao Yu and Michael T. Orchard are with the Department
of Electrical and Computer Engineering, Rice University, Houston,
TX 77005, USA e-mail: complexfiltering@gmail.com; orchard@rice.edu.}

\thanks{Kuida Liu is with TuSimple Inc., 9191 Towne Centre Dr, STE 600, San Diego, CA 92122, USA email: queta@foxmail.com.}}

\markboth{IEEE Transactions on Image Processing}%
{Shell \MakeLowercase{\textit{et al.}}: Bare Demo of IEEEtran.cls for IEEE Journals}

\maketitle

\IEEEpeerreviewmaketitle

\begin{abstract}
Manifold models consider natural-image patches to be on a low-dimensional manifold embedded in a high dimensional state space and each patch and its similar patches to approximately lie on a linear affine subspace. Manifold models are closely related to semi-local similarity, a well-known property of natural images, referring to that for most natural-image patches, several similar patches can be found in its spatial neighborhood. Many approaches to single image interpolation use manifold models to exploit semi-local similarity by two mutually exclusive parts: i) searching each target patch’s similar patches and ii) operating on the searched similar patches, the target patch and the measured input pixels to estimate the target patch. Unfortunately, aliasing in the input image makes it challenging for both parts. A very few works explicitly deal with those challenges and only ad-hoc solutions are proposed.

To overcome the challenge in the first part, we propose a carefully-designed adaptive technique to remove aliasing in severely aliased regions, which cannot be removed from traditional techniques. This technique enables reliable identification of similar patches even in the presence of strong aliasing. To overcome the challenge in the second part, we propose to use the aliasing-removed image to guide the initialization of the interpolated image and develop a progressive scheme to refine the interpolated image based on manifold models. Experimental results demonstrate that our approach reconstructs edges with both smoothness along contours and sharpness across profiles, and achieves an average Peak Signal-to-Noise Ratio (PSNR) significantly higher than existing model-based approaches.
\end{abstract}


\section{Introduction}
\label{sec:intro}

Single image interpolation addresses the problem of generating a high-resolution (HR) image from a single low-resolution (LR) image which is directly downsampled from the HR image without any blurring or noise.

Since the problem is ill-posed, the priors on the target HR image and/or on the relationship between the LR and HR images dictate any interpolation algorithm’s performance. Early image interpolation algorithms utilize spatially-invariant models to impose the smoothness constraint on the entire target HR image. For example, the bicubic interpolator~\cite{keys1981cubic} utilizes a polynomial approximation to estimate the missing pixel from its surrounding measured pixels. Lanczos interpolator~\cite{duchon1979lanczos} computes each missing pixel by filtering its nearby measured pixels with a fixed kernel. These techniques typically generate edges with blur, jaggies, and haloes due to their inability to adapt to non-smooth, fast-evolving structures.

Later proposed algorithms~\cite{li2001new, zhang2008image, mallat2010super, liu2011image, yu2018location} rely on adaptive models on the geometric regularity of image structures and have greatly improved the interpolation performance. For example, edge-directed algorithms~\cite{li2001new, zhang2008image, liu2011image} model and exploit local orientation-invariance between the LR and HR covariance structures in the vicinity of edges. Direction-based algorithms~\cite{mallat2010super} exploit the smoothness along contour directions and often estimate a missing pixel by involving the pixels far beyond its local neighborhood. A recently proposed location-directed algorithm~\cite{yu2018location} exploits the constraint that multiple frequency components representing the same edge shall indicate the same locations of edges. This constraint helps estimate unaliased bands of frequency coefficients representing the HR image from the aliased bands of frequency coefficients representing the LR image and thereby generates typical interpolated edges with high quality.

Recent approaches~\cite{guo2012multiscale, dong2013sparse, romano2014single, zhu2016image, sun2016image} to single image interpolation focus on models for image patches (i.e., blocks of pixel values), and have demonstrated impressive results. Their models can be understood to be based on a manifold model for image patches from natural images. The manifold assumes that $N\times N$-pixel patches from natural images are found on a low-dimensional manifold embedded in the $N^2$-dimensional state space of patches. Relevant characteristics of manifold include: i) the manifold only models the pixels values of each patch, and is ambivalent to the positions within the image at which the patch is located (e.g., nearby points on the manifold need not come from nearby positions in the image); ii) for any natural-image patch on the manifold, all neighboring natural-image patches that are within an infinitesimally small Euclidean distance approximately lie on a low-dimensional linear affine subspace (the tangent space to the manifold at that point); iii) the low-dimensional local structure of the manifold reflects an assumption that two neighboring natural-image patches are related to each other by small perturbations of some features that both patches contain, since the number of such features that any patch can contain is far fewer than the number of pixels in the patch, the local neighborhood of natural-image patches at any given patch is low-dimensional.

Semi-local similarity is a well-known and widely-used property of natural-image patches. It refers to the fact that for most natural-image patches, several similar patches can be found in its spatial neighborhood. When coupled with the manifold model, semi-local similarity suggests that for a given natural-image patch, by searching its spatial neighborhood in the image one can usually find several patches from its local neighborhood on the manifold. By collecting together the given patch along with several of its similar patches, it is possible to exploit the low-dimensional character of this local neighborhood of patches from the manifold to design algorithms addressing various image processing problems, such as denoising, inpainting, and deblurring, with impressive results. A key advantage of such approaches is that they exploit properties of collections of similar patches, without making any explicit assumptions on the structure on the patch itself. Thus, these methods often work well in regions with relatively simple contents (e.g., smoothly varying pixel values, edges, etc.) as well as those with more complicated contents (e.g., textures and patterns). Examples of algorithms that use collections of similar patches for processing a target patch would be i) projecting a noisy patch into a low-dimensional affine subspace spanned by the collection of similar patches to decrease noise while preserving signal structure; ii) modeling the collection of a target patch and its similar patches as linear combinations of a small number of atoms from a dictionary for various image processing objectives. These approaches, and many similar ones, leverage the low-dimensional character of the collection of similar patches, without making explicit assumptions about the contents of the patch.

A paradigm for exploiting semi-local similarity using the manifold model to address image processing problems involves refining the target image through iterations with each iteration consisting of two mutually exclusive parts. The first part collects a set of similar patches for each target patch of an image to be refined. The second part operates on the set of similar patches and the target patch to be refined as well as the measured input pixels to refine the target patch. The success of semi-local similarity-based approaches relies on the quality achieved in each of the two parts: the quality of the set of similar patches collected for each target patch, and the quality of the operator applied to the similar patches and the unrefined target patch to estimate the refined target patch.

This paradigm has demonstrated its success in dealing with denoising problems. However, when applying this paradigm to single image interpolation problems, both parts of the iterations described in the previous paragraph face challenges unique to this application. Namely, aliasing in the LR image presents unique challenges to i) identifying reliable similar patches and ii) estimating refined target patches given an identified set of similar patches. Existing approaches to this problem incorporate ad-hoc solutions to these challenges. The approach proposed in this paper formulates single image interpolation explicitly as addressing these two challenges in mutually exclusive parts: the identification of similar patches, and the estimation of target patches given identified similar patches, and we provide systematic solutions to handle these two challenges. In the rest of this section, we elaborate aliasing's influence on both parts and introduce our solutions.

\begin{figure}[!tb]
\begin{minipage}{0.32 \linewidth}
  \centering
  \centerline{\includegraphics[width=0.9\linewidth]{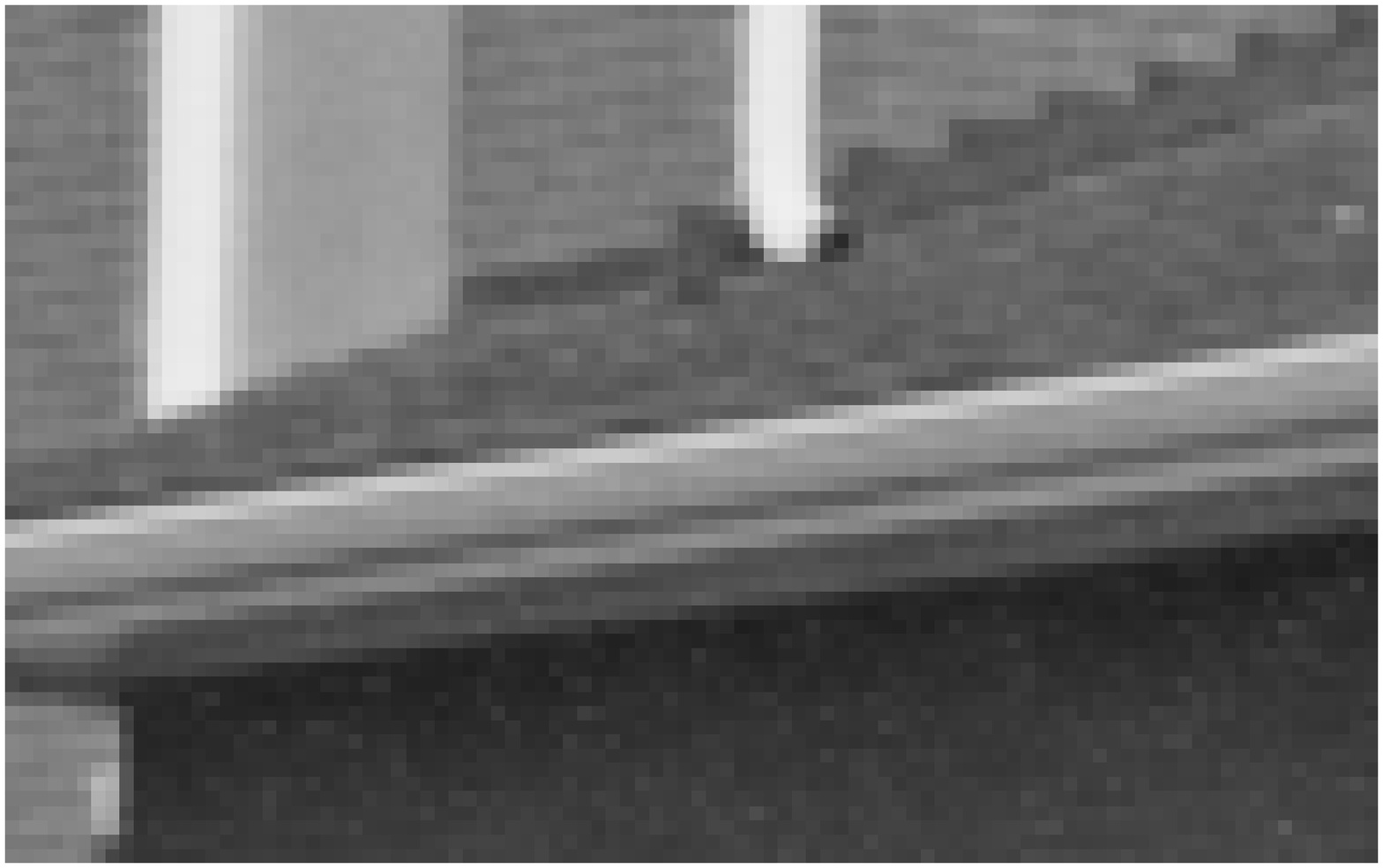}}
  \centerline{(a)}\medskip
\end{minipage}
\begin{minipage}{0.32 \linewidth}
  \centering
  \centerline{\includegraphics[width=0.9\linewidth]{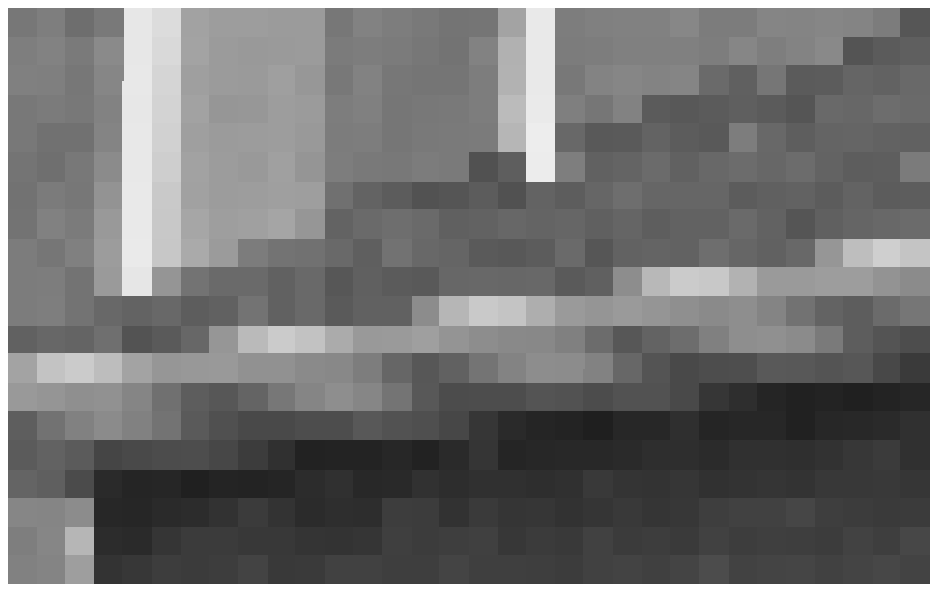}}
  \centerline{(b)}\medskip
\end{minipage}
\begin{minipage}{0.32 \linewidth}
\centering
\centerline{\includegraphics[width=0.9\linewidth]{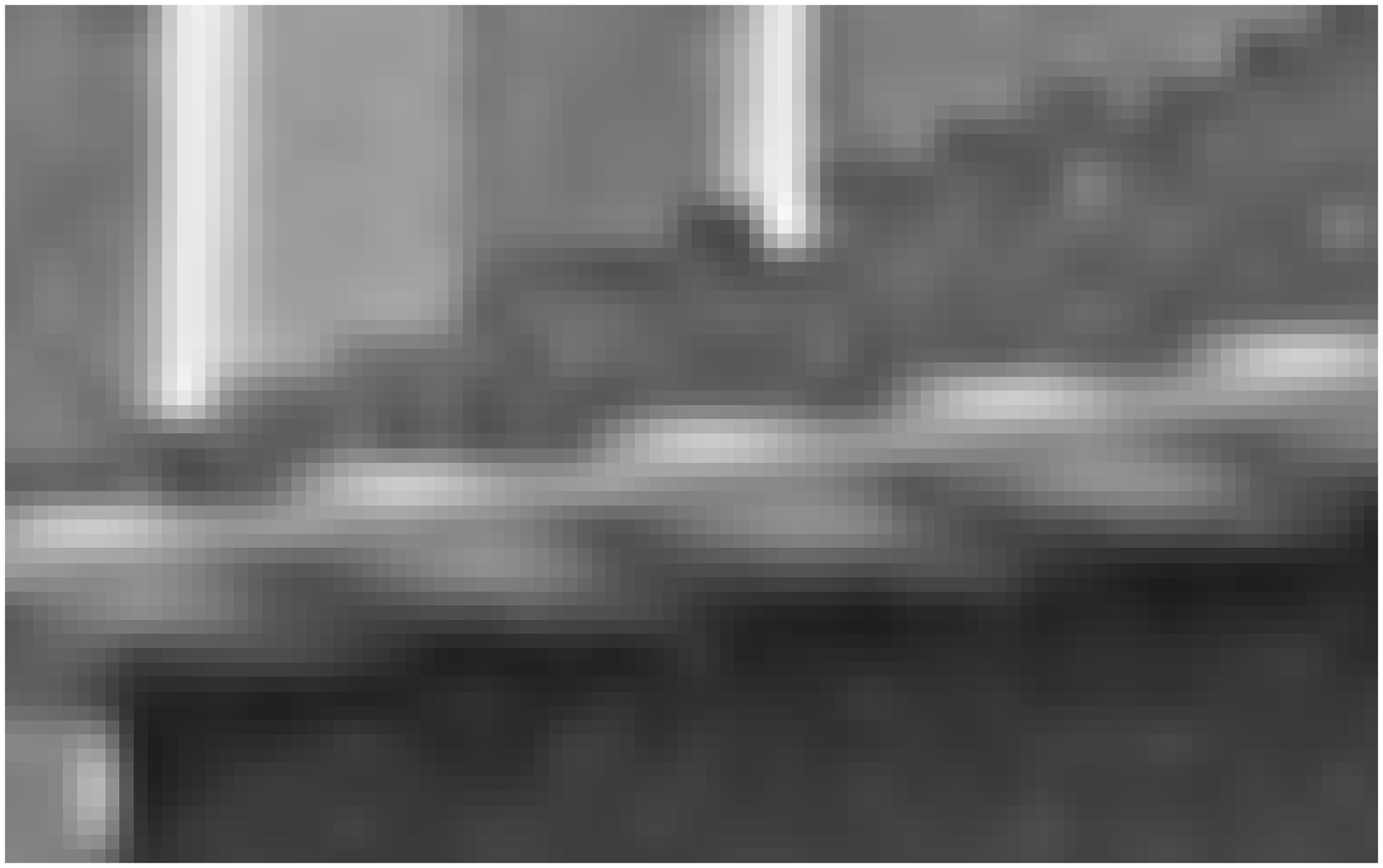}}
  \centerline{(c)}\medskip
\end{minipage} \\
\begin{minipage}{0.32 \linewidth}
  \centering
  \centerline{\includegraphics[width=0.9\linewidth]{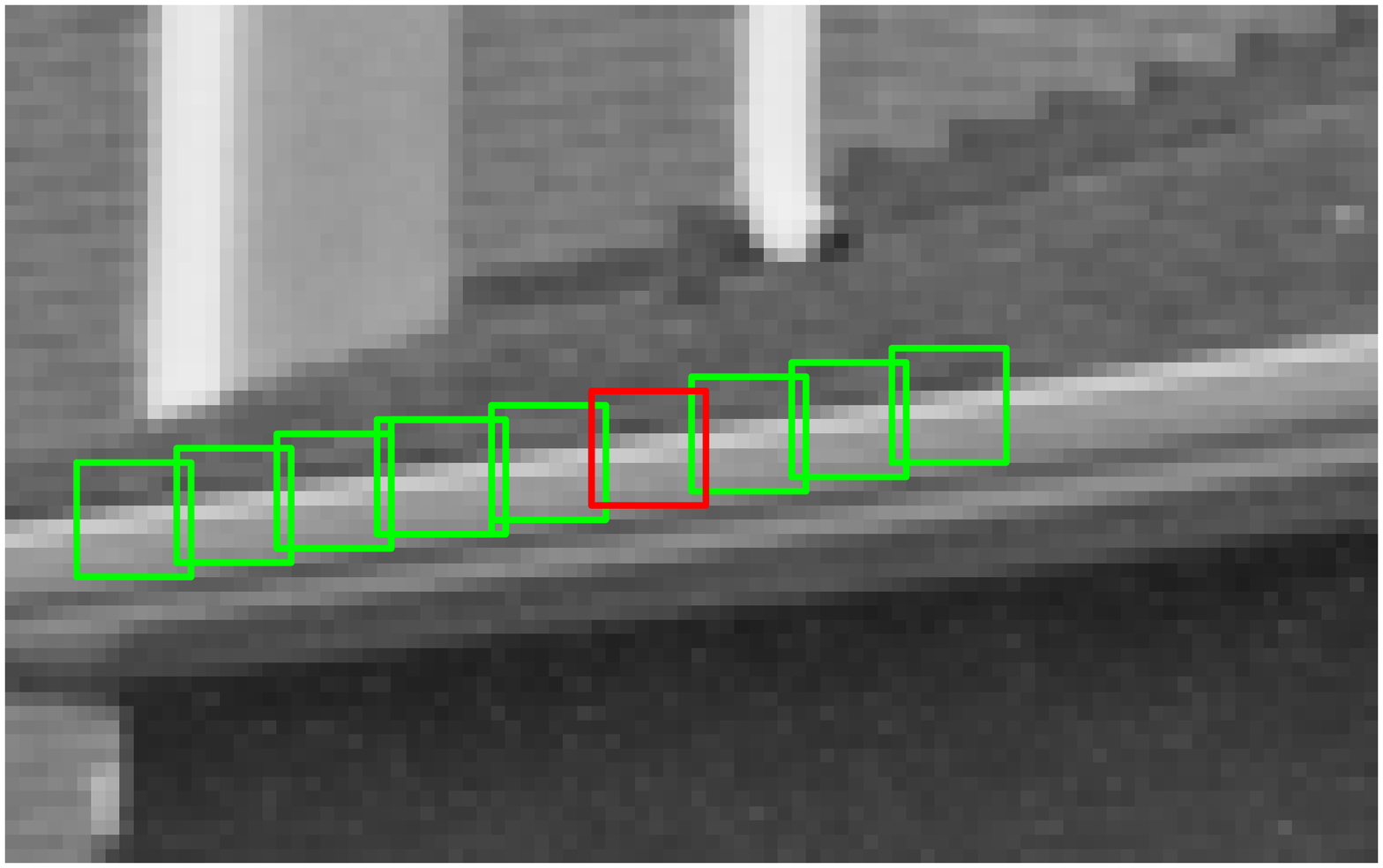}}
  \centerline{(d)}\medskip
\end{minipage}
\begin{minipage}{0.32 \linewidth}
  \centering
  \centerline{\includegraphics[width=0.9\linewidth]{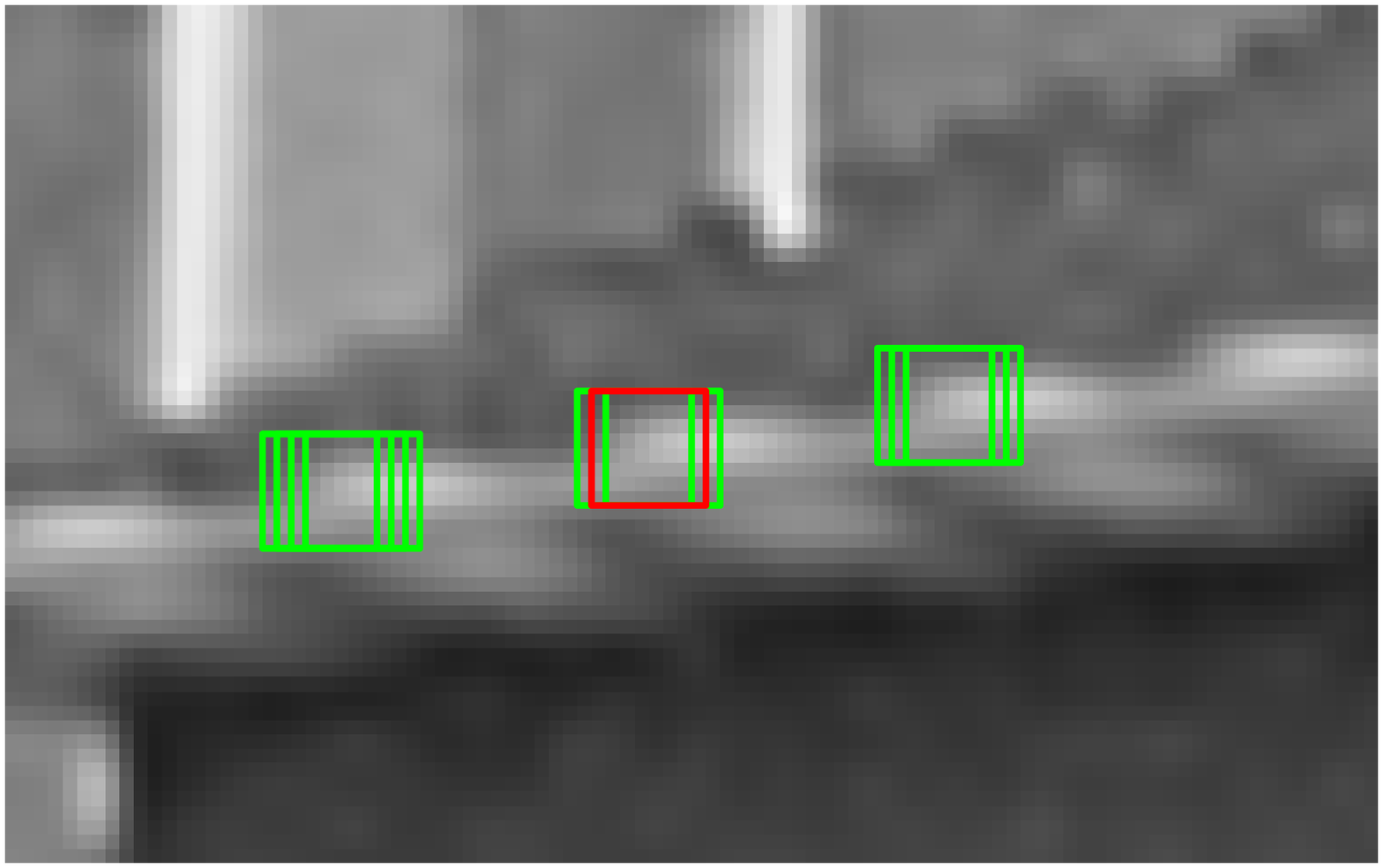}}
  \centerline{(e)}\medskip
\end{minipage}
\begin{minipage}{0.32 \linewidth}
  \centering
  \centerline{\includegraphics[width=0.9\linewidth]{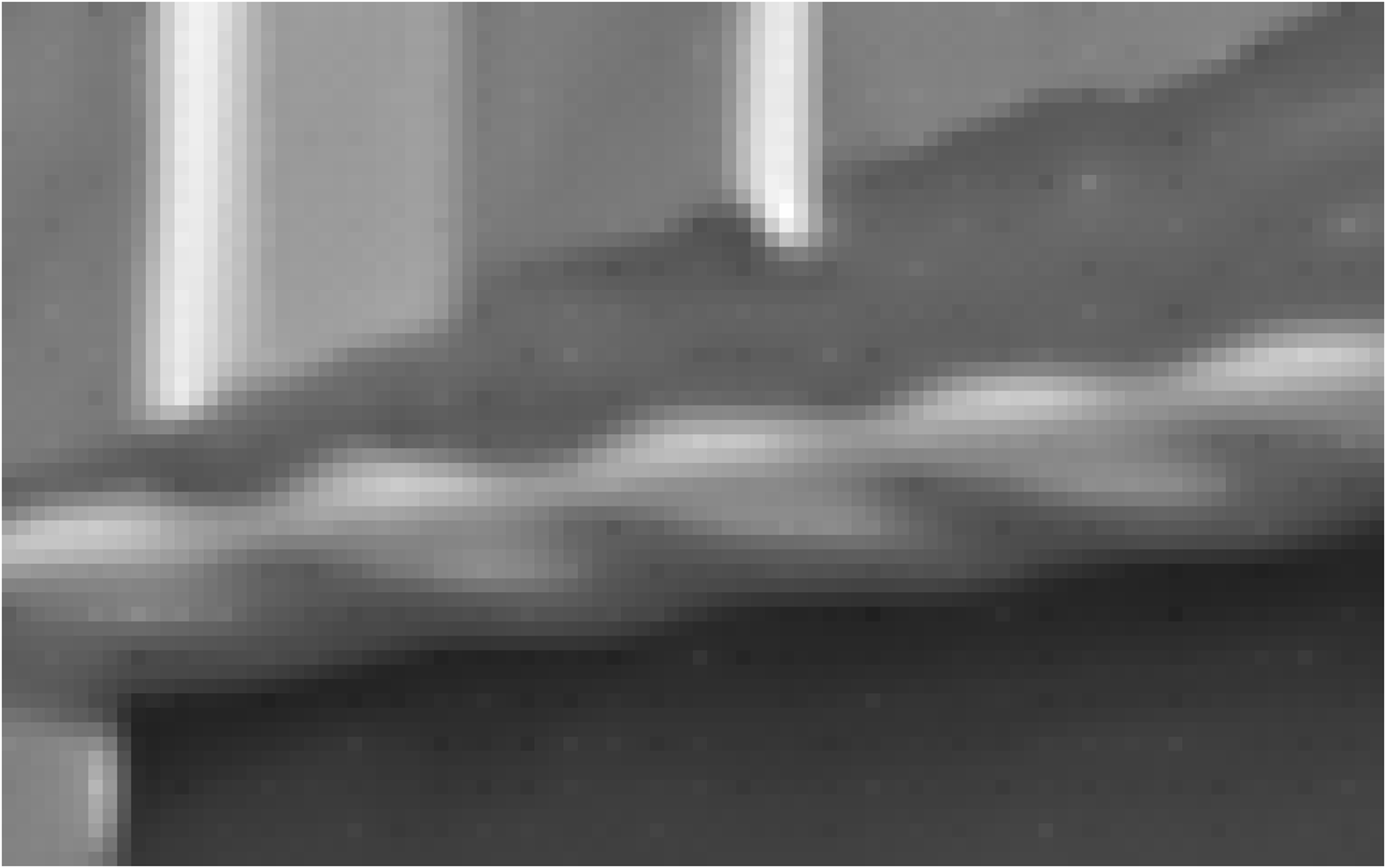}}
  \centerline{(f)}\medskip
\end{minipage} \\
\caption{An illustration of the bad choice of similar patches due to aliasing and the resulting artifacts in the interpolated image. (a) A region of interest in~\textit{House} image. (b) The downsample-by-$3$ version of (a)(Each pixel is magnified to fit the space). (c) The bicubic interpolated image of (b). (d) A target patch (the red block) and its $9$ most similar patches (the green blocks) searched in the original image. (e) A target patch (the red block) and its $9$ most similar patches (the green blocks) searched in the bicubic interpolated image. (f) The interpolated image using similar patches from (e) via NARM~\cite{dong2013sparse}.}
\label{fig:aliasing_effect}
\end{figure}

Identifying reliable similar patches of each target patch is challenging due to the presence of aliasing in the LR image. For example, if the original LR image is upsampled with zero-filling to initialize the reconstruction, two patches whose ground-truth, HR versions contain slightly shifted versions of nearly identical spatial structures may have a large Euclidean distance from each other, due to the grid of zero values in both patches. An ad-hoc solution is to calculate distances between patches of a bicubic interpolated version of the LR image. Unfortunately, high-frequency image structures like lines and contours typically have substantial aliasing energy in the LR image (i.e., energy from frequencies above the Nyquist limit appearing as periodic artifact patterns at lower frequencies) that cannot be completely removed with linear interpolation of the LR image. The presence of artifact patterns diverts similar patches from matching the actual image content to matching the aliasing patterns. The interpolated images depending upon these biased similar patches will also be biased and exhibit artifacts. We illustrate the aliasing patterns in the interpolated image in Fig.~\ref{fig:aliasing_effect}(f) as an example. These aliasing patterns result from the wrongly identified similar patches illustrated as green blocks in a highly aliased image in Fig.~\ref{fig:aliasing_effect}(e).

Aliasing from the LR image makes it difficult to estimate the refined target patches, even when correct similar patches are given. Current approaches either ignore aliasing in patches and continue to apply the aforementioned denoising paradigm, or mitigate aliasing but only in an ad-hoc manner. For example,~\cite{dong2013sparse,romano2014single} generate an initial estimate of the HR image from linear interpolation, model each refined target patch as a weighted sum of all pixels from similar patches, and compute the weights as functions of all the pixels from similar patches and unrefined target patches. The aliasing in these pixels often biases the weight estimation and subsequently brings artifacts to the refined target patches.~\cite{guo2012multiscale,sun2016image} mitigate the aliasing's influence by modeling each refined target patch as the weighted sum of only measured pixels from similar patches. However, the weights of measured pixels are still estimated as functions of these aliased pixels as in~\cite{dong2013sparse,romano2014single}.

To deal with aliasing in the first part, we propose to remove aliasing from the LR image. Low-pass filtering can remove aliasing in high-frequency components and has helped identified good similar patches in mildly-aliased regions~\cite{guo2012multiscale, yu2019single,yu2019spatially}, since high-frequency components are commonly more corrupted by aliasing than low-frequency components. Unfortunately, low-pass filtering can hardly deal with strong aliasing. Typical examples of strong aliasing include disjoint edges downsampled from their thin and continuous counterparts. This is because the original thin edge has substantial energy at the highest frequency and aliasing corrupts the zero-frequency components. In this case, the edge remains disjoint after low-pass filtering and the target patch containing discontinuous edges cannot find its similar patches containing continuous pieces (actual image content). To deal with strong aliasing, we recognize that many regions with strong aliasing contain directional localized structures and preserving the orientation information of these structures is sufficient to identify reliable similar patches. In this paper, we first propose an adaptive and aggressive operator that captures the energy along the dominant directions of local edges and attenuates the energy of aliasing and the energy along secondary directions of localized edges. This operator inevitably introduces artifacts in the vicinity of localized edges along secondary directions. Through follow-up refinement, our algorithm robustly removes aliasing both in low and high frequency components of the LR image and suppresses the artifacts along secondary directions.

To deal with aliasing in the second part, in the initial iteration, we use only the measured pixels from similar patches as the bases to estimate the unrefined pixels in the unrefined target patch. In addition, we estimate the weights of these measured pixels as functions of patches from the aliasing-removed image described in the previous paragraph. Using measured pixels precludes the influence of aliasing on the bases and estimating weights using the aliasing-removed image mitigates the influence of aliasing on the weights. The initial estimates of the target patches are unavoidably coarse, since the aliasing-removed image has too little high-frequency energy to precisely estimate the weights. For this reason, we propose follow-up iterations using appropriate manifold models to progressively refine the target patches and subsequently generate the interpolated image.



We organize the rest of paper as follows: Section~\rom{2} introduces the scheme to remove aliasing from the LR image for collecting reliable similar patches. Section~\rom{3} introduces the iterations for estimating the target patches given the identified similar patches. Section~\rom{4} introduce the overall scheme of our algorithm. In Section~\rom{5}, we test the algorithm on standard test images and compare it with state-of-the-art single image interpolation algorithms. Section~\rom{6} concludes this paper.

\section{Remove Aliasing for Finding Reliable Similar Patches}
\label{sec:aliasing_removal}

Removing aliasing in both low and high frequencies from the LR image lies at the heart of collecting reliable similar patches. This section details the aliasing removal algorithm.

\subsection{Remove Aliasing in High Frequencies}
\label{subsubsec:lowpass_filtering}
We remove aliasing in high frequencies via applying Gaussian low-pass filtering to $\bm{I}_L$:
\begin{equation}
\label{eq:convolution_lpf}
\bm{I}_L^{lp}=\bm{I}_L \ast \bm{F},
\end{equation}
where $\ast$ denotes the convolution operator. The filter $\bm{F}$, of size $d \times d$, is a unit-gain, 2D Gaussian with standard deviation $\sigma$.

\begin{figure}[tb]
\centering
\begin{minipage}{0.45 \linewidth}
  \centering
  \centerline{\includegraphics[width=0.95\linewidth]{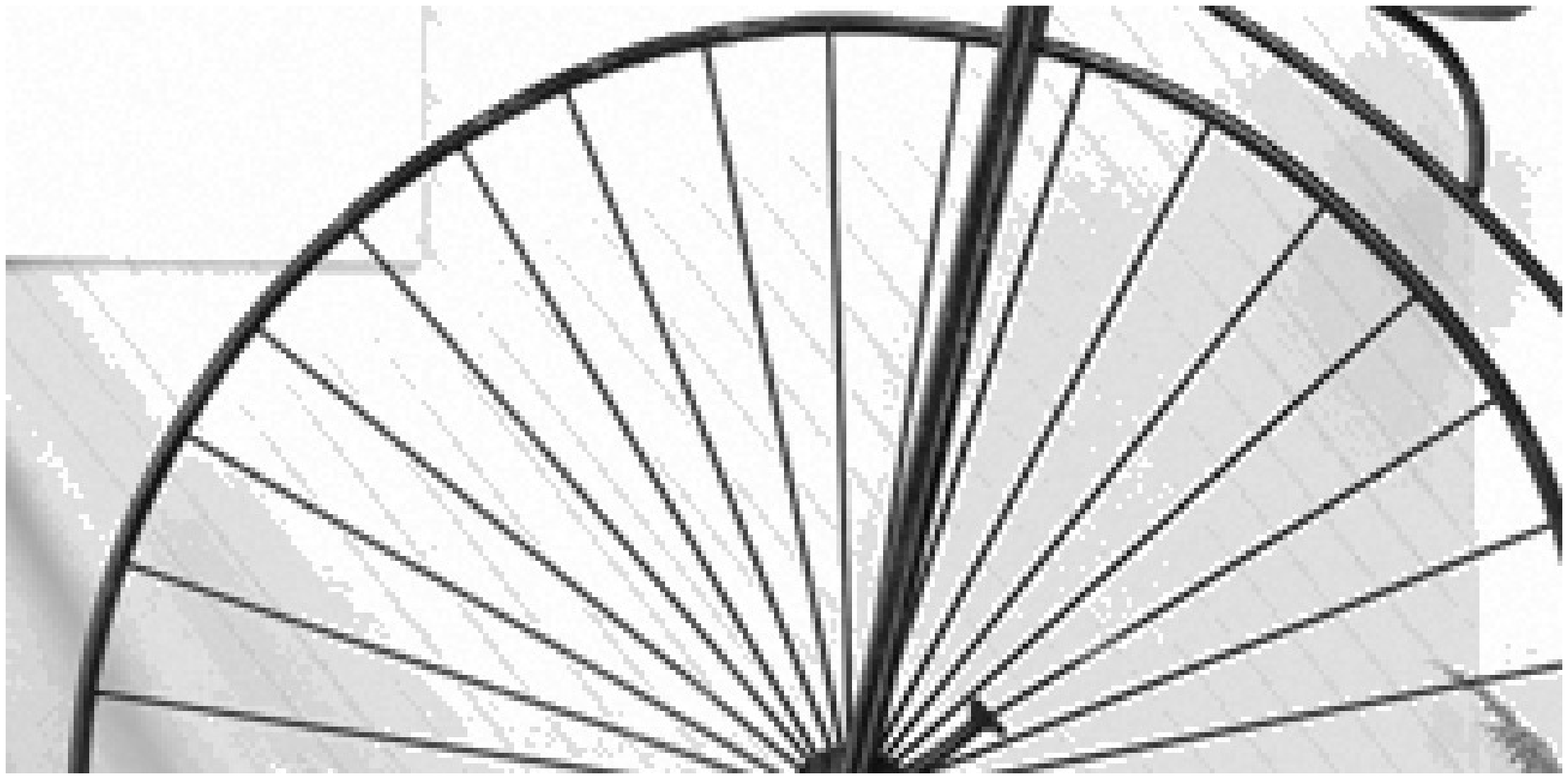}}
  \centerline{(a)} \medskip
\end{minipage}
\begin{minipage}{0.45 \linewidth}
  \centering
  \centerline{\includegraphics[width=0.95\linewidth]{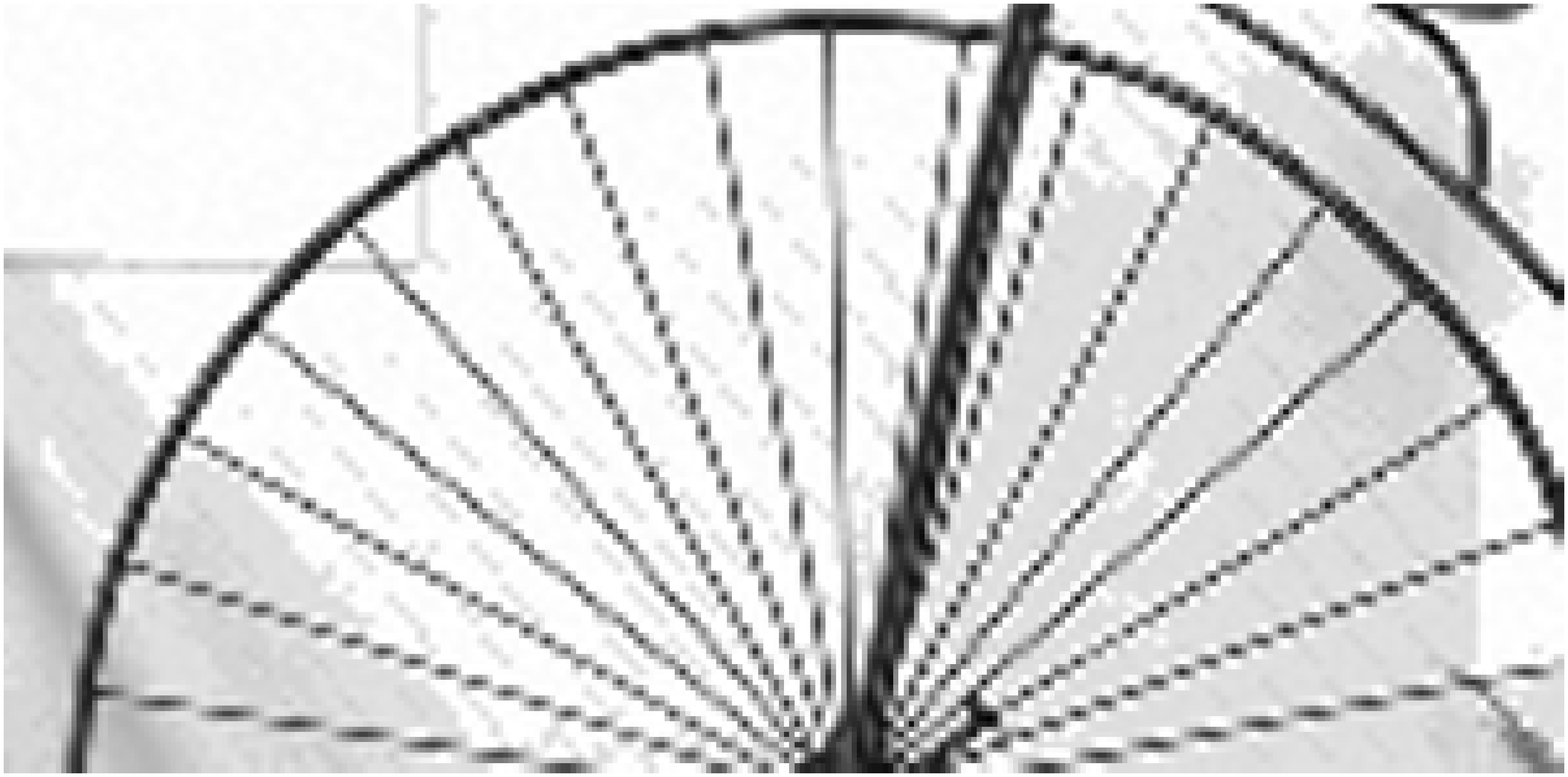}}
  \centerline{(b)} \medskip
  \end{minipage} \\
\begin{minipage}{0.45 \linewidth}
  \centering
  \centerline{\includegraphics[width=0.95\linewidth]{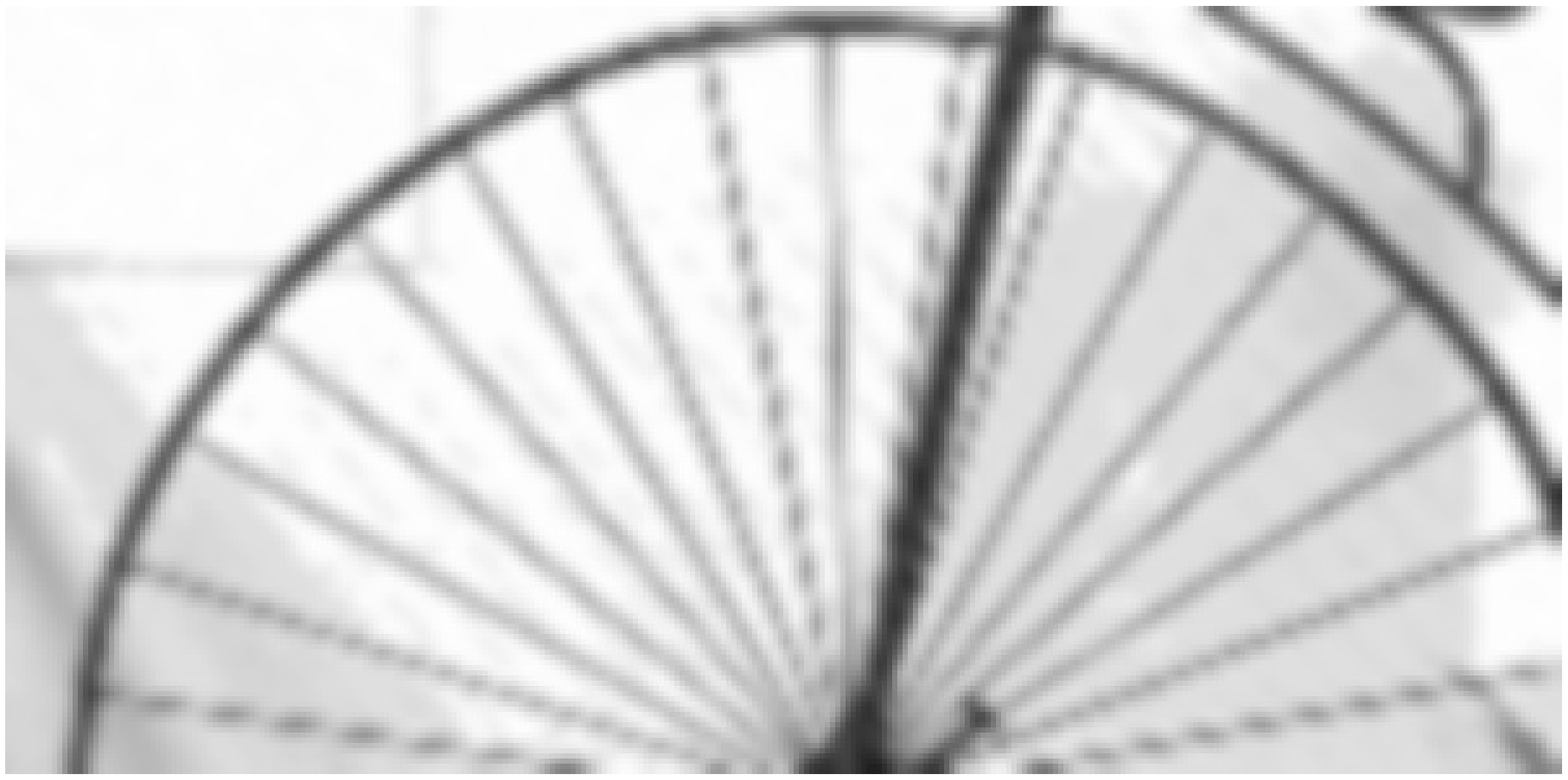}}
  \centerline{(c)}\medskip
\end{minipage}
\begin{minipage}{0.45 \linewidth}
  \centering
  \centerline{\includegraphics[width=0.95\linewidth]{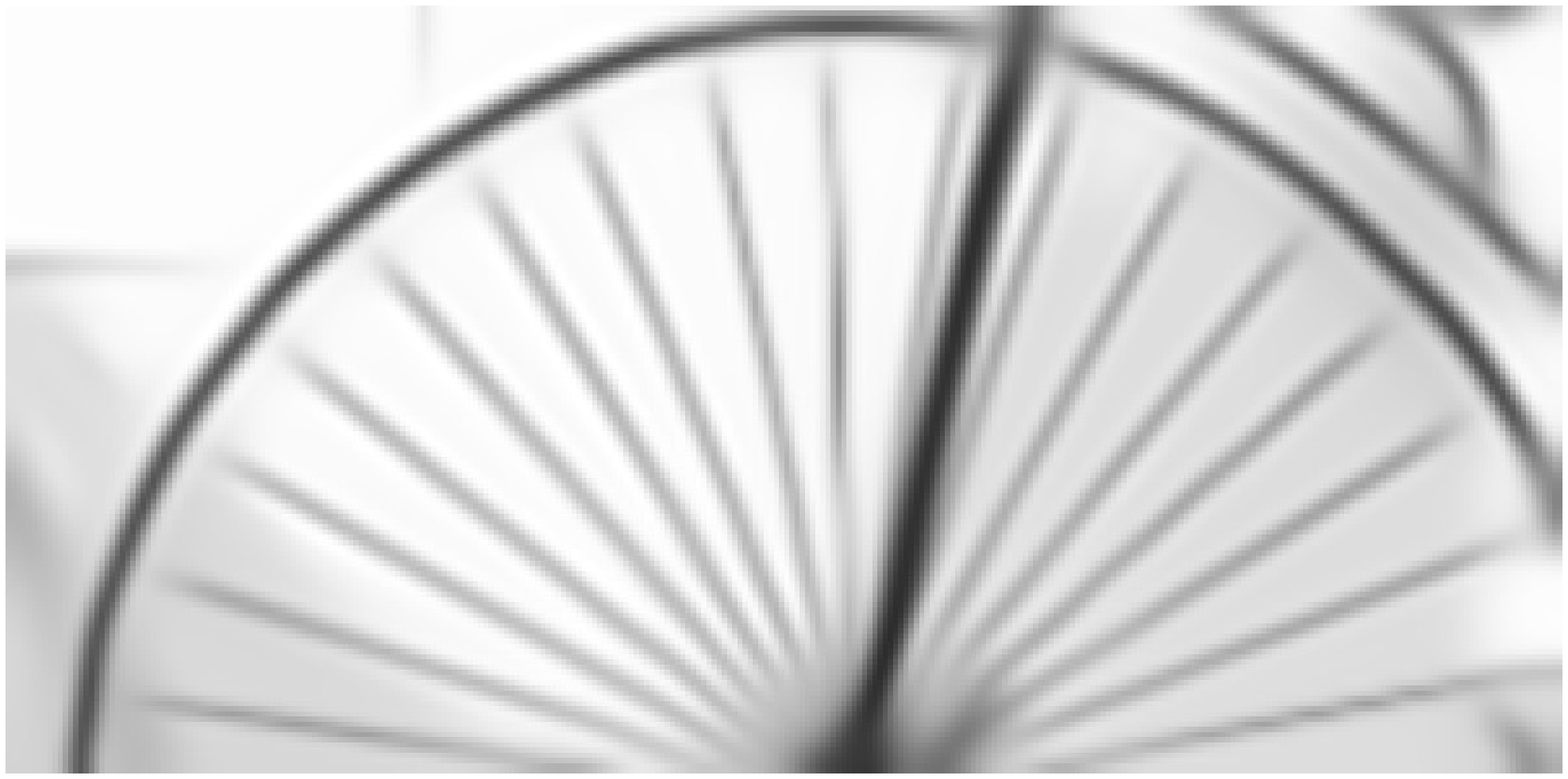}}
  \centerline{(d)}\medskip
\end{minipage}
\caption{The aliasing removal effect of our techinique. (a) The ground-truth~\textit{Wheel} image. (b) The bicubic interpolated version of the downsampled-by-2 \textit{Wheel} image. (c) The processed image using a spatially-invariant technique (low-pass filtering and bicubic interpolation). (d) The processed image using our spatially-variant aliasing removal algorithm (Algorithm~\ref{alg:stage_algorithm_removal}).}
\label{fig:guide_image_comparison}
\end{figure}

\subsection{Remove Aliasing in Low Frequencies}
\label{subsec:svp}

Low-pass filtering with a fixed blur kernel can only handle mild aliasing. When aliasing is so strong that corrupts the low-frequency components, typical structures such as the originally continuous spokes in Fig.~\ref{fig:guide_image_comparison}~(a) turn discontinuous in Fig.~\ref{fig:guide_image_comparison} (b). Applying spatially-invariant low-pass filtering on the input image can hardly recover the continuity since the low frequency components have already been corrupted. In Fig.~\ref{fig:guide_image_comparison} (c), the target patch  containing disjointed spokes can hardly find its similar patches containing a continuous piece of spoke. Thus, the spokes in the refined target patch will often remain disjoint, which severely affects the visual perception.

The regions with strong aliasing often contain directional localized structures. Preserving the orientations of these structures is the key to preserving the continuity and to identifying reliable similar patches. This motivates us to design an adaptive operator to extract the orientation-related characterizations of each patch and to represent the patch by such characterizations. Principal components of a group of similar patches in the vicinity of directional structures are such characterizations, since for a few bases to capture the vast majority of the energy of these patches, their manifested directions shall be aligned with the directions of the localized structure.

To robustly estimate the bases, the influence of aliasing in $\boldsymbol{I}_L^{lp}$ shall be attenuated. Here we borrow the idea of denoising by averaging. Specifically, when computing the principal components, instead of using a group of patches similar to the $i$-th target patch, $\boldsymbol{P}^{lp}_i$, from $\boldsymbol{I}_L^{lp}$, we use the union of groups of patches. Each group of $K_{ar}$ patches is similar to one of the $9$ patches within the $3 \times 3$ neighborhood of $\boldsymbol{P}^{lp}_i$'s position. The reason why involving the union of $9$ groups of patches can attenuate aliasing is that the aliasing components are often distinct at $9$ positions yet the orientations are nearly identical. For notational convenience, we denote the union of grouped matrix as $\boldsymbol{Y}_i$, of size $\tilde{K}\times n^2$, where $\tilde{K}$ is the cardinality of the union of the $9$ sets and $n$ is the patch's size length. We also denote its centered version as $\tilde{\boldsymbol{Y}}_i$, so that
\begin{equation}
\label{eq:center_Y}
\tilde{\boldsymbol{Y}}_i=\boldsymbol{Y}_i-\bar{\boldsymbol{Y}}_i,
\end{equation}
where each row of $\bar{\bm{Y}}_i$, $\bar{\boldsymbol{y}}_i^{\top}$, is the mean of $\tilde{K}$ patches. In addition, we denote $\tilde{\boldsymbol{y}}_i \in \mathbb{R}^{n^2 \times 1}$ as:
\begin{equation}
\label{eq:mean_Y}
  \tilde{\boldsymbol{y}}_i=\boldsymbol{P}^{lp}_i-\bar{\boldsymbol{y}}_i.
\end{equation}

Given $\tilde{\boldsymbol{Y}}_i$, we solve for its top $k$ principal components through SVD in order for them to represent the local directions. $\tilde{\boldsymbol{Y}}_i$ is decomposed as:
\begin{equation}
\label{eq:svd_Y}
\boldsymbol{U}_i\boldsymbol{D}_i\boldsymbol{V}_i^{\top}=\tilde{\boldsymbol{Y}}_i,
\end{equation}
where $\boldsymbol{U}_i$ is a $\tilde{K} \times \tilde{K}$ unitary matrix; $\boldsymbol{D}_i$ is a $\tilde{K} \times n^2 $ diagonal matrix with non-negative real numbers on the diagonal; $\boldsymbol{V}_i$ is a $n^2 \times n^2$ unitary matrix. The $j$-th columns of $\boldsymbol{V}_i$ is denoted as as $\boldsymbol{v}^j_i$, where $1\leqslant j \leqslant k$. Aiming to preserve the directional structures, we update $\tilde{\bm{y}}_i$ as $\tilde{\bm{y}}^{u}_i$, the linear combination only of the top $k$ principal components through:
\begin{equation}
  \label{eq:projection}
   \tilde{\boldsymbol{y}}^{u}_i=\sum_{j=1}^k\langle\tilde{\boldsymbol{y}}_i,\boldsymbol{v}^j_i\rangle\boldsymbol{v}^j_i.
\end{equation}

By averaging the contribution of all the overlapped updated target patches relevant to each pixel, we generate an intermediate version of aliasing-removed image via:
\begin{equation}
\label{eq:synthesize}
\left(\sum_{i=1}^{l}\boldsymbol{R}_{i}^{\top} \hat{\boldsymbol{P}}_i\right)./\left(\sum_{i=1}^{l} \boldsymbol{R}_{i}^{\top} \boldsymbol{1}_{n^2\times 1}\right),
\end{equation}
where $\hat{\boldsymbol{P}}_i = \tilde{\boldsymbol{y}}^{u}_i+\bar{\boldsymbol{y}}_i$; $l$ is the number of target patches in $\bm{I}_L^{lp}$; $\boldsymbol{R}_{i}(\cdot)$ is an operator extracting $n^2$ pixels from the $i$-th patch from an image; $\boldsymbol{R}^{\top}_{i}(\cdot)$ puts back $n^2$ pixels into the $i$-th position and padds with zeros elsewhere; $./$ stands for the element-wise division of two matrices;  $\boldsymbol{1}_{n^2\times 1}$ is a vector of size $n^2\times 1$ with all the elements being $1$.

By repeating the above procedures for another time and applying bicubic interpolation to the output computed from~\eqref{eq:synthesize}, we generate $\boldsymbol{{I}}_{ar}$ as the image with removed aliasing in low and high frequencies of the same size as the HR image. A typical example of $\boldsymbol{{I}}_{ar}$ is illustrated in Fig.~\ref{fig:guide_image_comparison}(d) where the spokes retain the smoothness in the original HR image. $\boldsymbol{{I}}_{ar}$ well preserves the dominant directions of localized edges but removes the secondary directions of localized edges. We will further refine $\boldsymbol{{I}}_{ar}$ in Section~\ref{subsec:handling_artifacts} to better estimate the similar patches in the vicinity of localized edges along secondary directions. Prior to generating the refined version of $\boldsymbol{{I}}_{ar}$, $\boldsymbol{{I}}_{ar}$ will first serve as a preliminary aliasing-removed image to initialize the selection of similar patches and to estimate the weights of similar patches in Section~\ref{sec:mister_algorithm}.

\begin{algorithm}[tb]
\label{alg:stage_algorithm_removal}
    \SetKwInOut{Input}{Input}
    \SetKwInOut{Output}{Output}
    \Input{input image $\bm{I}_L$; the number of target patches in $\bm{I}_L$, $l$; patch size $n$; number of each group of similar patches $K_{ar}$; search window size $W$; Gaussian filter size $d$; Gaussian filter standard deviation $\sigma$; the number of top principal components $k$; the number of iterations $\mathscr{I}$.}
    \Output{the aliasing-removed image $\bm{I}_{ar}$.}
    Compute $\bm{I}_L^{lp}$ via~\eqref{eq:convolution_lpf}.\\
    \For{iteration $\leftarrow 1$ \KwTo $\mathscr{I}$}{
    	\For{i $\leftarrow 1$ \KwTo $l$}{
    	Extract $\bm{P}_i^{lp}$ from $\bm{I}_L^{lp}$ as the $i$-th target patch. \\
    	Group the union of $9K_{ar}$ similar patches of $\bm{P}_i^{lp}$ in $\bm{Y}_i$ via $\ell_1$ distance.\\
    	Compute $\bar{\bm{y}}_i$ as the mean of the union of $9K_{ar}$ similar patches. \\
    	Compute $\tilde{\bm{Y}}_i$, $\tilde{\bm{y}}_i$ via~\eqref{eq:center_Y} and~\eqref{eq:mean_Y}, respectively. \\
    	Compute $\boldsymbol{V}_i$ given $\tilde{\boldsymbol{Y}}_i$ via~\eqref{eq:svd_Y}. \\
    	\For{j $\leftarrow 1$ \KwTo $k$}{
			Extract $\bm{v}^j_i$ as the $j$-th column of $\boldsymbol{V}_i$.
			}
		Compute $\tilde{\bm{y}}^{u}_i$ given $\tilde{\bm{y}}_i$ and all $\bm{v}^j_i$ via~\eqref{eq:projection}.
    	}
    	$\hat{\boldsymbol{P}}_i = \tilde{\boldsymbol{y}}^{u}_i+\bar{\boldsymbol{y}}_i$. \\
    	Update $\bm{I}_L^{lp}$ given all $\hat{\boldsymbol{P}}_i$ via~\eqref{eq:synthesize}.
    }
	Compute $\bm{{I}}_{ar}$ as the bicubic interpolation of $\bm{I}_L^{lp}$. \\
    \caption{Spatially-Variant Aliasing Removal}
\end{algorithm}

\section{Estimate Target Patches Given Their Identified Similar Patches}
\label{sec:mister_algorithm}

In this section, given the collected similar patches, we focus on introducing how to mitigate aliasing's influence on estimating the target patches in the initial iteration and how to progressively refine the target patches in later iterations.

\subsection{The Basic Interpolation Scheme}
We assume the low-resolution input image $\bm{I_L}$ of size $ M \times N$ to be the directly downsampled version of a high-resolution image $\bm{I_H}$ of size $2M \times 2N$. The downsampling grid starts with the upper-left pixel in $\bm{I_H}$, i.e.
\begin{equation}
\label{eq:dsp}
\bm{I_L}(p,q)=\bm{I_H}(2p-1,2q-1),
\end{equation}
where $1 \leqslant p \leqslant M, 1 \leqslant q \leqslant N$. The interpolated image is labeled as $\bm{\hat{I}_H}$ which aims to be as close to $\bm{I_H}$ as possible. In the iterations to generate $\bm{\hat{I}_H}$ from $\bm{I_L}$, there exists an unrefined target patch with even side length $n$. We label the $i$-th unrefined target patch as $\boldsymbol{P}_i~(\boldsymbol{P}_i \in \mathbb{R}^{n^2 \times 1})$.
\begin{figure}[b]
\centering
\begin{minipage}{.45\linewidth}
\centerline{\includegraphics[width=4.0 cm]{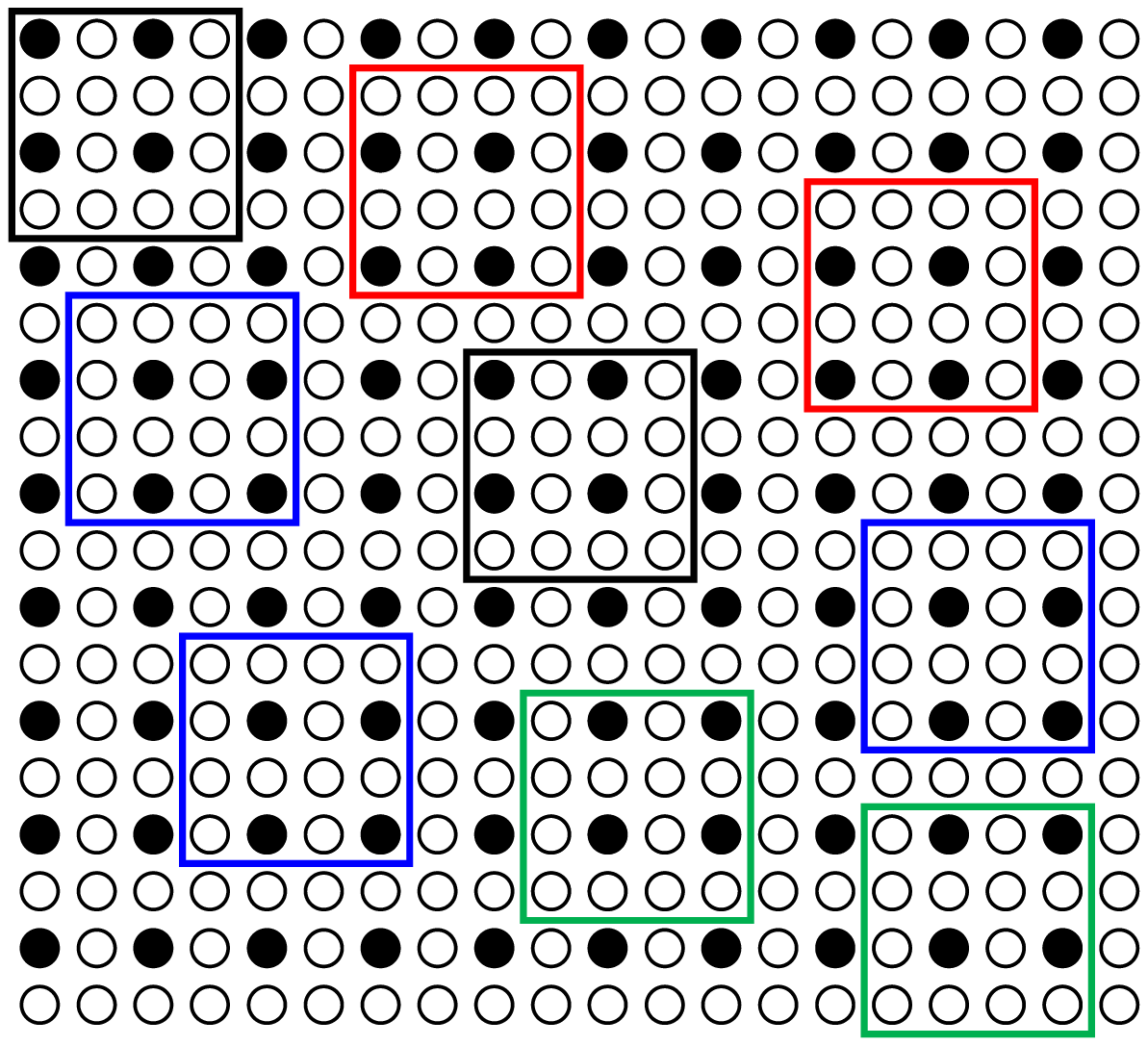}}
\centerline{(a) }\medskip
\end{minipage}
\centering
\begin{minipage}{0.45\linewidth}
\vspace{0.2 cm}
  \centering
  \centerline{\includegraphics[width=3.8 cm]{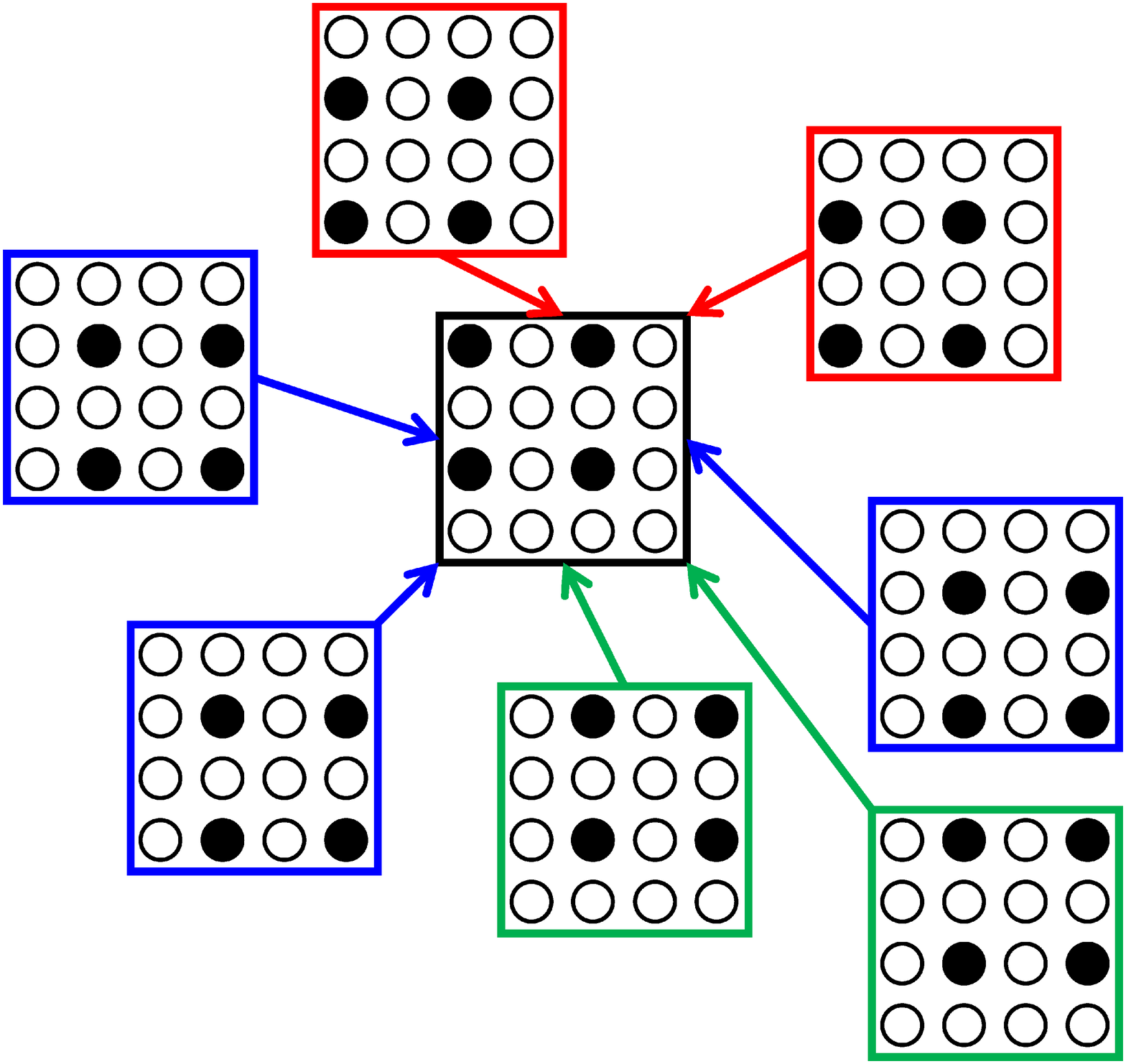}}
  \centerline{(b) }\medskip
\end{minipage}
\caption{The basic interpolation scheme for estimating the unrefined pixels in a target patch. The solid circles in (a,b) are measured pixels from the LR image. The empty circles in (a,b) are pixels to be refined. (a) The target patch is labeled by a black frame at the center of with its upper-left pixel at OO grid. Its similar patches whose upper-left pixels are OE-, EO-, and EE-located are labeled by green, red, and blue frames, respectively. (b) Each similar patch has measured pixels that corresponds to the pixels to be refined in the target patch. Similar schemes can also be found in~\cite{guo2012multiscale, sun2016image}.}
\label{fig:illustration_2x_interp}
\end{figure}

For notational convenience, we categorize the pixels in $\bm{I_H}$ as OO-, OE-, EO-, EE-located, depending on the parity (O stands for odd, E stands for even) of its vertical and horizontal coordinates. As mentioned in~\eqref{eq:dsp}, the measured pixels are OO-located. In addition, we denote the patches with its upper-left pixels be OO-, OE-, EO-, EE-located as OO, OE, EO, EE patches. Regarding each patch as a $n \times n$ image whose upper-left pixel's vertical and horizontal coordinates are both odd, we categorize OO-, OE-, EO-, EE-located pixels within a patch as OO-, EO-, OE-, EE-phase pixels. Assuming each target patch is OO-located, as illustrated in Fig.~\ref{fig:illustration_2x_interp}, among all similar patches, OE, EO, EE patches will have OE-, EO-, EE-phase pixels(all measured pixels) similar to unrefined OE-, EO-, EE-phase pixels in the target patch, respectively. In other words, all the unrefined pixels in a target patch have their similar pixels found as the measured pixels. To estimate the target patch's unrefined pixels (OE-, EO-, EE-phase pixels), we use their corresponding measured pixels from OE, EO, EE similar patches as bases and express the refined target patch as:
\begin{align}
\begin{split}
\hat{\bm{P}}_i = & \bm{M}^{OO} \bm{P}_i + \sum_{j=1}^{K^{OE}}\omega_{i, j}^{OE}\bm{M}^{OE} \bm{P}_{i,j}^{OE} + \\
& \sum_{j=1}^{K^{EO}}\omega_{i, j}^{EO}\bm{M}^{EO} \bm{P}_{i,j}^{EO} + \sum_{j=1}^{K^{EE}}\omega_{i,j}^{EE}\bm{M}^{EE} \bm{P}_{i,j}^{EE}, \\
\end{split}
\label{eq:piupdate}
\end{align}
where $\hat{\bm{P}}_i$, $\bm{P}_{i,j}^{OE}$, $\bm{P}_{i,j}^{EO}$, $\bm{P}_{i,j}^{EE}$ $\in$ $\mathbb{R}^{n^2 \times 1}$ are the vectorized forms of the refined target patch, the target patch's $j$-th similar OE, EO and EE patches, respectively; $\omega_{i,j}^{OE}$, $\omega_{i,j}^{EO}$, $\omega_{i,j}^{EE}$ are the weights of $j$-th similar OE, EO, EE patch of $\bm{P}_i$, respectively; each type of similar patches has $K^{OE}$, $K^{EO}$, $K^{EE}$ patches, respectively; $\bm{M}^{OO}$, $\bm{M}^{OE}$, $\bm{M}^{EO}$, $\bm{M}^{EE}$ $\in$ $\mathbb{R}^{n^2 \times n^2}$ are matrices that select OO-, OE-, EO-, EE-phase pixels from a patch and sets the rest of pixels $0$. After updating each target patch, we average all target patches and generate the refined image via~\eqref{eq:synthesize} (here $l$ refers to the number of OO patches).

\subsection{The Interpolation Scheme in the Initial Iteration (Stage~$1$A)}
\label{subsec:1st_iteration}

Mitigating the influence of aliasing on estimating the refined patches is the key task in the initial iteration. We use the measured pixels from similar patches as the bases and estimate their weights using patches from an aliasing-removed image.

In the initial iteration, we denote the aliasing-removed image as $\bm{I}_{G}$ and use $\bm{I}_{G}$ (guide image) to search the positions of similar patches of each target patch and to compute their corresponding weights for estimating the target patch. Specifically, similar patches are searched in a $W\times W$ ($W$ is odd) square window centered at the target patch's upper-left pixel coordinate. The similarity metric is defined as:
\begin{equation}
\label{eq:similarity_metric1}
S(\bm{P}^G_i,\bm{P}^{Gf}_{i,j})=e^{-\|\bm{P}^G_i-\bm{P}^{Gf}_{i,j}\|_1/c_w},
\end{equation}
where $\bm{P}^G_i$ denotes the $i$-th patch in $\bm{I}_{G}$ sharing the same position of the unrefined $i$-th target patch $\bm{P}_i$; $\bm{P}^{Gf}_{i,j}$ denotes the $j$-th patch within the neighorhood of $\bm{P}^G_i$ belonging to OE (f refers to OE), EO (f refers to EO) or EE (f refers to EE) patches; $\|\bm{P}^G_i-\bm{P}^{Gf}_{i,j}\|_1$ denotes the $\ell_1$ distance between two patches in $\bm{I}_{G}$; and $c_w$ is a constant that controls the decay of similarity values as a function of the $\ell_1$ distance. For OE, EO or EE similar patches, the similarity values of top $K^f$ similar patches are stored in descending order in vector $\bm{S}_i^{OE}$, $\bm{S}_i^{EO}$ and $\bm{S}_i^{EE}$, respectively.

Given the positions of $\bm{P}_i$'s similar patches at each phase, the weights of these similar patches~($\bm{\omega}_i^f$) are computed by approximating the OO-phase pixels in $\bm{P}^G_{i}$ via their corresponding OO-phase pixels from similar patches. The reason we only use OO-phase pixels here is because in early iterations, the pixels on measurement grid are more reliable than the pixels off the measurement grid. $\bm{\omega}_i^f$ is computed with a closed-form solution by solving a regularized least squares problem:
\begin{equation}
\label{eq:grr}
\bm{\omega}_i^{f} = \arg\min_{\bm{\omega}_i} \left[\|\bm{M}^{OO}_c\bm{Q}^{f}_i\bm{\omega}_i-\bm{M}^{OO}_c{\bm{P}^G_i} \|^2_2+\lambda{\bm{\omega}_i}^{\top}\bm{\Sigma}^f_i\bm{\omega}_i \right],
\end{equation}
where $\bm{Q}^f_i$, of size $n^2 \times K^f$, stores one type (OE, EO or EE) of top-$K^f$ similar patches; $\bm{M}_c^{OO}$ is a $\frac{n^2}{4}\times n^2$ matrix that selects OO-phase pixels; $\lambda$ is a scalar that controls the regularization strength; $\bm{\Sigma}^f_i=\diag\left(\frac{{\bm{S}^f(1)}}{{\bm{S}^f(1)}},\dots,\frac{{\bm{S}^f(1)}}{{\bm{S}^f(i)}},\dots, \frac{{\bm{S}^f(1)}}{{\bm{S}^f(K)}}\right)$ is a diagonal matrix that penalizes the weights of less similar patches. Given~\eqref{eq:grr},~\eqref{eq:piupdate}, we compute a refined patch $\hat{\bm{P}}_i$. By averaging all the refined patches, we generate an intermediately interpolated image via~\eqref{eq:synthesize}.


\subsection{The Interpolation Scheme in Later Iterations (Stage~$1$B)}
\label{subsec:2nd_iteration}

The interpolated image from the initial iteration is coarsely estimated since both the collected similar patches and their weights are determined by $\bm{I}_G$ which has few high-frequency details. To refine the coarsely interpolated image, we treat the previously interpolated image as $\bm{I}_G$ to identify similar patches and compute their weights in the next iteration and to use a smaller patch size. By several iterations, we generate the output image $\bm{I}_{\mathrm{S1}}$ from Stage~$1$B with its details in Algorithm~\ref{alg:stage1}. Due to the similarity between the initial iteration (Stage $1$A) and the follow-up iterations (Stage $1$B), we name these iterations as Stage $1$. Owing to significance of the initial iteration (Stage $1$A), we will further discuss it in Section~\ref{subsec:ablation_study}.

\begin{algorithm}[tb]
\label{alg:stage1}
\SetKwInOut{Input}{Input}
\SetKwInOut{Output}{Output}
\Input{input image $\bm{I}_L$; guide image $\bm{I}_G$; the number of target patches in the upsampled and zero-filled $\bm{I}_L$, $l$; patch size $n_A$, $n_B$ in the first and later iterations; the number of each target patch's OE, EO or EE similar patches $K$; search window size $W_A$, $W_B$ in the first and later iterations; the penalty constants $\lambda_A$, $\lambda_B$ in the first and later iterations; the parameter in the similar metric $c_w$; the number of iterations $\mathscr{I}$.}
\Output{the output image from Stage 1: $\bm{I}_{S1}$.}
\For{iteration $\leftarrow 1$ \KwTo $\mathscr{I}$}{
\eIf{iteration $=1$}{
Use $n_A$, $W_A$, $\lambda_A$.}
{
 Use $n_B$, $W_B$, $\lambda_B$.}
\For{i $\leftarrow 1$ \KwTo $l$}{
Extract $\bm{P}_i$ from upsampled and zero-filled $\bm{I}_{L}$ as the $i$-th target patch. \\
Extract $\bm{P}_i^{G}$ from $\bm{I}_{G}$ at $\bm{P}_i$'s position. \\
\For{f $\leftarrow OE, EO, EE$ }{
Find $K$ $f$ similar patches to $\bm{P}_i^{G}$ via~\eqref{eq:similarity_metric1}.\\
Store similar patches in $\bm{Q}^f_i$ and the similarity measures in $\bm{\Sigma}^f_i$. \\
Compute $\bm{\omega}_i^{f}$ given $\bm{P}_i^{G}$, $\bm{\Sigma}^f_i$, $\bm{Q}^f_i$ and $\lambda$ via~\eqref{eq:grr}.\\
}
Generate ${\hat{\bm{P}}_i}$ given $\bm{\omega}_i^{OE}$, $\bm{\omega}_i^{OE}$, $\bm{\omega}_i^{OE}$ and $\bm{P}_i$ via~\eqref{eq:piupdate}.
}
Generate $\bm{I}_{G}$ given all $\hat{\boldsymbol{P}}_i$ via~\eqref{eq:synthesize}.
}
$\bm{{I}}_{S1} = \bm{I}_{G}$. \\
\caption{Stage~$1$}
\end{algorithm}

\subsection{The Interpolation Scheme in Later Iterations (Stage~$2$)}
The major drawback of Stage~$1$ is that estimating the weights of similar patches is imprecise: only a quarter of pixels are used in the target patch and similar patches. In Stage~$2$, we estimate the weights by involving all the pixels in similar patches to approximate all the pixels in the target patch. The weights $\bm{\omega}^f_i$ is solved via:
\begin{equation}
\label{eq:grr_solution_s2}
\bm{\omega}^f_i=\left({\bm{Q}^f_i}^{\top}\bm{Q}^f_i+\lambda \bm{\Sigma}_i^{f} \right)^{-1} \left({\bm{Q}^f_i}^{\top}{{\bm{P}^G_i}} \right),
\end{equation}
where the notations are identical to those in~\eqref{eq:grr}. By several iterations, we obtain the output image $\bm{I}_{\mathrm{S2}}$ from Stage~$2$. The details of Stage~$2$ is described in Algorithm~\ref{alg:stage2}.

\begin{algorithm}[tb]
\label{alg:stage2}
\SetKwInOut{Input}{Input}
\SetKwInOut{Output}{Output}
\Input{input image $\bm{I}_{S1}$; the number of target patches in $\bm{I}_{S1}$, $l$; patch size $n$; the number of each target patch's OE, EO or EE similar patches $K$; search window size $W$; the penalty constant $\lambda$; the parameter in the similar metric $c_w$, the number of iterations $\mathscr{I}$.}
\Output{the output image from Stage 2: $\bm{I}_{S2}$.}
$\bm{I}_G = \bm{I}_{S1}$. \\
\For{iteration $\leftarrow 1$ \KwTo $\mathscr{I}$}{
\For{i $\leftarrow 1$ \KwTo $l$}{
Extract $\bm{P}_i^{G}$ from $\bm{I}_{G}$ as the $i$-th target patch. Set $\bm{P}_i = \bm{P}_i^{G}$.\\
\For{f $\leftarrow OE, EO, EE$ }{
Find $K$ $f$ similar patches to $\bm{P}_i^{G}$ via~\eqref{eq:similarity_metric1}.\\
Store similar patches in $\bm{Q}^f_i$ and the similarity measures in $\bm{\Sigma}^f_i$. \\
Compute $\bm{\omega}_i^{f}$ given $\bm{P}_i^{G}$, $\bm{\Sigma}^f_i$, $\bm{Q}^f_i$ and $\lambda$ via~\eqref{eq:grr_solution_s2}.\\
}
Generate ${\hat{\bm{P}}_i}$ given $\bm{\omega}_i^{OE}$, $\bm{\omega}_i^{OE}$, $\bm{\omega}_i^{OE}$ and $\bm{P}_i$ via~\eqref{eq:piupdate}.
}
Generate $\bm{I}_{G}$ given all $\hat{\boldsymbol{P}}_i$ via~\eqref{eq:synthesize}.
}
$\bm{{I}}_{S2} = \bm{I}_{G}$.
\caption{Stage~$2$}
\end{algorithm}

\subsection{The Interpolation Scheme in Later Iterations (Stage~$3$)}

\begin{algorithm}[tb]
\label{alg:stage3}
\SetKwInOut{Input}{Input}
\SetKwInOut{Output}{Output}
\Input{input image $\bm{I}_{S2}$; $\bm{I}_{S2}$'s mean, $c$; the number of target patches in $\bm{I}_{S2}$, $l$; patch size $n_A$, $n_B$ in early and later iterations; the number of each target patch's similar patches $K$; search window size $W_A$, $W_B$ in early and later iterations; the penalty constants $\lambda_A$, $\lambda_B$ in early and later iterations; the number of early and later iterations $\mathscr{I}_A$, $\mathscr{I}_B$.}
\Output{the output image from Stage 3: $\bm{I}_{S3}$.}
$\bm{I}_G = \bm{I}_{S2} - c$. \\
\For{iteration $\leftarrow 1$ \KwTo $\mathscr{I}_A + \mathscr{I}_B$}{
\eIf{iteration $<=\mathscr{I}_A$}{
Use $n_A$, $W_A$, $\lambda_A$.}
{
 Use $n_B$, $W_B$, $\lambda_B$.}
\For{i $\leftarrow 1$ \KwTo $l$}{
Extract $\bm{P}_i^{G}$ from $\bm{I}_{G}$ as the $i$-th target patch. Set $\bm{P}_i = \bm{P}_i^{G}$.\\
Find $K$ similar patches to $\bm{P}_i^{G}$ based on~\eqref{eq:similarity_metric2}.\\
Store similar patches in $\bm{Q}_i$ and the similarity measures in $\bm{\Sigma}_i$. \\
Compute $\bm{\omega}_i$ given $\bm{P}_i^{G}$, $\bm{\Sigma}_i$, $\bm{Q}_i$ and $\lambda$ via~\eqref{eq:grr_solution_s3}.\\
Generate ${\hat{\bm{P}}_i}$ given $\bm{\omega}_i$ and $\bm{P}_i$ via~\eqref{eq:piupdate2}.
}
Generate $\bm{I}_{G}$ given all $\hat{\boldsymbol{P}}_i$ via~\eqref{eq:synthesize}.
}
$\bm{{I}}_{S3} = \bm{I}_{G} + c$.
\caption{Stage~$3$}
\end{algorithm}

The major drawback of Stage~$1$ and Stage~$2$ is that for each target patch, only OE, EO and EE similar patches are used and OO similar patches are ignored. To address the drawback, Stage~$4$ exploits OO similar patches to refine the pixels off the measurement grid in each target patch since these pixels are close to ground-truth after iterations in the previous two stages. Thus, we approximate each target patch as a linear combination of all the similar patches regardless of OO, OE, EO or EE patches. Further, the target patch is not limited to a OO patch, but can be any OO, OE, EO, EE patch.

Mathematically, we use a target patch and its similar patches ($K$ patches in total) to approximate the target patch itself. The similar patches is searched based on a new similarity metric:
\begin{equation}
\label{eq:similarity_metric2}
S(\bm{P}^G_i,\bm{P}^{G}_{i,j})=\left({\bm{P}^G_i}^{\top}{\bm{P}^{G}_{i,j}}\right)/\left({\left\| \bm{P}^G_i\right\|_{2}}{\left\| \bm{P}^{G}_{i,j}\right\|_{2}}\right).
\end{equation}
The motivation of the new similarity metric is to involve correlated patches as the similar patches. It is likely that the patches of the edges along the same directions, of the same sharpness, but with different polarities can be considered as the same patch. Prior to computing the similarity metric, we subtract the input image by the mean of the measured pixel values to involve similar patches with different polarities.

The weights of $K$ patches, $\bm{\omega}_i$, is computed via:,
\begin{equation}
\label{eq:grr_solution_s3}
\bm{\omega}_i=\left({\bm{Q}_i}^{\top}\bm{Q}_i+\lambda \bm{\Sigma}_i \right)^{-1} \left({\bm{Q}_i}^{\top}{{\bm{P}^G_i}} \right),
\end{equation}
where $\bm{Q}_i$ refers to the matrix of size $n^2\times K$ that stores all the $K$ patches; $\bm{\Sigma}_i=\diag\left(\frac{{\bm{S}(1)}}{{\bm{S}(1)}},\dots,\frac{{\bm{S}(1)}}{{\bm{S}(i)}},\dots, \frac{{\bm{S}(1)}}{{\bm{S}(K)}}\right)$ is a diagonal matrix that penalizes the weights of less similar patches in $\bm{I}_G$. Given $\bm{\omega}_i$, the updated patch $\hat{\bm{P}}_i$ is solved by:
\begin{equation}
\label{eq:piupdate2}
\hat{\bm{P}}_i= \bm{M}^{r}\sum_{j=1}^{K}\omega_{i,j}  \bm{P}_{i,j}^{G}+ \bm{M}^{OO} \bm{P}^G_i,
\end{equation}
where $\bm{M}^{r}=\bm{M}^{OE}+\bm{M}^{EO}+\bm{M}^{EE}$; $\omega_{i,j}$ is a scalar that denotes the weight of $j$-th similar patch of $\bm{P}^G_i$. After synthesizing all updated patches, we obtain a new image and complete one iteration. After a few iterations, we add the aforementioned mean and obtain the output image with the details shown in Algorithm~\ref{alg:stage3}. Note that we choose a smaller patch size ($n_B<n_A$) after $i_A$ iterations in Algorithm~\ref{alg:stage3} to exploit semi-local similarity at a finer scale.

\subsection{The Interpolation Scheme in Later Iterations (Stage~$4$)}
In previous stages, our approach to estimating the weights of less similar patches is ad-hoc: we simply assign smaller weights. To better handle less similar patches, we consider a similar patch as a combination of similar and dissimilar structures to the target patch. Dissimilar structures may include mild aliasing pattern distinct from the target patch and shall capture a smaller portion of energy compared with similar structures among the group of similar patches and the target patch. This motivates us to represent the group of patches via a few bases by applying SVD to the group while only preserving dominant principal components. We denote the group corresponding to the $i$-th target patch as $\bm{X}_i \in \mathbb{R}^{k\times n^2}$ with $k$ patches and each patch is stored as a row with $n^2$ pixels. Also we denote the centered version of $\bm{X}_i$ as $\tilde{\bm{X}}_i$. $\tilde{\bm{X}}_i=\bm{X}_i-\bar{\bm{X}}_i$, where each row of $\bar{\bm{X}}_i$ is the mean of $k$ patches. We also denote the target group as $\tilde{\bm{X}}^{lr}_i$. $\tilde{\bm{X}}^{lr}_i$ is spanned by only a few principal components and can be computed via low-rank approximation.

\begin{algorithm}[tb]
\label{alg:stage4}
\SetKwInOut{Input}{Input}
\SetKwInOut{Output}{Output}
\Input{input image $\bm{I}_{S3}$; the number of target patches in $\bm{I}_{S3}$, $l$; patch size $n_A$, $n_B$ in early and later iterations; variance threshold of each target patch $\mathrm{th}_A$, $\mathrm{th}_B$ in early and later iterations; the number of each target patch's similar patches $K$; search window size $W$; the penalty constants $\alpha_A$, $\lambda_B$ in early and later iterations; the number of early and later iterations $\mathscr{I}_A$, $\mathscr{I}_B$.}
\Output{the output image from Stage 4: $\bm{I}_{S4}$.}
\For{iteration $\leftarrow 1$ \KwTo $\mathscr{I}_A + \mathscr{I}_B$}{
\eIf{iteration $<=\mathscr{I}_A$}{
Use $n_A$, $\alpha_A$, $\mathrm{th}_A$.}
{
 Use $n_B$, $\alpha_B$, $\mathrm{th}_B$.}
\For{i $\leftarrow 1$ \KwTo $l$}{
Extract $\bm{P}_i^{G}$ from $\bm{I}_{G}$ as the $i$-th target patch.\\
Group Searched $K$ similar patches of $\bm{P}_i^{G}$ into $\bm{X}_i$. Compute $\tilde{\bm{X}}_i$, $\bar{\bm{X}}_i$. \\
\eIf{$\mathrm{var}(\bm{P}_i^{G})>\mathrm{th}$}{
Compute $\tilde{\bm{X}}^{lr}_i$ given $\tilde{\bm{X}}_i$ via~\eqref{eq:low_rank3}.}
{
 $\tilde{\bm{X}}^{lr}_i=\tilde{\bm{X}}_i$.}
}
Generate $\bm{I}_{G}$ given all $\tilde{\bm{X}}^{lr}_i$ and $\bar{\bm{X}}_i$ via~\eqref{eq:synthesize}. \\
 Force $\bm{I}_{G}$ to have the same pixel values on the downsampling grid as $\bm{I}_{S3}$ has.
}
$\bm{{I}}_{S4} = \bm{I}_{G}$. \\
\caption{Stage~$4$}
\end{algorithm}

Noticing that a principal component corresponding to a larger singular value carries fewer dissimilar structures than a principal component corresponding to a smaller singular value does, we choose to less shrink a larger singular value and more shrink a smaller singular value. Thus, we use the Weighted Nuclear Norm Minimization (WNNM)~\cite{gu2014weighted} as below:
\begin{equation}
\label{eq:low_rank2}
\min_{\tilde{\bm{X}}^{lr}_i}\frac{1}{2}\|\tilde{\bm{X}}^{lr}_i-\tilde{\bm{X}}_i\|^2_\mathrm{F}+\alpha\sum_j \omega_j \|\sigma_j(\tilde{\bm{X}}^{lr}_i)\|_1,
\end{equation}
where $\sigma_j(\tilde{\bm{X}}^{lr}_i)$ is the $j$-th singular value of $\tilde{\bm{X}}^{lr}_i$; $\omega_j$ is the weight on the $j$-th singular value; $\alpha$ is a parameter that controls the regularization strength. Since a larger singular values deserves less shrinkage, we choose a smaller weight. As mentioned in~\cite{dong2012nonlocal}, we the empirical distribution of singular values can be modeled as a Laplacian. Given this prior and inspired by BayesShrink in~\cite{chang2000adaptive}, we choose the threshold as:
\begin{equation}
\label{eq:weight_low_rank}
  \omega_j =\frac{ \alpha }{\hat{\sigma}_j+\epsilon},
\end{equation}
where $ \omega_j$ is the weight corresponding to the $j$-th singular value of $\tilde{\bm{X}}^{lr}_i$; $\hat{\sigma}_j=\sqrt{(\sigma_j(\tilde{\bm{X}}^{lr}_i)^2/n^2)}$ is the estimation of signal standard deviation given the $i$-th singular value; $\epsilon$ is a small constant. Based on~\cite{gu2014weighted}, the solution of~\eqref{eq:low_rank2} is:
\begin{equation}
\label{eq:low_rank3}
\tilde{\bm{X}}^{lr}_i=\bm{U}_i{\mathcal{S}}_{\bm{\omega}_i}(\bm{D}_i)\bm{V}_i^{\top},
\end{equation}
where $\bm{U}_i\bm{D}_i\bm{V}_i^{\top}=\tilde{\bm{X}}_i$ and ${\mathcal{S}}_{\bm{\omega}_i}$ is the generalized soft-thresholding operator with the weighted vector $ {\bm{\omega}_i}$ as the threshold to threshold the diagonal elements of $\bm{D}_i$, i.e. $ \left[{\mathcal{S}}_{\bm{\omega}_i}(\bm{D}_i)\right]_{j,j}=\max(\left[\bm{D}_i\right]_{j,j}-\alpha\omega_j,0)$.

The estimated image will be computed via averaging the contribution of relevant groups of patches to each pixel:
\begin{equation}
\label{eq:synthesize_group}
\left(\sum_{i=1}^{l}\boldsymbol{R}_{G_i}^{\top} (\tilde{\boldsymbol{X}}^{lr}_{i}+{\bar{\boldsymbol{X}}}_{i}) \right)./\left(\sum_{i=1}^{l} \boldsymbol{R}_{G_i}^{\top} \boldsymbol{1}_{n^2\times k}\right),
\end{equation}
where $\boldsymbol{R}_{G_i}(\cdot)$ extracts the group of $k$ patches corresponding to the $i$-th target patch; $\boldsymbol{R}^{\top}_{G_i}(\cdot)$ puts back the $i$-th group of patches into their original positions and pads with zeros elsewhere; $./$ stands for element-wise division of two matrices; $l$ is the number of grouped patches in the input image; $\boldsymbol{1}_{n^2\times k}$ is a matrix of size $n^2\times k$ with all the elements being $1$. After multiple iterations, we generate the output image from Stage~$4$. The details of Stage~$4$ are in Algorithm~\ref{alg:stage4}.

\section{The Overall Interpolation Scheme}
\label{sec:mister_overall}

\begin{figure}[htbp]
\begin{minipage}{\linewidth}
\centering
\centerline{\includegraphics[width=0.8\linewidth]{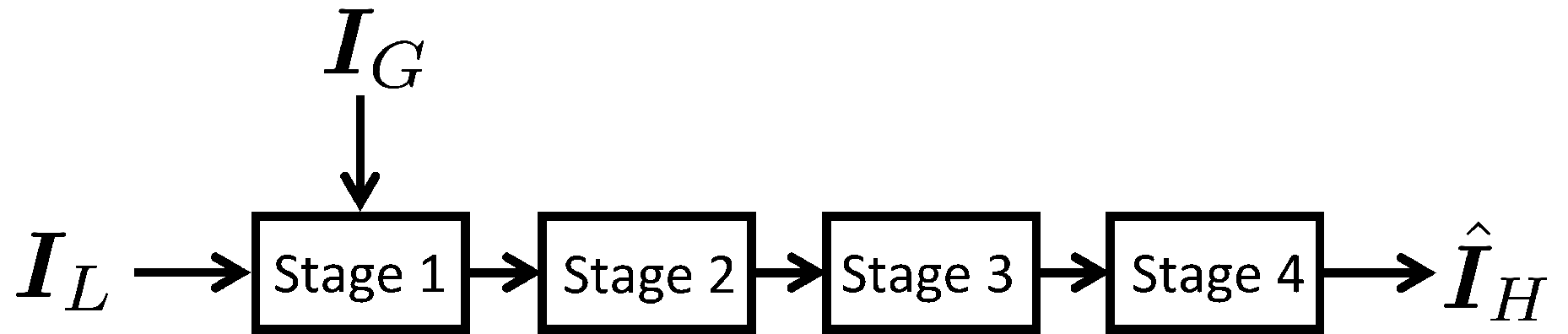}}
\end{minipage}
\caption{The interpolation scheme for interpolating an image by a factor of $2$.}
\label{fig:whole_stage_x2}
\end{figure}

In light of the aforementioned schemes, we summarize our algorithm in Fig.~\ref{fig:whole_stage_x2}. We first generate a guide image $\bm{I}_G$ and then use $\bm{I}_G$ to initialize the positions and weights of similar patches for Stage~$1$. Through the cascade of interpolation stages, we generate the interpolated image $\hat{\bm{I}}_H$.

\subsection{Handling the Bias Towards Dominant Edges}
\label{subsec:handling_artifacts}
\begin{figure}[htbp]
\centering
\begin{minipage}{0.25 \linewidth}
  \centering
  \centerline{\includegraphics[width=0.95\linewidth]{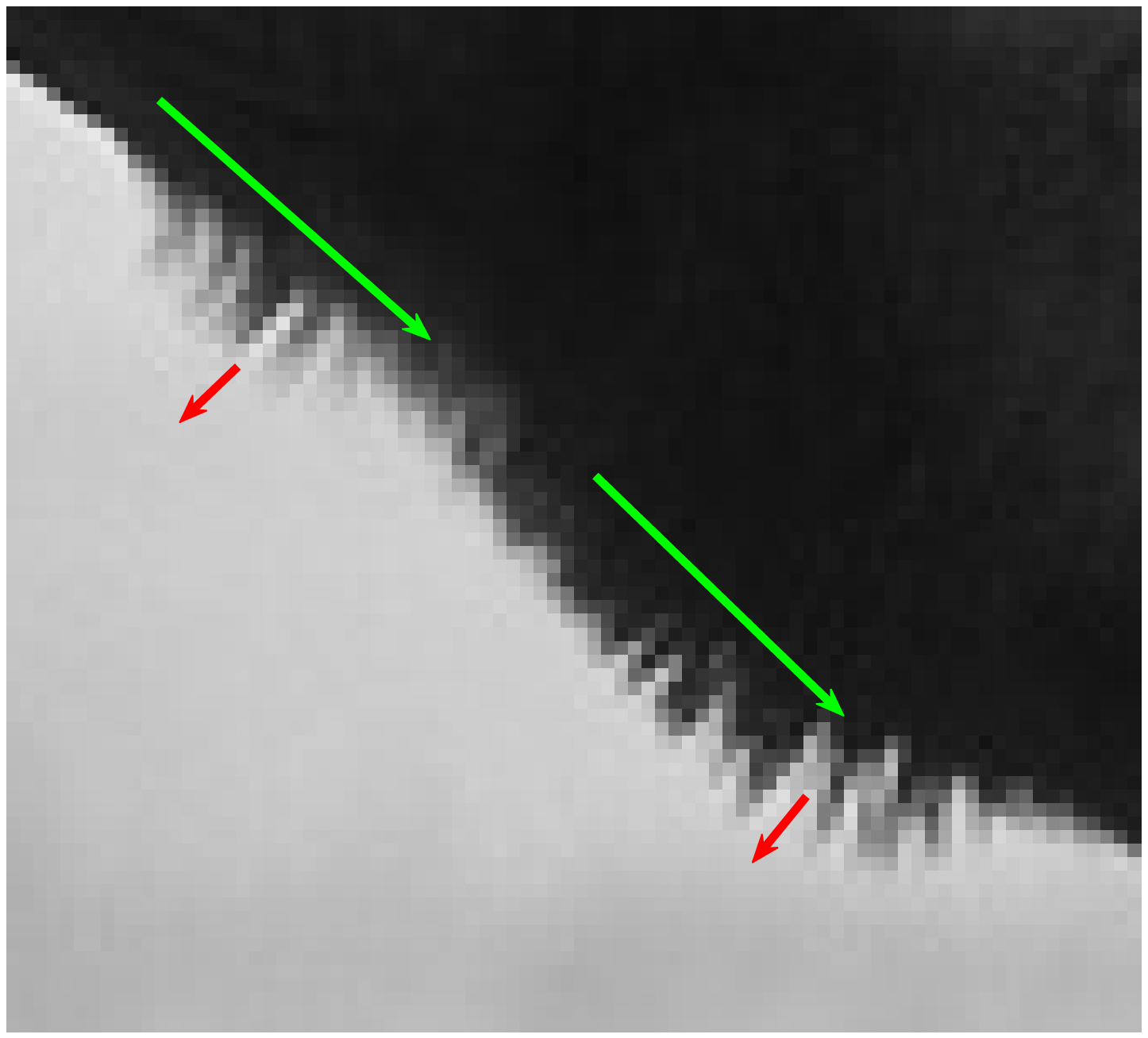}}
  \centerline{(a)} \medskip
  \end{minipage}
\begin{minipage}{0.25 \linewidth}
  \centering
  \centerline{\includegraphics[width=0.95\linewidth]{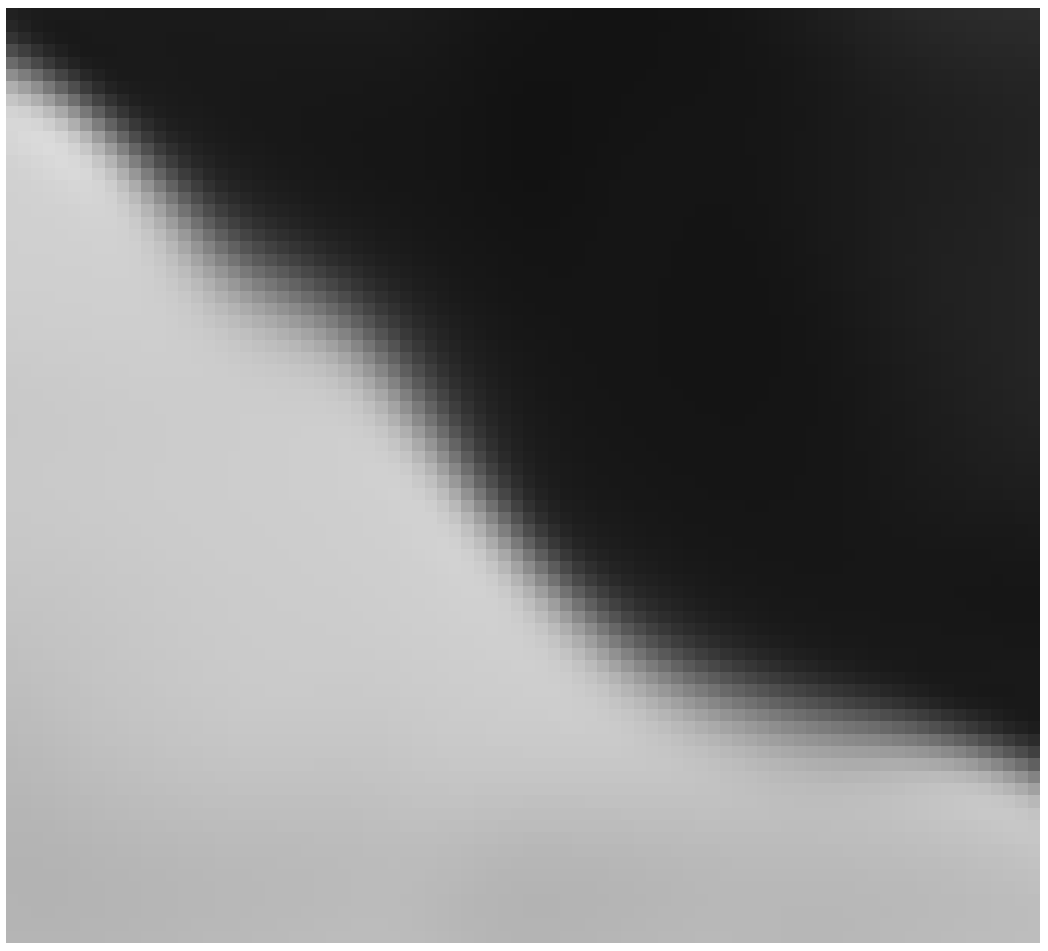}}
  \centerline{(b)}\medskip
\end{minipage}
\begin{minipage}{0.25 \linewidth}
  \centering
  \centerline{\includegraphics[width=0.95\linewidth]{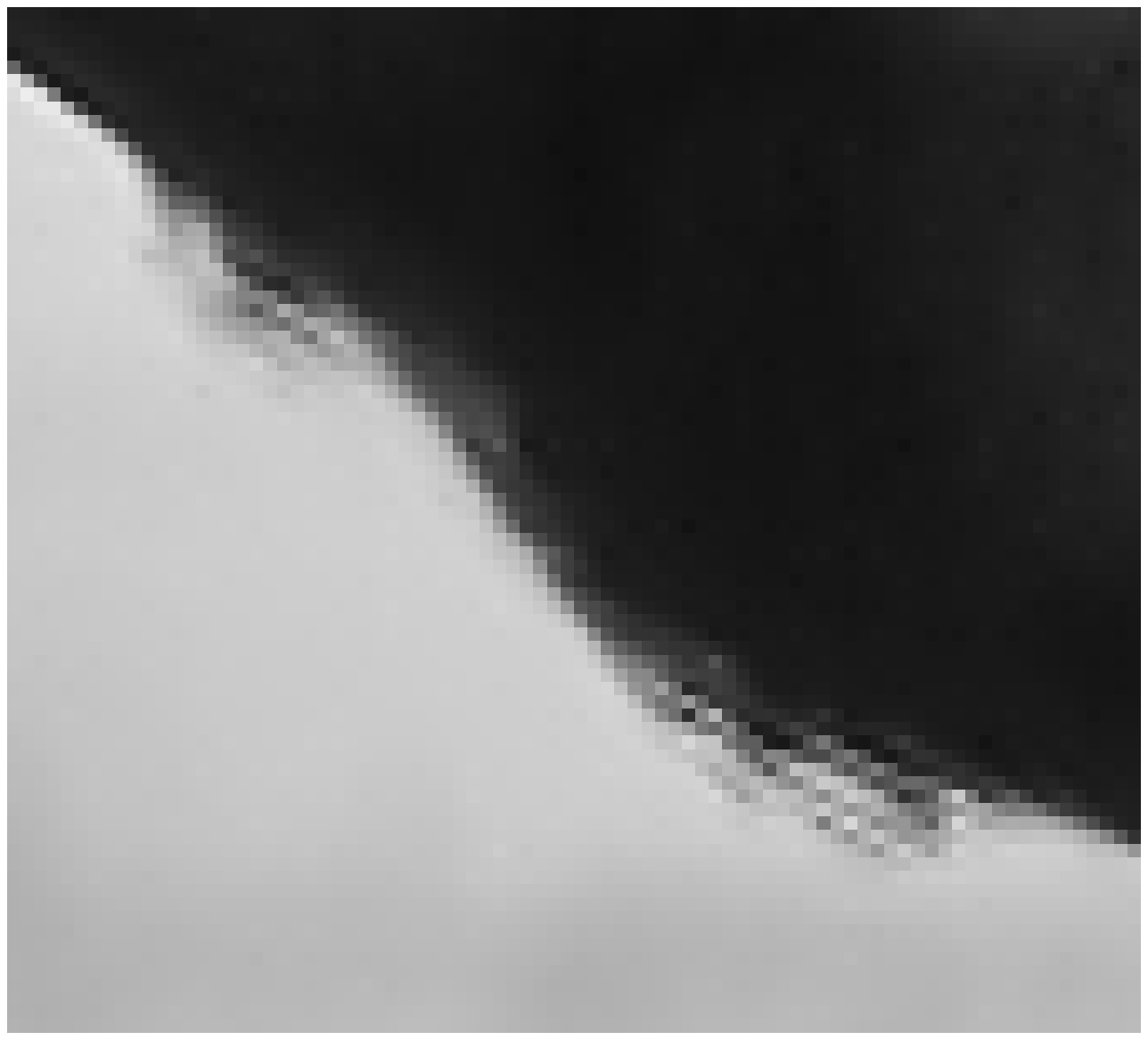}}
  \centerline{(c)}\medskip
\end{minipage}
\caption{The artifacts from using $\bm{I}_{ar}$ as the guide image in Fig.~\ref{fig:whole_stage_x2}. (a) The region of interest in ground-truth~\textit{Elk} image. The green arrows refer to the directions of the dominant edge (neck) and the red arrows refer to the directions of the secondary edges (fur). (b) The region of interest in $\bm{I}_{ar}$ after applying Algorithm~\ref{alg:stage_algorithm_removal}. (c) The region of interest in the output image via using $\bm{I}_{ar}$ as~$\bm{I}_{G}$ (guide image) in Fig.~\ref{fig:whole_stage_x2}.}
\label{fig:artifacts}
\end{figure}

Since the spatially-variant projection in Algorithm~\ref{alg:stage_algorithm_removal} discards the energy spanned by less dominant components, the less dominant directional features will be removed when secondary edges carry distinct directions to directions of the dominant edges. We denote the phenomenon as the bias towards dominant edges and demonstrate it in Fig.~\ref{fig:artifacts}(b). The original local directions~(the red arrows in Fig.~\ref{fig:artifacts}(a)) whose directions are drastically different from the dominant directions~(the green arrows in Fig.~\ref{fig:artifacts}(a)) become invisible in Fig.~\ref{fig:artifacts}(b). Using Fig.~\ref{fig:artifacts}(b) as the guide image will result to wrong similar patches and subsequently gives rise to visual artifacts in the interpolated images. Fig.~\ref{fig:artifacts}(c) demonstrates such artifacts in the interpolated image through Stage~$1-4$ by treating $\bm{I}_{ar}$ as the guide image $\bm{I}_G$ in Fig.~\ref{fig:whole_stage_x2}.
\begin{figure}[b]
\begin{minipage}{\linewidth}
\centering
\centerline{\includegraphics[width=0.65\linewidth]{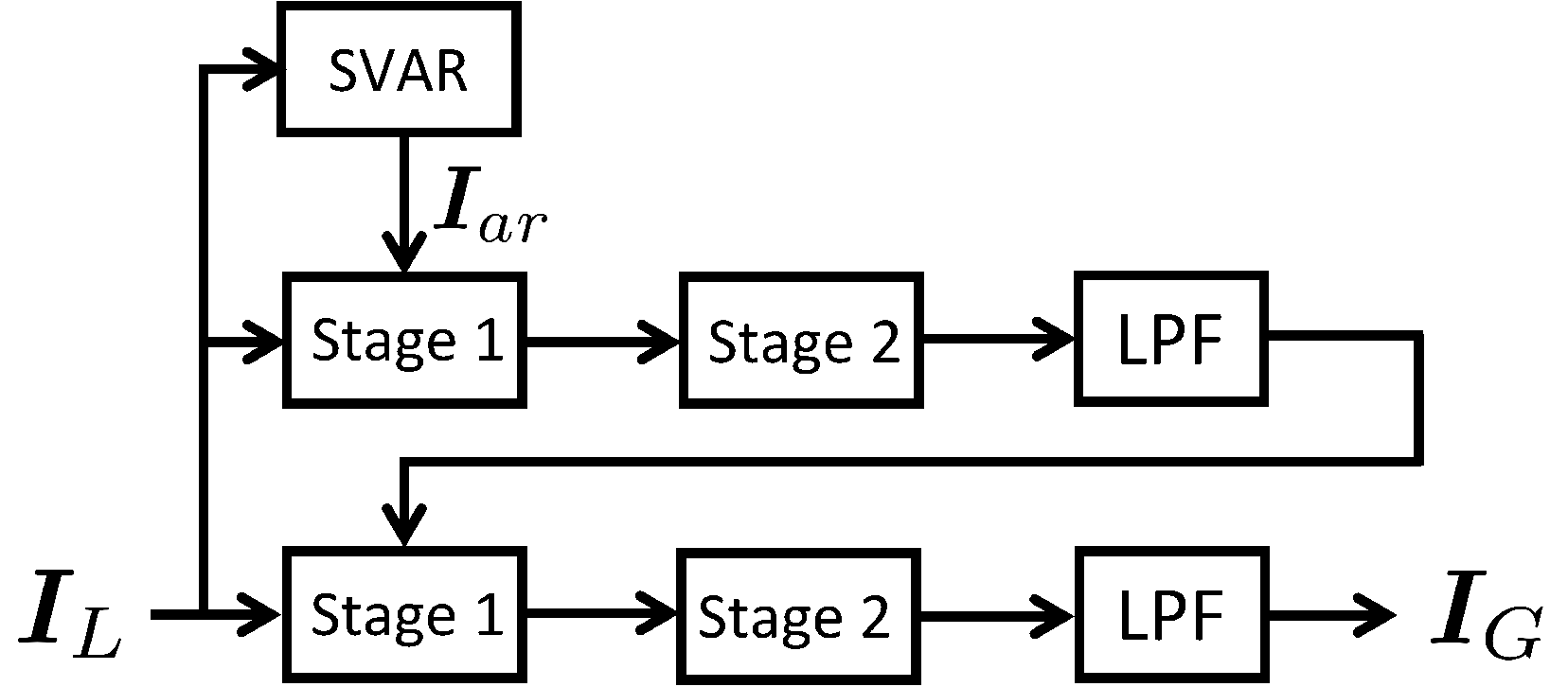}}
\end{minipage}
\caption{The aliasing removal scheme to generate the guide image~$\bm{I}_G$. SVAR block stands for spatially-variant aliasing removal in Algorithm~\ref{alg:stage_algorithm_removal}. LPF block stands for the spatially-invaraint low-pass filtering in~\eqref{eq:convolution_lpf}.}
\label{fig:aliasing_removal}
\end{figure}

To address this bias, we first use $\bm{I}_{ar}$ as the preliminary guide image to generate an interpolated image, blur the interpolated image, and then use the blurred image as a refined guide image to regenerate the interpolated image. The rationale behind this blur operator is that the pixels in the vicinity of removed directional features often carry very high frequency (shown in Fig.~\ref{fig:artifacts}(c)), due to the difference between the interpolated dominant directions and the secondary directions formed by the measured pixels at the downsampling grid.

The scheme to handle the bias is shown in Fig.~\ref{fig:aliasing_removal}. This scheme follows the idea described in the above paragraph but with one more interpolation procedure to more robustly generate the guide image $\bm{I}_{G}$. We treat $\bm{I}_{G}$ from Fig.~\ref{fig:aliasing_removal} as the guide image to guide the initial iteration in Fig.~\ref{fig:whole_stage_x2}.

\subsection{Interpolation by a Factor of $3$}

The scheme of interpolating by a factor of $3$ is mostly analogous to that of interpolating by a factor of $2$ in Fig.~\ref{fig:whole_stage_x2}, except that the scheme of interpolating by a factor of $3$ discards Stage~$4$ and only uses Stage~$1-3$. The rationale behind not using Stage~$4$ is that, the input LR image downsampled by a factor of $3$ from the original HR image is severely aliased and our interpolation algorithm may select wrong similar patches. The performance drop due to ``refining'' the regions with wrong similar patches generally outweighs the gain due to refining the regions with reliable similar patches.

\section{EXPERIMENTS AND RESULTS}

\begin{figure*}[htbp]
\resizebox{1.0\textwidth}{!}{%
\setlength\tabcolsep{0.2 pt}
\begin{tabular}{@{}ccccccccc@{}}
\includegraphics[height=18mm]{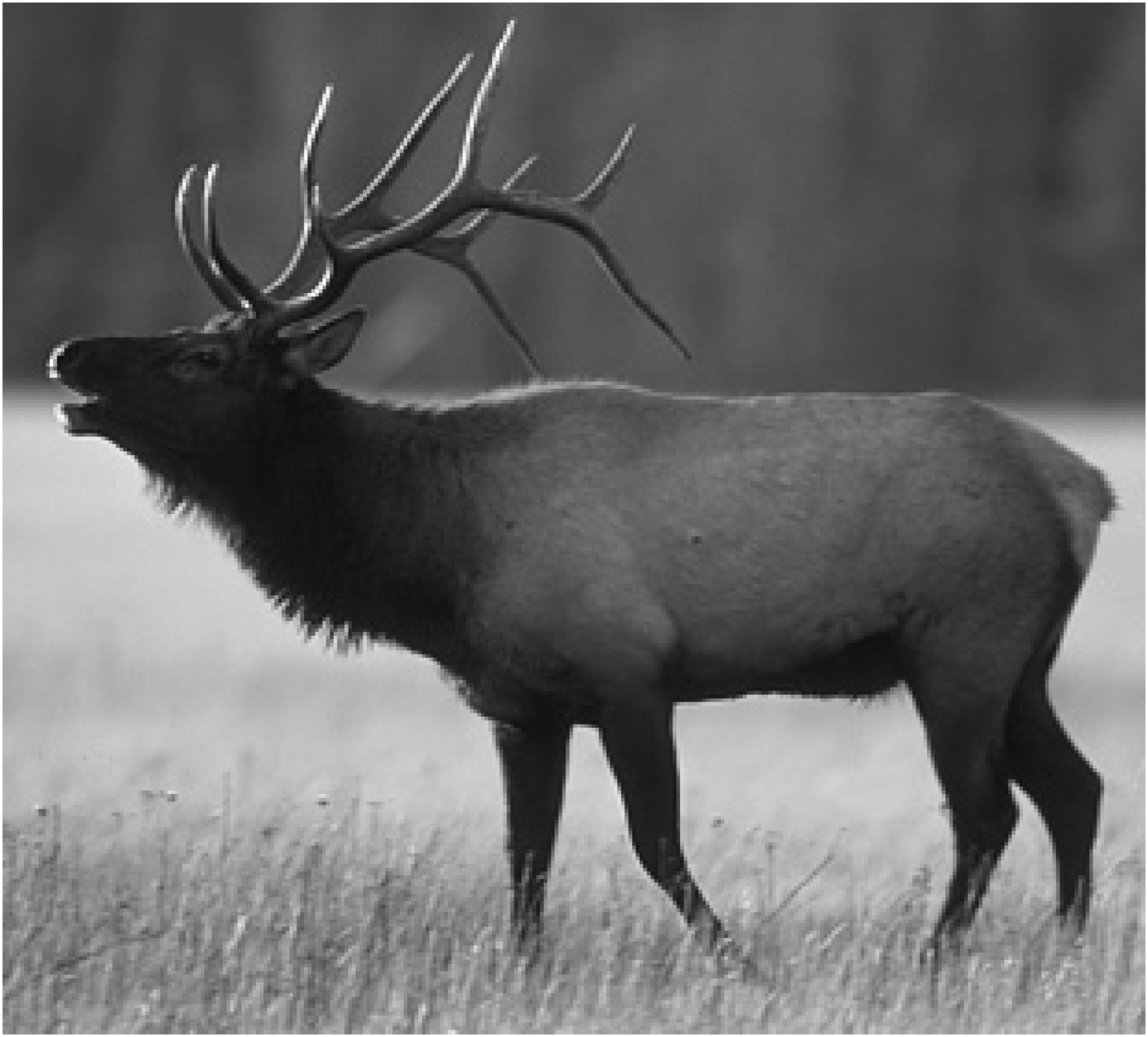} &
\includegraphics[height=18mm]{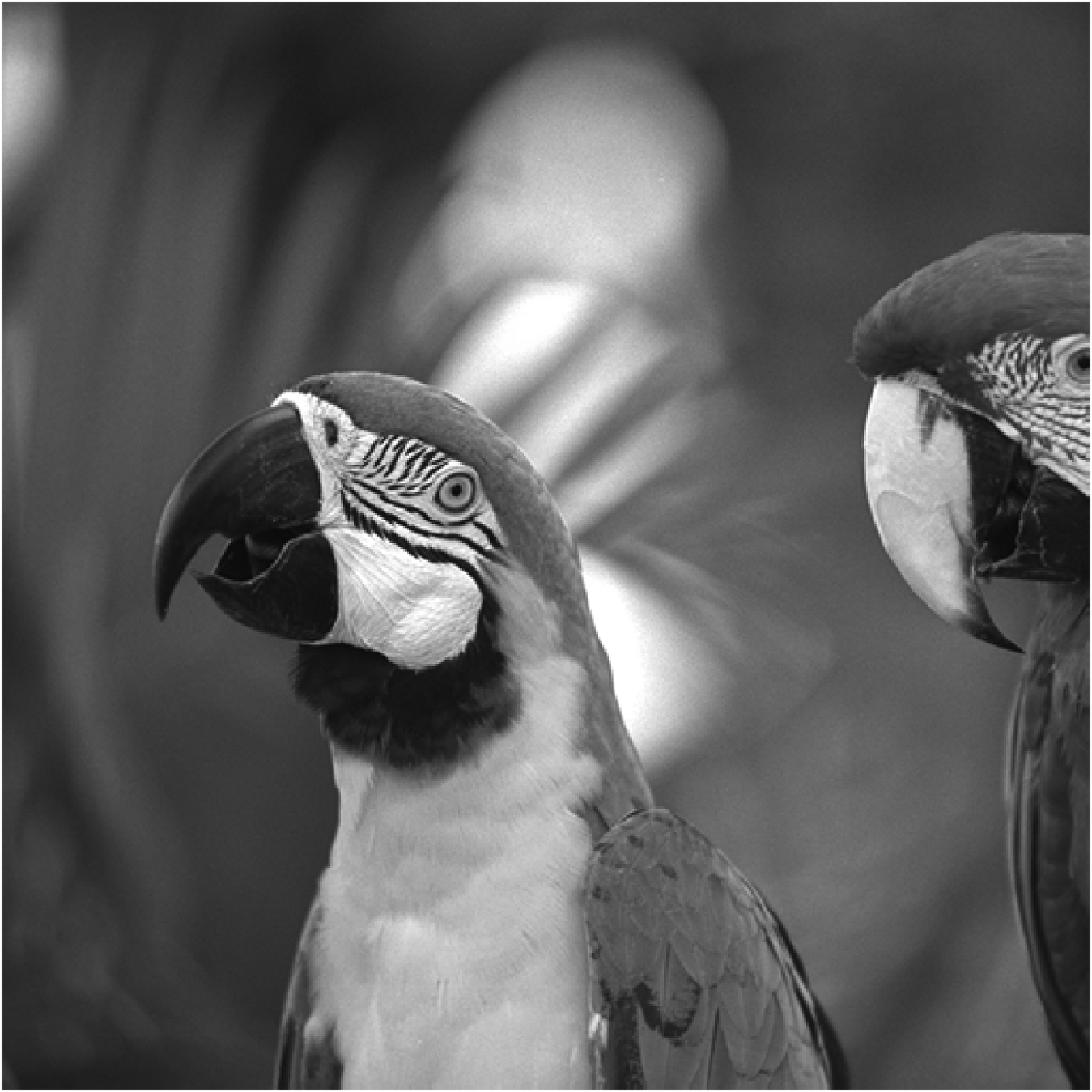} &
\includegraphics[height=18mm]{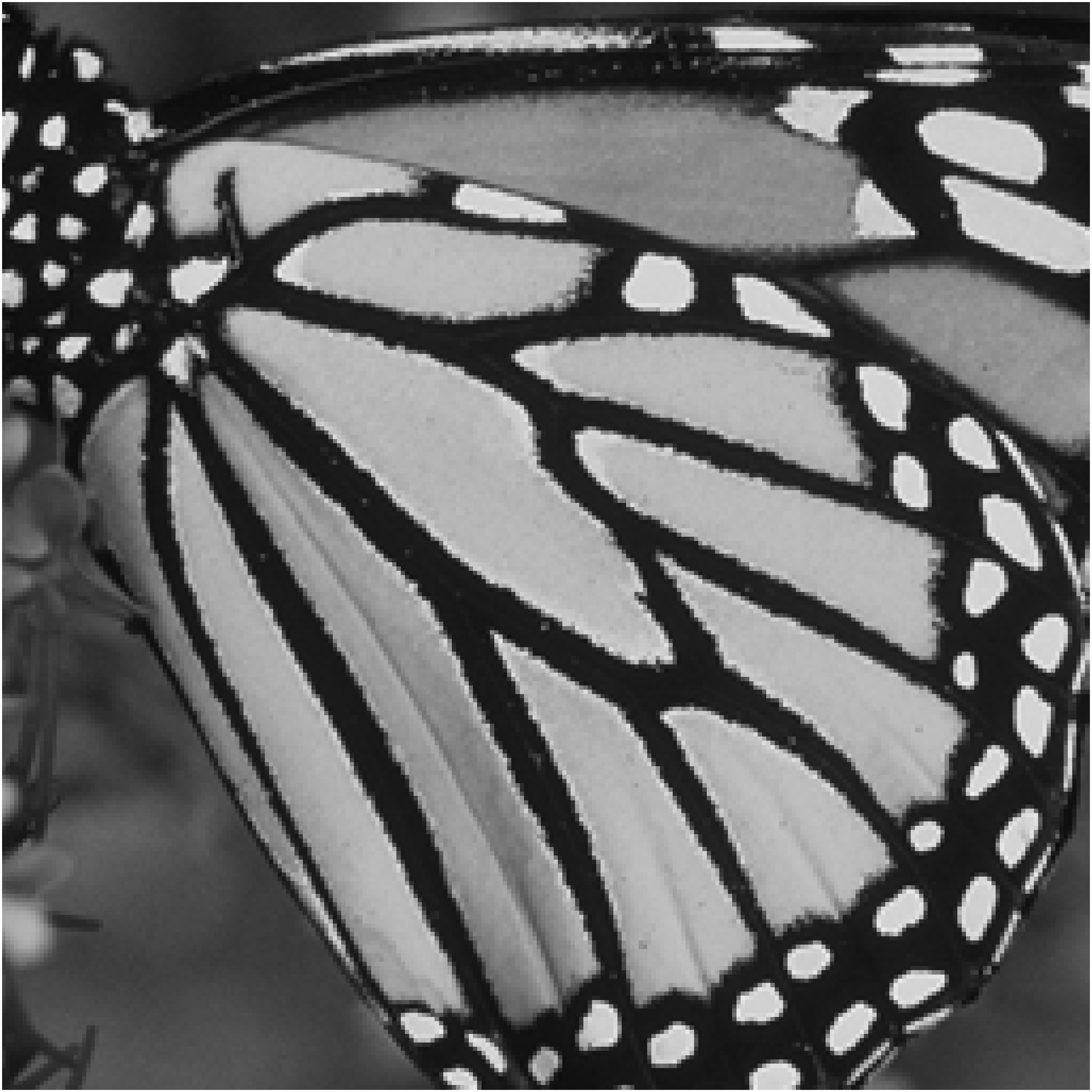} &
\includegraphics[height=18mm]{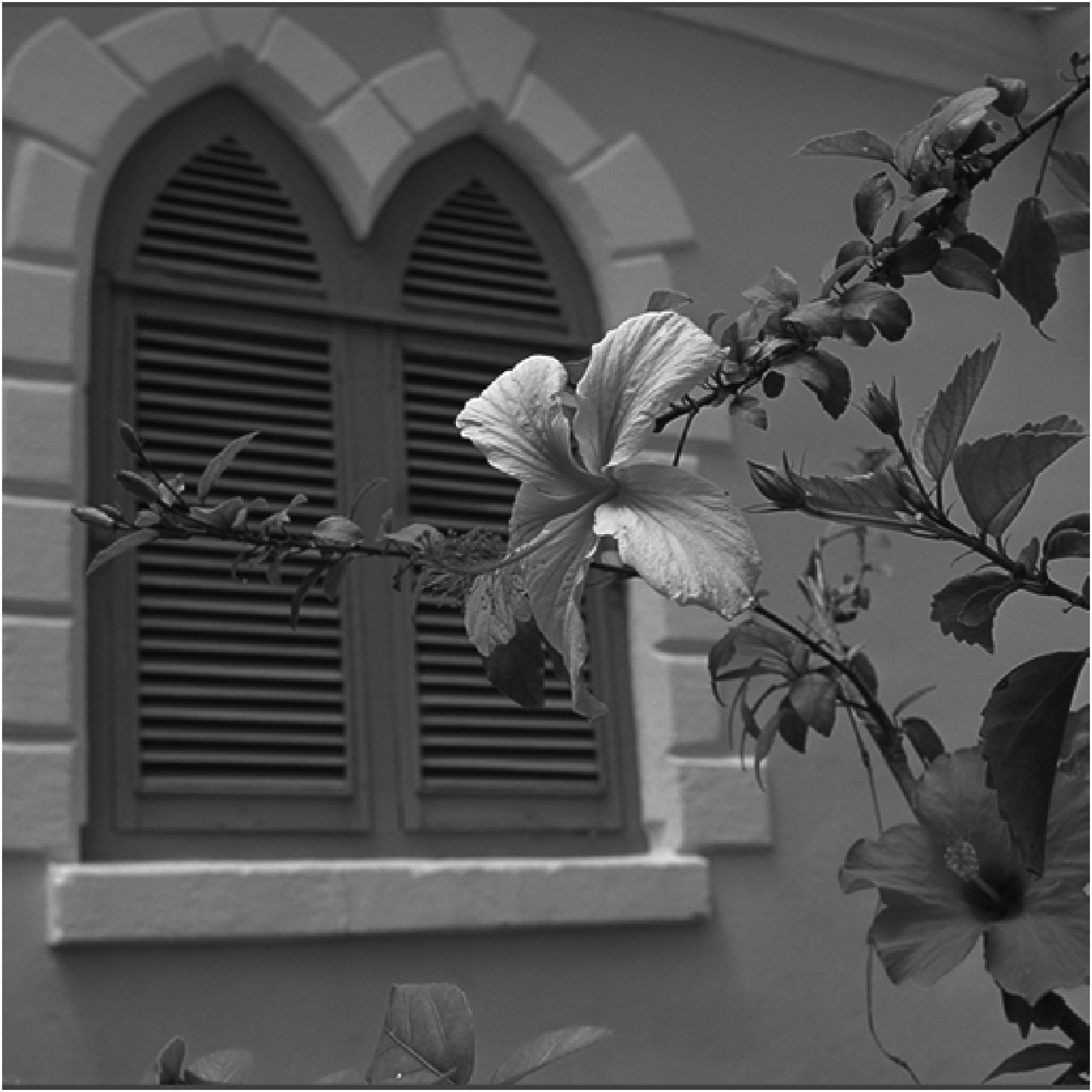} &
\includegraphics[height=18mm]{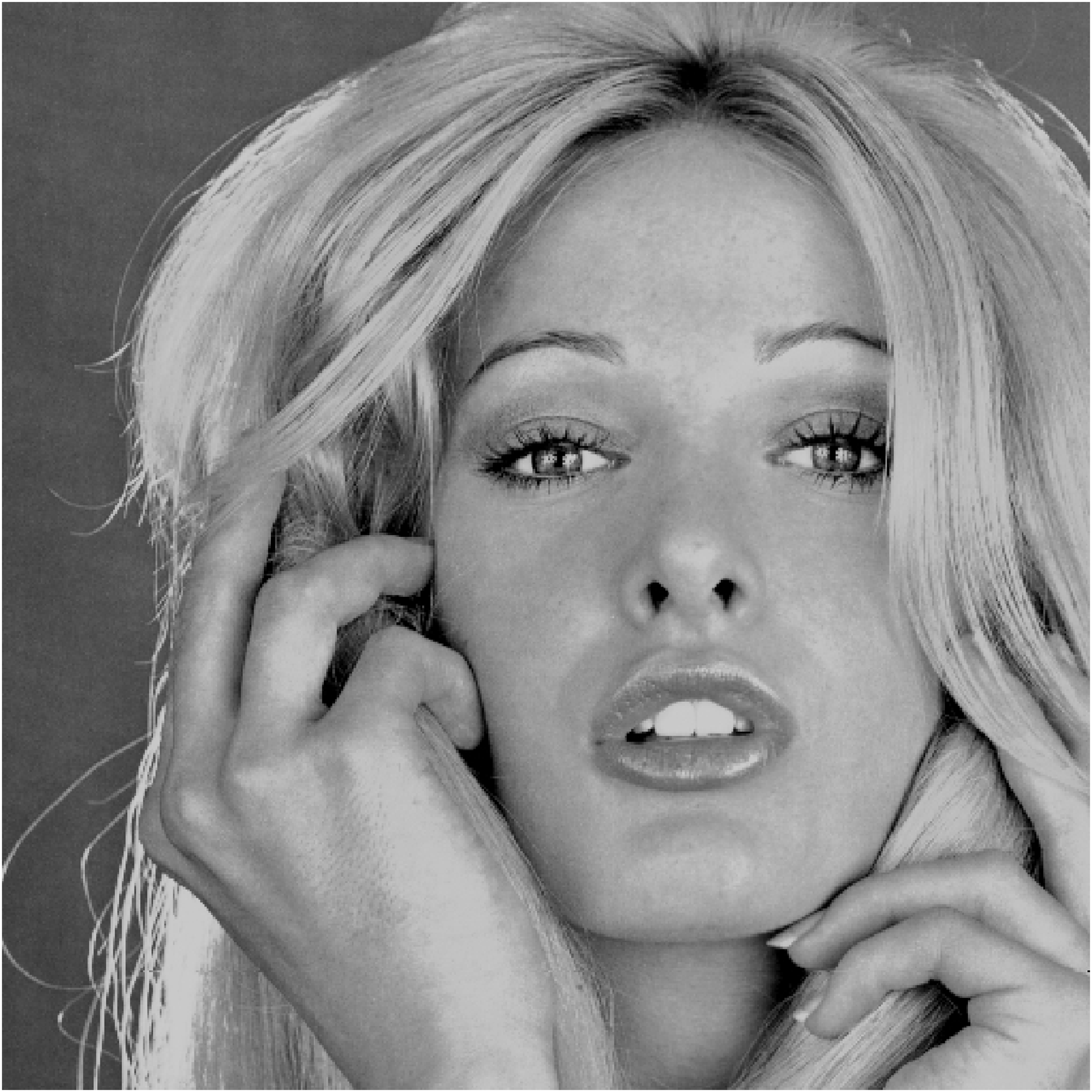} &
\includegraphics[height=18mm]{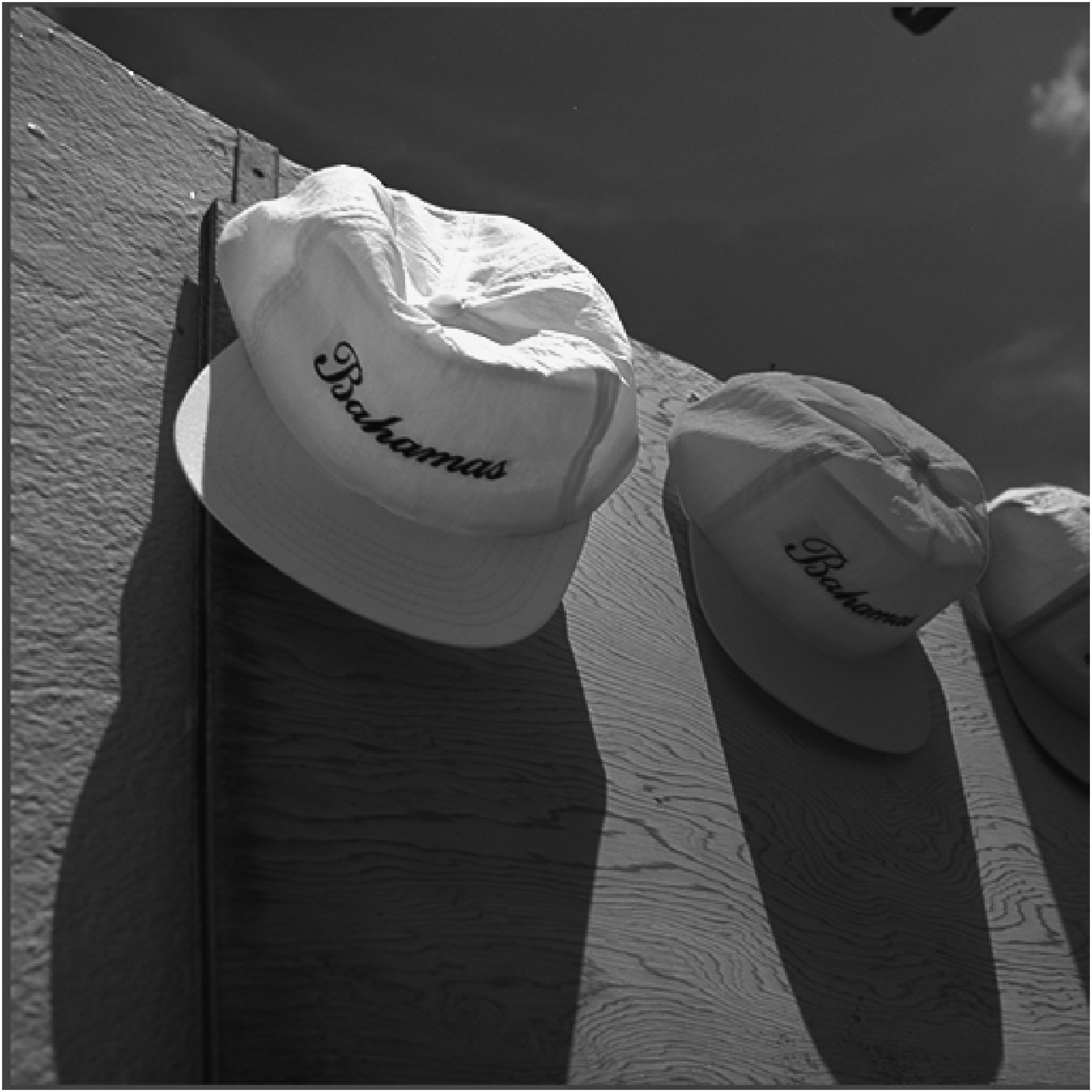} &
\includegraphics[height=18mm]{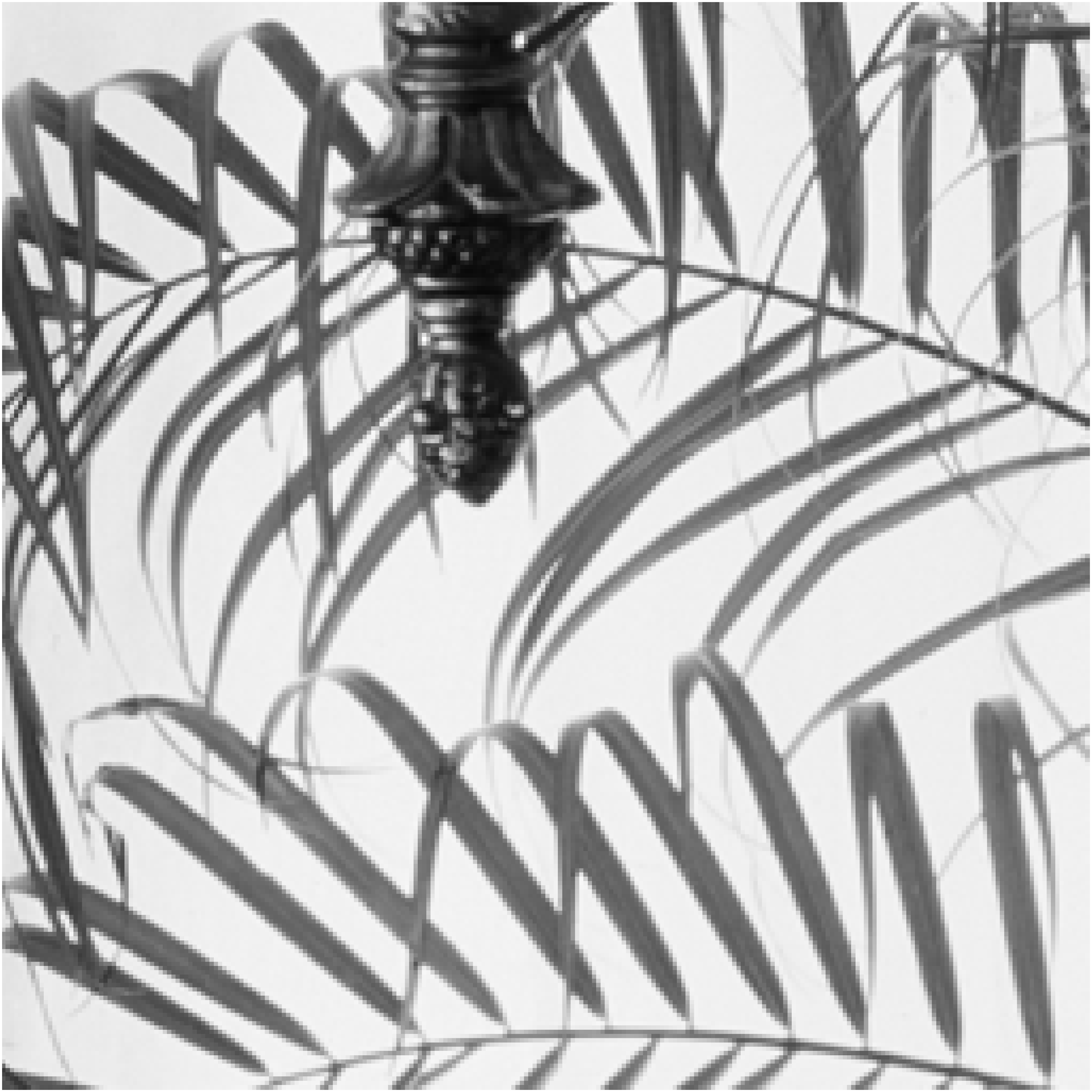} &
\includegraphics[height=18mm]{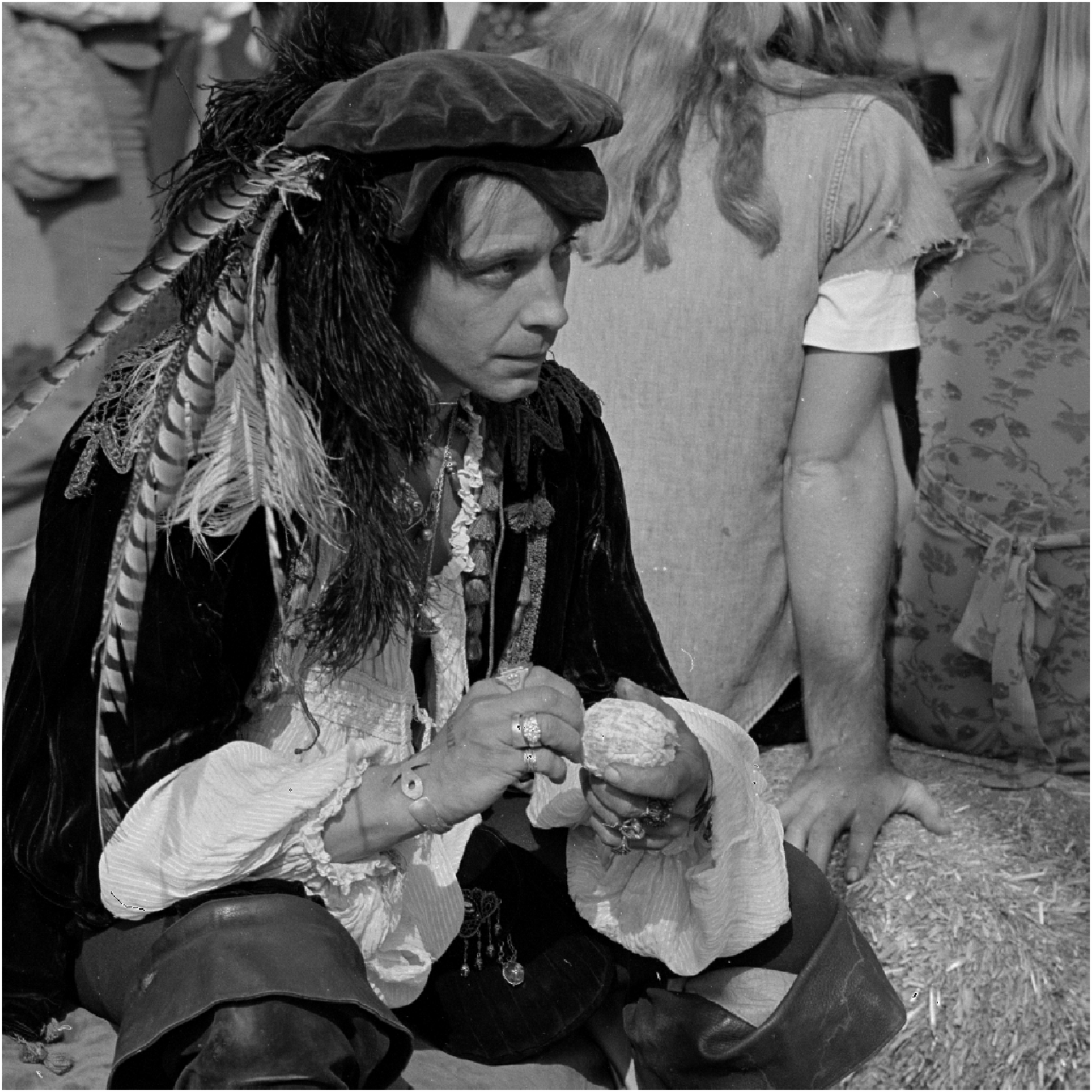} &
\includegraphics[height=18mm]{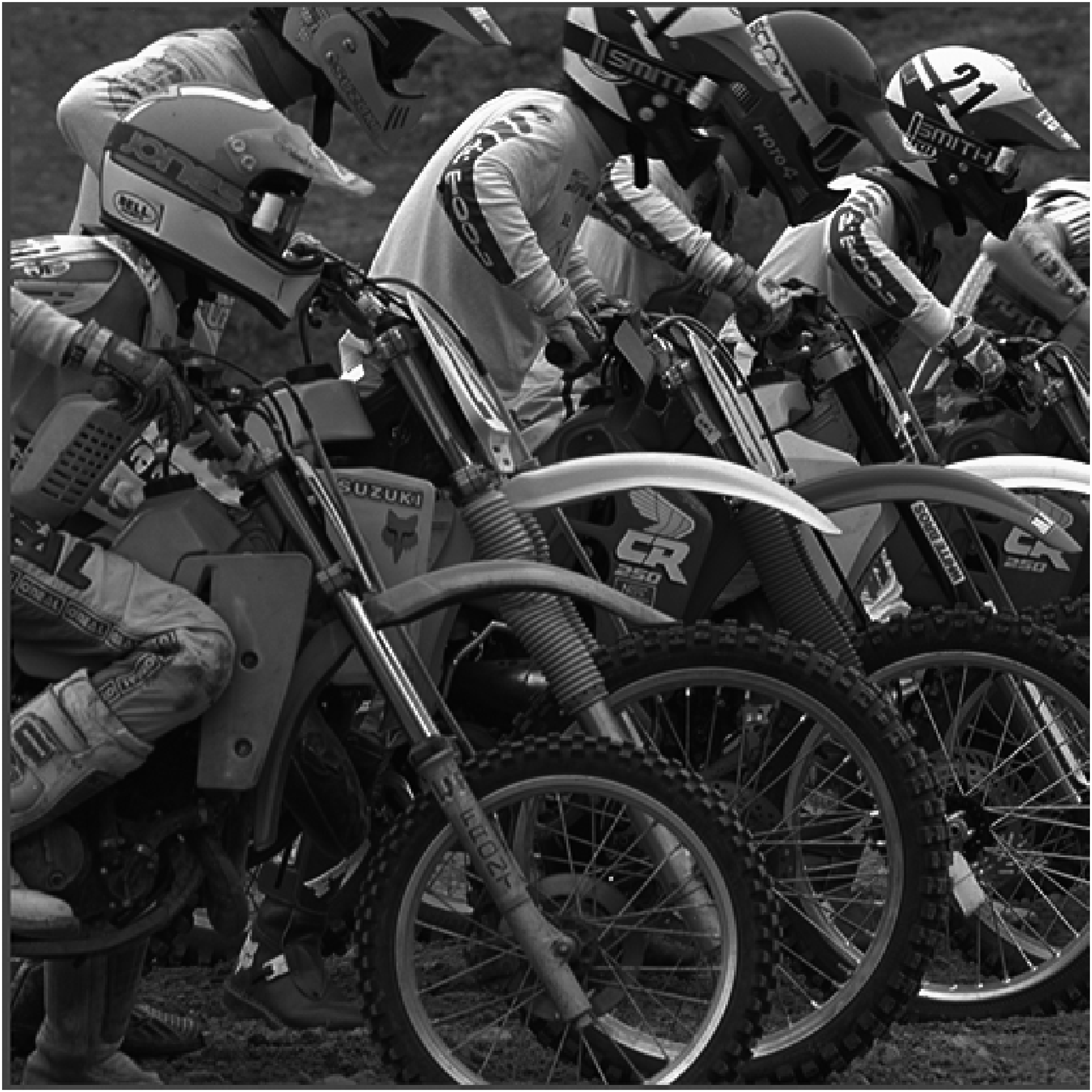} \\
\includegraphics[height=18mm]{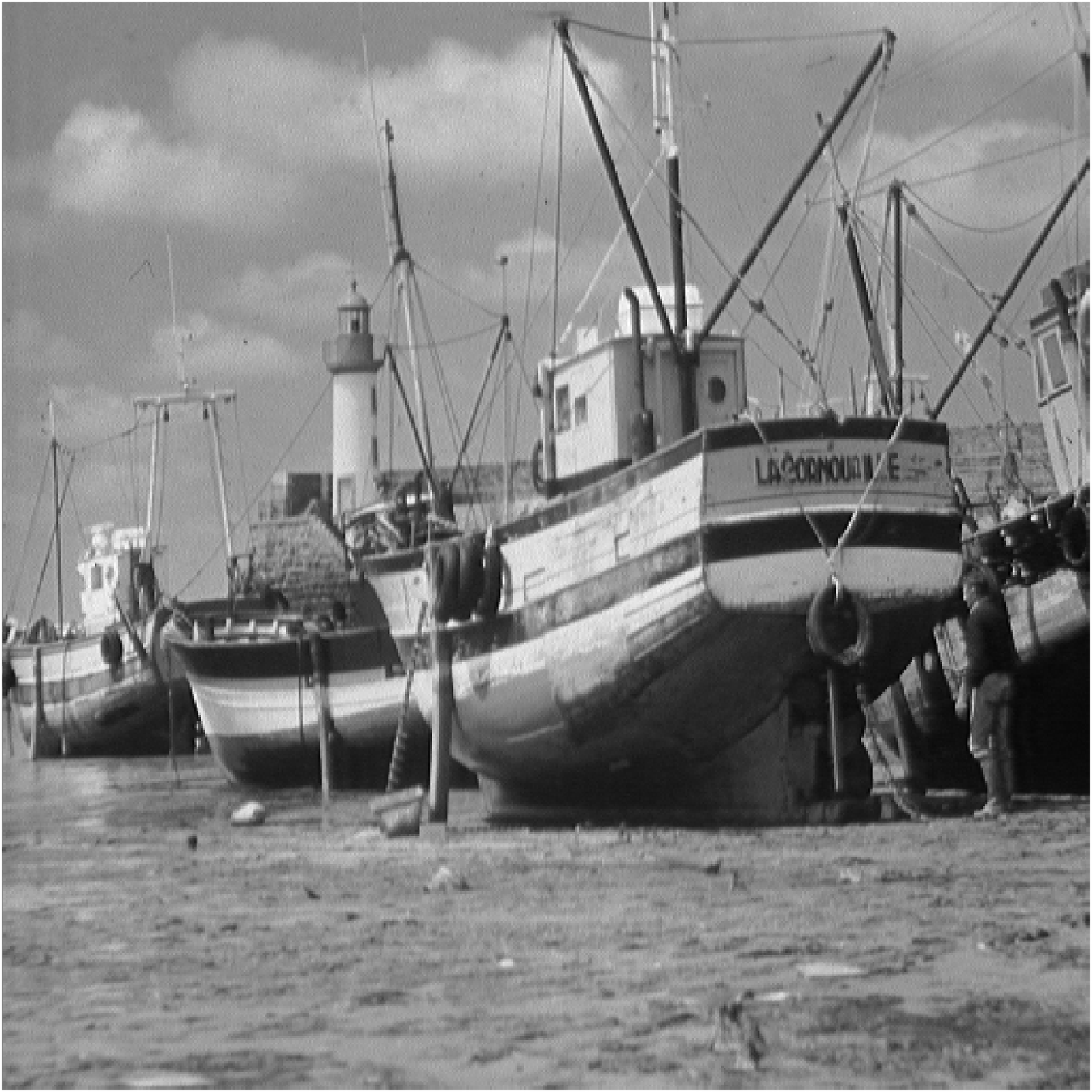} &
\includegraphics[height=18mm]{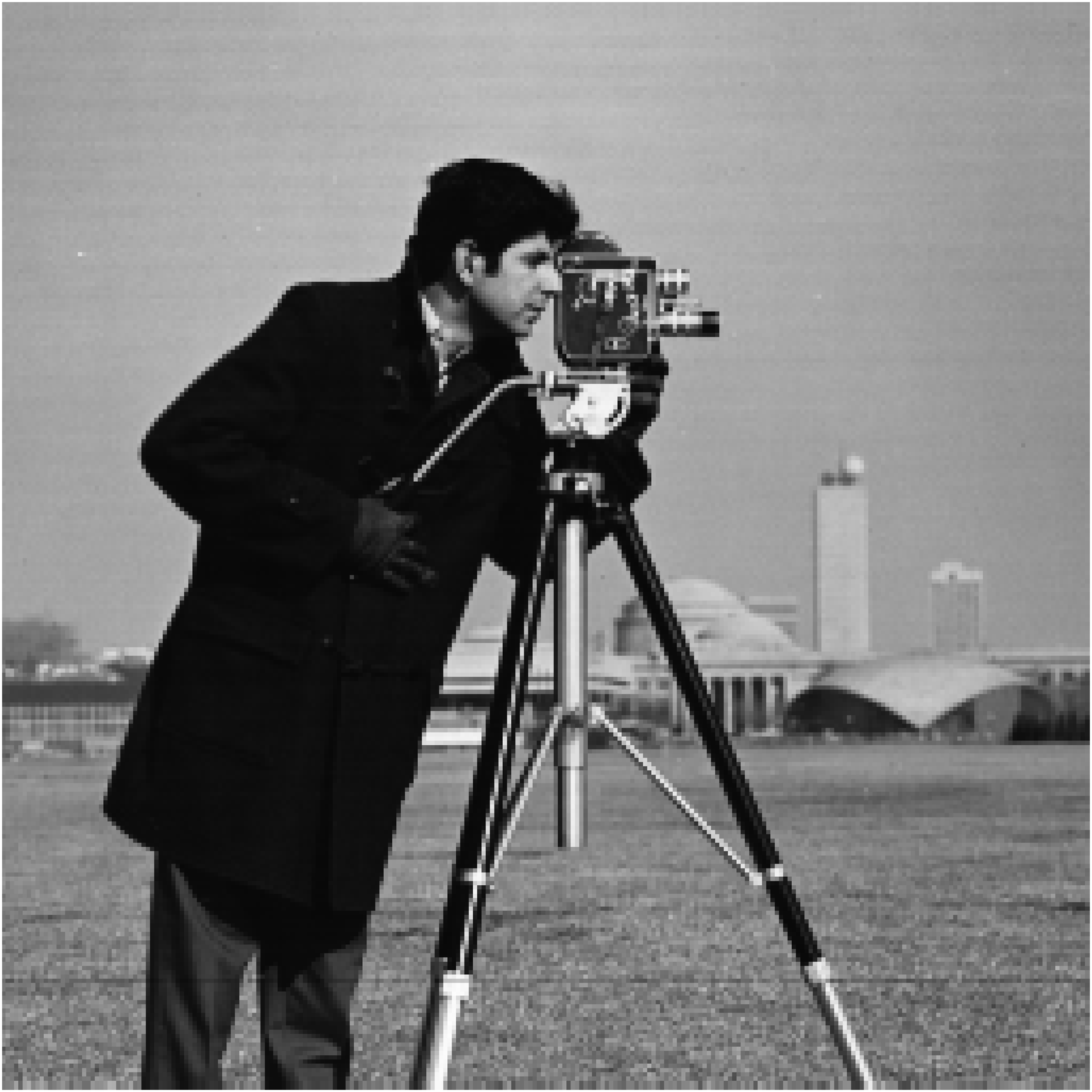} &
\includegraphics[height=18mm]{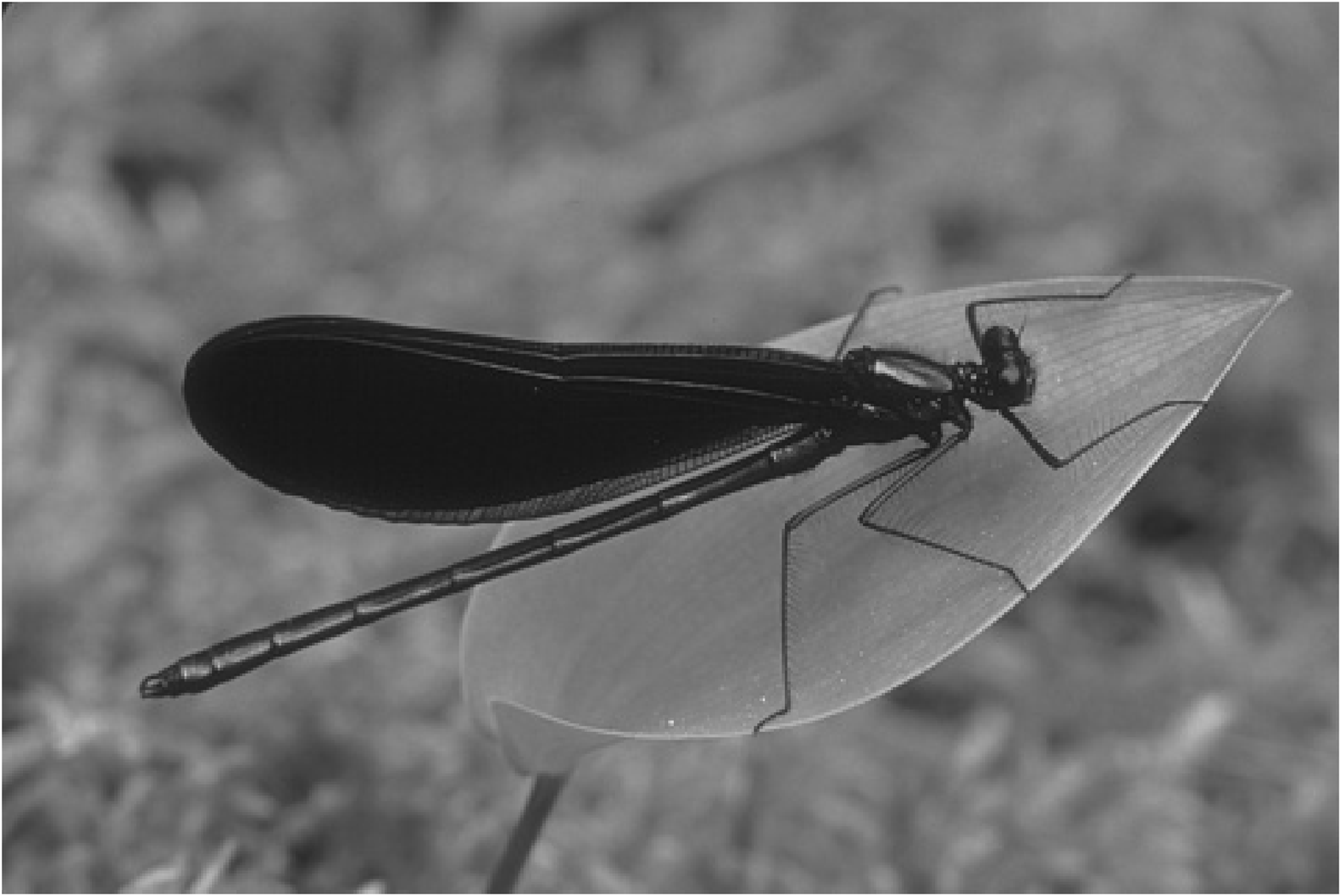} &
\includegraphics[height=18mm]{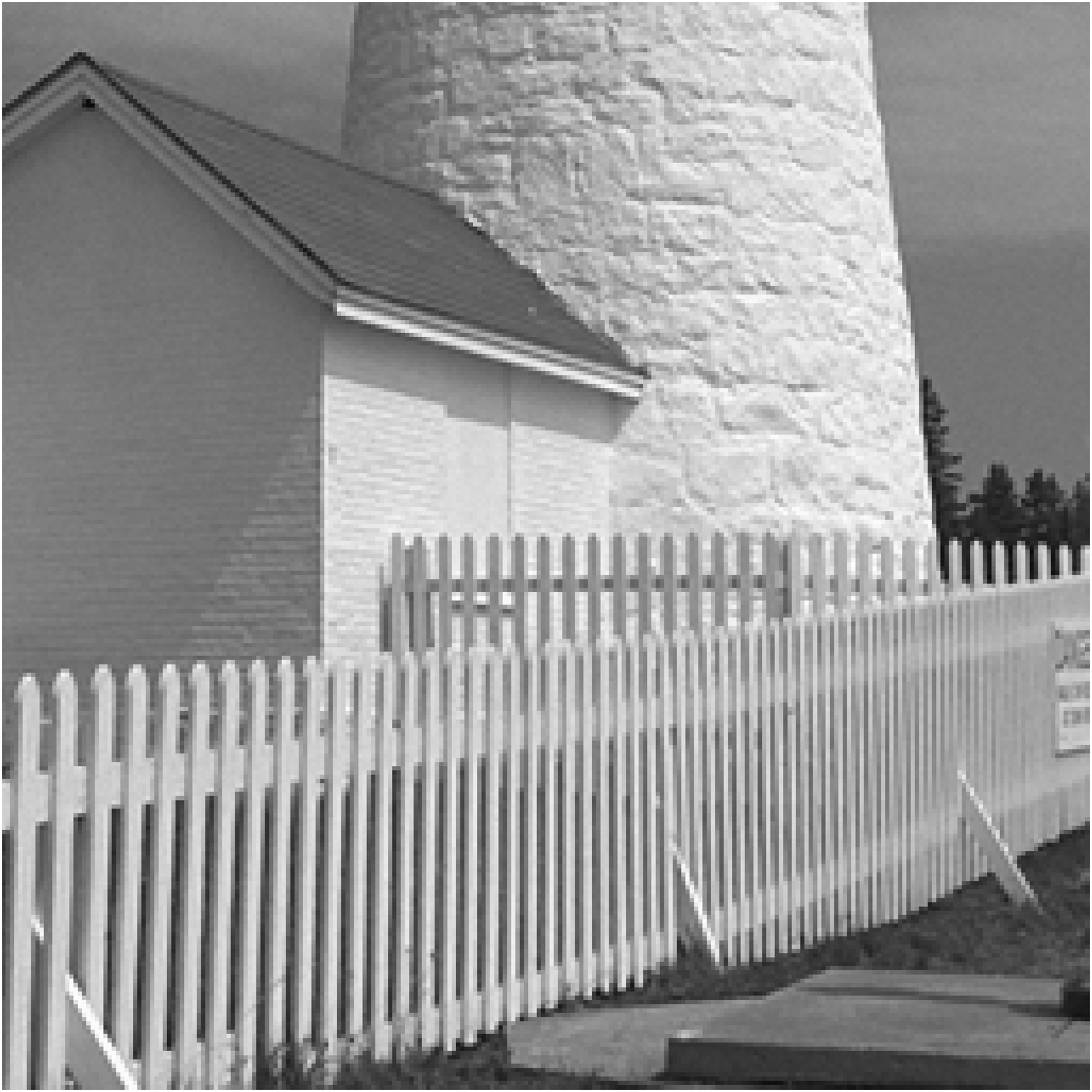} &
\includegraphics[height=18mm]{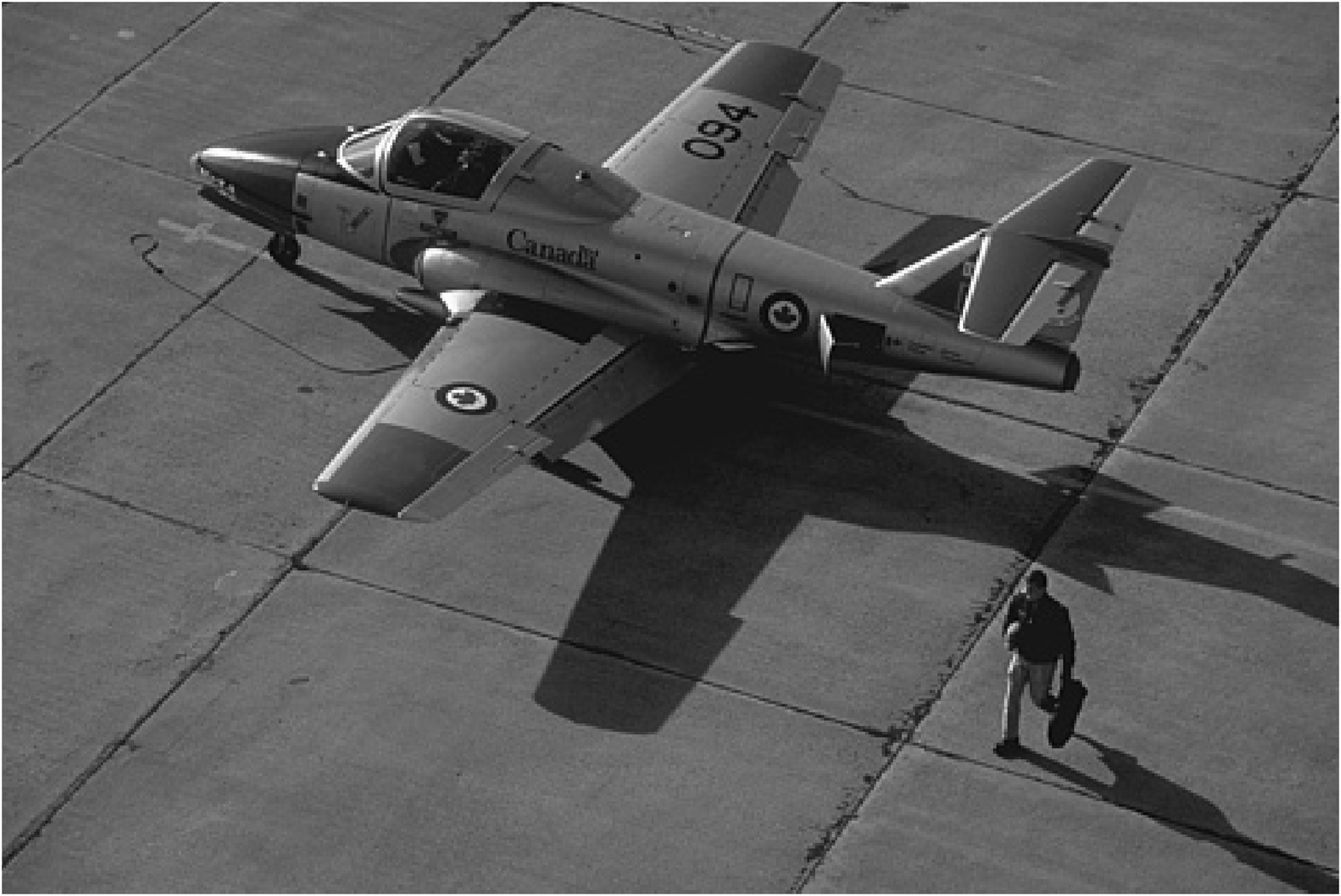} &
\includegraphics[height=18mm]{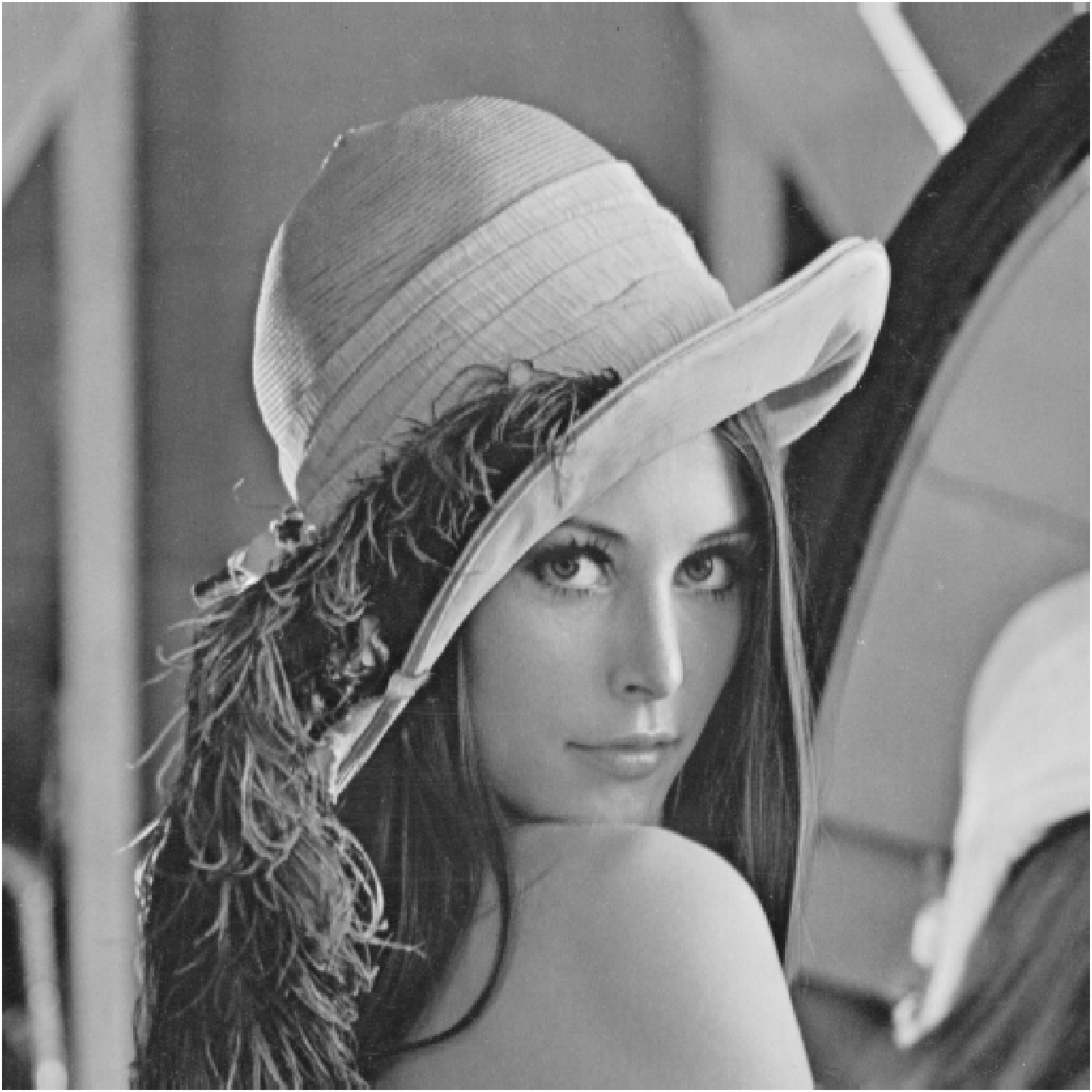} &
\includegraphics[height=18mm]{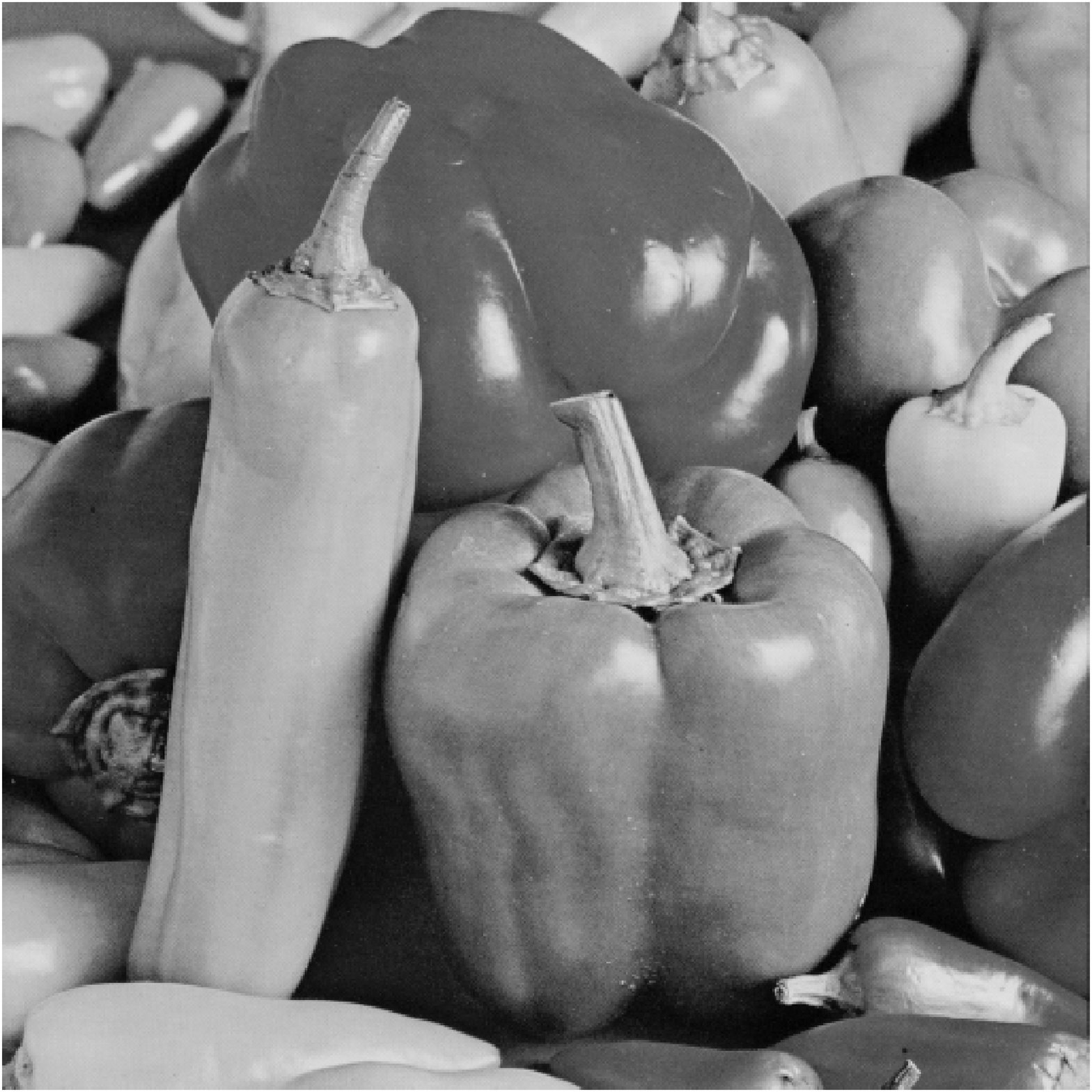} &
\includegraphics[height=18mm]{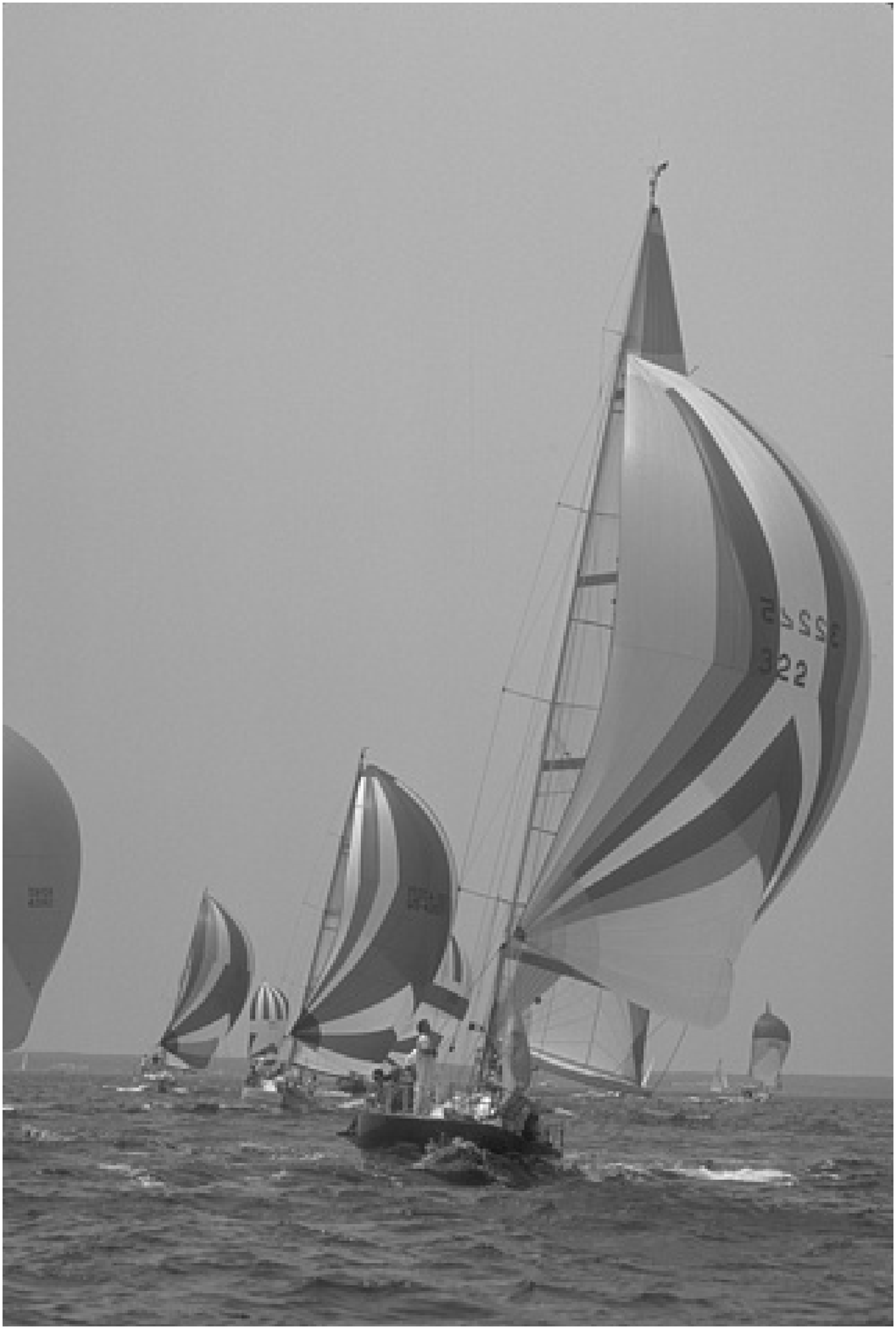} &
\includegraphics[height=18mm]{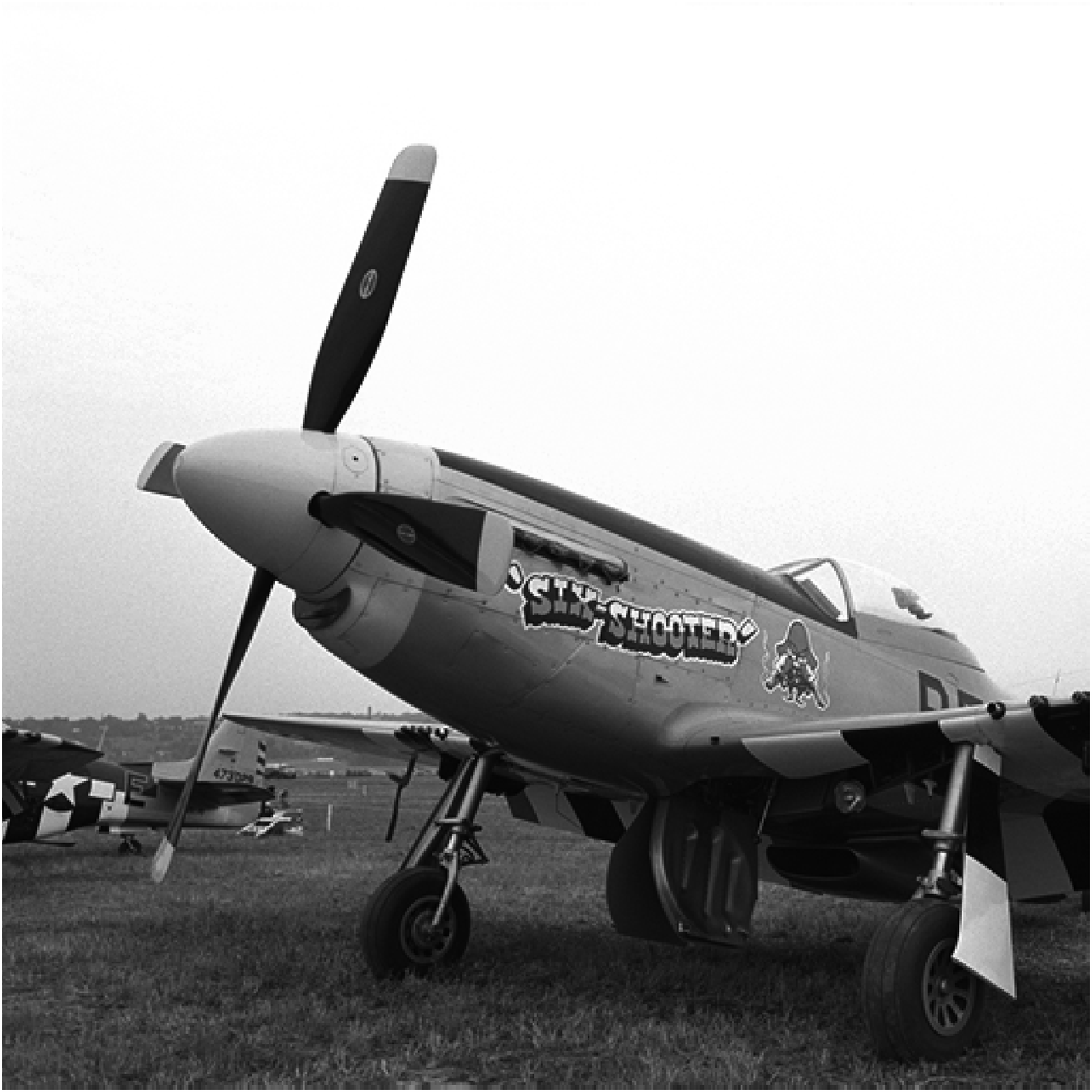} \\
\includegraphics[height=18mm]{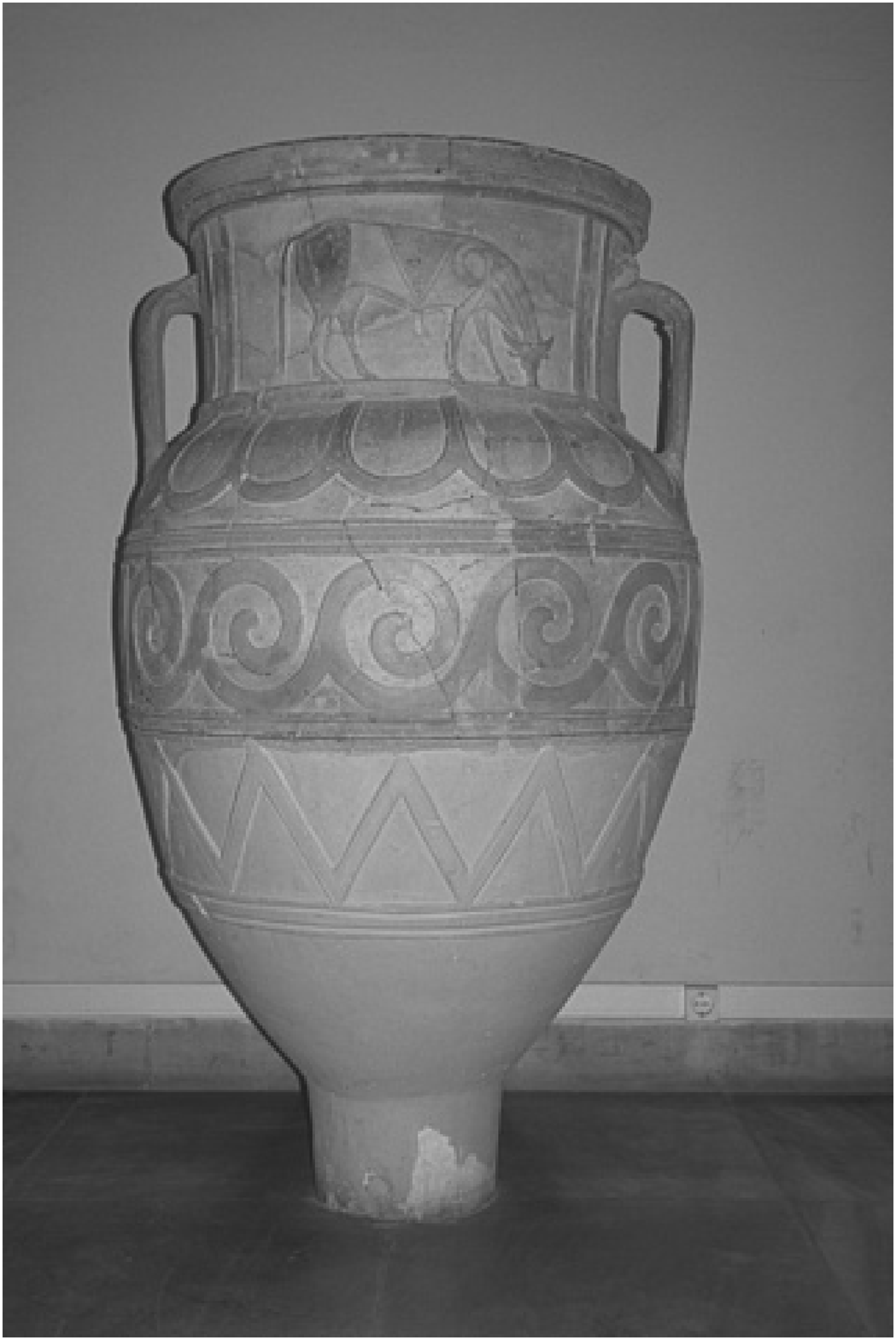} &
\includegraphics[height=18mm]{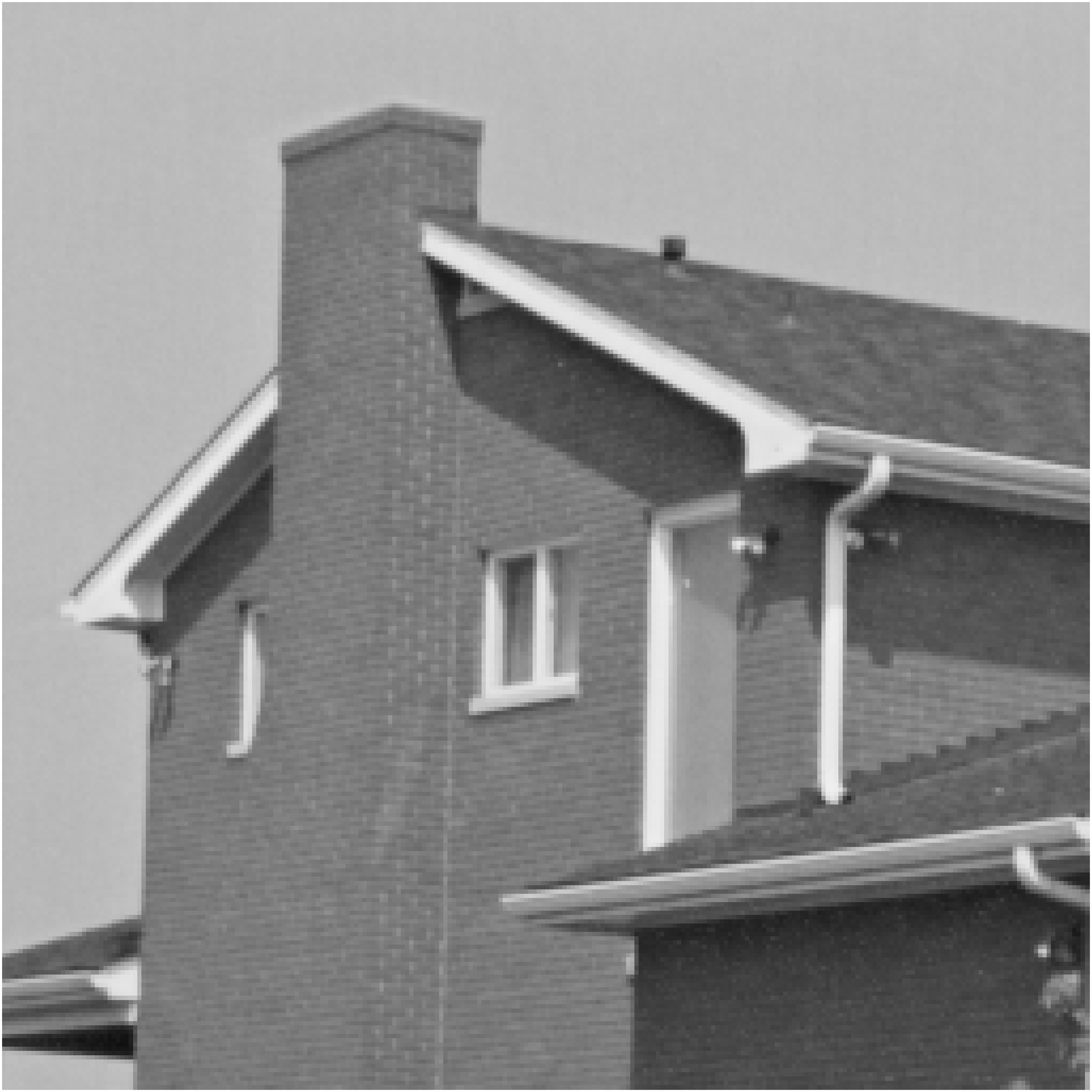} &
\includegraphics[height=18mm]{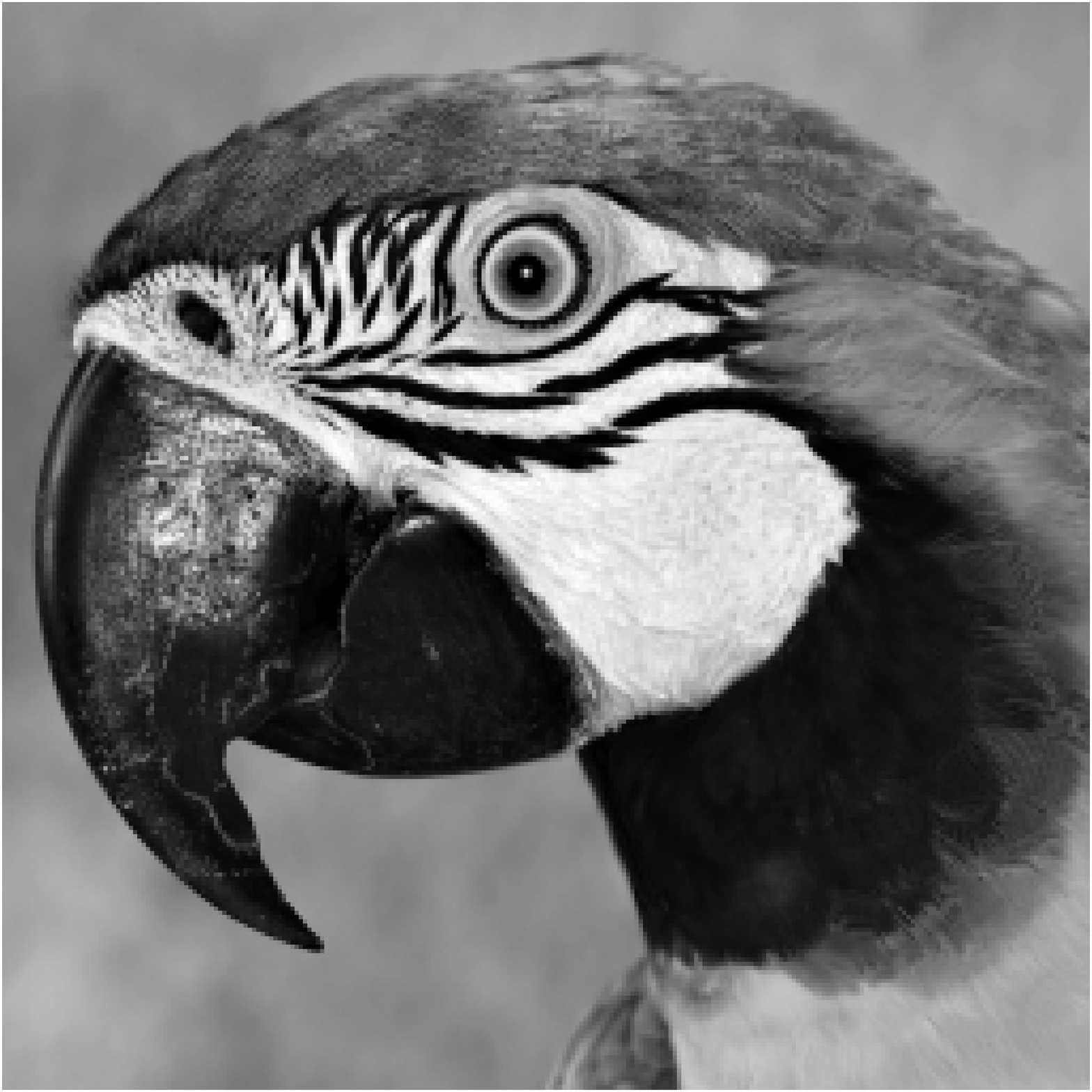} &
\includegraphics[height=18mm]{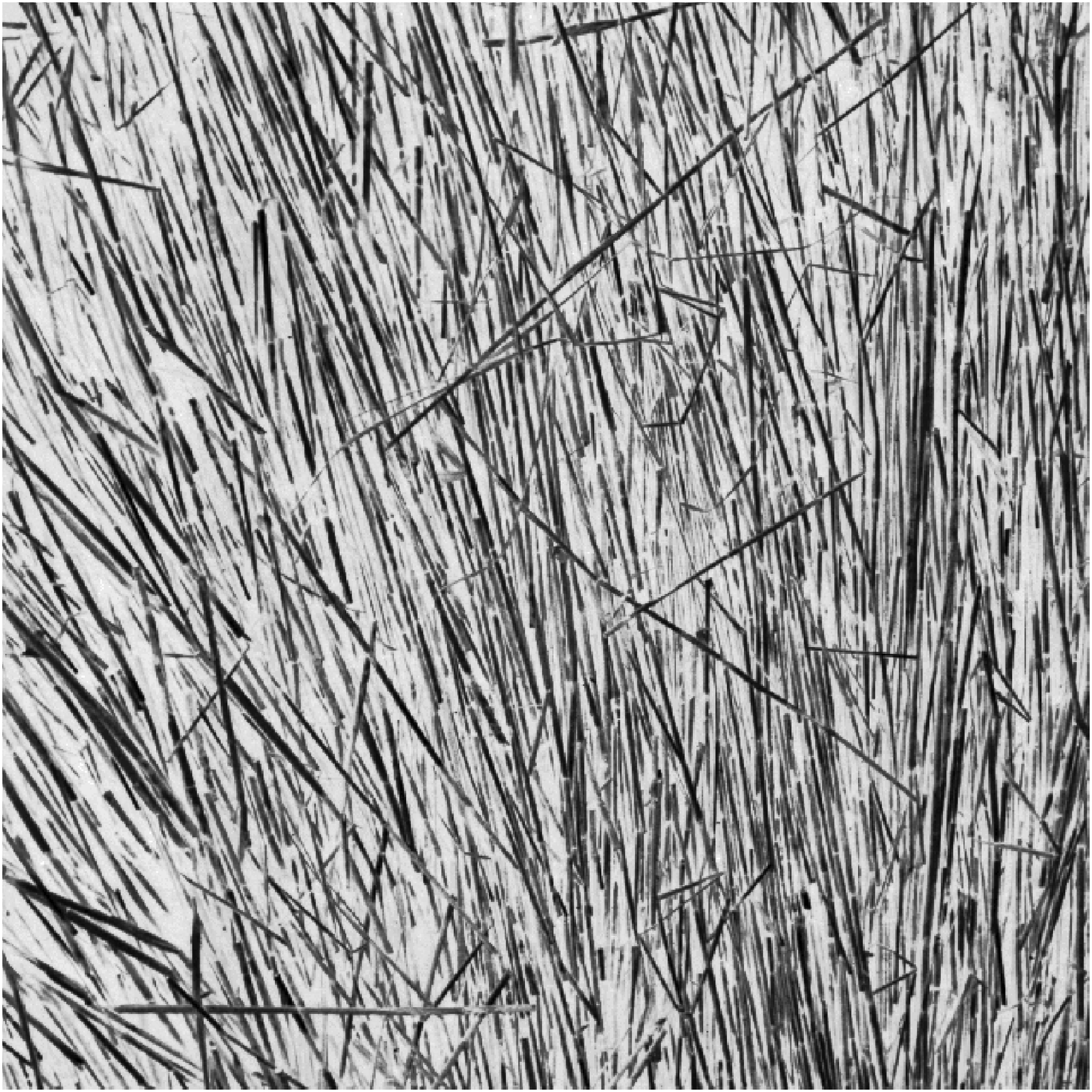} &
\includegraphics[height=18mm]{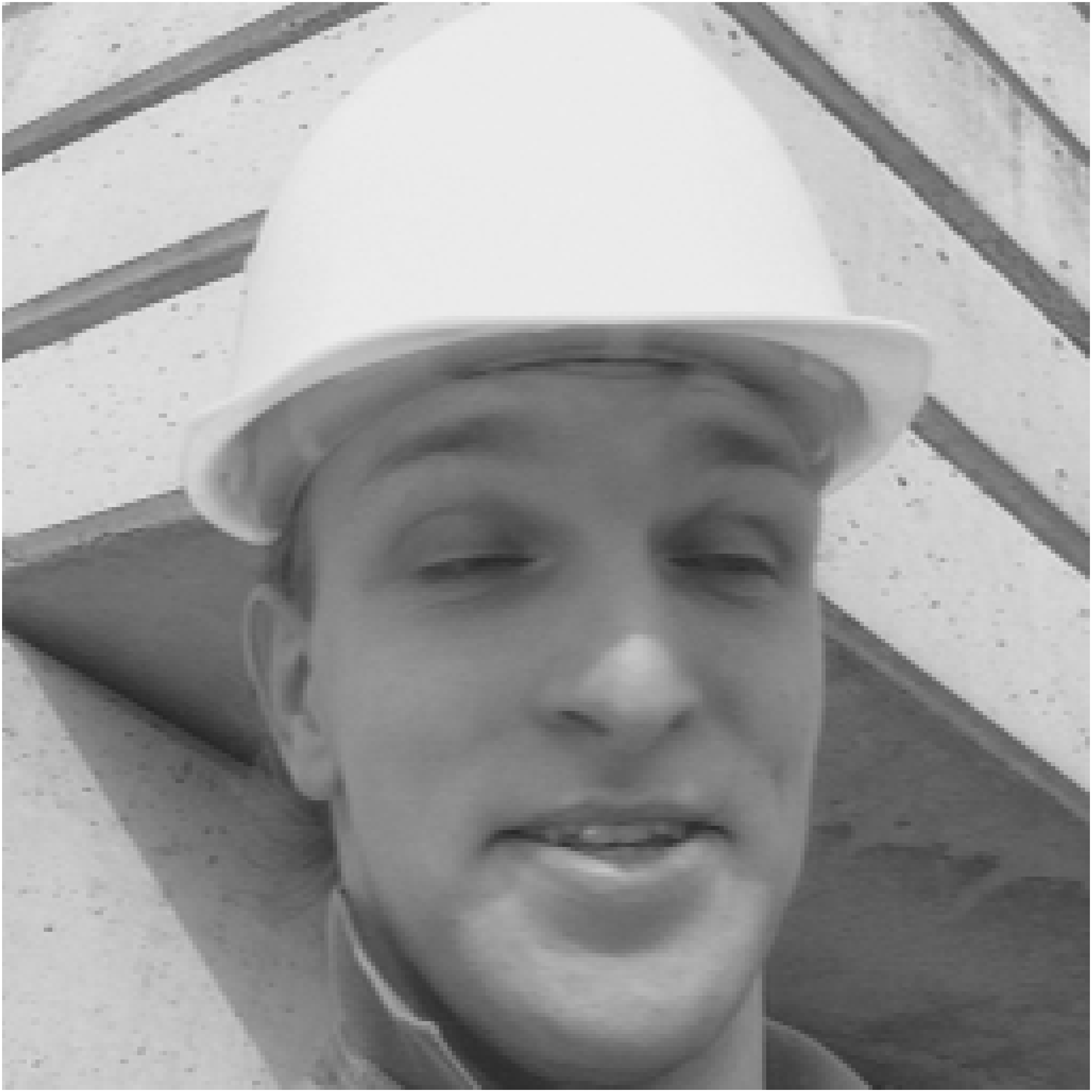} &
\includegraphics[height=18mm]{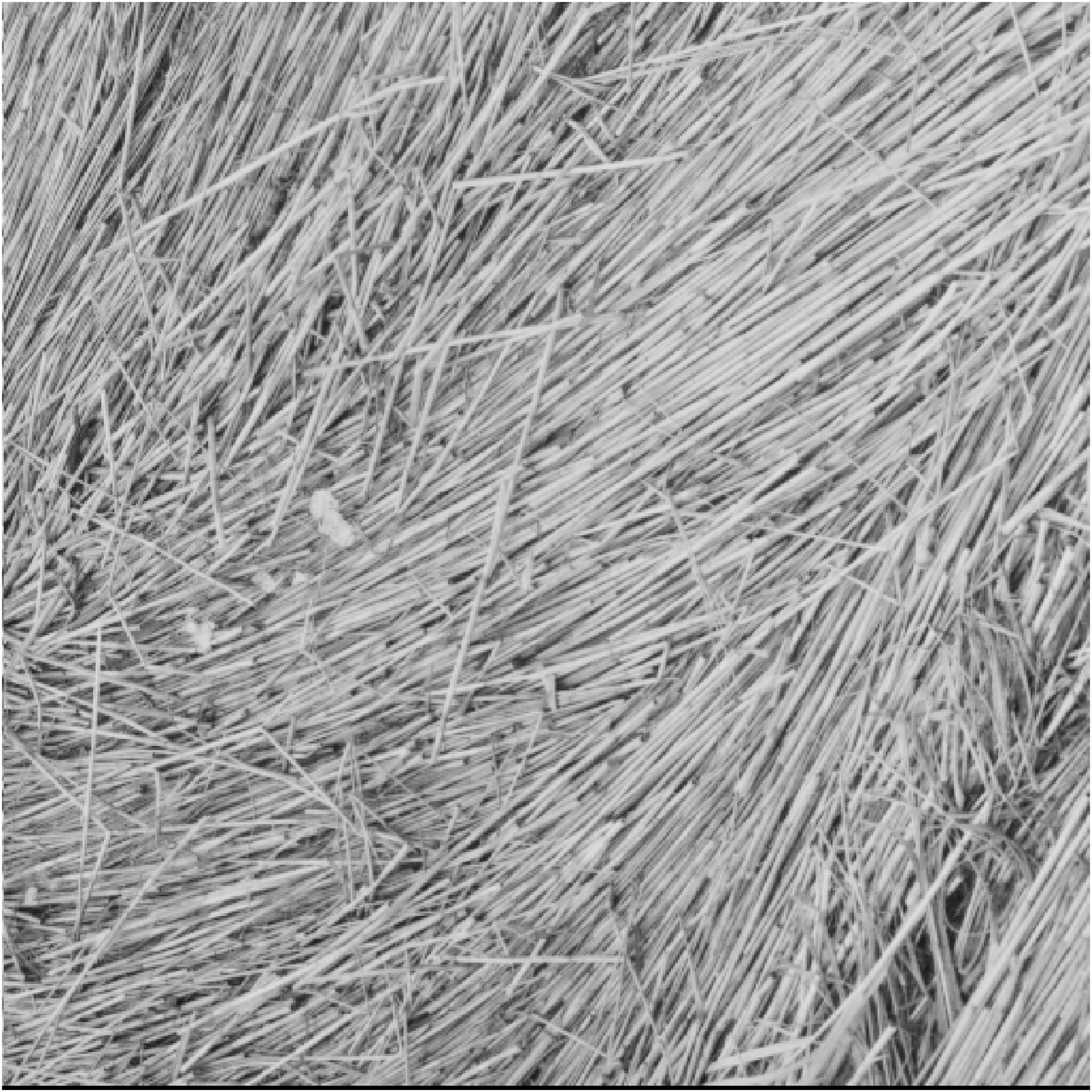} &
\includegraphics[height=18mm]{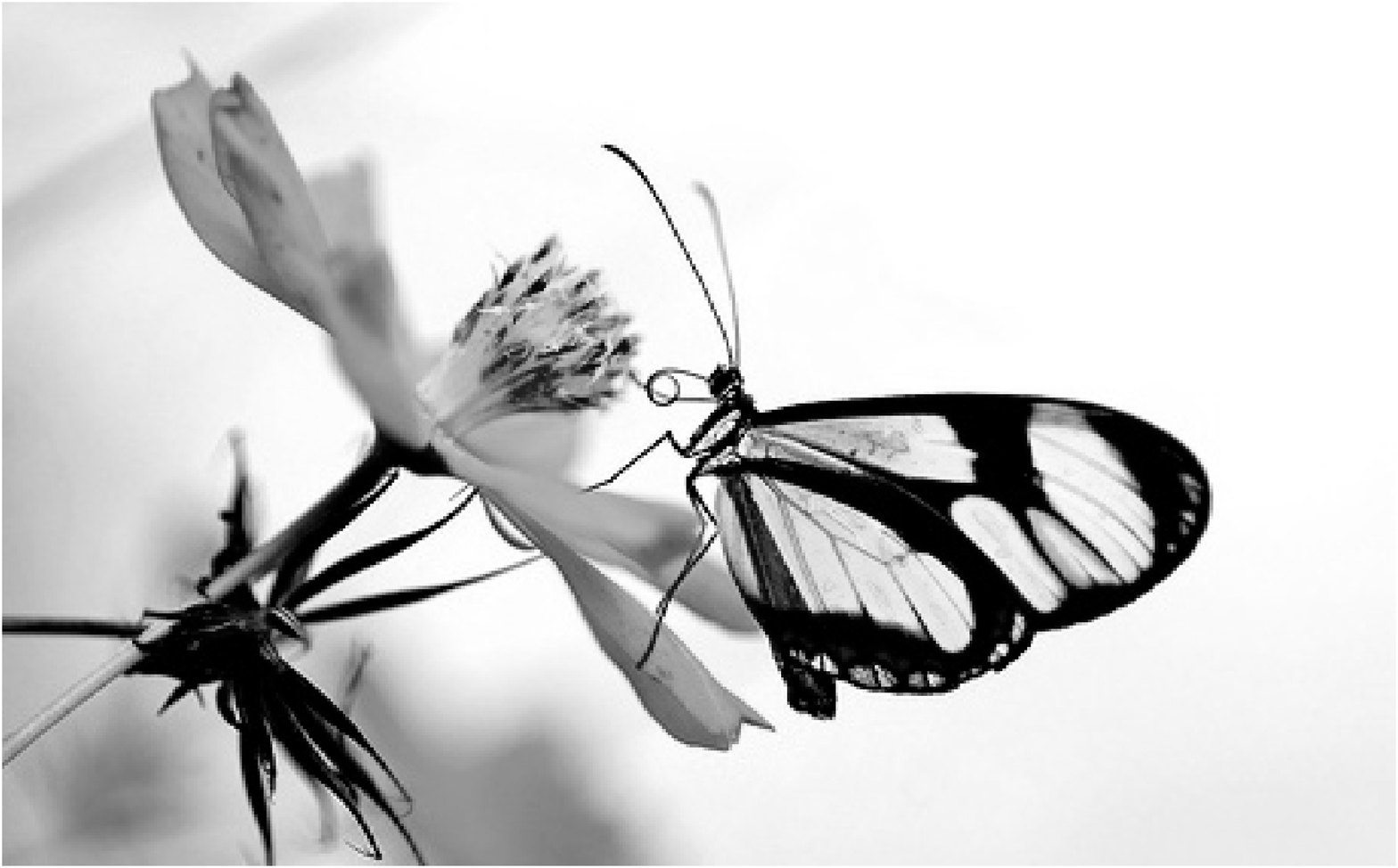} &
\includegraphics[height=18mm]{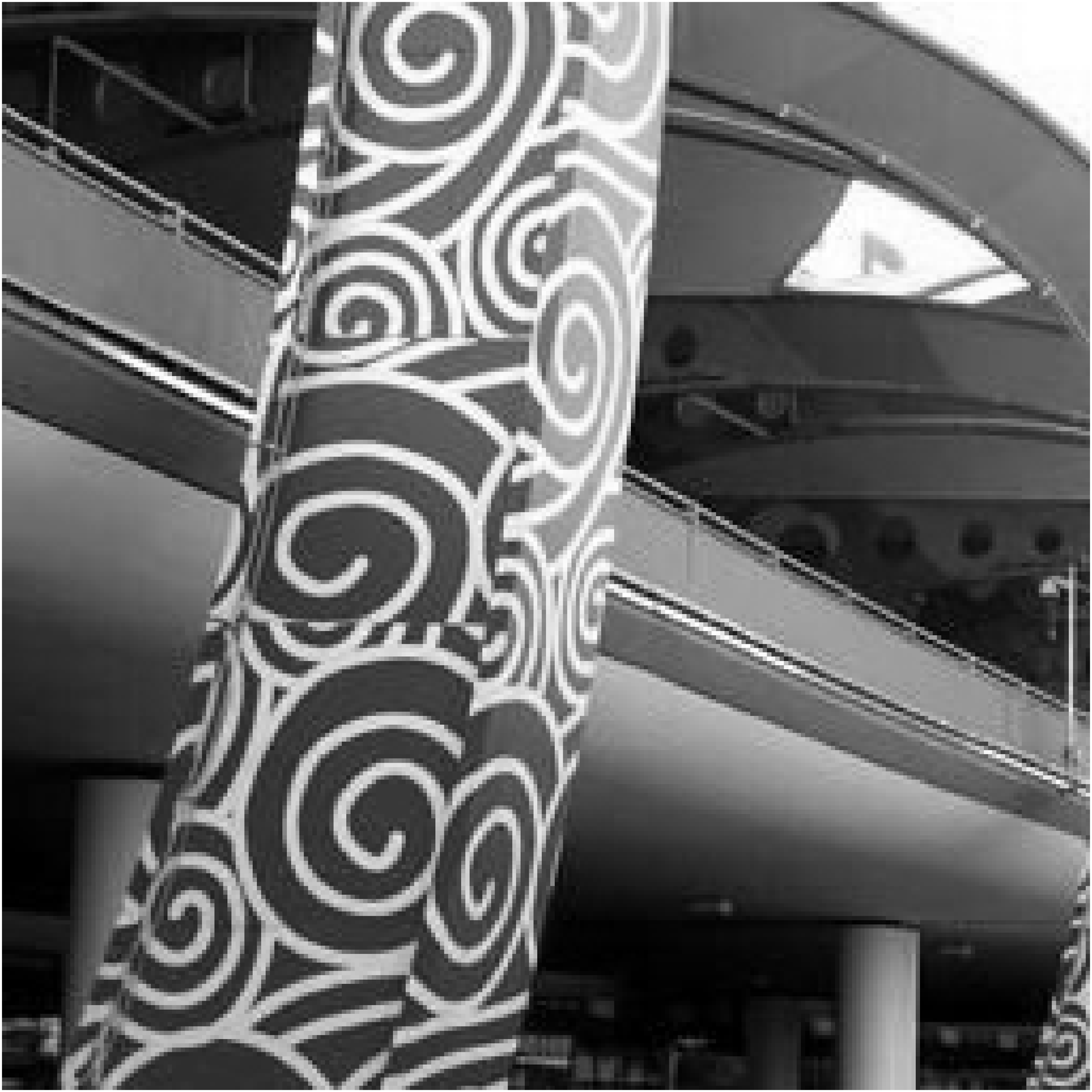} &
\includegraphics[height=18mm]{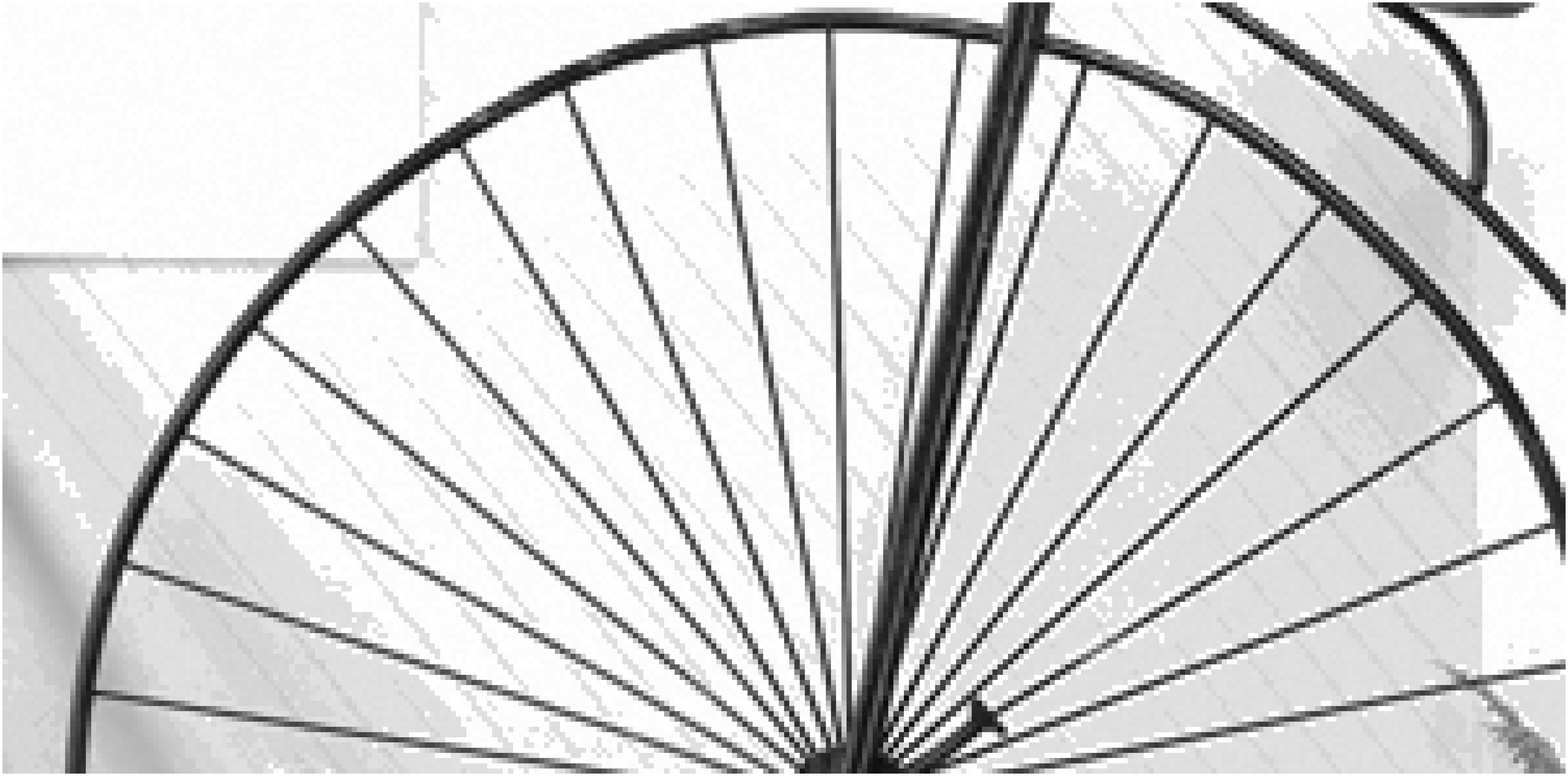}
\end{tabular}
}
\caption{$27$ test images from left to right and from top to bottom: \textit{Elk}, \textit{Birds}, \textit{Monarch}, \textit{Flower}, \textit{Woman}, \textit{Hats}, \textit{Leaves}, \textit{Male}, \textit{Motorbike}, \textit{Boat}, \textit{Cameraman}, \textit{Dragonfly}, \textit{Fence}, \textit{Fighter}, \textit{Lena}, \textit{Peppers}, \textit{Sail}, \textit{Plane}, \textit{Vase}, \textit{House}, \textit{Parrot}, \textit{Texture}, \textit{Foreman}, \textit{Straws}, \textit{Butterfly}, \textit{Station} and \textit{Wheel}.}
\label{fig:testset}
\end{figure*}

We dub our algorithm MISTER (Manifold-Inspired Single image inTERpolation algorithm). MISTER runs on $27$ commonly used grayscale test images in the literature~\cite{dong2013sparse,romano2014single,sun2016image,zhong2019predictor} shown in Fig.~\ref{fig:testset}. For each image, MISTER interpolates its downsampled-by-$2$ version by a factor of $2$ and its downsampled-by-$3$ version by a factor of $3$, both horizontally and vertically. The downsampling grids of both tasks start with the pixel at the upper-left corner of each test image. To deal with the pixels near boundaries, we perform interpolation at a size greater than the HR image via reflective padding and crop the extended pixels on the enlarged interpolated image to evaluate the interpolation performance. The evaluation metric we use is Peak Signal-to-Noise Ratio ($\mathrm{PSNR}$) defined as:
\begin{equation}
\label{eq:psnr_def}
\mathrm{PSNR} =20 \log_{10} \left(255/\sqrt{\mathrm{MSE}}\right),
\end{equation}
where $\mathrm{MSE}=\left\|\bm{\hat{I}}_H-{\bm{I}}_H\right\|^2_\mathrm{F}/\#$, $\#$ is the number of pixels, $\|\cdot\|_\mathrm{F}$ denotes the Frobenius norm. To rate the overall performance, we use the average of $27$ $\mathrm{PSNR}$ values. 

\subsection{Ablation Study}
\label{subsec:ablation_study}

\begin{table}[tb]
\renewcommand{\arraystretch}{1.1}
\caption{Average $\mathrm{PSNR}$ Evolution (in decibels) of the Outputs from MISTER's Stage~$1$A (S-1A), Stage~$1$B (S-1B), Stage~$2$ (S-2), Stage~$3$ (S-3), Stage~$4$ (S-4), as well as Average $\mathrm{PSNR}$ of the Outputs from four controlled experiments.}
\setlength\tabcolsep{1.8 pt}
\centering
\begin{tabular}{||c||c|c|c|c|c||c|c|c|c||}
\hline
\multirow{2}{*}{Experiment} &
\multicolumn{5}{c||}{$\times 2$} &
\multicolumn{4}{c||}{$\times 3$} \\
\cline{2-10}
 & S-1A & S-1B & S-2  & S-3  & S-4 & S-1A & S-1B & S-2  & S-3  \\
\hline
E-C1  & \multicolumn{4}{c|}{\multirow{4}{*}{\diagbox[innerwidth=2.8cm,  height = 10ex]{}{}}}  & 30.72 &  \multicolumn{3}{c|}{\multirow{4}{*}{\diagbox[innerwidth=2.1cm,  height = 10ex]{}{}}}  & 26.78\\
 \cline{1-1}
 \cline{6-6}
 \cline{10-10}
E-C2  & \multicolumn{4}{c|}{} & 30.83 &  \multicolumn{3}{c|}{} & 27.14\\
 \cline{1-1}
 \cline{6-6}
 \cline{10-10}
E-C3  & \multicolumn{4}{c|}{} & 31.04 & \multicolumn{3}{c|}{} & 27.03\\
 \cline{1-1}
 \cline{6-6}
 \cline{10-10}
E-C4  & \multicolumn{4}{c|}{} & 31.10 & \multicolumn{3}{c|}{} & 27.11\\
 \hline
MISTER & 30.70 & 30.82 & 31.00 & 31.12 & 31.20 & 26.94 & 26.63  & 27.11  & 27.26\\
 \hline
\end{tabular}
\label{tab:psnr_stages}
\end{table}

We conducted the following ablation studies to access the importance of various features of MISTER. This study records the stage-wise $\mathrm{PSNR}$ evolution for MISTER and the $\mathrm{PSNR}$ of final results of four controlled experiments. We  first describe the four controlled experiments, and explain the motivation for each of them. Finally, we present and discuss the results.

\begin{itemize}
    \item The first controlled experiment (E-C1) uses the bicubic interpolated version of $\bm{I}_L$ as the guide image $\bm{I}_G$ in Fig.~\ref{fig:whole_stage_x2}. This guide image will be utilized to initialize the selection of similar patches and to compute the weights of similar patches in Stage $1$A;
    \item The second controlled experiment (E-C2) uses $\bm{I}_G$ computed from the scheme in Fig.~\ref{fig:aliasing_removal} as the guide image $\bm{I}_G$ in Fig.~\ref{fig:whole_stage_x2}. This guide image will be used only to initialize the selection of similar patches. Given these selected similar patches, we use the bicubic interpolated version of $\bm{I}_L$ to compute the weights of similar patches in Stage $1$A;
    \item The third controlled experiment (E-C3) uses the bicubic interpolated version of $\bm{I}_L^{lp}$ computed from~\eqref{eq:convolution_lpf} with fine-tuned filter parameters as the guide image $\bm{I}_G$ in Fig.~\ref{fig:whole_stage_x2}. This guide image will be utilized to initialize the selection of similar patches and to compute the weights of similar patches in Stage~$1$A;
    \item The fourth controlled experiment (E-C4) uses $\bm{I}_{ar}$ computed from Algorithm~\ref{alg:stage_algorithm_removal} as the guide image $\bm{I}_G$ in Fig.~\ref{fig:whole_stage_x2}. This guide image will be utilized to initialize the selection of similar patches and to compute the weights of similar patches in Stage~$1$A.
\end{itemize}

E-C1 highlights the combined influence of aliasing both on choosing reliable similar patches and on estimating the weights of similar patches in the final result.
The guide image of the initial iteration in E-C1 is significantly more aliased than that of MISTER.
E-C2 isolates the influence of aliasing on estimation of the weights of similar patches in the final result.
The initial iteration of E-C2 uses the same initial set of similar patches as MISTER does but uses the aliased pixels to estimate the weights.
E-C3 highlights the influence of aliasing in the low-frequency components on the final result.
E-C3 uses a guide image in the initial iteration with aliasing removed from its high-frequency components, but with significantly more low-frequency aliasing compared with MISTER.
E-C4 examines the influence of the bias towards the dominant edges in removing aliasing on the final result.
E-C4 uses $\bm{I}_{ar}$ as the guide image in the initial iteration, with fewer secondary edges compared with MISTER.
For MISTER, E-C1, E-C2, E-C3 and E-C4, we show their $\mathrm{PSNR}$ evolution by stage in Table~\ref{tab:psnr_stages}.

Comparing the E-C1 row with the MISTER row in Table~\ref{tab:psnr_stages}, we find that the combined effect of the initial aliasing is substantial.
The E-C1 $\mathrm{PSNR}$ in the final result drops from the MISTER $\mathrm{PSNR}$ both by $0.48$ dB for the tasks of interpolating by a factor of $2$ and $3$.
Comparing the E-C2 row of Table~\ref{tab:psnr_stages} to the MISTER row, we find that the effect of aliasing isolated to the estimation of weights is also quite significant.
After all iterations, E-C2's $\mathrm{PSNR}$ drops from MISTER by $0.37$ dB and by $0.12$ dB in the tasks of interpolating by a factor of $2$ and $3$, respectively.
Comparing the E-C3 row with the MISTER row in Table~\ref{tab:psnr_stages}, we find that the aliasing in the low-frequency components has a non-trivial influence on the final result. After all iterations, E-C3's $\mathrm{PSNR}$ drops from MISTER by $0.16$ dB and by $0.23$ dB in the task of interpolating by a factor of $2$ and $3$, respectively.
Comparing the E-C4 row with the MISTER row in Table~\ref{tab:psnr_stages}, we find that the bias towards dominant edges in removing aliasing has a non-trivial influence on the final result. E-C4's $\mathrm{PSNR}$ after all the iteration drops from MISTER by $0.10$ dB and by $0.15$ dB in the tasks of interpolating by a factor of $2$ and $3$, respectively.

In addition to studying the aforementioned features, we explore the significance of the refinement after the first iteration, i.e. the combination of Stage $1$B, Stage $2$, Stage $3$, Stage $4$. This refinement process in MISTER offers $0.50$ dB and $0.32$ dB $\mathrm{PSNR}$ gains in the task of interpolating by a factor of $2$ and $3$, respectively. Each of those stages in the interpolation by a factor of $2$ task offers an non-trivial $\mathrm{PSNR}$ gain. Similarly, in the interpolation by a factor of $3$ task, those stages are trending to improve the $\mathrm{PSNR}$.

In conclusion, each of the features of MISTER, i.e. i) removing the aliasing in the low-frequency components; ii) mitigating the bias towards dominant edges in removing aliasing; iii) using aliasing-removed image to guide the selection of similar patches in the initial iteration; iv) using aliasing-removed image to compute the weights of similar patches in the initial iteration; and v) the sequential use of the manifold models for exploiting semi-local similarity, plays a significant role in achieving the interpolation quality of MISTER.

\subsection{Interpolation by a factor of $2$}

\begin{table*}[tb]
\caption{Comparison of $\mathrm{PSNR}$s (in decibels) of the results between MISTER and ~\cite{li2001new,mallat2010super,zhang2008image,liu2011image,guo2012multiscale,zhu2016image,dong2013sparse,romano2014single,sun2016image,huang2015fast,ji2020image} in the Task of Interpolation by a Factor of $2$. The result with the highest $\mathrm{PSNR}$ among model-based approaches is highlighted in dark bold and the result with the highest $\mathrm{PSNR}$ is highlighted in red bold.}
\centering
\setlength\tabcolsep{2 pt}
\begin{tabularx}{\textwidth}{ ||c| *{13}{Y|} | c||}
 \hline
\textbf{Images} &  \textbf{Bicubic} & \textbf{NEDI} & \textbf{SME} & \textbf{SAI} & \textbf{RLLR} & \textbf{MSIA} & \textbf{NGSDG} & \textbf{NARM} & \textbf{ANSM}  & \textbf{NLPC} & \textbf{FIRF} & \textbf{MAIN} & \textbf{MISTER} \\
\hline
\textbf{AVERAGE}   &  29.00 & 29.37  & 29.86  & 29.91  & 29.88  & 29.89  & 29.82  & 30.23  & 30.59  & 30.48  & 30.69  & $\textcolor{red}{\bm{32.12}}$ & $\bm{31.20}$ \\     
 \hline
\end{tabularx}
\label{tab:test_2x}
\end{table*}

For the task of interpolating by a factor of $2$, in Table~\ref{tab:test_2x}, we compare MISTER with related works. These methods include (\romannumeral1) bicubic interpolation (Bicubic) (\romannumeral2) New Edge-Directed Interpolation (NEDI)~\cite{li2001new}, (\romannumeral3) Sparse Mixing Estimation (SME)~\cite{mallat2010super}, (\romannumeral4) Soft-Decision and Adaptive Interpolator (SAI)~\cite{zhang2008image}, (\romannumeral5) Regularized Local Linear Regression (RLLR)~\cite{liu2011image}, (\romannumeral6) Multiscale Semilocal Interpolation with Antialiasing (MSIA)~\cite{guo2012multiscale}, (\romannumeral7) Nonlocal Geometric Similarities and Directional Gradients (NGSDG)~\cite{zhu2016image}, (\romannumeral8) Nonlocal Auto-Regressive Modeling (NARM)~\cite{dong2013sparse}, (\romannumeral9) Adaptive Non-local Sparsity Modeling (ANSM)~\cite{romano2014single}, (\romannumeral10) Non-Local Patch Collaging (NLPC)~\cite{sun2016image},  (\romannumeral11) Fast Image Interpolation via Random Forests (FIRF)~\cite{huang2015fast} and (\romannumeral12) Multi-scale Attention-aware Inception Network (MAIN)~\cite{ji2020image}. Note that FIRF and MAIN are data-driven, learning-based approaches that learn the model parameters from an external set of training images. FIRF exploits random forests to partition the space of patches into subspaces and trains a regression model for each subspace to compute the HR image's patches from their counterparts in the LR image. MAIN uses deep learning based, attention-aware inception (AIN) networks to interpolate images and fuse the interpolated images from multi scales to generate the output image.  Involving FIRF and MAIN is to explore the performance difference between model-based approaches and data-driven, learning-based approaches. Table~\ref{tab:test_2x} demonstrates that compared with model-based algorithms, MISTER achieves the highest average $\mathrm{PSNR}$ and the highest $\mathrm{PSNR}$ in all the $27$ images. Specifically, compared with patch-based methods~\cite{dong2013sparse,romano2014single,sun2016image}, MISTER has an average $\mathrm{PSNR}$ that is $0.97$ dB better than NARM, $0.61$ dB better than ANSM and $0.72$ dB better than NLPC. MISTER outperforms NEDI, SME, SAI, RLLR, MSIA, NGSDG and bicubic interpolation with an average gain of $1.83$ dB, $1.34$ dB, $1.29$ dB, $1.32$ dB, $1.31$ dB, $1.38$ dB and $2.20$ dB, respectively. Compared with FIRF and MAIN, MISTER's average $\mathrm{PSNR}$ is higher than FIRF by $0.51$ dB, but lower than MAIN by $0.92$ dB.

\begin{figure*}[ht]
\begin{minipage}[b]{0.06 \linewidth}
\centering
\centerline{\includegraphics[width=1.1 cm]{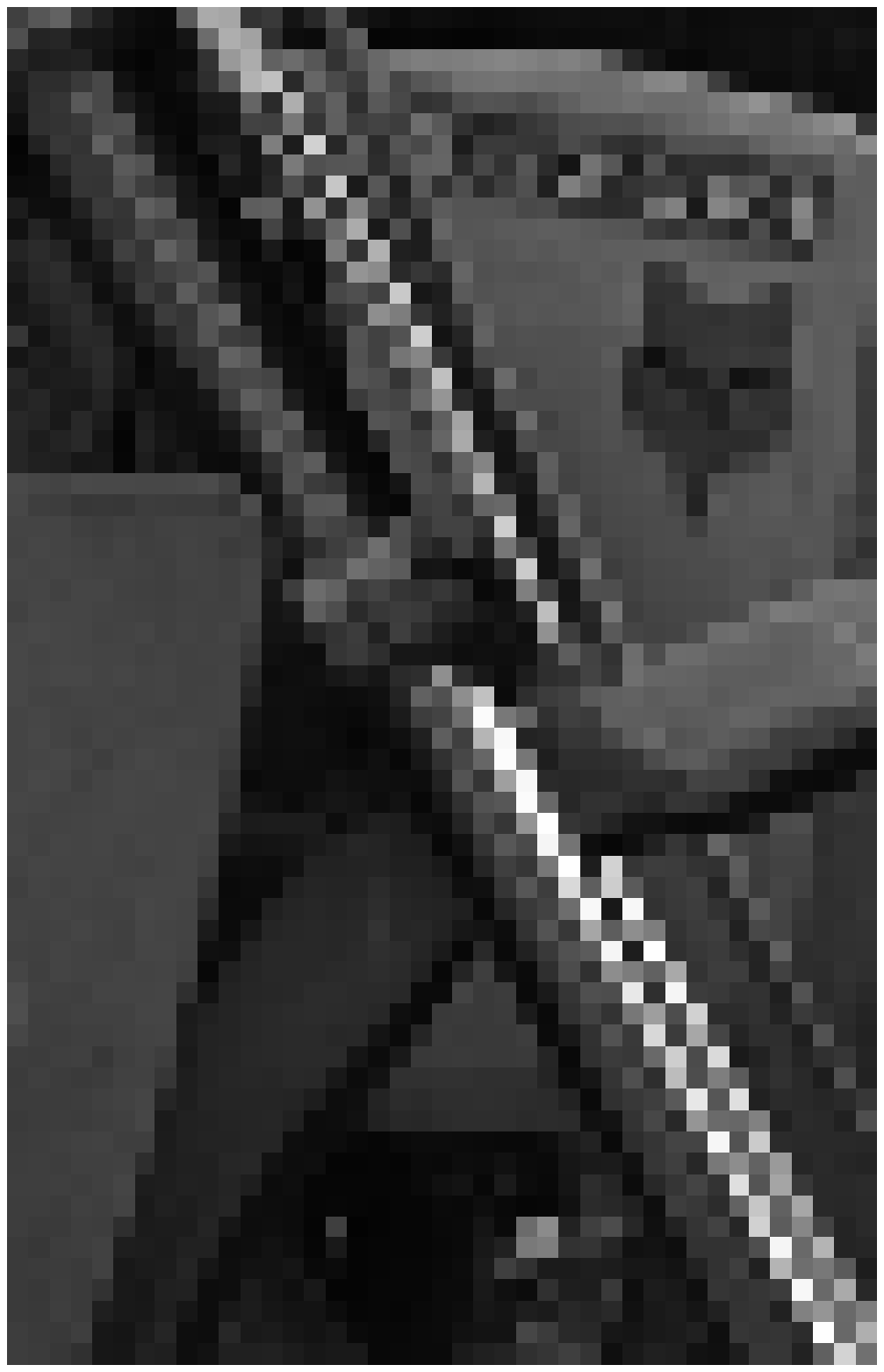}}
\centerline{(a)~LR}
\end{minipage}
\begin{minipage}[b]{0.12 \linewidth}
\centerline{\includegraphics[width=2.2 cm]{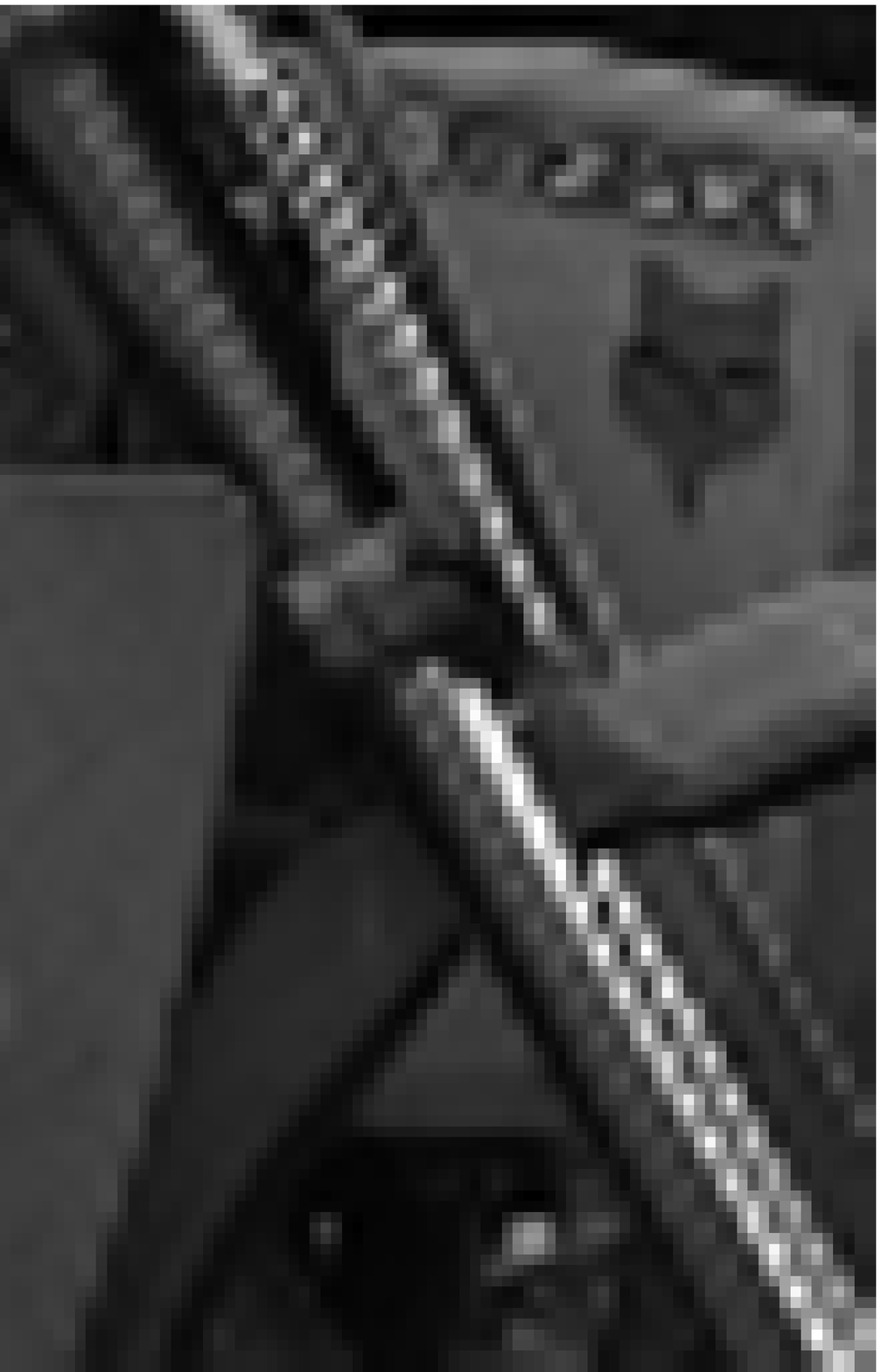}}
\centerline{(b)~Bicubic}
\end{minipage}
\begin{minipage}[b]{0.12 \linewidth}
\centering
\centerline{\includegraphics[width=2.2 cm]{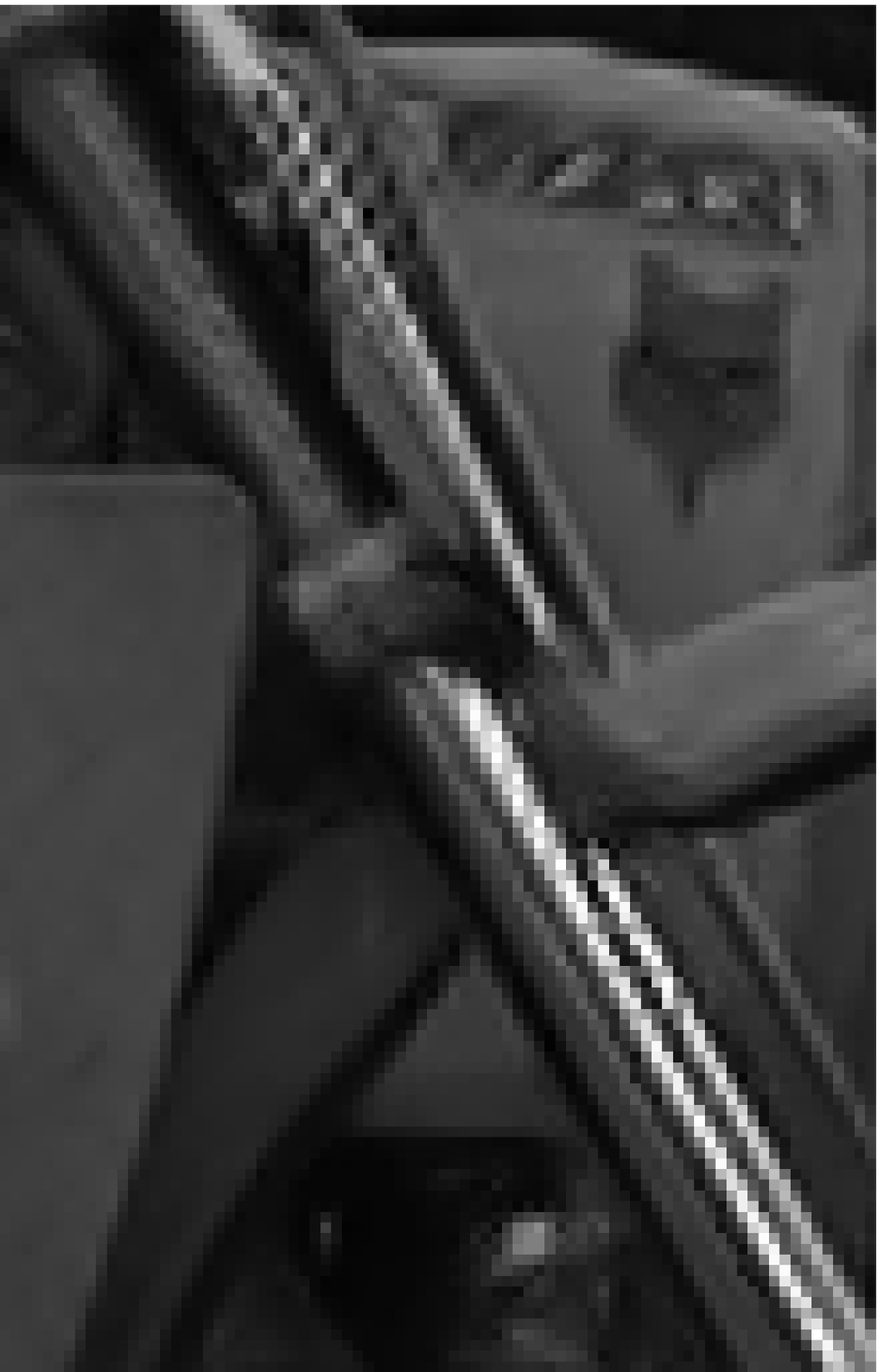}}
\centerline{(c)~NEDI}
\end{minipage}
\begin{minipage}[b]{0.12 \linewidth}
\centering
\centerline{\includegraphics[width=2.2 cm]{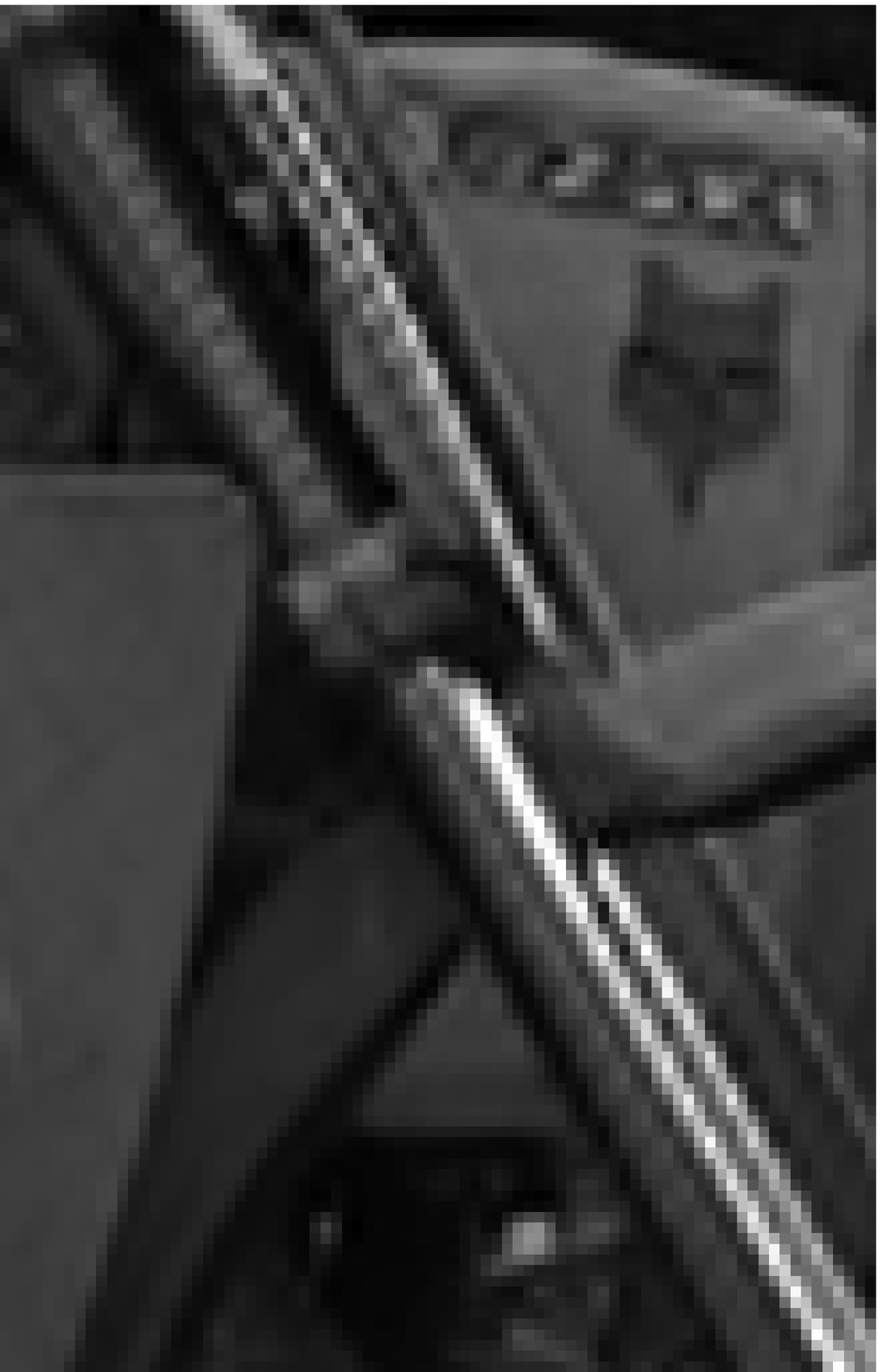}}
\centerline{(d)~SME}
\end{minipage}
\begin{minipage}[b]{0.12 \linewidth}
\centering
\centerline{\includegraphics[width=2.2 cm]{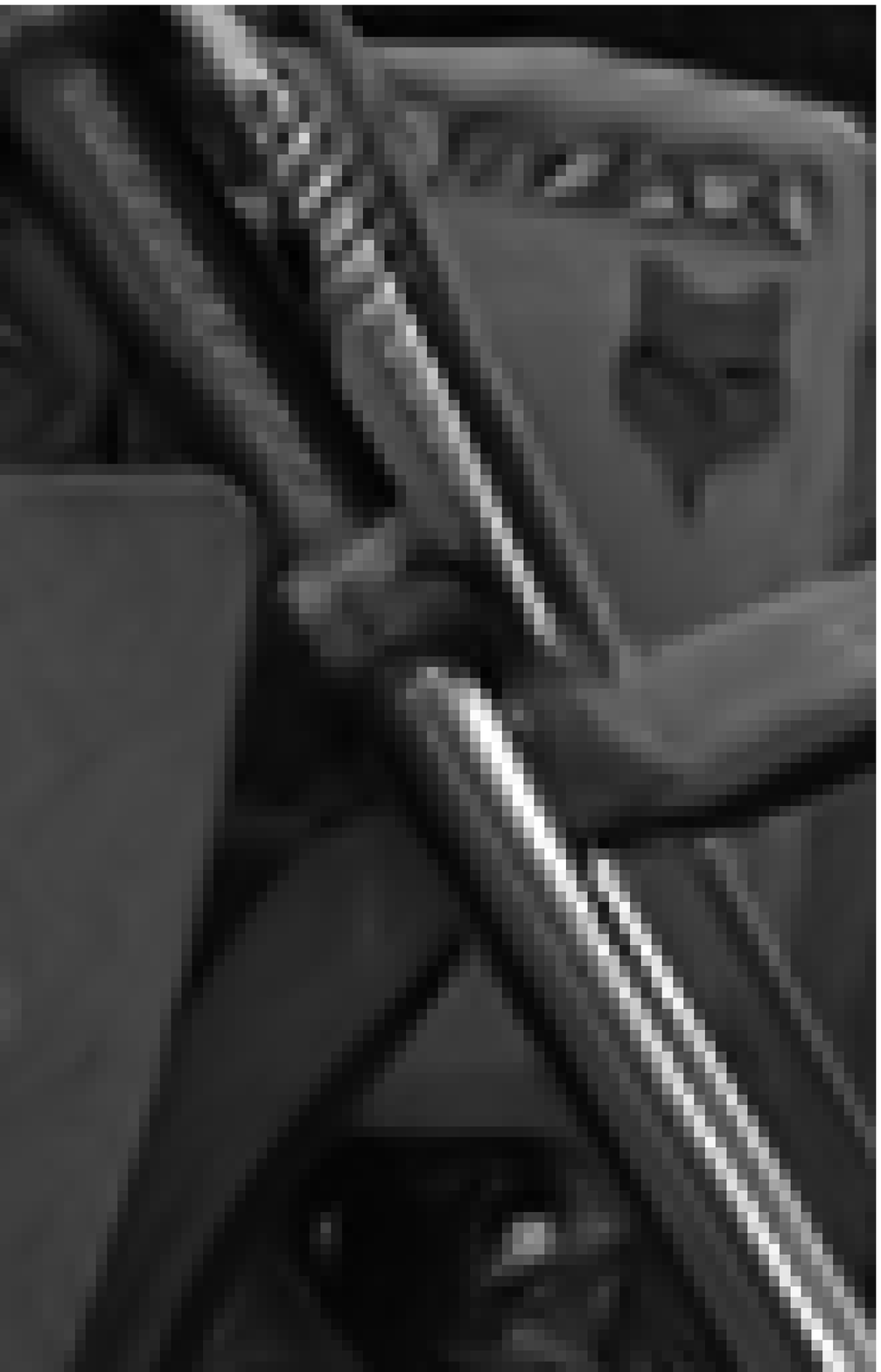}}
\centerline{(e)~SAI}
\end{minipage}
\begin{minipage}[b]{0.12 \linewidth}
\centering
\centerline{\includegraphics[width=2.2 cm]{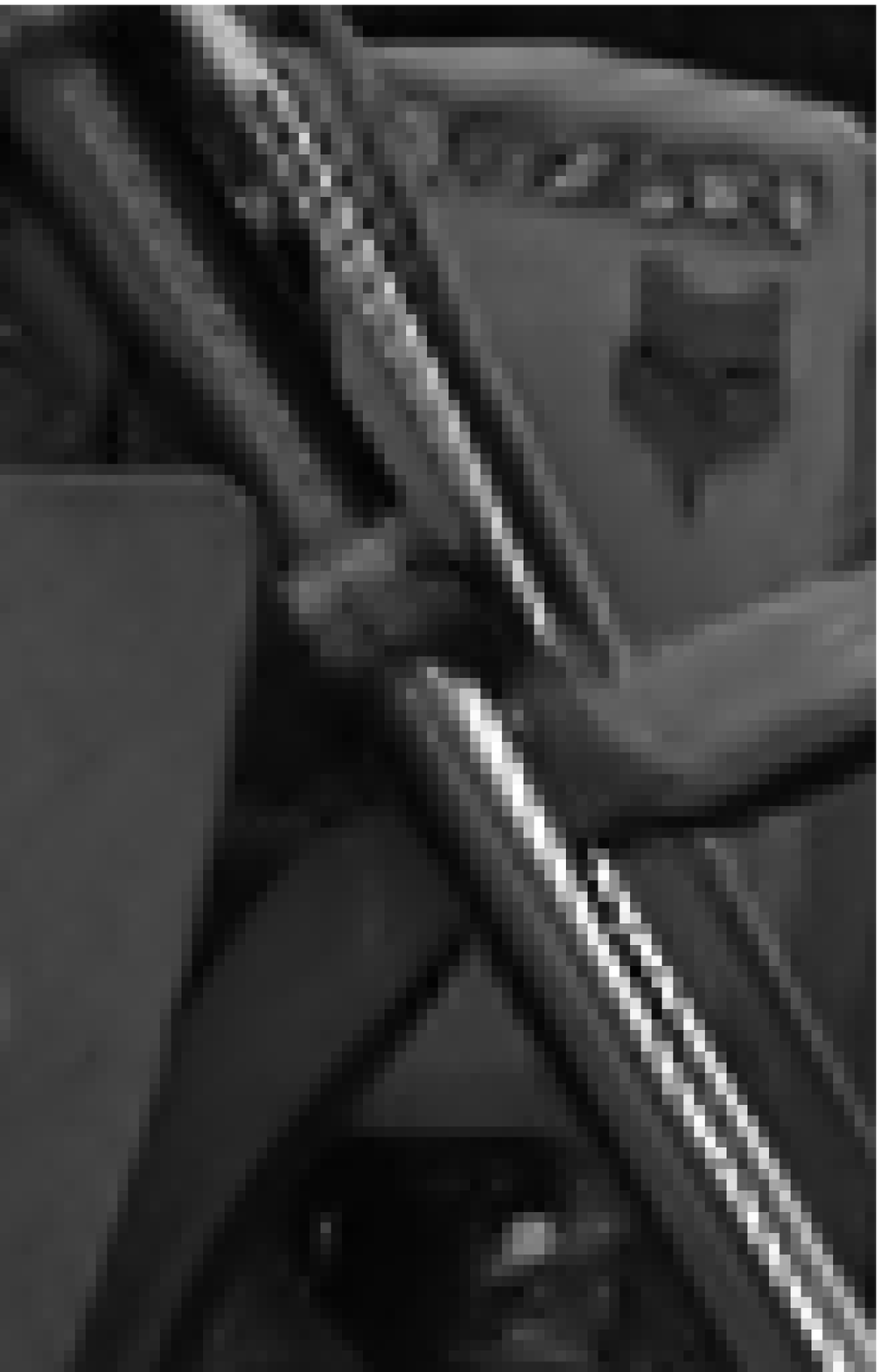}}
\centerline{(f)~RLLR}
\end{minipage}
\begin{minipage}[b]{0.12 \linewidth}
\centering
\centerline{\includegraphics[width=2.2 cm]{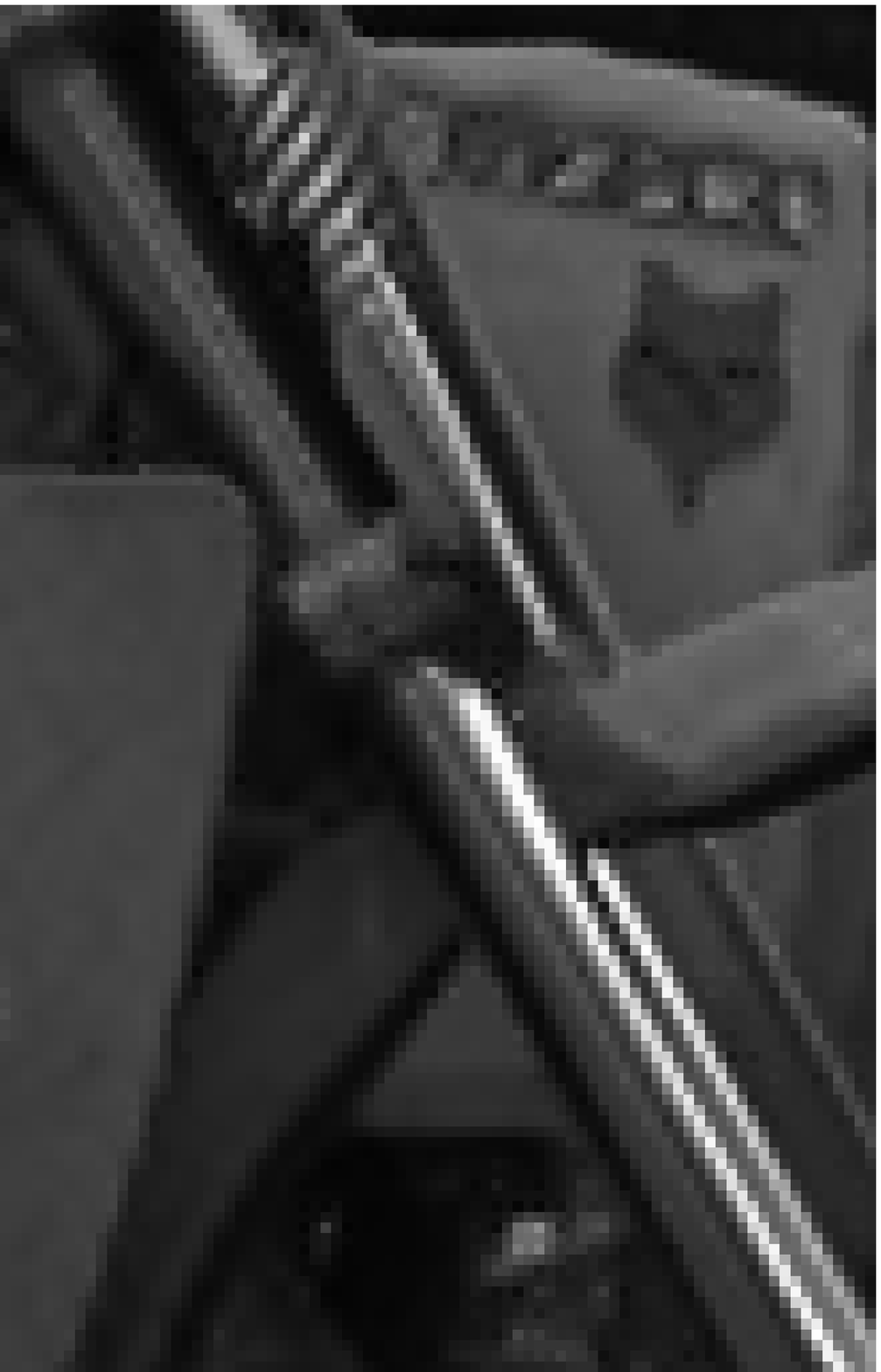}}
\centerline{(g)~MSIA}
\end{minipage}
\begin{minipage}[b]{0.12 \linewidth}
\centerline{\includegraphics[width=2.2 cm]{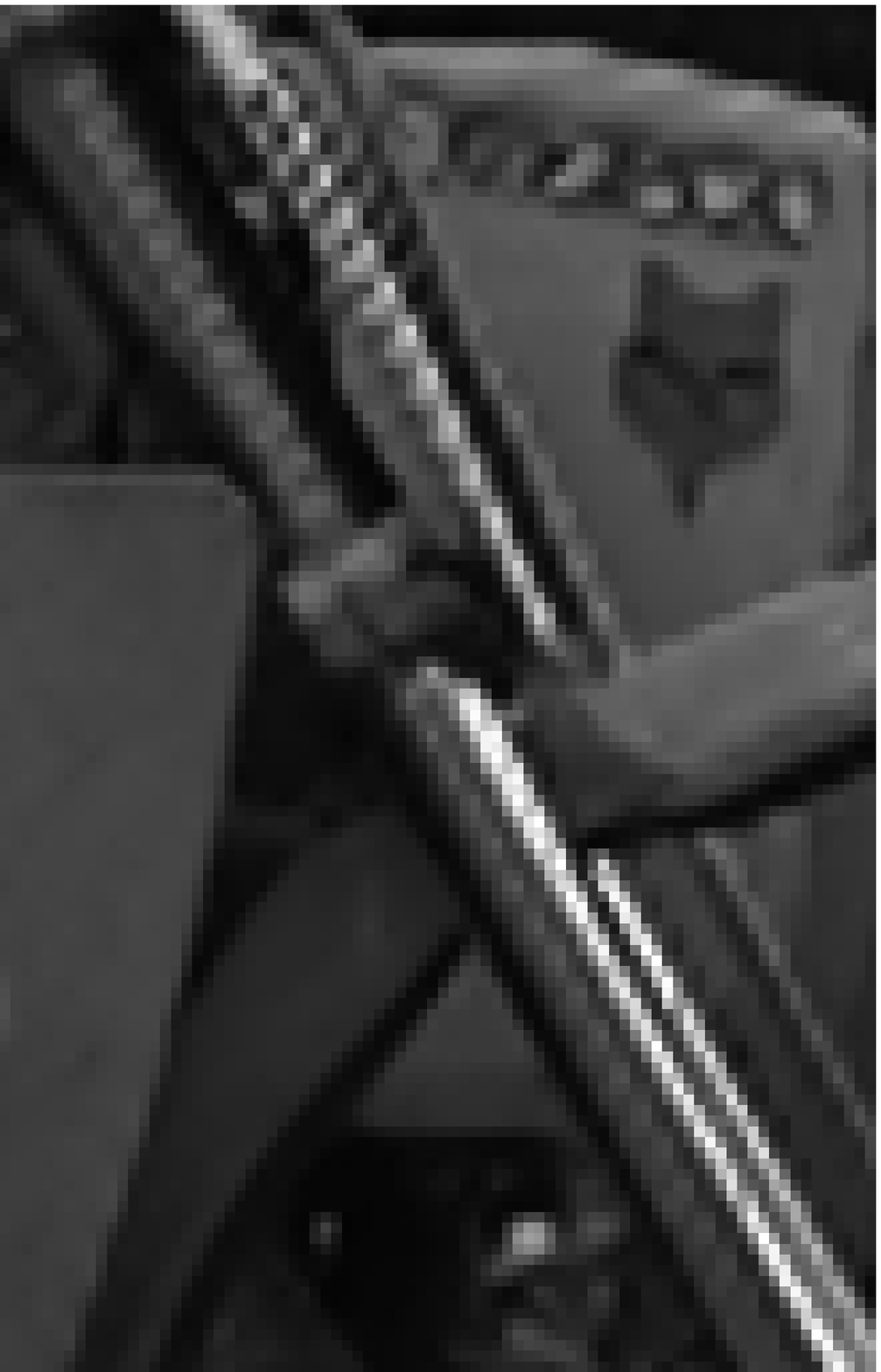}}
\centerline{(h)~NGSDG}
\end{minipage} \\
\centering
\begin{minipage}[b]{0.12 \linewidth}
\centerline{\includegraphics[width=2.2 cm]{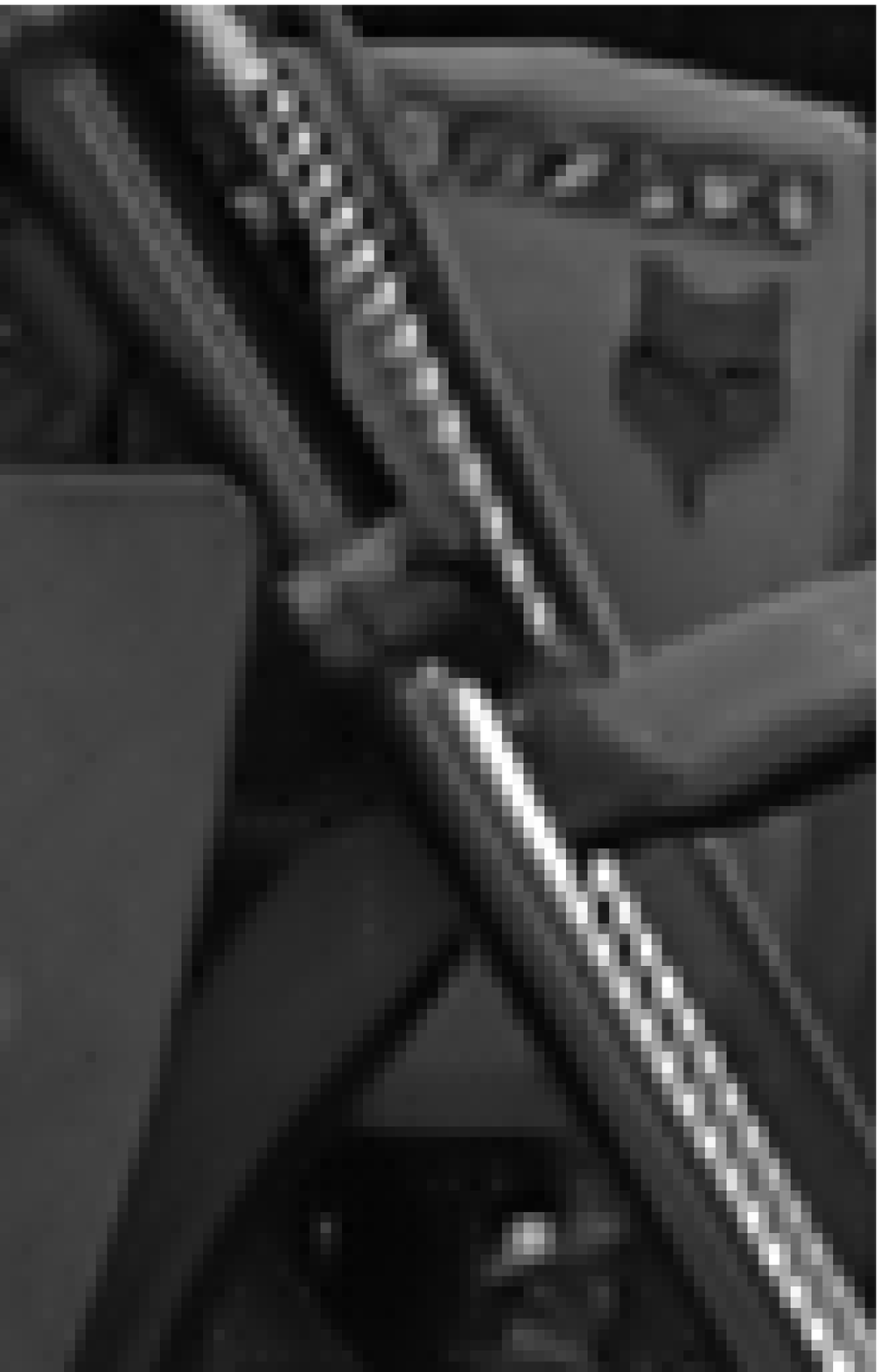}}
\centerline{(i)~NARM}
\end{minipage}
\begin{minipage}[b]{0.12 \linewidth}
\centerline{\includegraphics[width=2.2 cm]{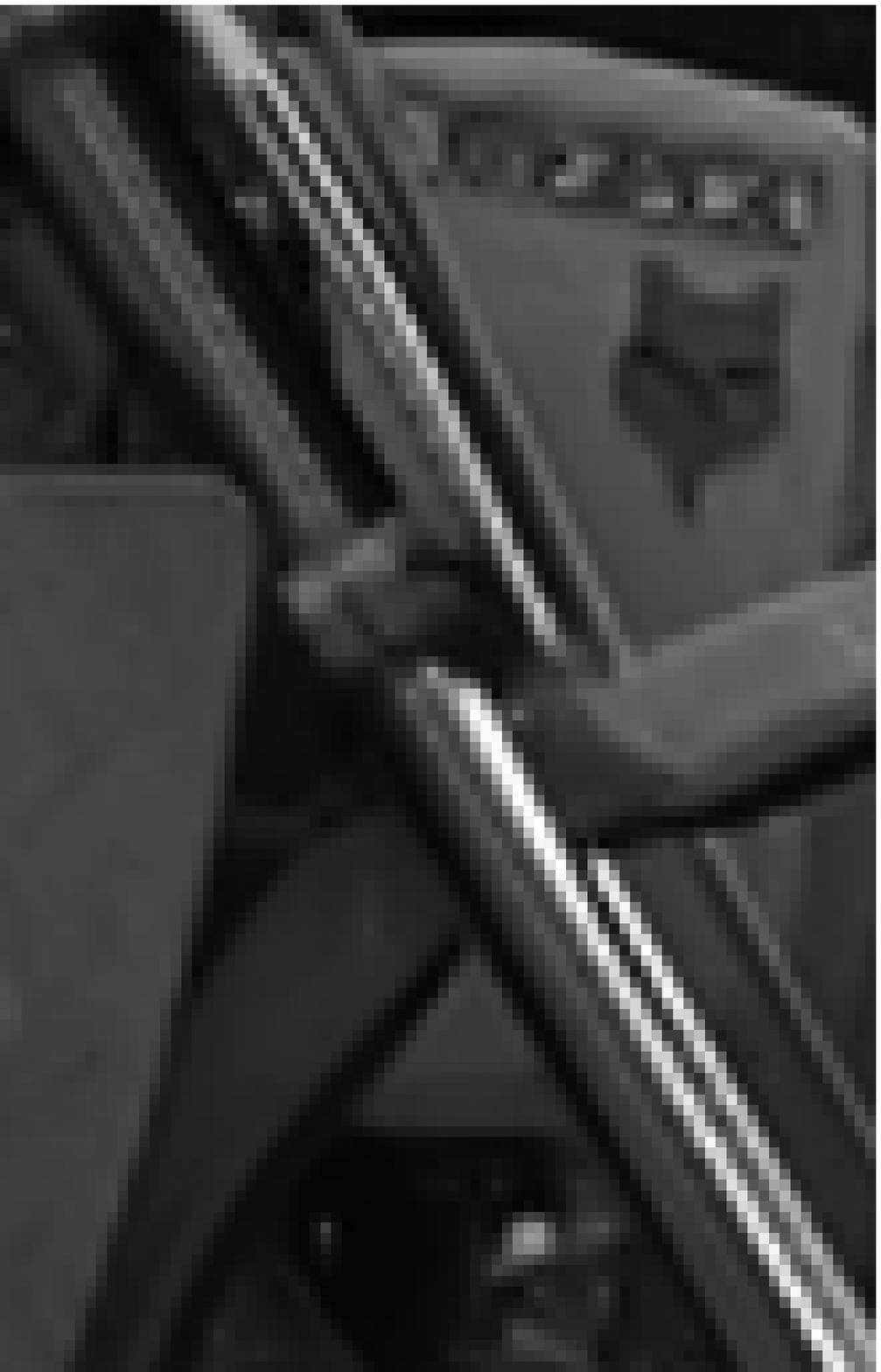}}
\centerline{(j)~ANSM}
\end{minipage}
\begin{minipage}[b]{0.12 \linewidth}
\centerline{\includegraphics[width=2.2 cm]{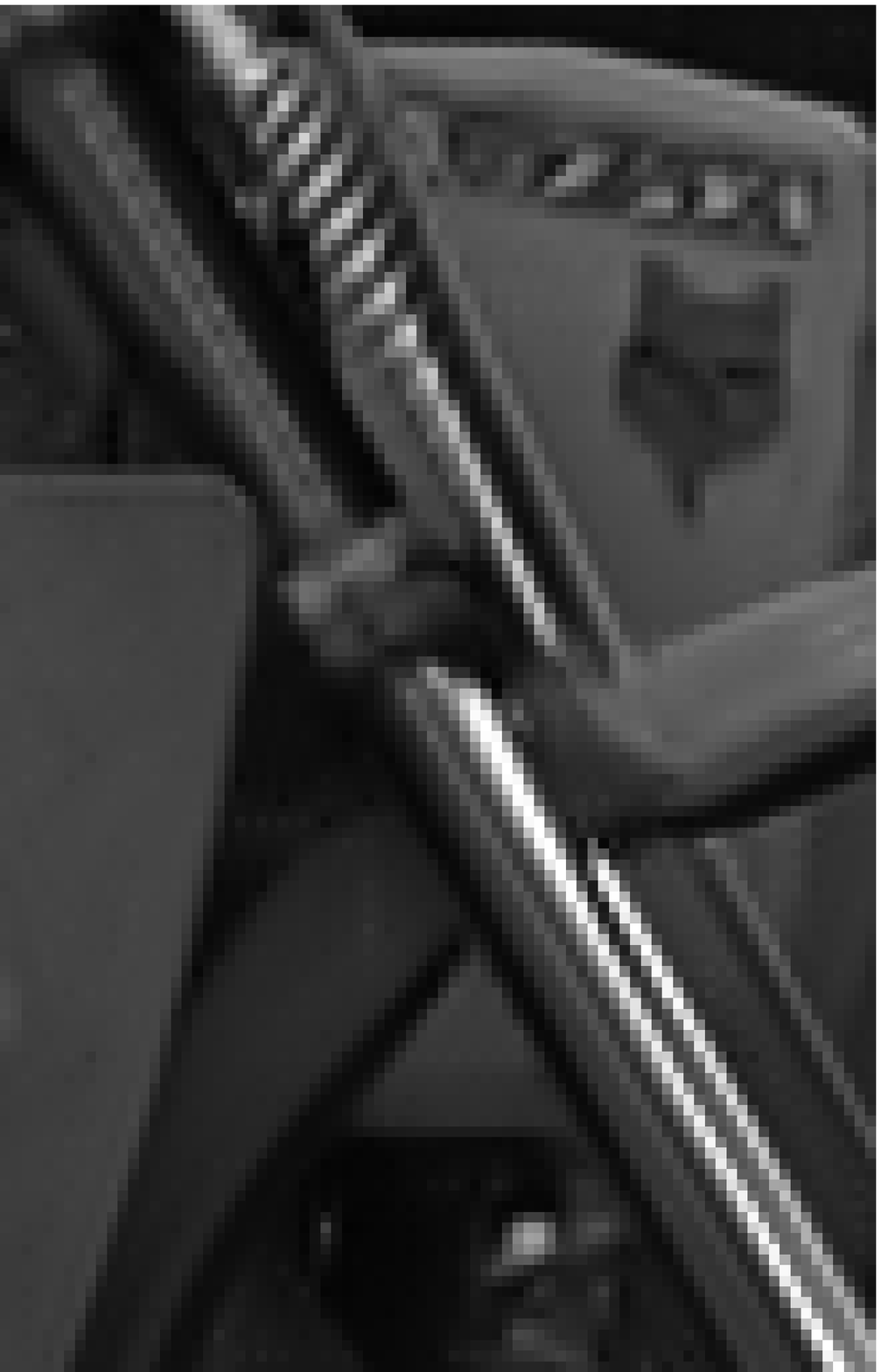}}
\centerline{(k)~NLPC}
\end{minipage}
\begin{minipage}[b]{0.12 \linewidth}
\centerline{\includegraphics[width=2.2 cm]{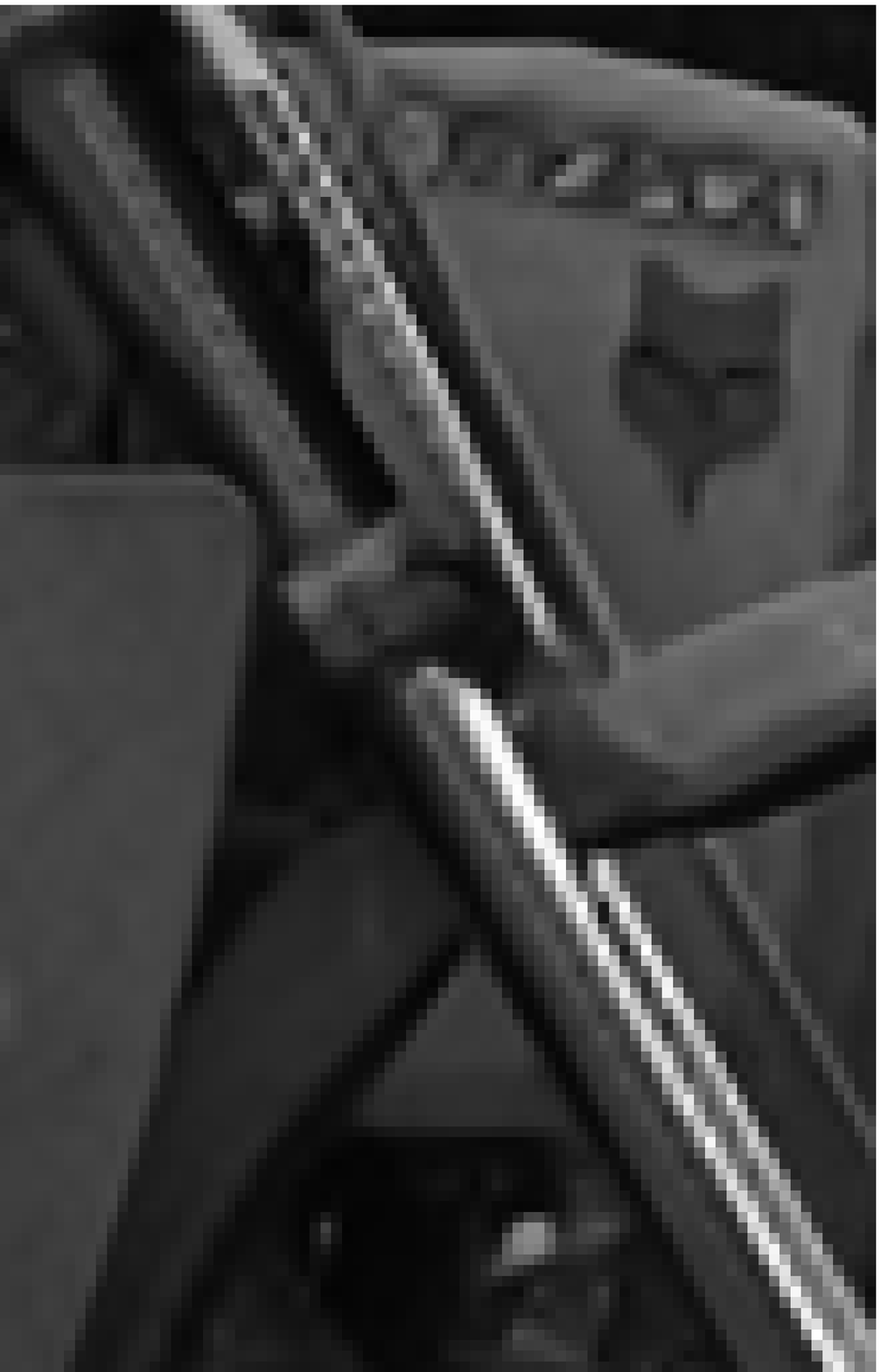}}
\centerline{(l)~FIRF}
\end{minipage}
\begin{minipage}[b]{0.12 \linewidth}
\centerline{\includegraphics[width=2.2 cm]{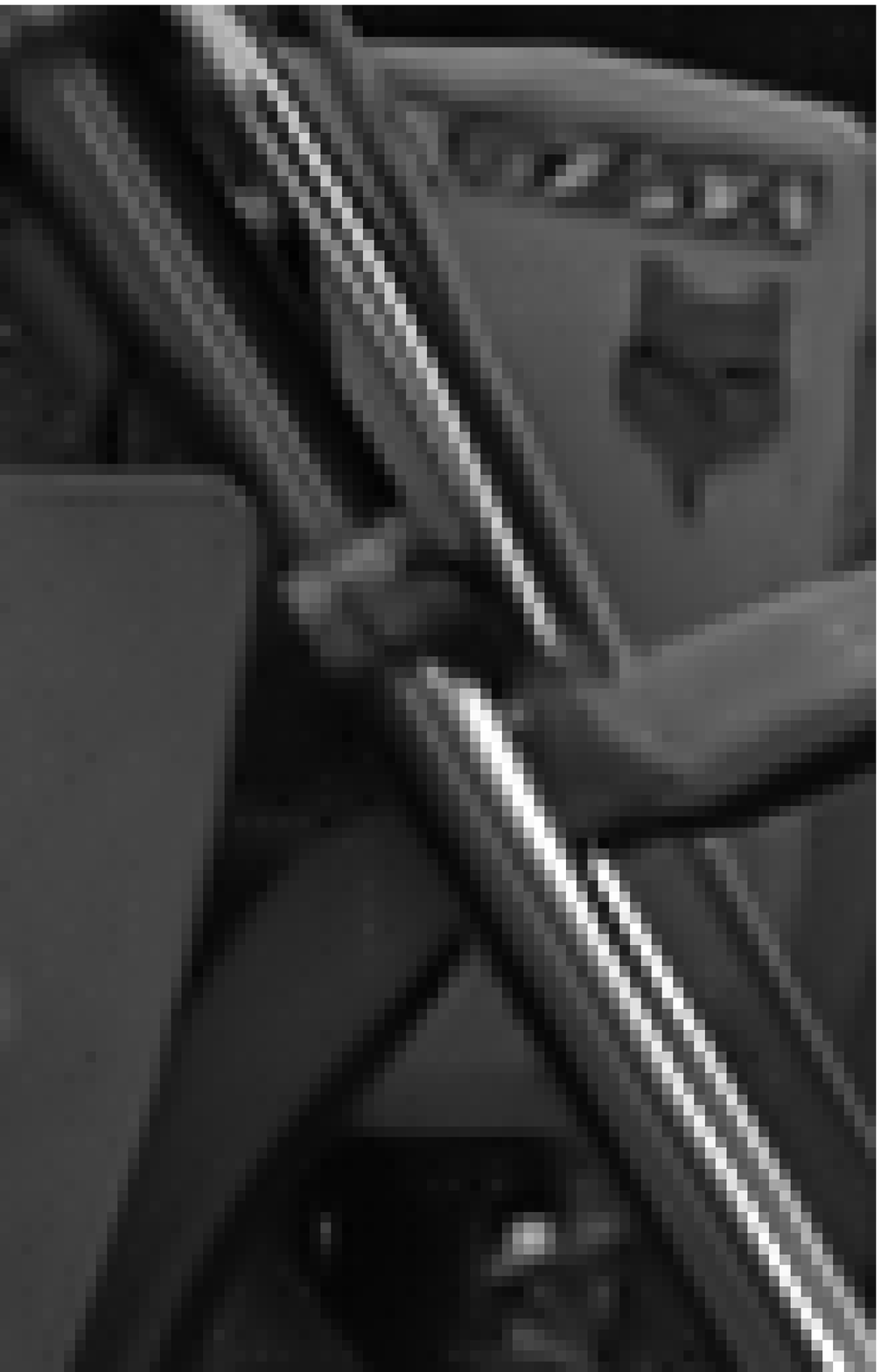}}
\centerline{(m)~MAIN}
\end{minipage}
\begin{minipage}[b]{0.12 \linewidth}
\centerline{\includegraphics[width=2.2 cm]{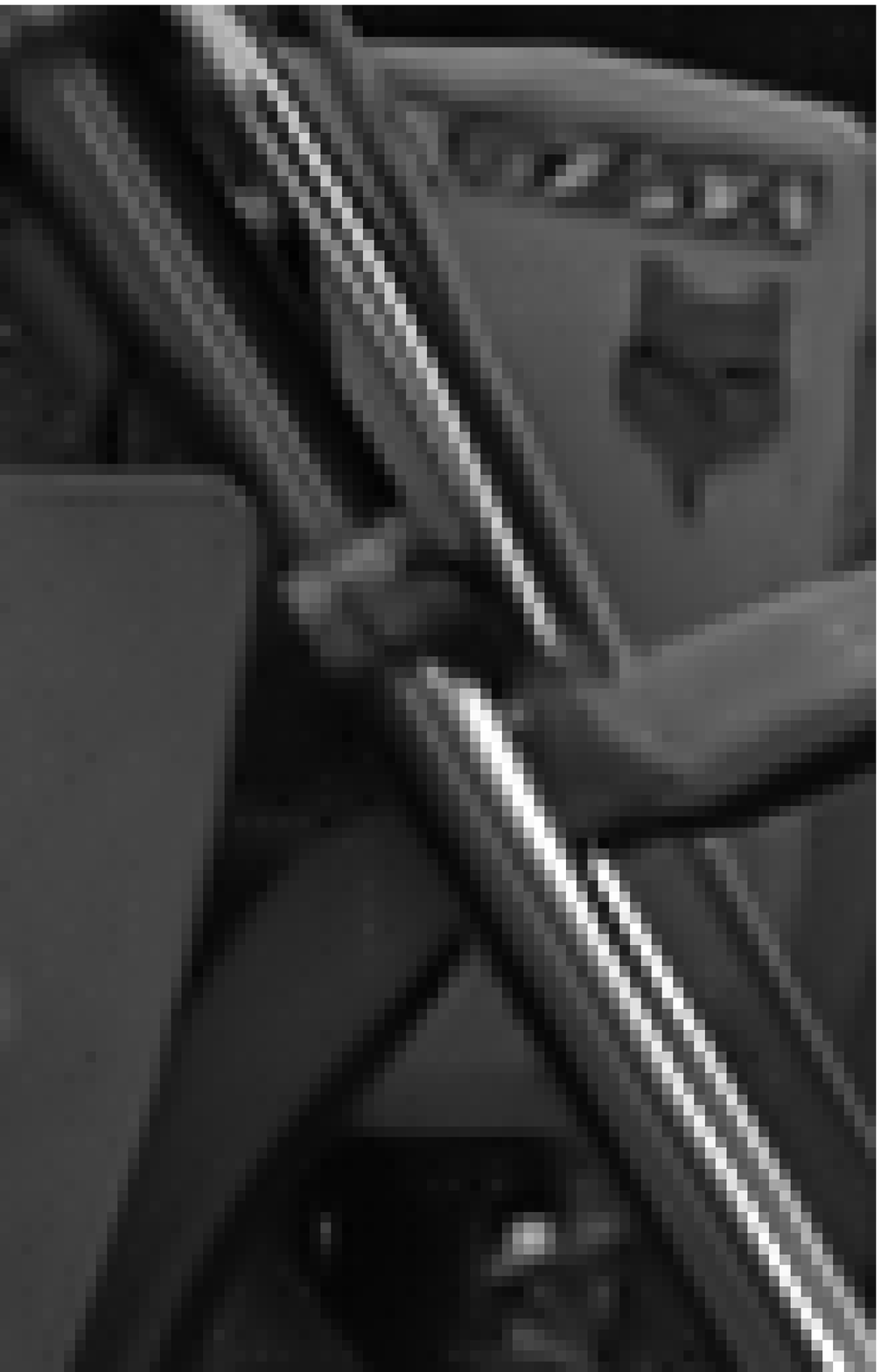}}
\centerline{(n)~MISTER}
\end{minipage}
\begin{minipage}[b]{0.12 \linewidth}
\centerline{\includegraphics[width=2.2 cm]{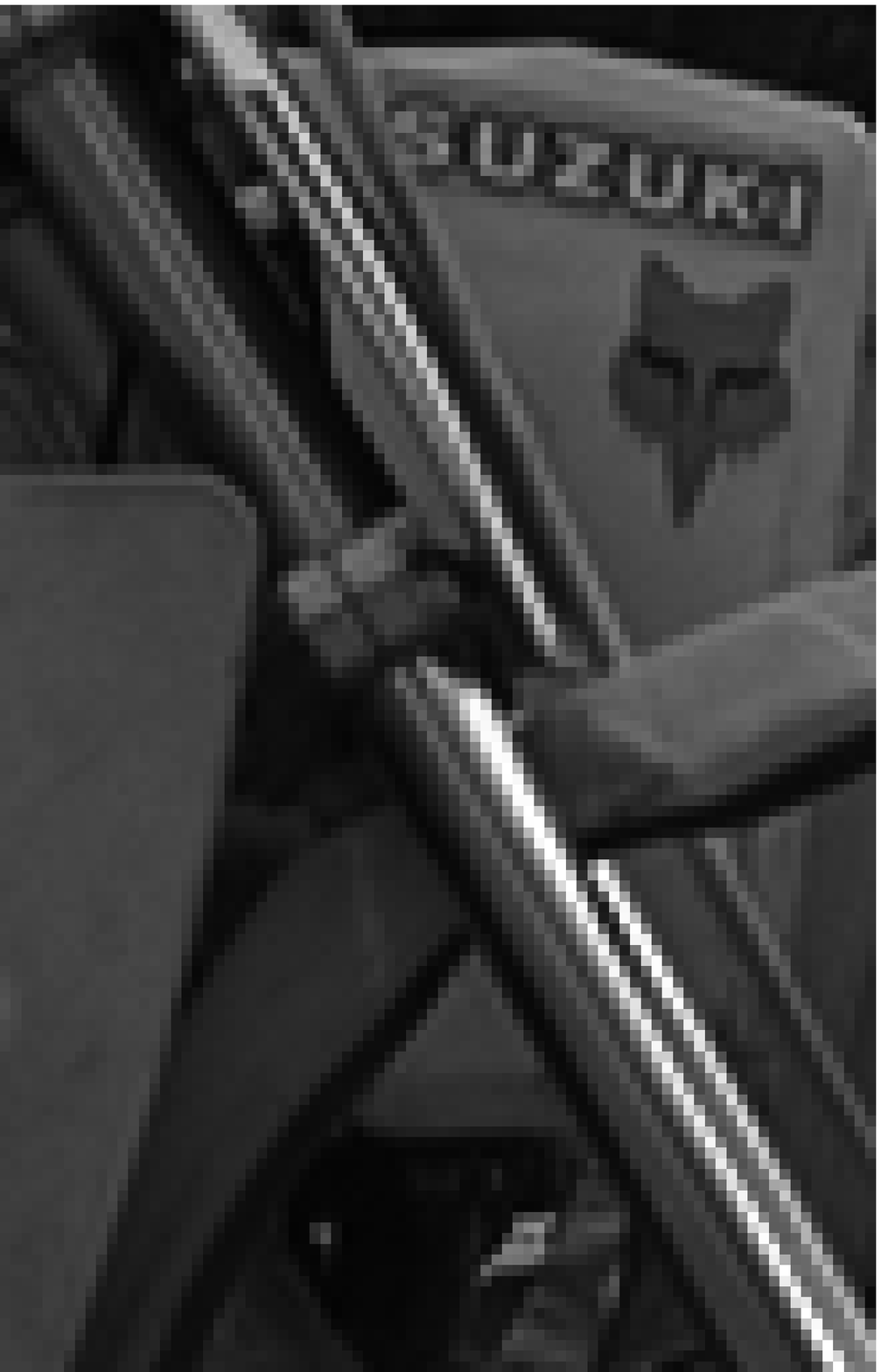}}
\centerline{(o)~Ground Truth}
\end{minipage}
\caption{The zoom-in comparison of crop from \textit{Motorbike} in the interpolation task by a factor of $2$.}
\label{fig:edge_motorbike}
\end{figure*}

\begin{figure*}[ht]
\begin{minipage}[b]{0.06 \linewidth}
\centering
\centerline{\includegraphics[width=1.1 cm]{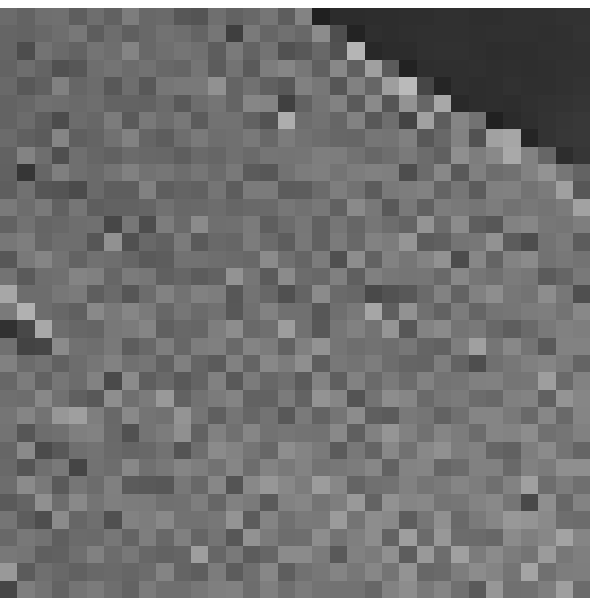}}
\centerline{(a)~LR}
\end{minipage}
\begin{minipage}[b]{0.12 \linewidth}
\centerline{\includegraphics[width=2.2 cm]{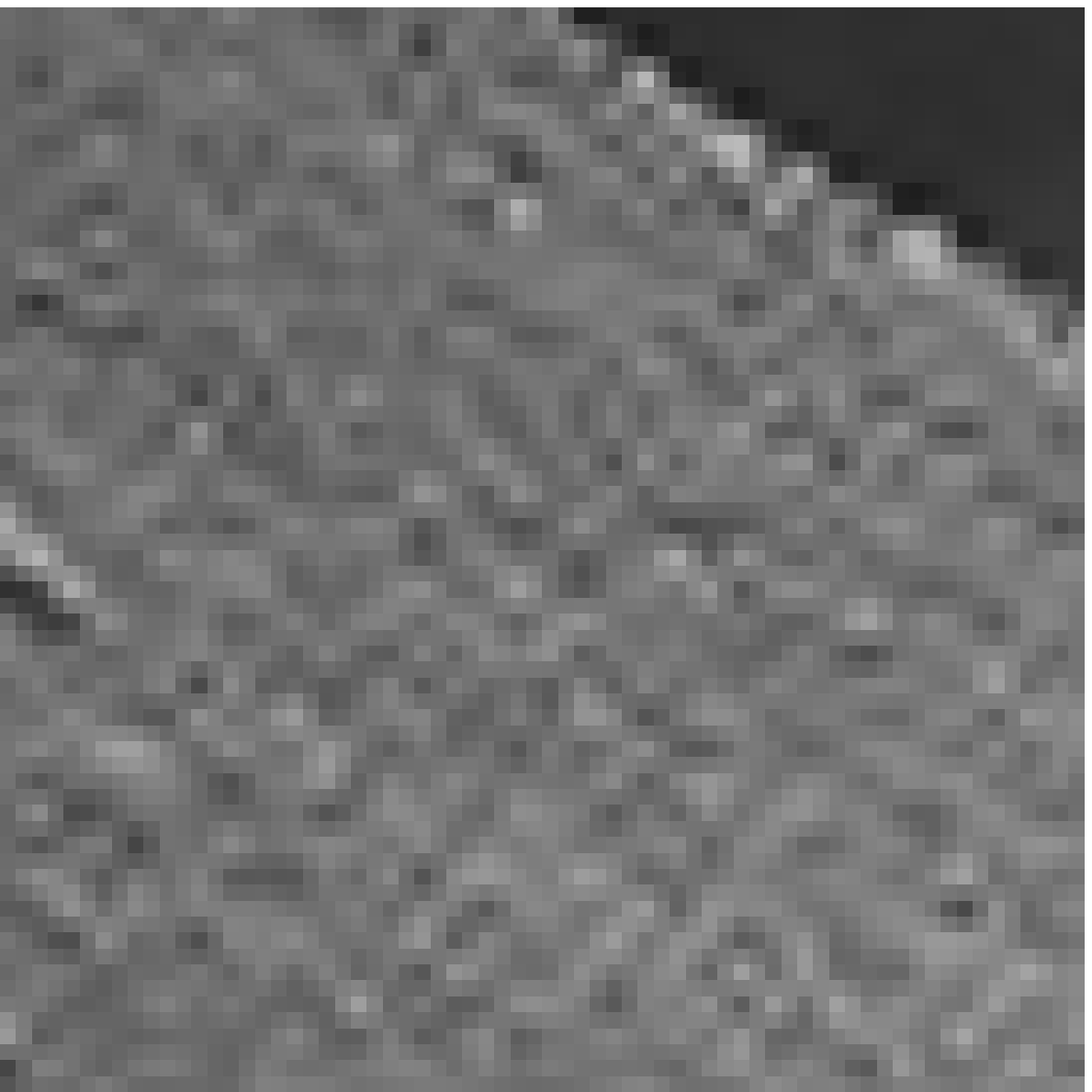}}
\centerline{(b)~Bicubic}
\end{minipage}
\begin{minipage}[b]{0.12 \linewidth}
\centering
\centerline{\includegraphics[width=2.2 cm]{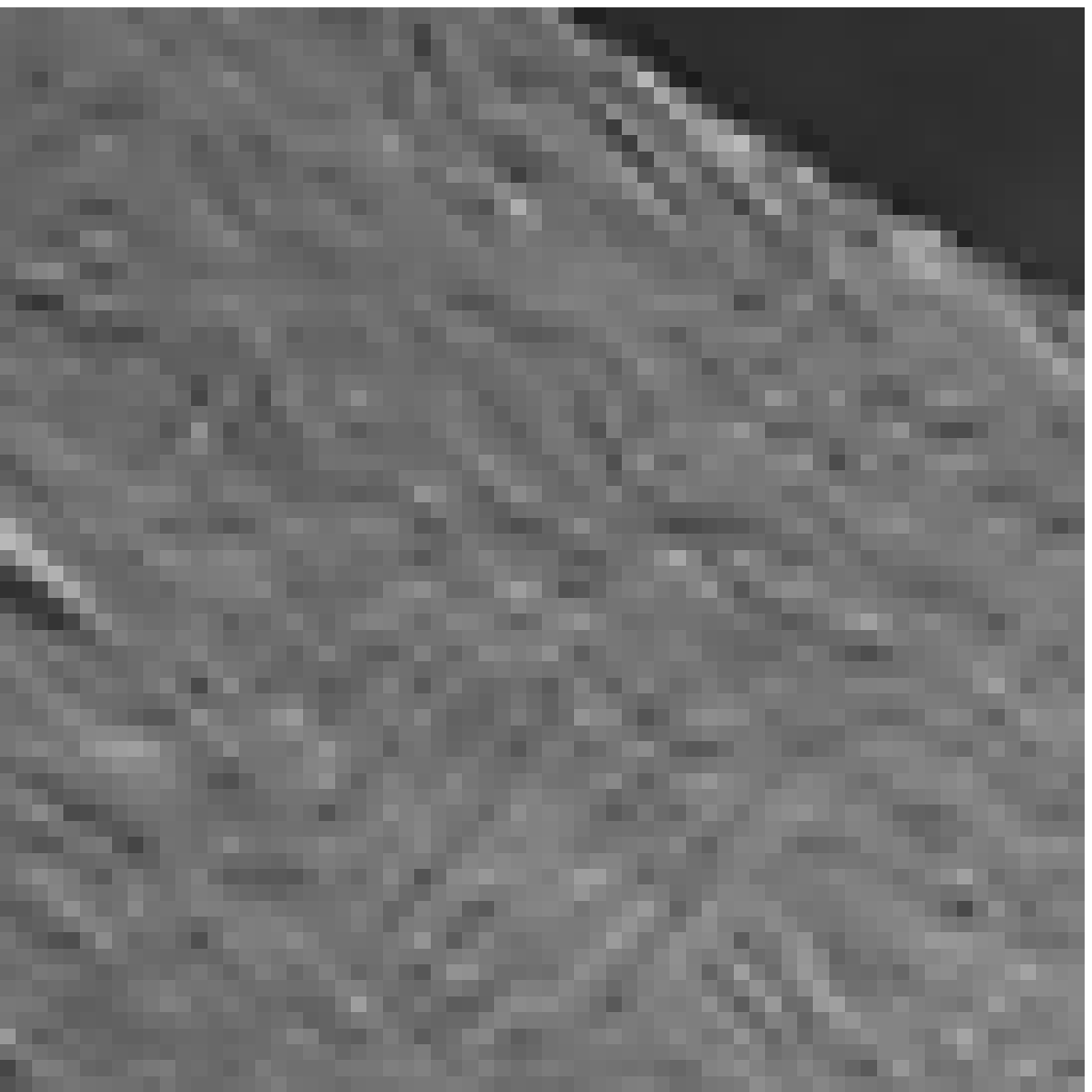}}
\centerline{(c)~NEDI}
\end{minipage}
\begin{minipage}[b]{0.12 \linewidth}
\centering
\centerline{\includegraphics[width=2.2 cm]{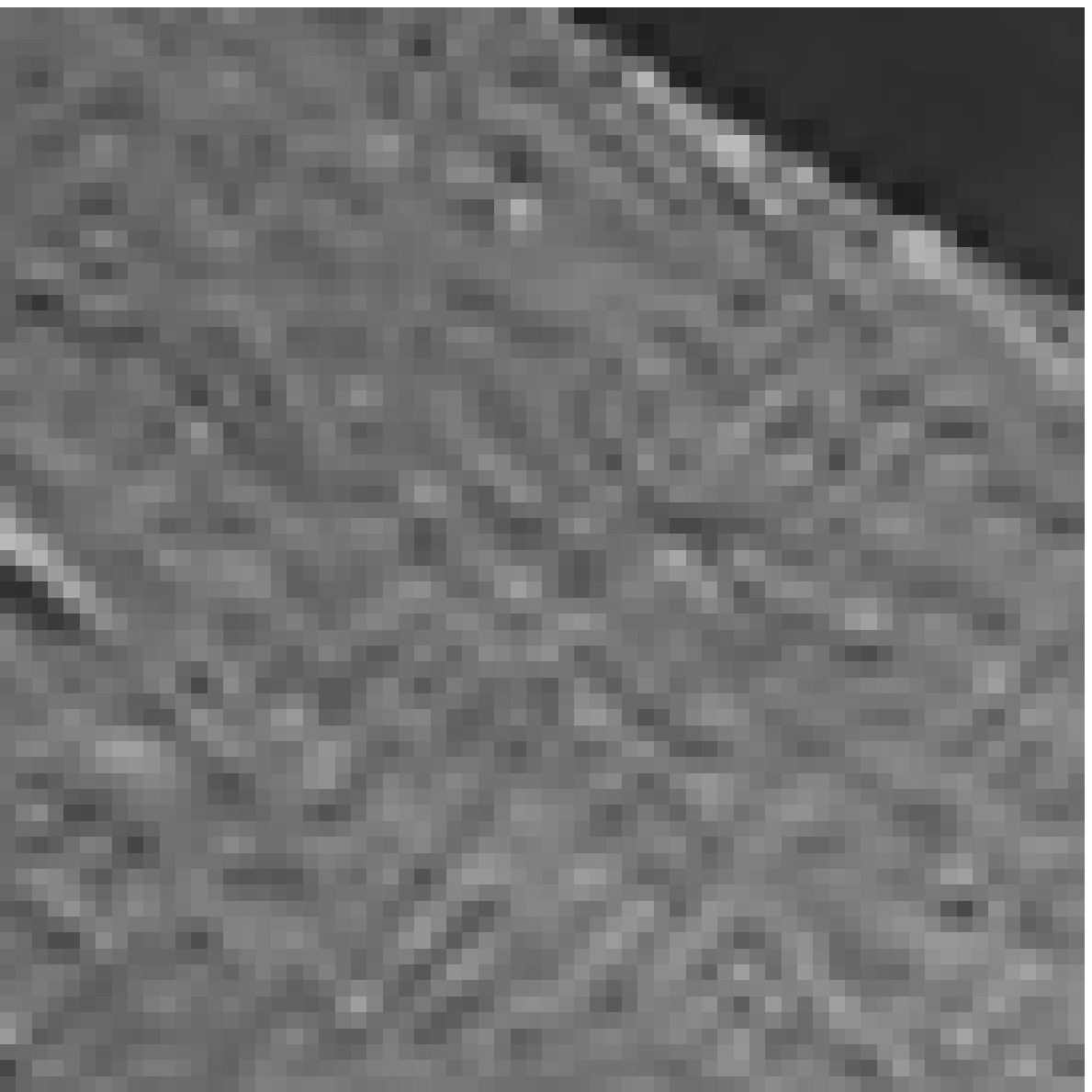}}
\centerline{(d)~SME}
\end{minipage}
\begin{minipage}[b]{0.12 \linewidth}
\centering
\centerline{\includegraphics[width=2.2 cm]{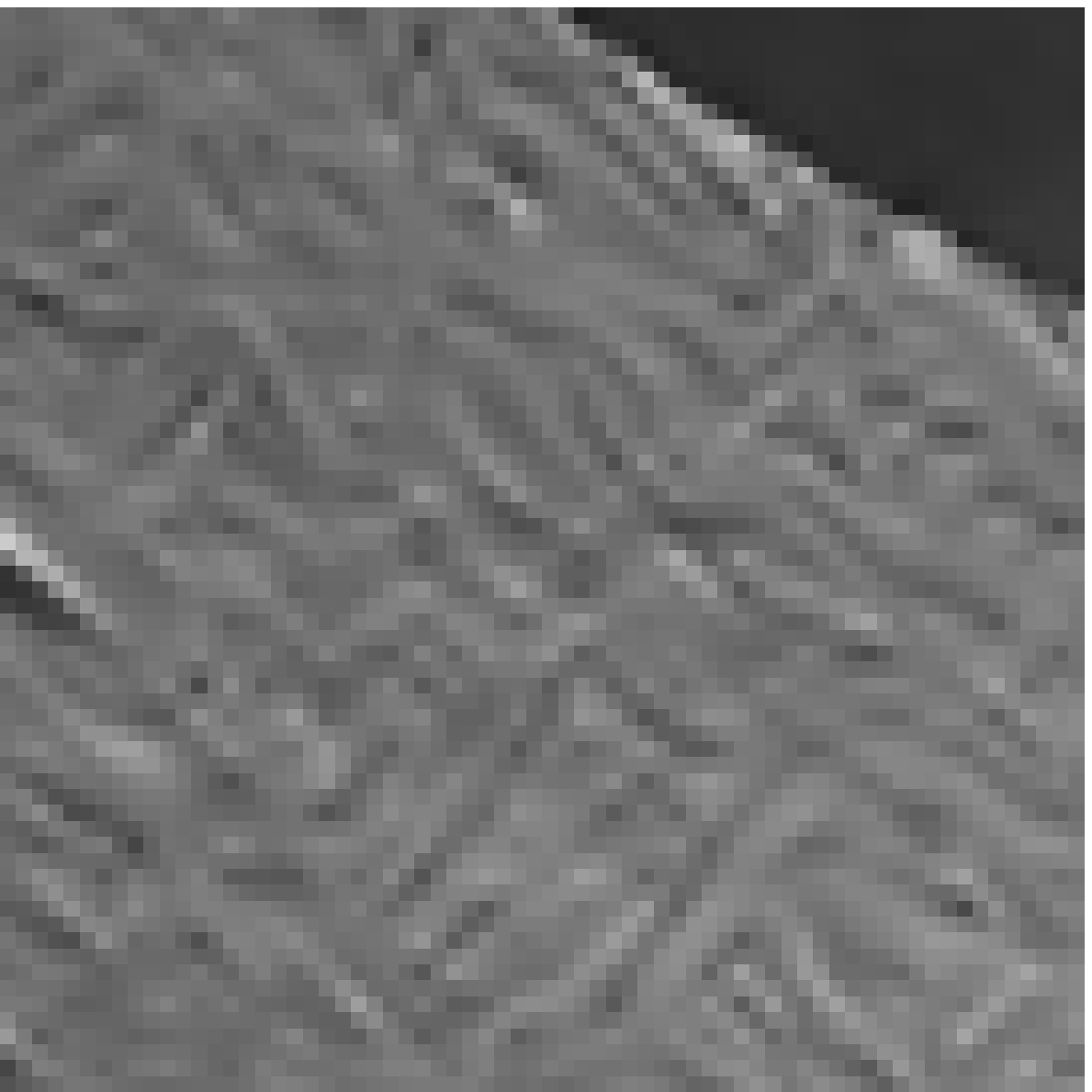}}
\centerline{(e)~SAI}
\end{minipage}
\begin{minipage}[b]{0.12 \linewidth}
\centering
\centerline{\includegraphics[width=2.2 cm]{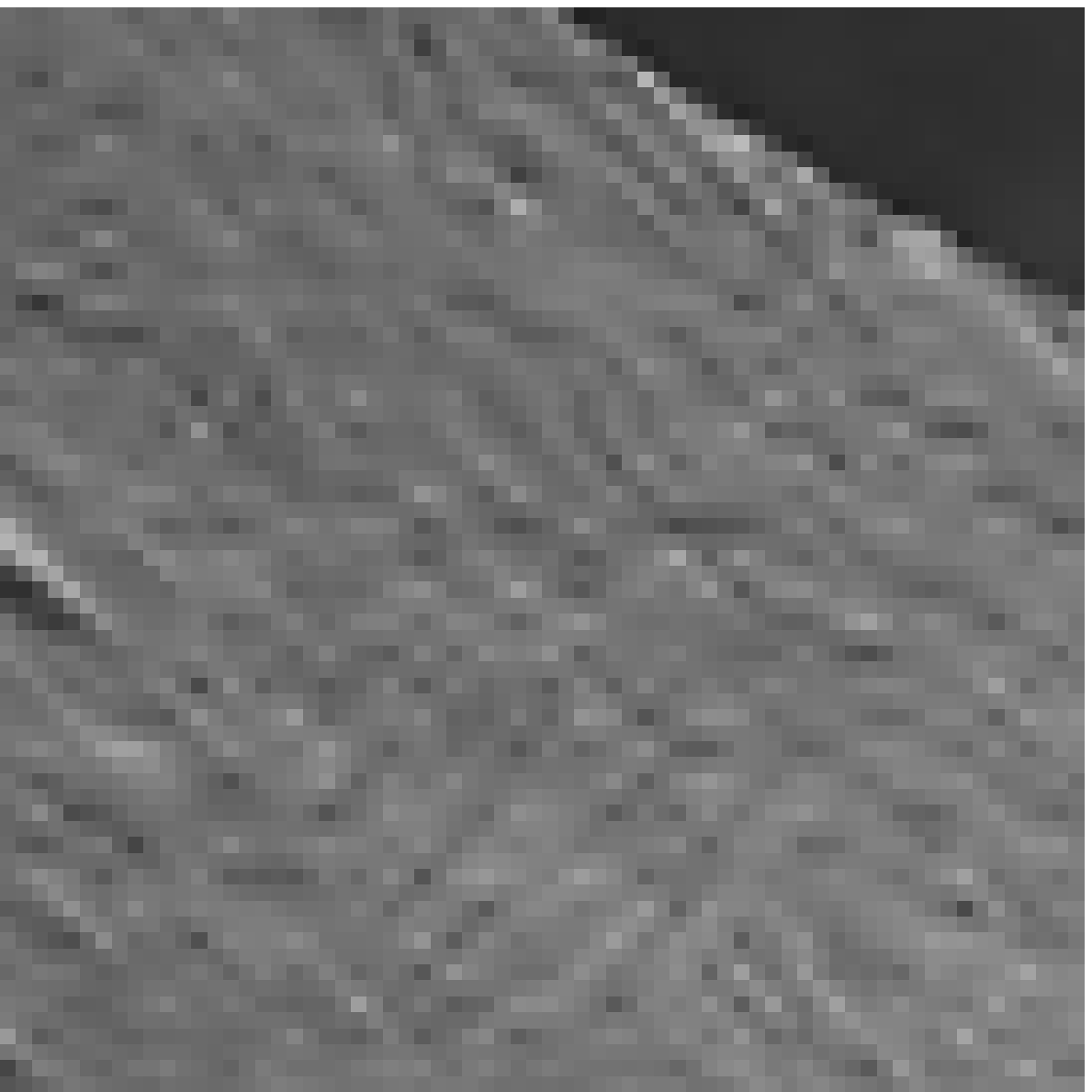}}
\centerline{(f)~RLLR}
\end{minipage}
\begin{minipage}[b]{0.12 \linewidth}
\centering
\centerline{\includegraphics[width=2.2 cm]{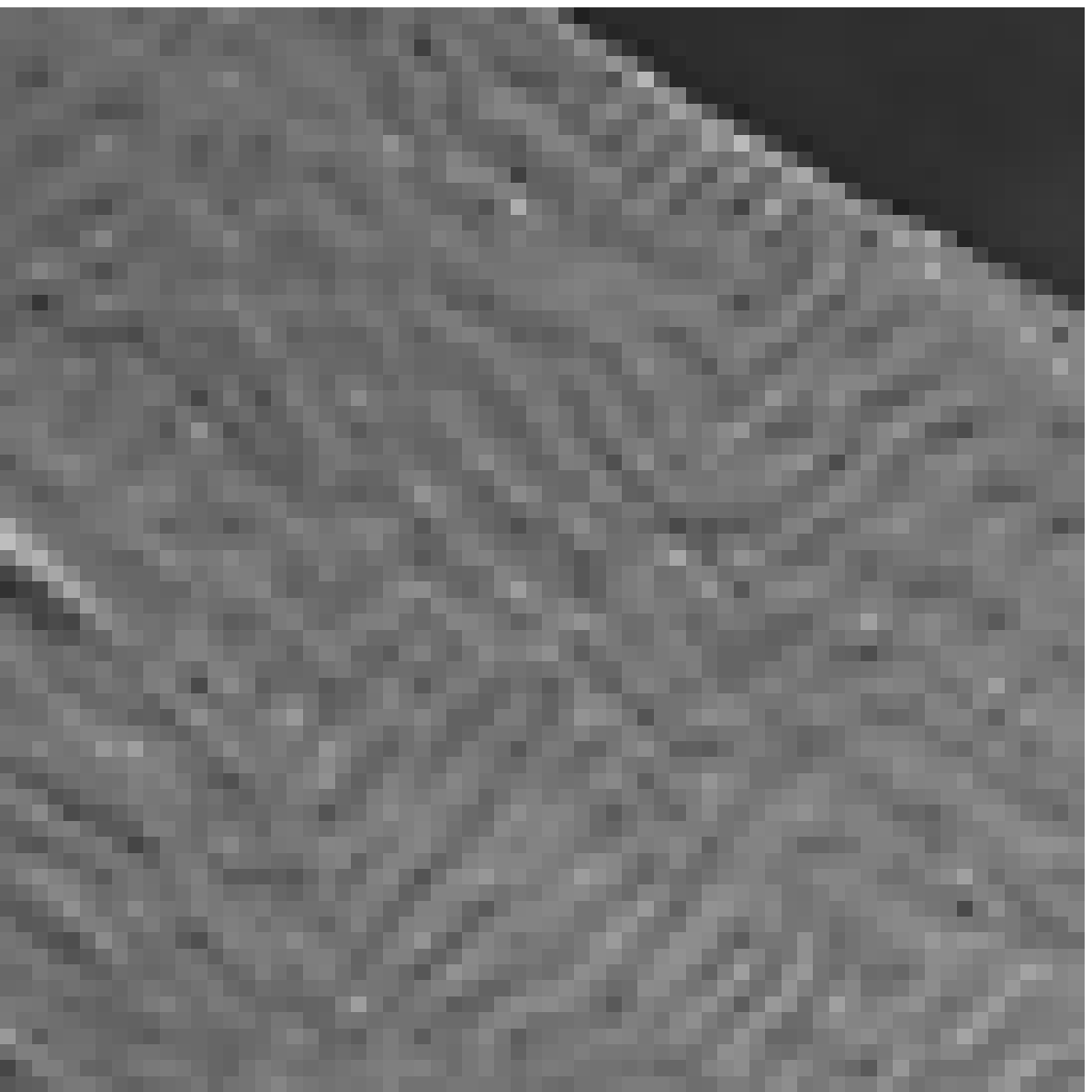}}
\centerline{(g)~MSIA}
\end{minipage}
\begin{minipage}[b]{0.12 \linewidth}
\centering
\centerline{\includegraphics[width=2.2 cm]{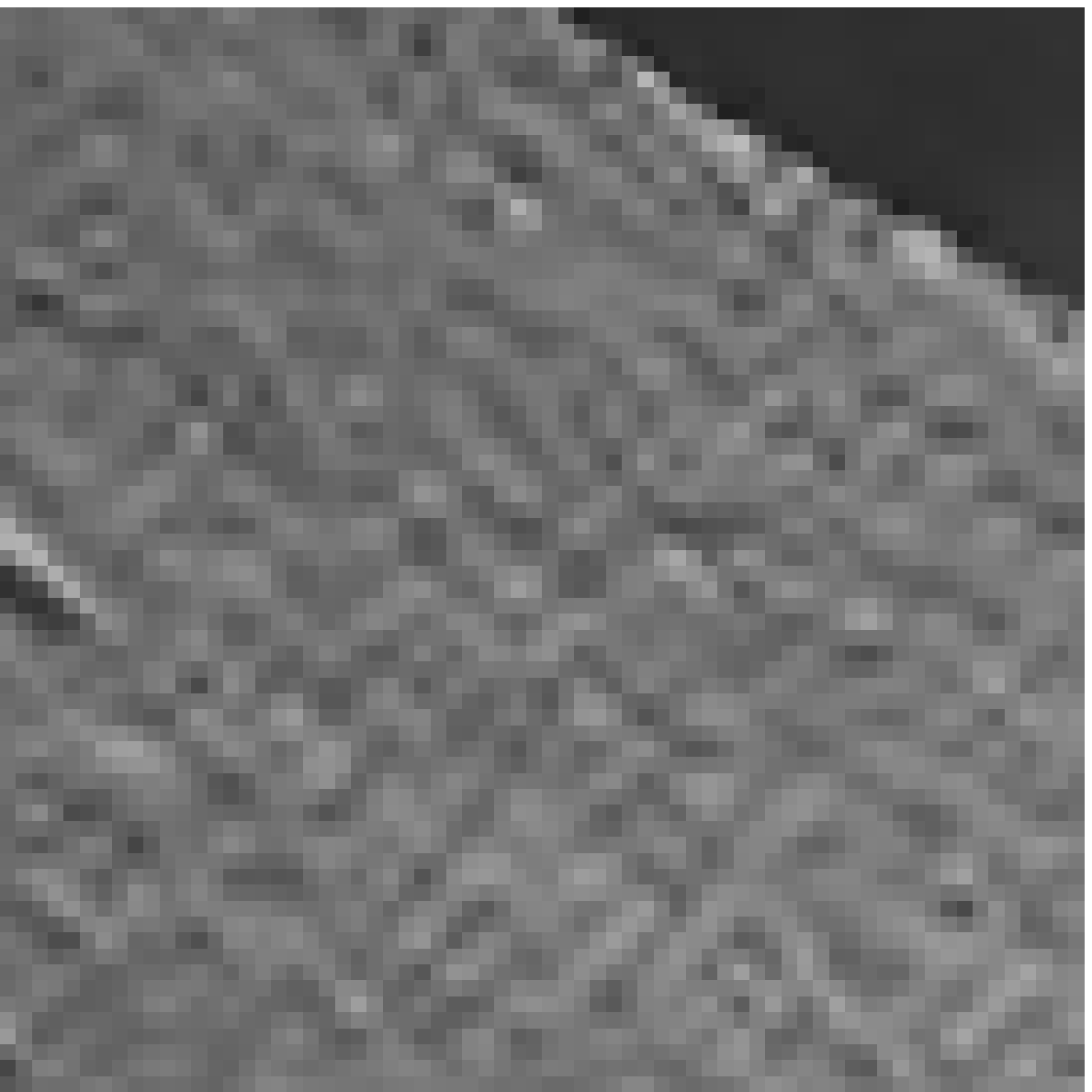}}
\centerline{(h)~NGSDG}
\end{minipage} \\
\centering
\begin{minipage}[b]{0.12 \linewidth}
\centerline{\includegraphics[width=2.2 cm]{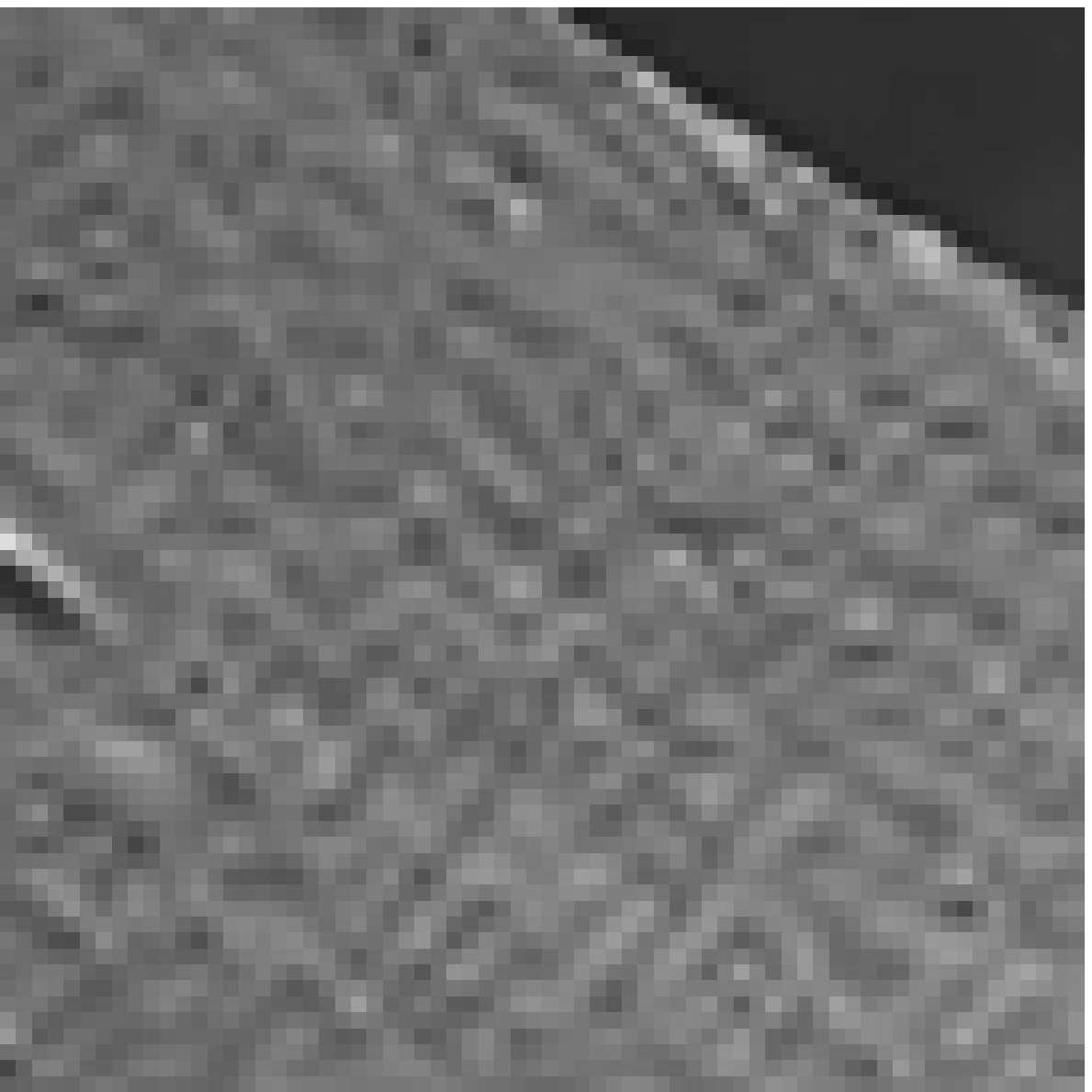}}
\centerline{(i)~NARM}
\end{minipage}
\begin{minipage}[b]{0.12 \linewidth}
\centerline{\includegraphics[width=2.2 cm]{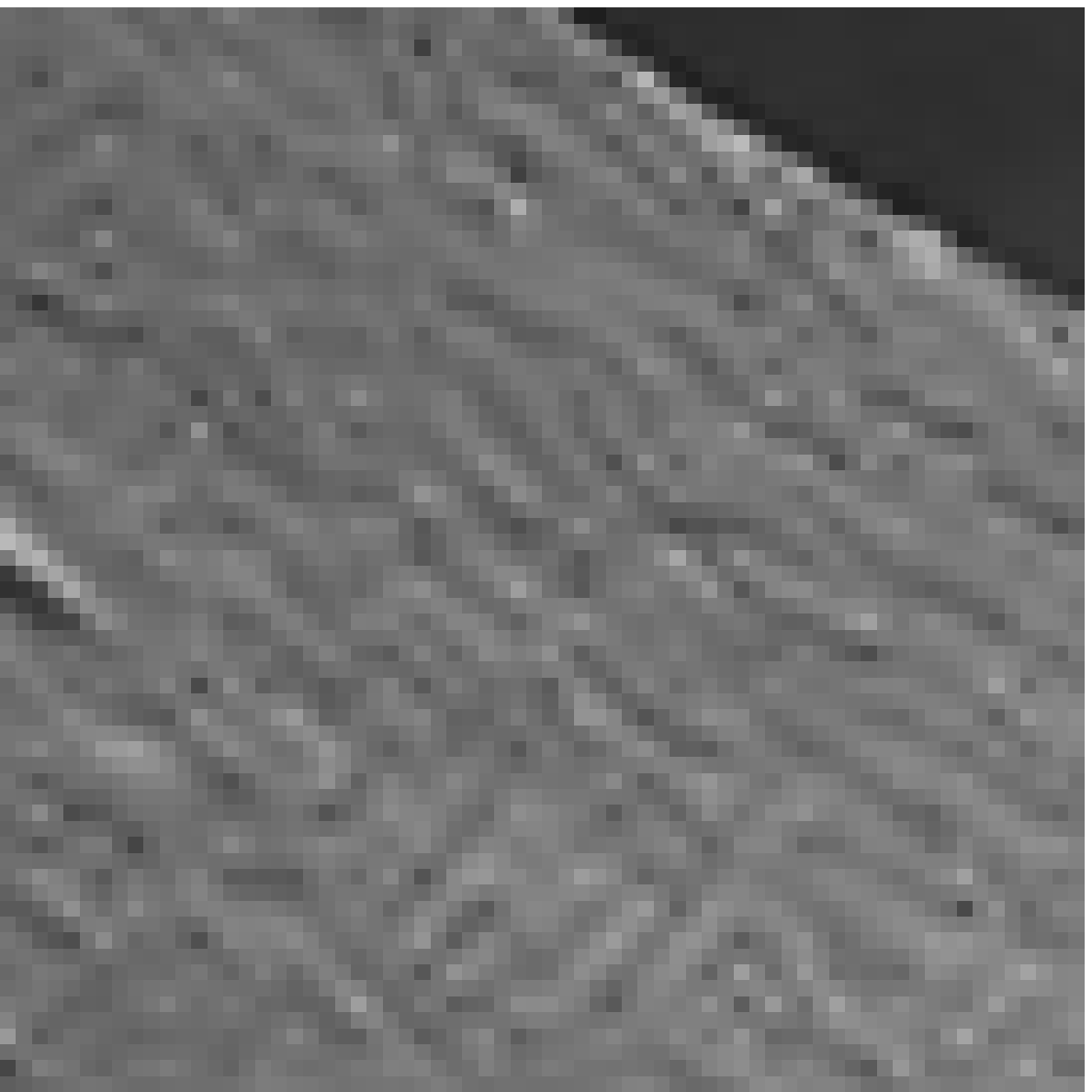}}
\centerline{(j)~ANSM}
\end{minipage}
\begin{minipage}[b]{0.12 \linewidth}
\centerline{\includegraphics[width=2.2 cm]{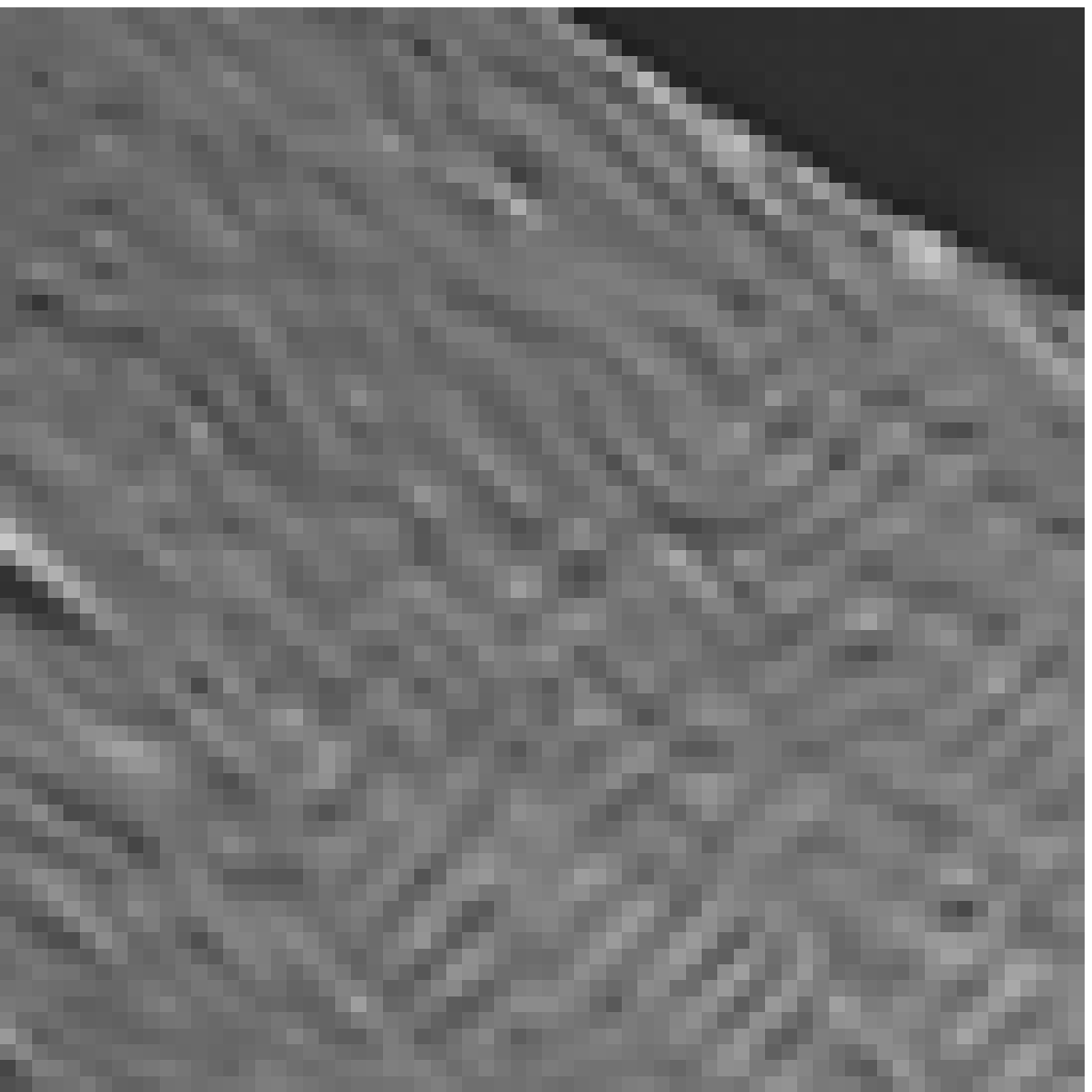}}
\centerline{(k)~NLPC}
\end{minipage}
\begin{minipage}[b]{0.12 \linewidth}
\centerline{\includegraphics[width=2.2 cm]{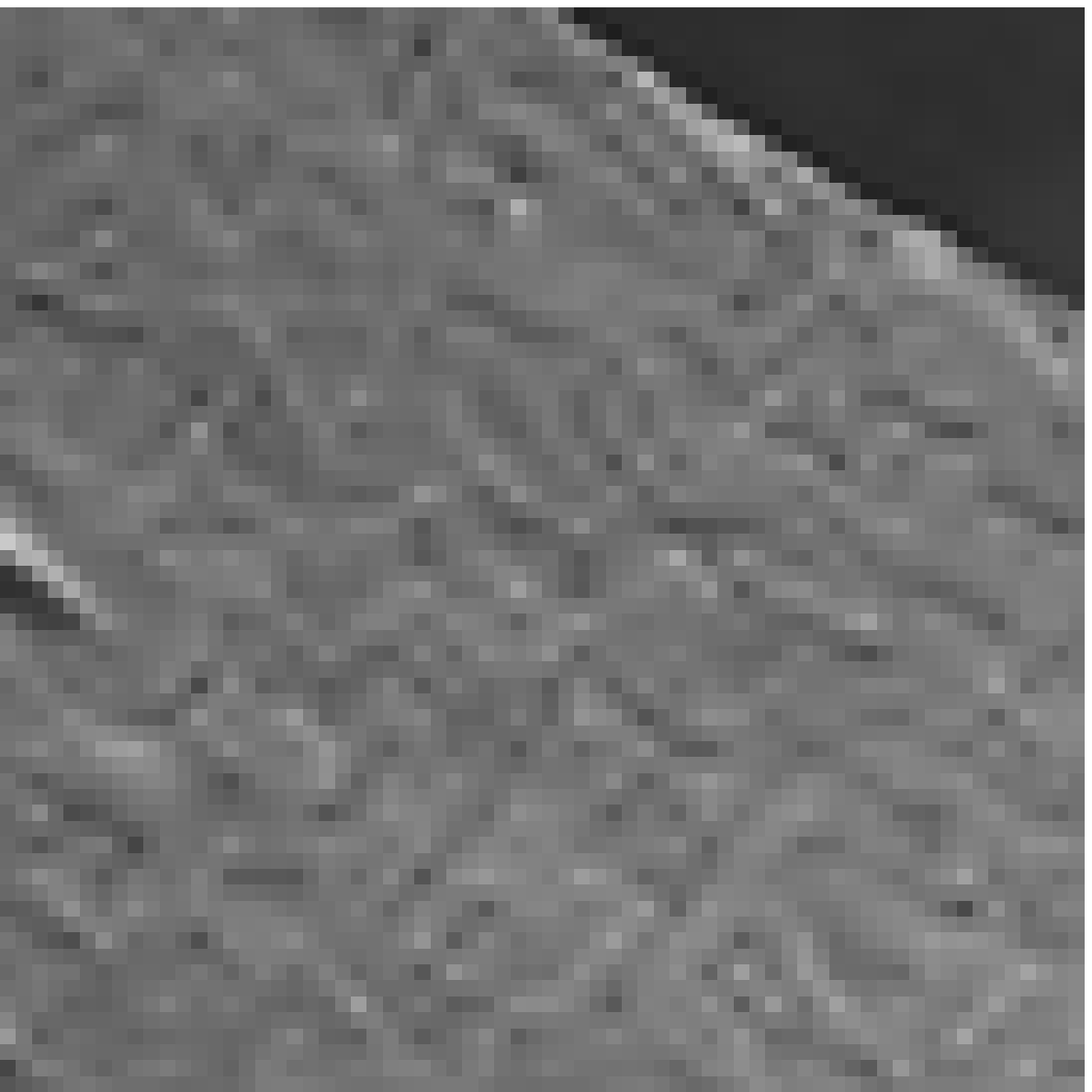}}
\centerline{(l)~FIRF}
\end{minipage}
\begin{minipage}[b]{0.12 \linewidth}
\centerline{\includegraphics[width=2.2 cm]{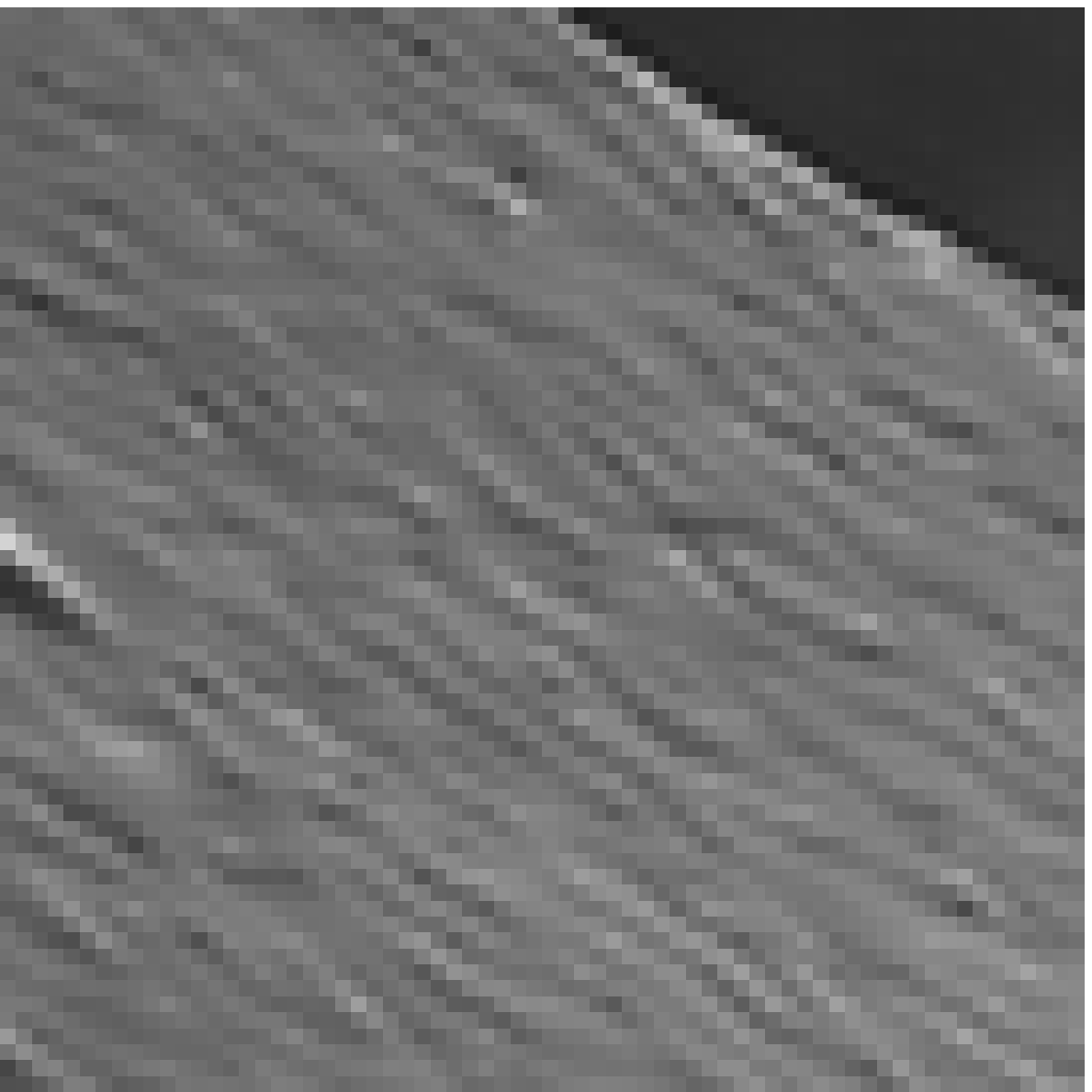}}
\centerline{(m)~MAIN}
\end{minipage}
\begin{minipage}[b]{0.12 \linewidth}
\centerline{\includegraphics[width=2.2 cm]{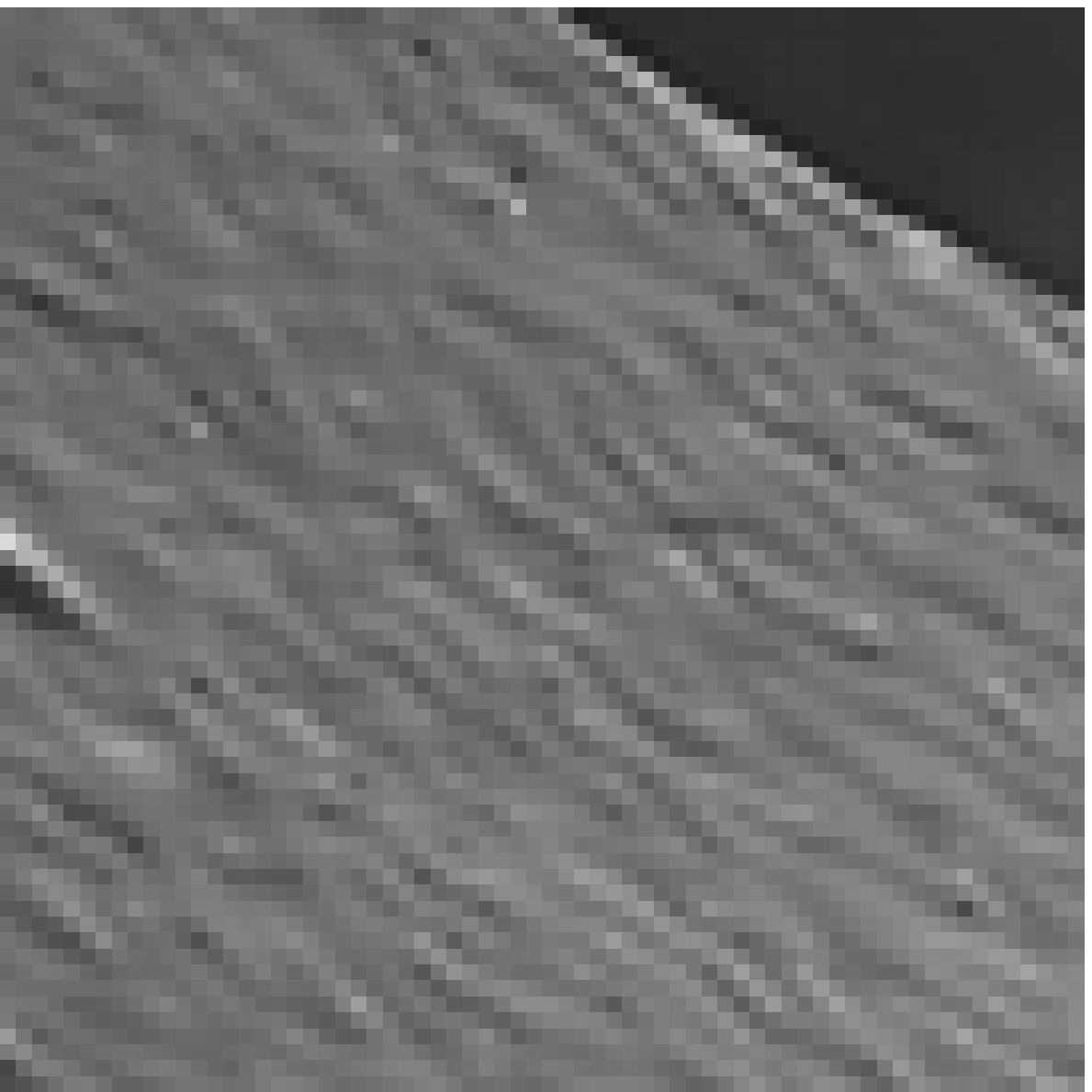}}
\centerline{(n)~MISTER}
\end{minipage}
\begin{minipage}[b]{0.12 \linewidth}
\centerline{\includegraphics[width=2.2 cm]{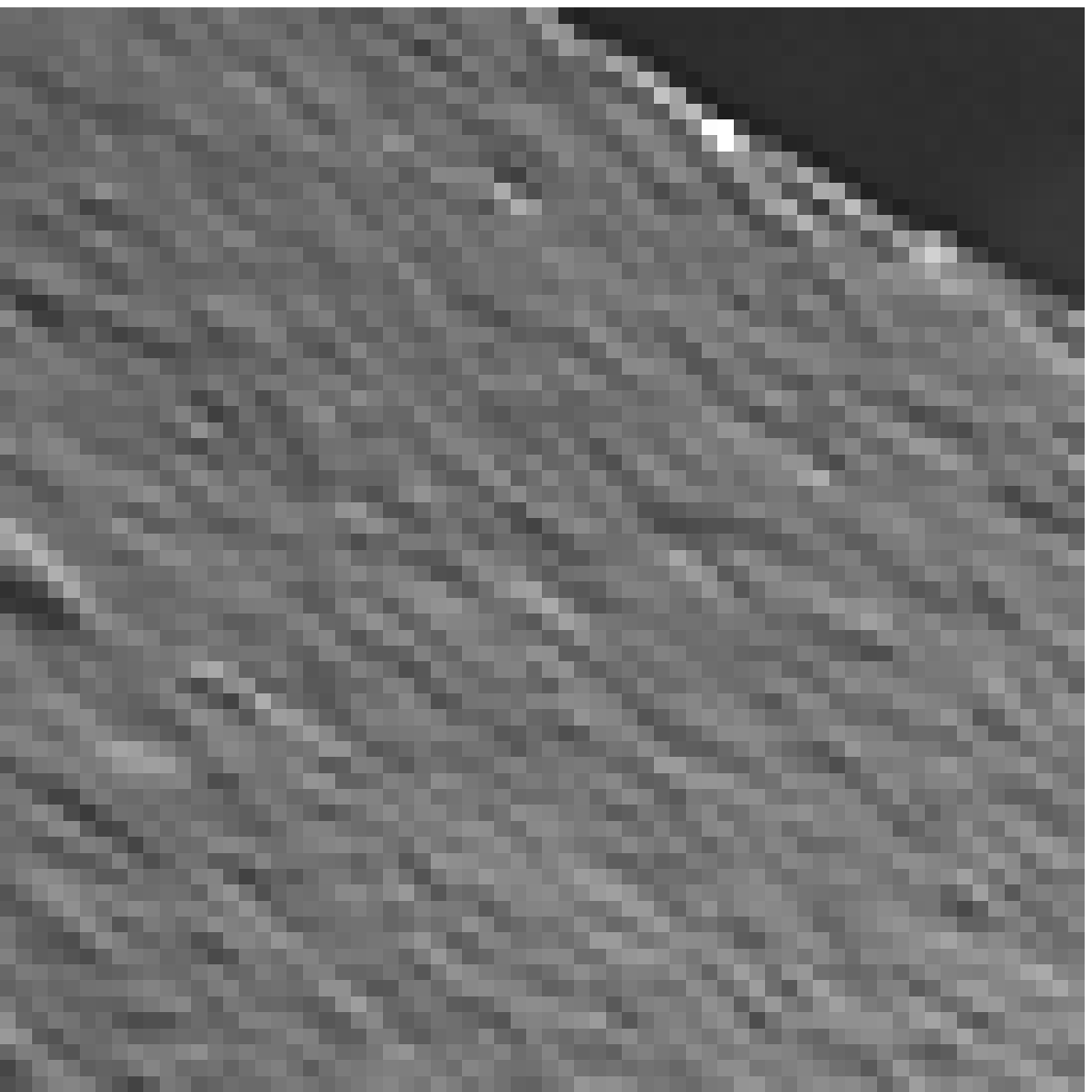}}
\centerline{(o)~Ground Truth}
\end{minipage}
\caption{The zoom-in comparison of crop from \textit{Hats} in the interpolation task by a factor of $2$.}
\label{fig:texture_hats}
\end{figure*}

\begin{figure}[htbp]
\centering
\begin{minipage}[b]{0.485 \linewidth}
\centerline{\includegraphics[width=2.2 cm]{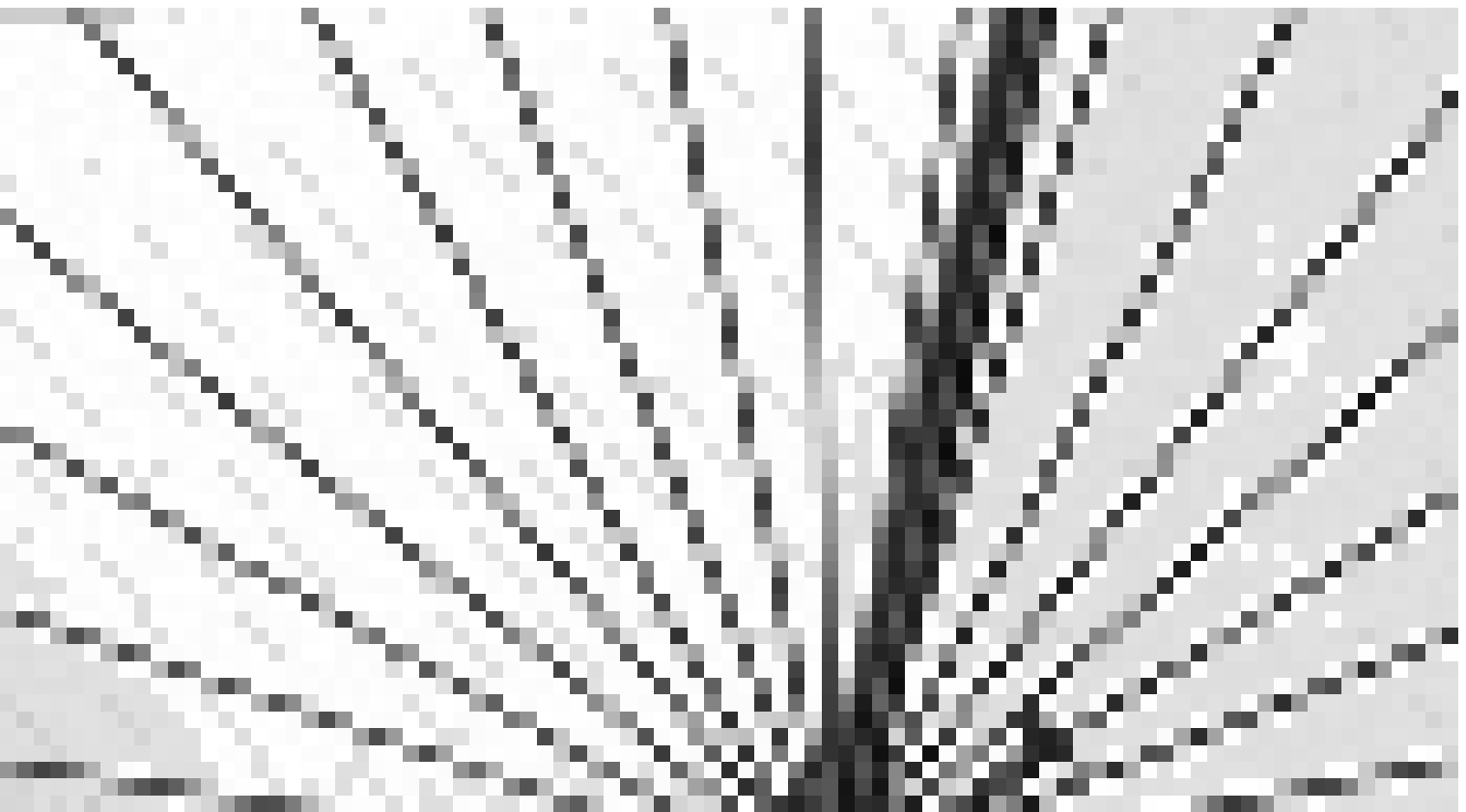}}
\centerline{(a)~LR}
\end{minipage}\\
\begin{minipage}[b]{0.485 \linewidth}
\centerline{\includegraphics[width=4.4 cm]{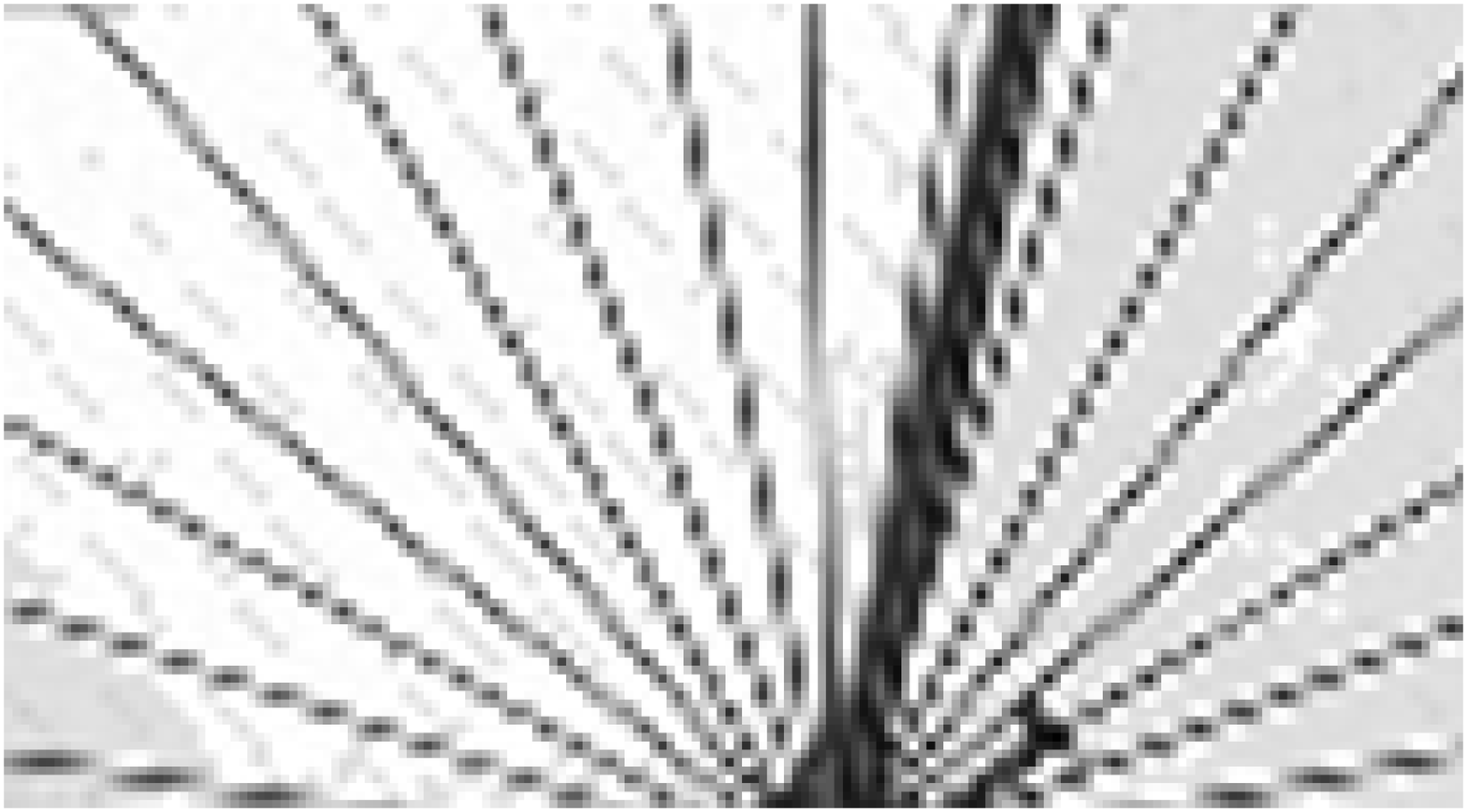}}
\centerline{(b)~Bicubic}
\end{minipage}
\begin{minipage}[b]{0.485 \linewidth}
\centering
\centerline{\includegraphics[width=4.4 cm]{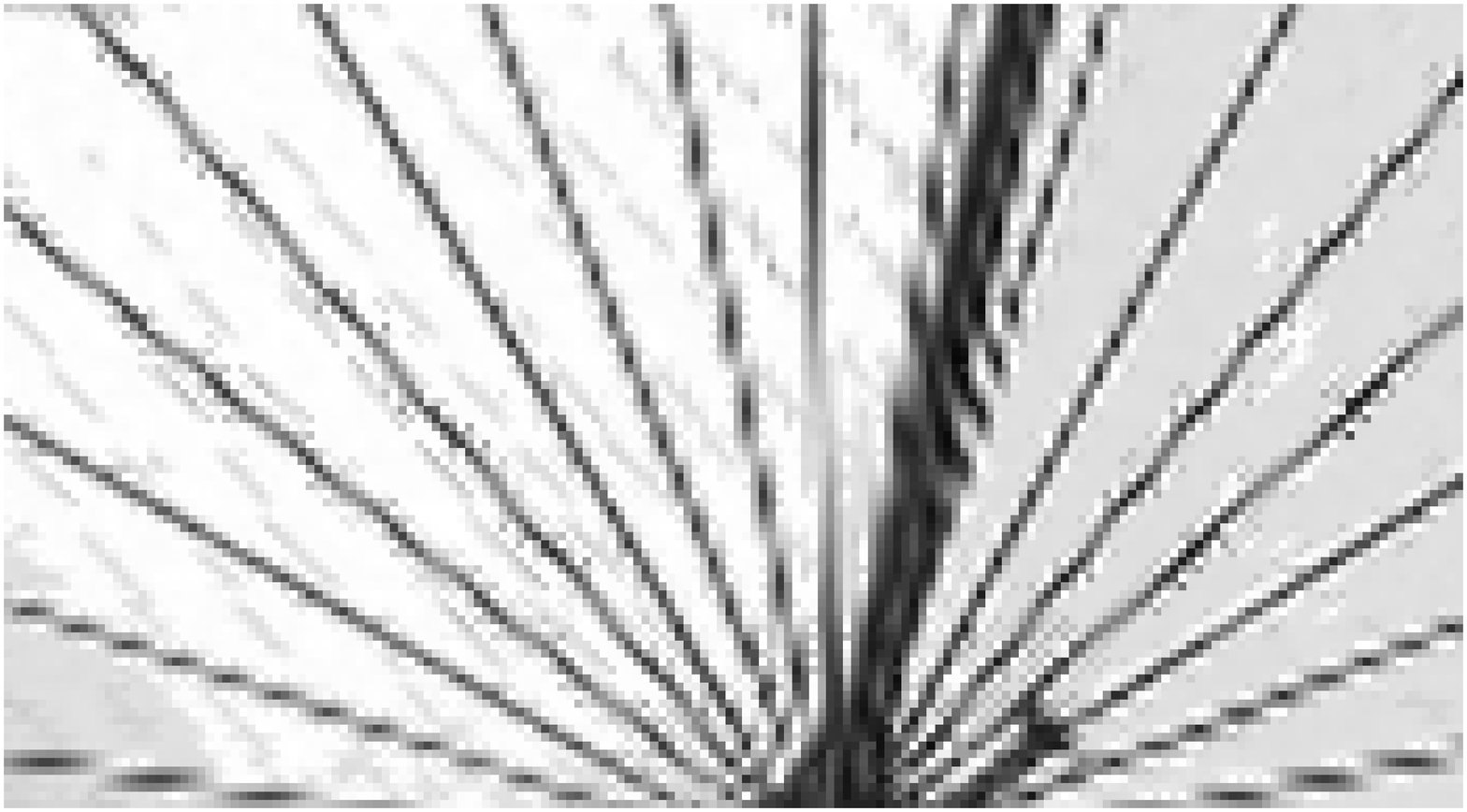}}
\centerline{(c)~NEDI}
\end{minipage} \\
\begin{minipage}[b]{0.485 \linewidth}
\centering
\centerline{\includegraphics[width=4.4 cm]{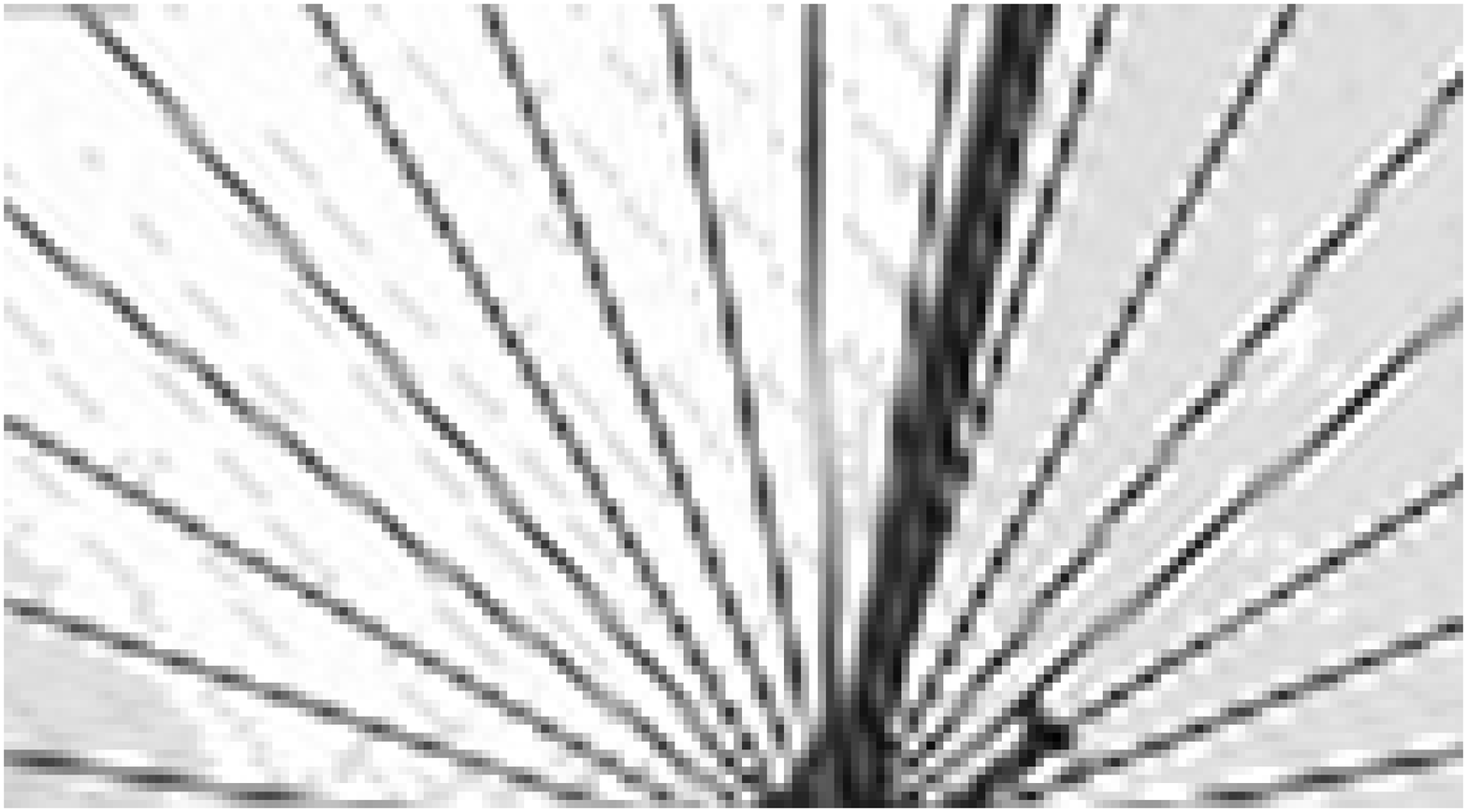}}
\centerline{(d)~SME}
\end{minipage}
\begin{minipage}[b]{0.485 \linewidth}
\centering
\centerline{\includegraphics[width=4.4 cm]{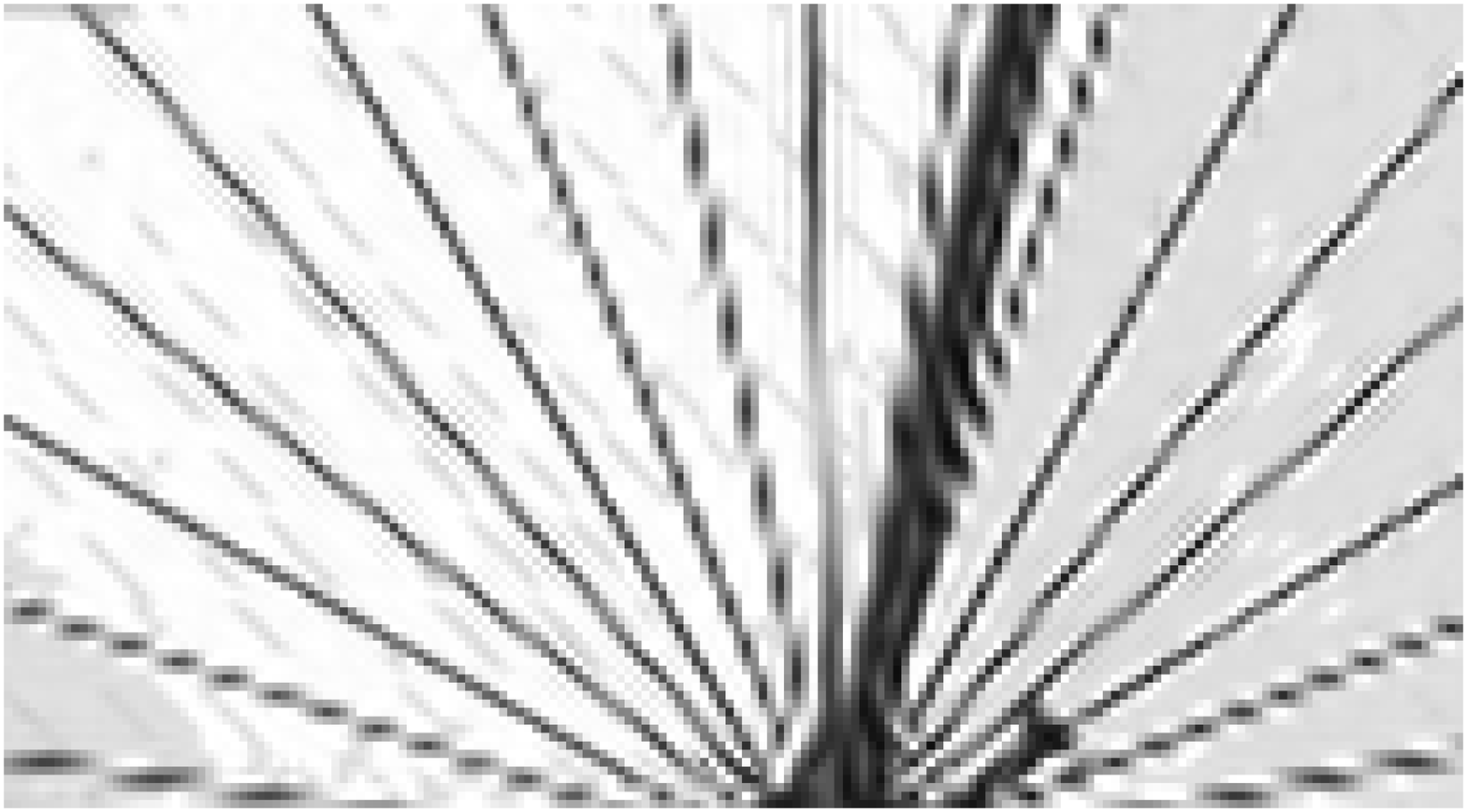}}
\centerline{(e)~SAI}
\end{minipage} \\
\begin{minipage}[b]{0.485 \linewidth}
\centering
\centerline{\includegraphics[width=4.4 cm]{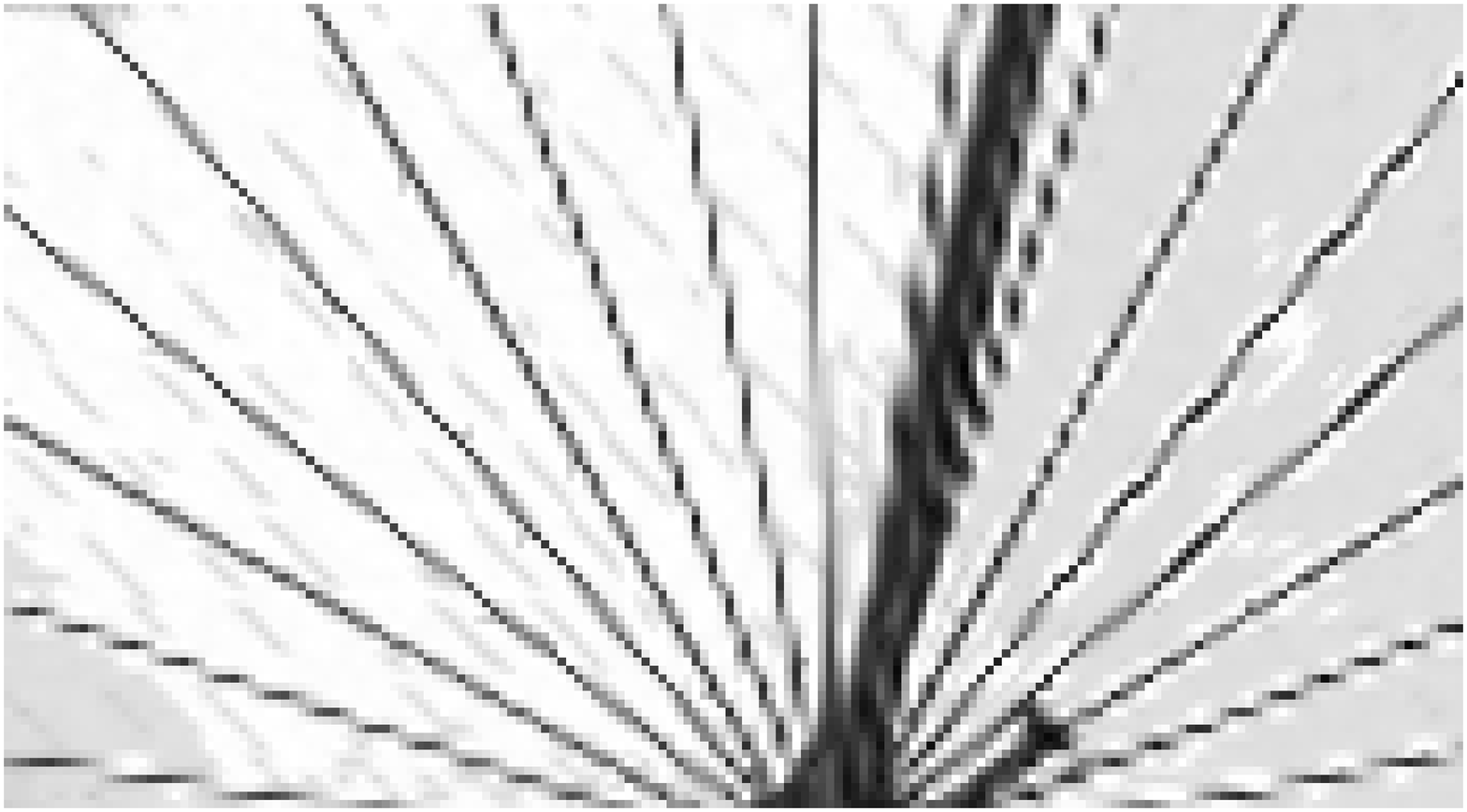}}
\centerline{(f)~RLLR}
\end{minipage}
\begin{minipage}[b]{0.485 \linewidth}
\centering
\centerline{\includegraphics[width=4.4 cm]{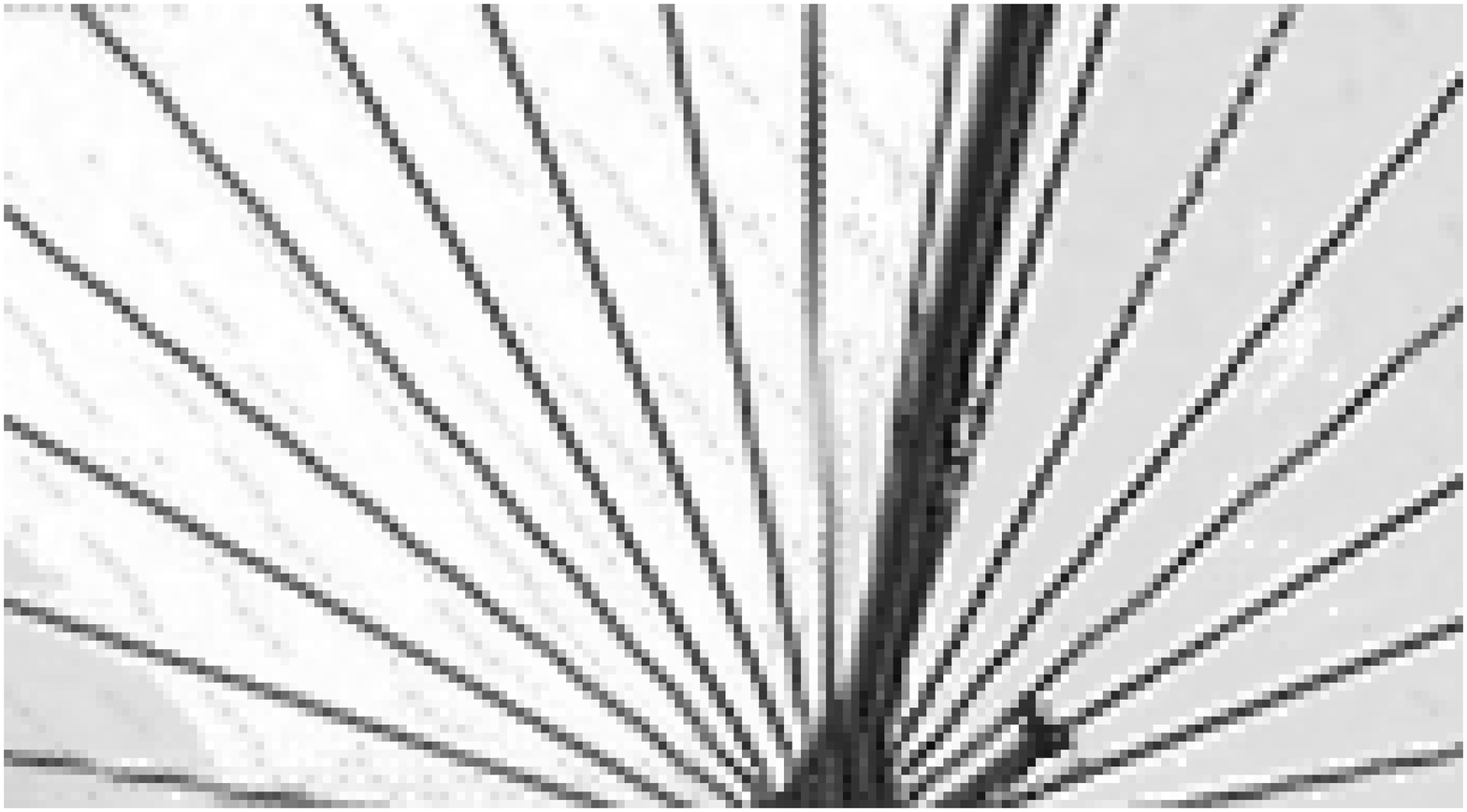}}
\centerline{(g)~MSIA}
\end{minipage} \\
\begin{minipage}[b]{0.485 \linewidth}
\centerline{\includegraphics[width=4.4 cm]{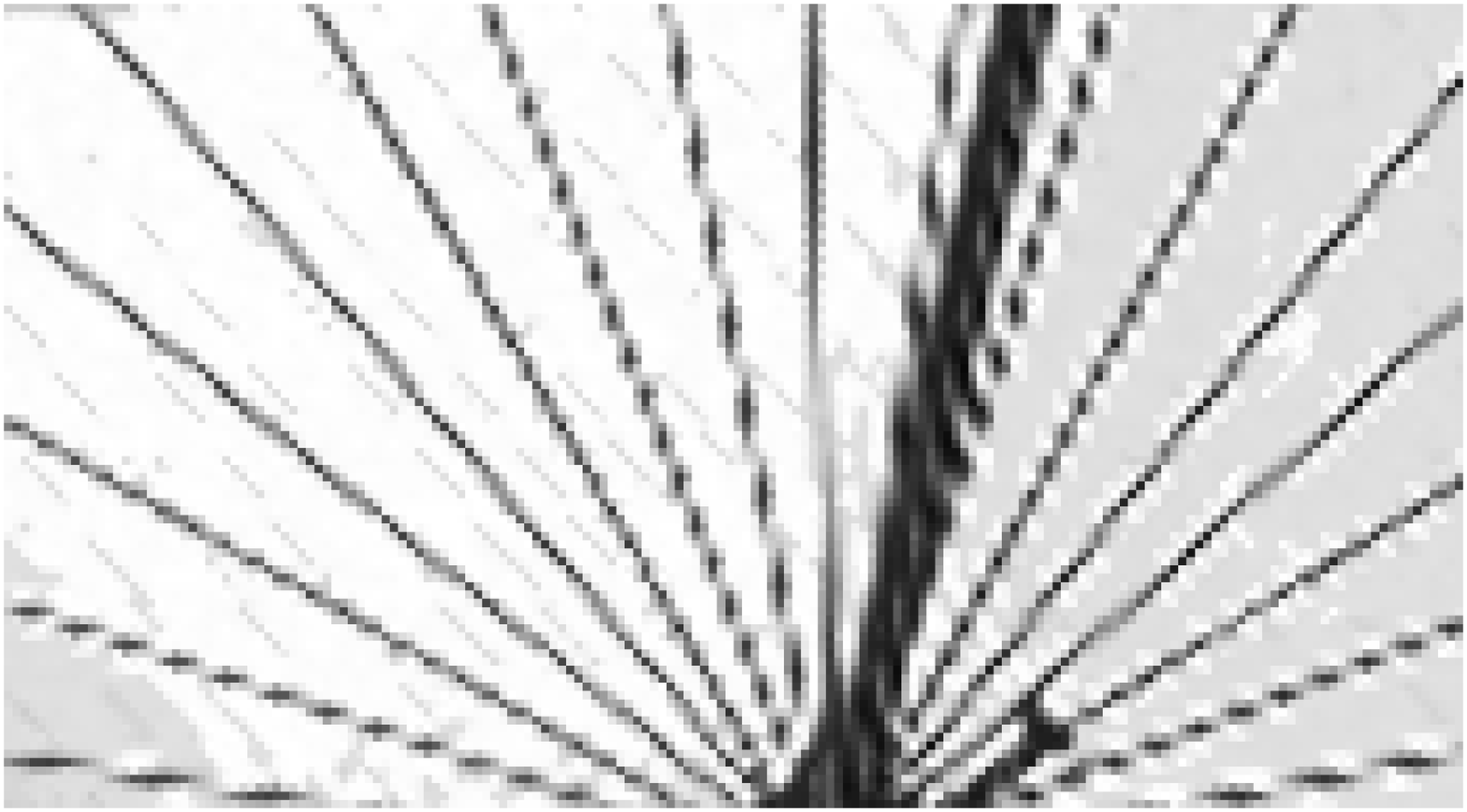}}
\centerline{(h)~NGSDG}
\end{minipage}
\centering
\begin{minipage}[b]{0.485 \linewidth}
\centerline{\includegraphics[width=4.4 cm]{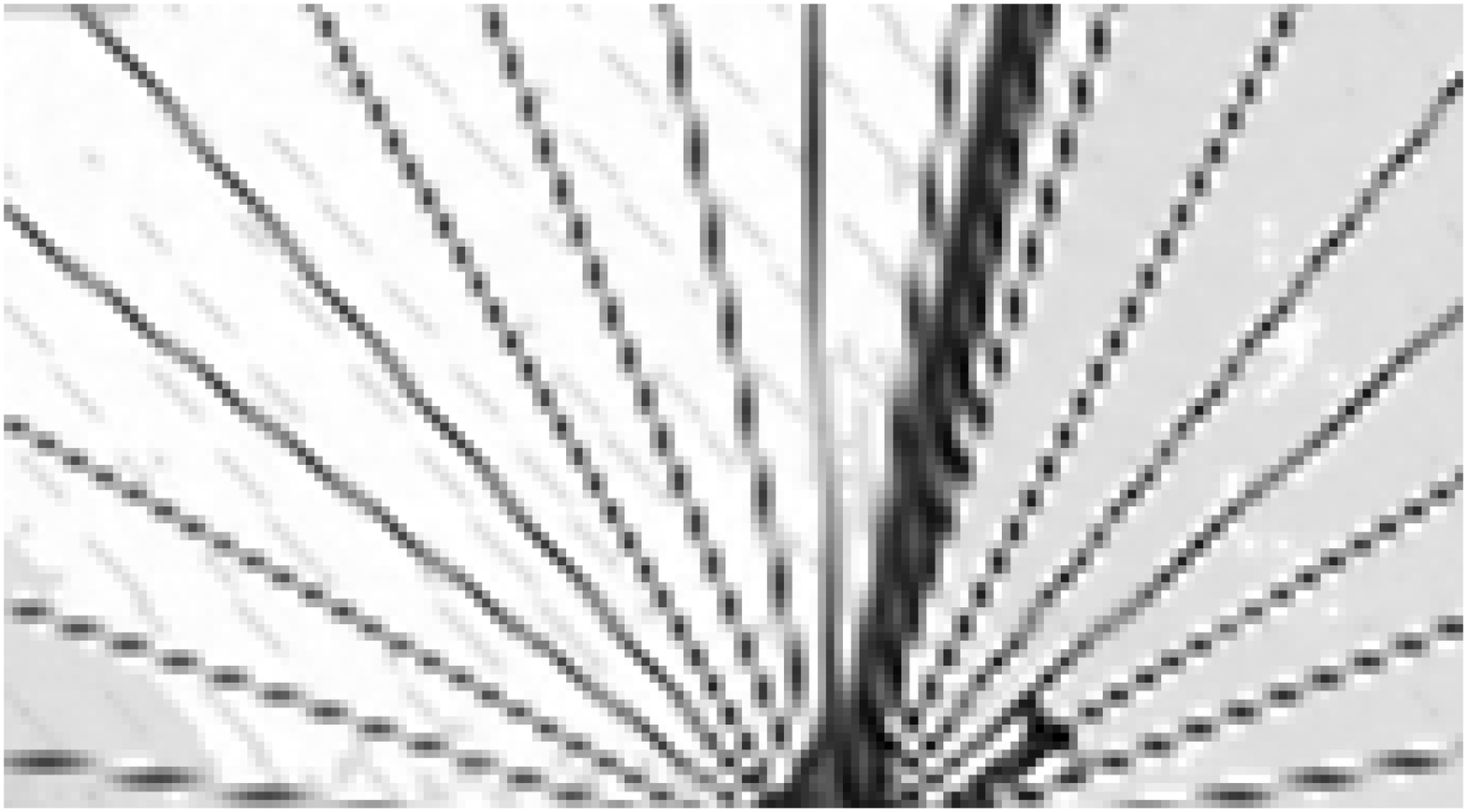}}
\centerline{(i)~NARM}
\end{minipage} \\
\begin{minipage}[b]{0.485 \linewidth}
\centerline{\includegraphics[width=4.4 cm]{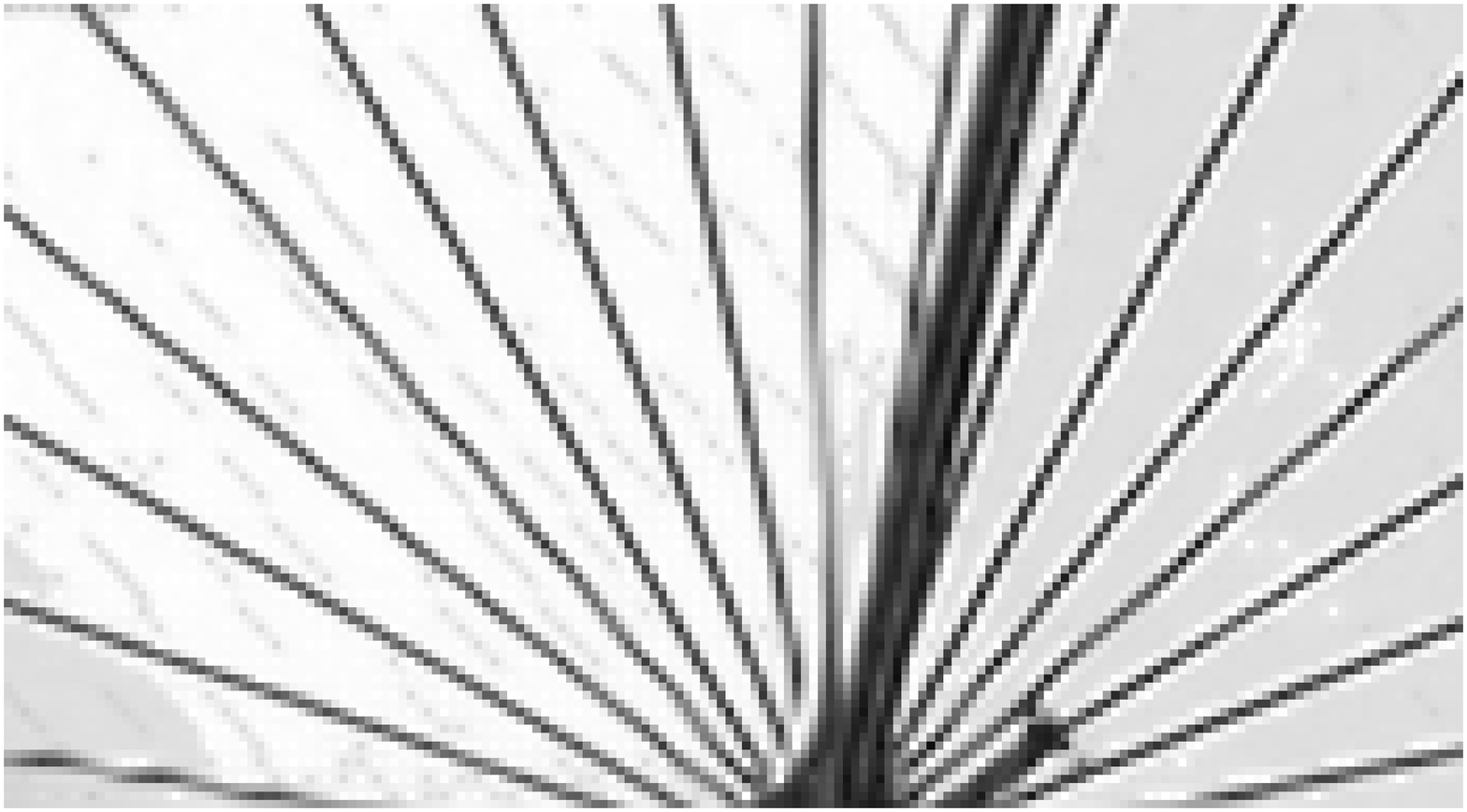}}
\centerline{(j)~ANSM}
\end{minipage}
\begin{minipage}[b]{0.485 \linewidth}
\centerline{\includegraphics[width=4.4 cm]{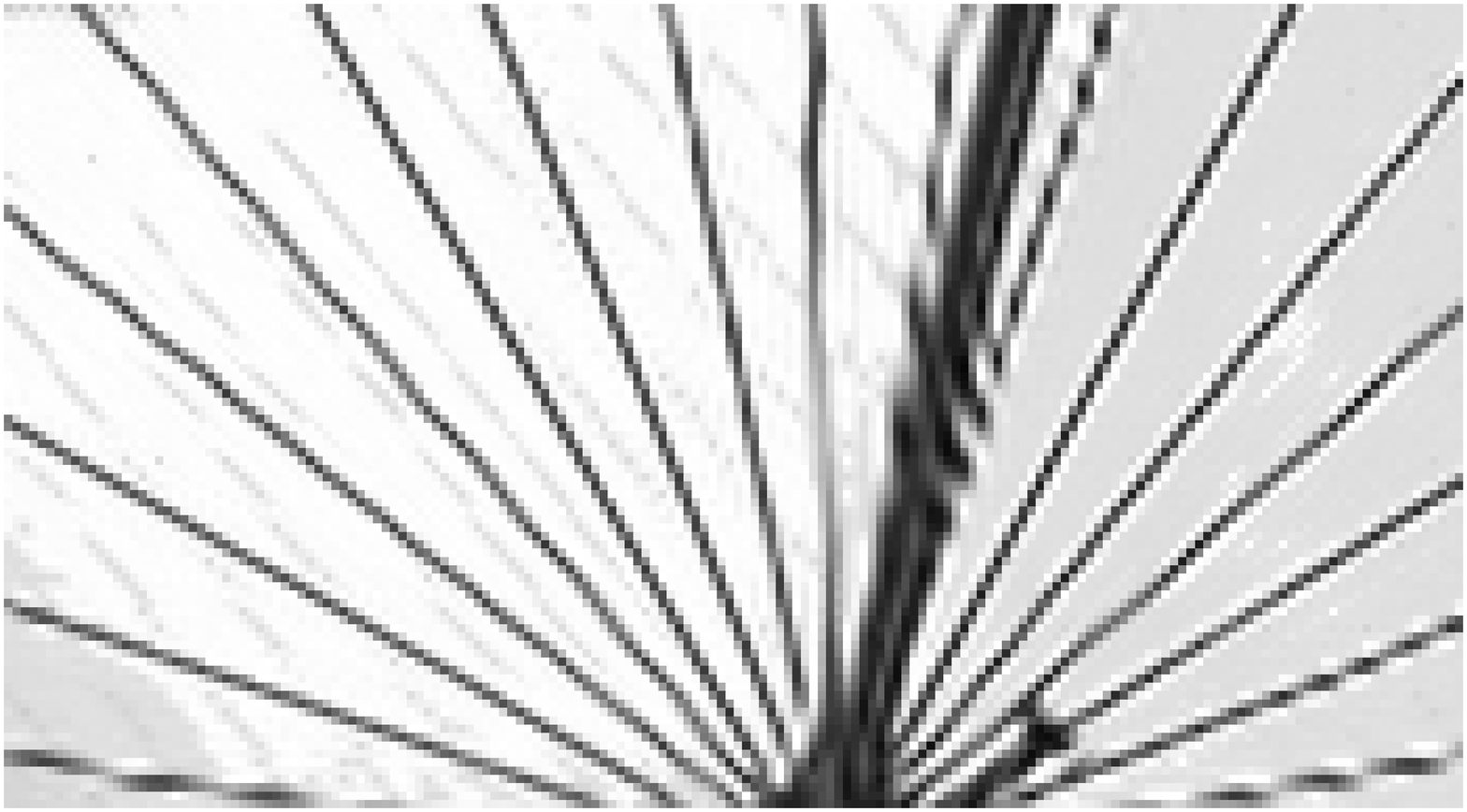}}
\centerline{(k)~NLPC}
\end{minipage} \\
\begin{minipage}[b]{0.485 \linewidth}
\centerline{\includegraphics[width=4.4 cm]{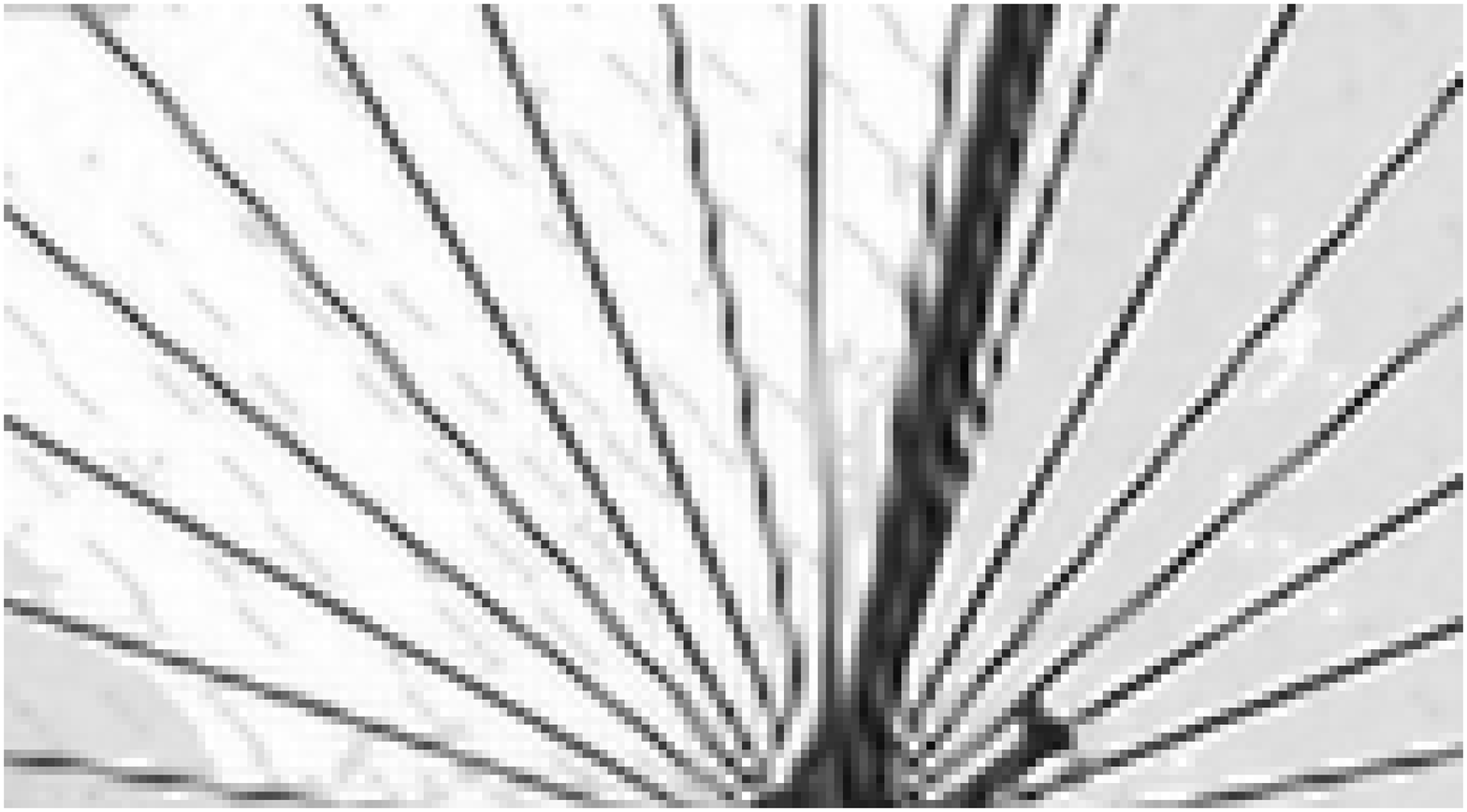}}
\centerline{(l)~FIRF}
\end{minipage}
\begin{minipage}[b]{0.485 \linewidth}
\centerline{\includegraphics[width=4.4 cm]{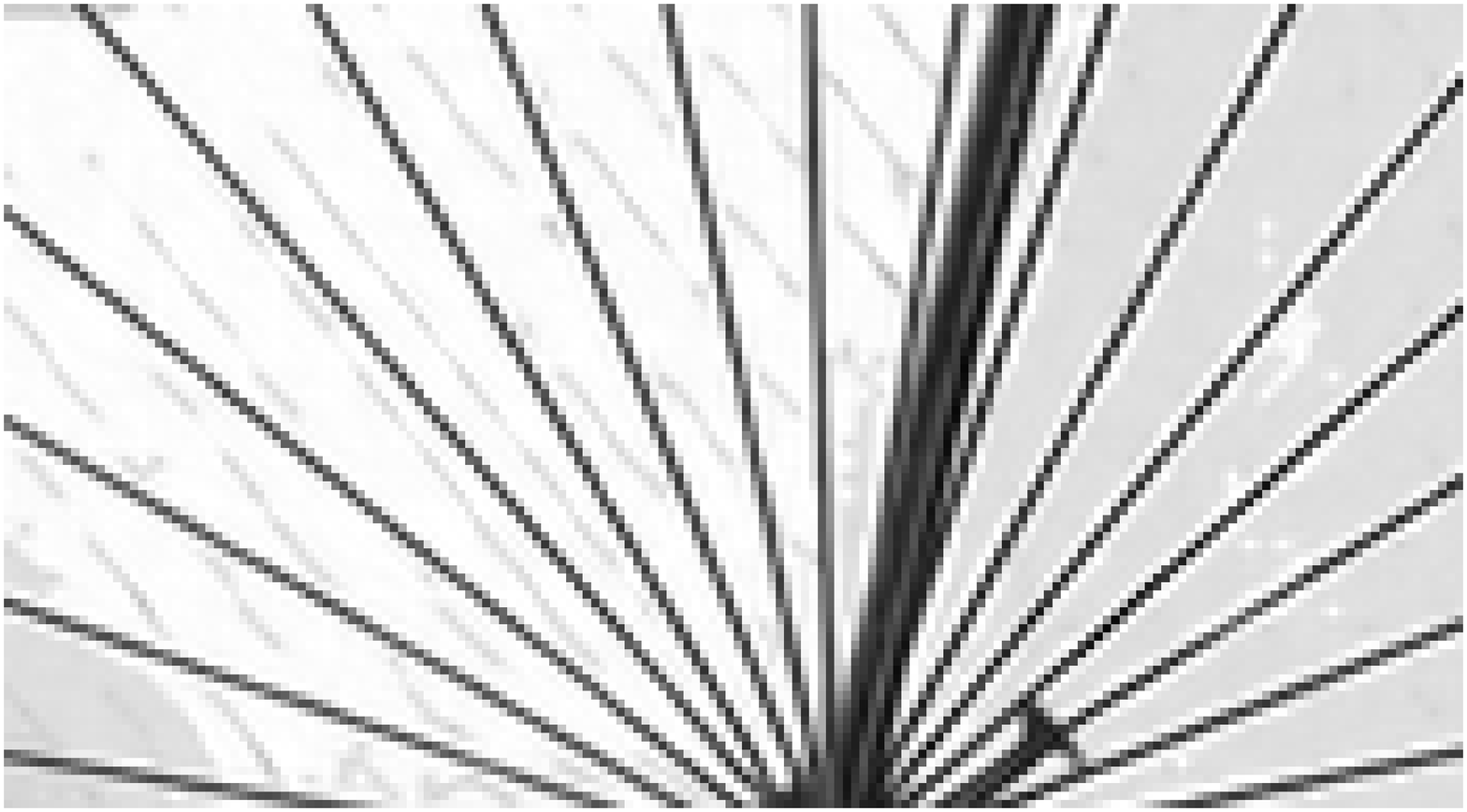}}
\centerline{(m)~MAIN}
\end{minipage} \\
\begin{minipage}[b]{0.485 \linewidth}
\centerline{\includegraphics[width=4.4 cm]{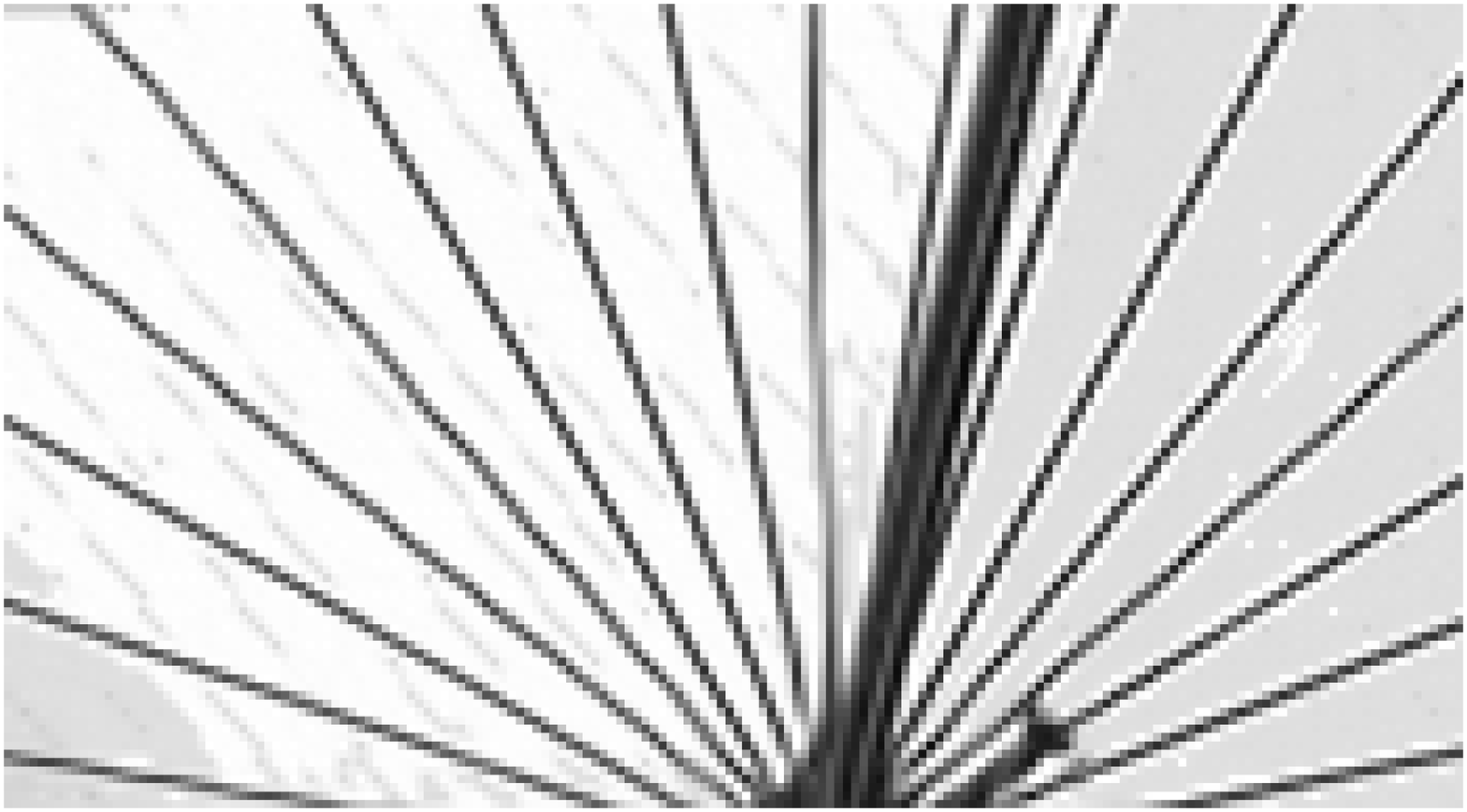}}
\centerline{(n)~MISTER}
\end{minipage}
\begin{minipage}[b]{0.485 \linewidth}
\centerline{\includegraphics[width=4.4 cm]{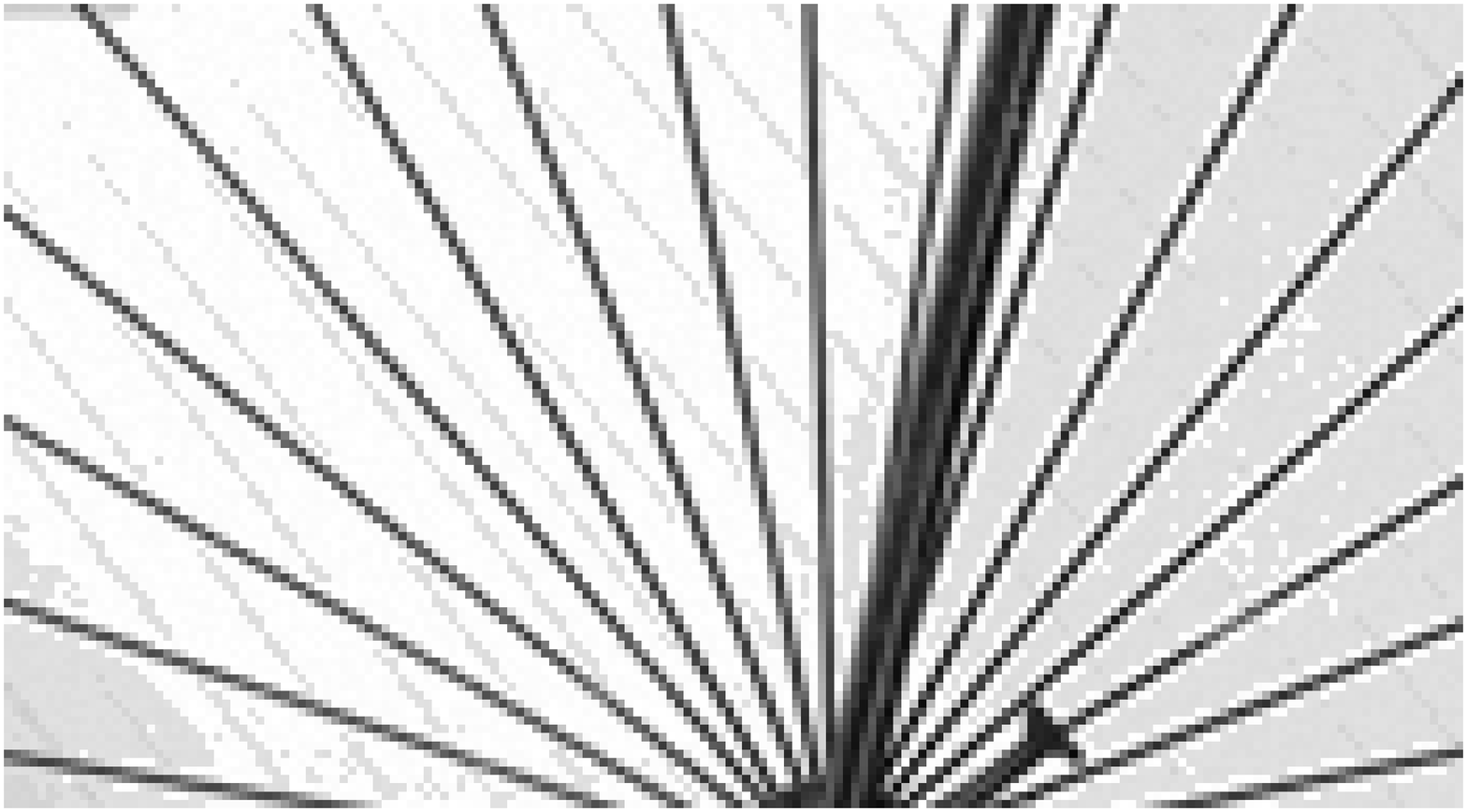}}
\centerline{(o)~Ground Truth}
\end{minipage}
\caption{The zoom-in comparison of crop from \textit{Wheel} in the interpolation task by a factor of $2$.}
\label{fig:edge_wheel}
\end{figure}

Fig.~\ref{fig:edge_motorbike},~\ref{fig:texture_hats},~\ref{fig:edge_wheel} demonstrate visual comparisons between the above algorithms' results. In Fig.~\ref{fig:edge_motorbike}(n), MISTER's edges along upper-left to bottom-right orientation preserve both smoothness along the contour direction and sharpness across the profile direction, without noticeable artifacts. Among the interpolated edges from model-based methods, the edges from ANSM have the most similar visual quality to ours, but with noticeable jaggies along the contour direction. In Fig.~\ref{fig:texture_hats}(n), MISTER's textures preserve the correct directions (from upper-left to bottom-right) as the ground-truth image. However, the textures from the other model-based algorithms generates distinctly different directions from the ground-truth. In Fig.~\ref{fig:edge_wheel}(n), MISTER's edges preserve the contour continuity of all the spokes along various directions. Among the interpolated edges from all the model-based algorithms, the edges from ANSM and MSIA have the most similar visual quality to ours. Yet, MSIA's spokes exhibit zipper artifacts and jaggies and ANSM's spokes exhibit mild but noticeable jaggies.

It is interesting that although FIRF achieves higher $\mathrm{PSNR}$ than recent model-based single image interpolation algorithms, strong artifacts are noticeable from Fig.~\ref{fig:edge_motorbike}(l),~\ref{fig:texture_hats}(l),~\ref{fig:edge_wheel}(l). For example, the contours along upper-left to bottom-right directions are not smooth in Fig.~\ref{fig:edge_motorbike}(l) and Fig.~\ref{fig:edge_wheel}(l). The directions of the textures in Fig.~\ref{fig:texture_hats}(l) are hardly noticeable.

\subsection{Interpolation by a factor of $3$}
For the task of interpolation by a factor of $3$, due to the availability of the source code, we compare MISTER only with bicubic interpolation, NARM, ANSM, NLPC and MAIN in Table~\ref{tab:test_3x}. Table~\ref{tab:test_3x} demonstrates that among model-based single image interpolation algorithms, MISTER achieves the highest average $\mathrm{PSNR}$ and the highest $\mathrm{PSNR}$ in $25$ out of $27$ images. Specifically, MISTER outperforms bicubic interpolation, NARM, ANSM and NLPC with an average $\mathrm{PSNR}$ improvement of $1.74$ dB, $0.96$ dB, $0.65$ dB and $0.54$ dB, respectively. Compared with the deep-learning based MAIN, MISTER’s average $\mathrm{PSNR}$ is $0.92$ dB lower.

\begin{table}[htbp]
\caption{Comparison of $\mathrm{PSNR}$s (in decibels) of the results between MISTER and~\cite{dong2013sparse,romano2014single,sun2016image,ji2020image} in the task of interpolation by a factor of $3$. The result with the highest $\mathrm{PSNR}$ among model-based approaches is highlighted in dark bold and the result with the highest $\mathrm{PSNR}$ is highlighted in red bold.}
\centering
\resizebox{0.5\textwidth}{!}{%
\begin{tabular}{||c|c|c|c|c|c|c||}
 \hline
\textbf{Images}  & \textbf{Bicubic} & \textbf{NARM} & \textbf{ANSM}  & \textbf{NLPC}  & \textbf{MAIN} & \textbf{MISTER} \\
\hline
\textbf{AVERAGE}  & 25.52  & 26.30  & 26.61  & 26.72 & $\textcolor{red}{\bm{28.18}}$ & $\bm{27.26}$ \\     
 \hline
\end{tabular}}
\label{tab:test_3x}
\end{table}

\begin{figure*}[tb]
\centering
\begin{minipage}[b]{0.07 \linewidth}
\centerline{\includegraphics[width=1.2 cm]{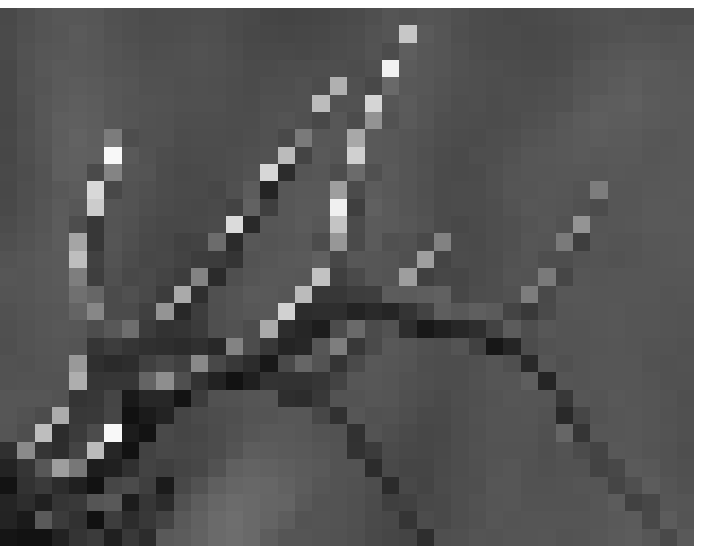}}
\centerline{(a)~LR}
\end{minipage}
\begin{minipage}[b]{0.21 \linewidth}
\centerline{\includegraphics[width=3.6 cm]{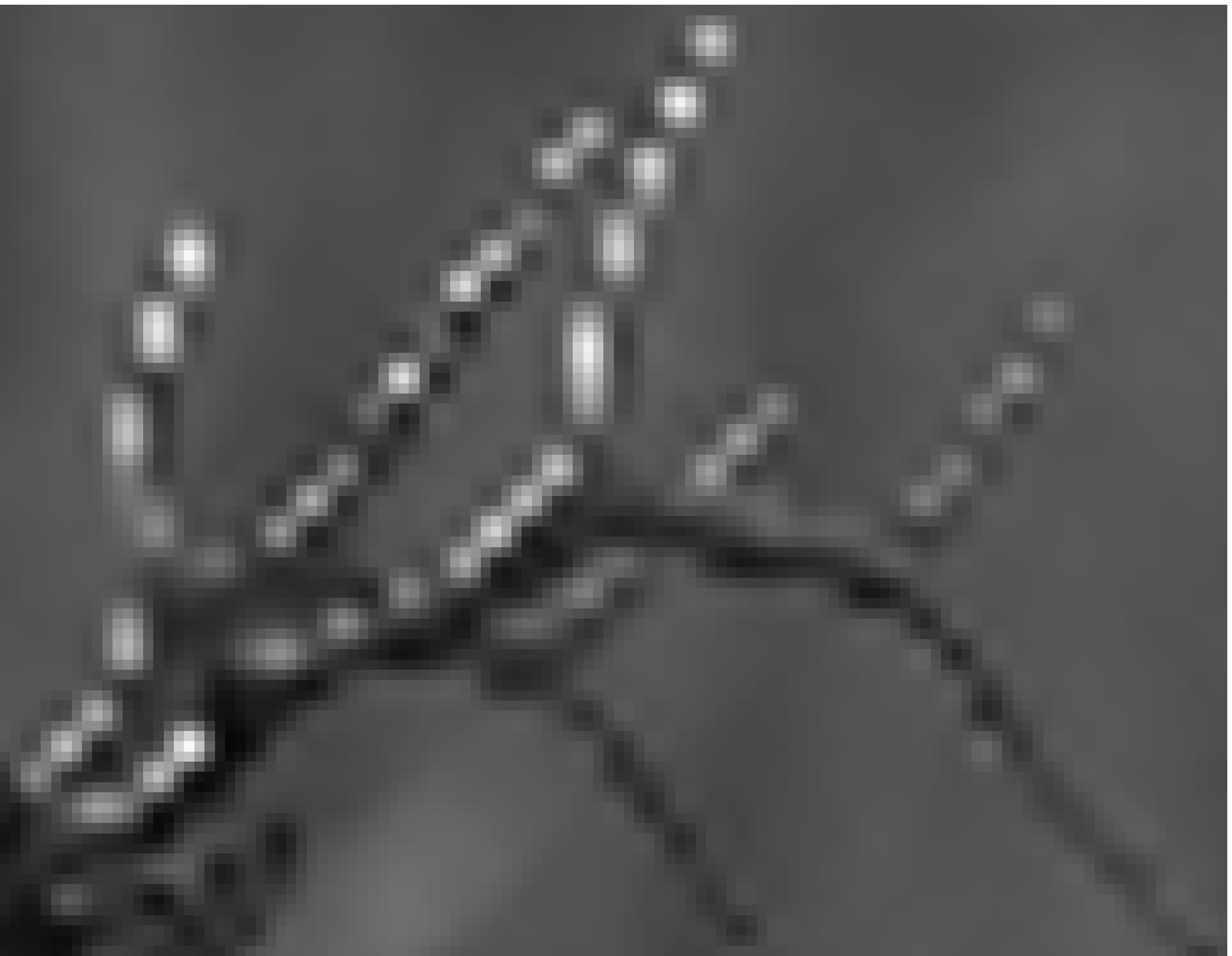}}
\centerline{(b)~Bicubic}
\end{minipage}
\begin{minipage}[b]{0.21 \linewidth}
\centerline{\includegraphics[width=3.6 cm]{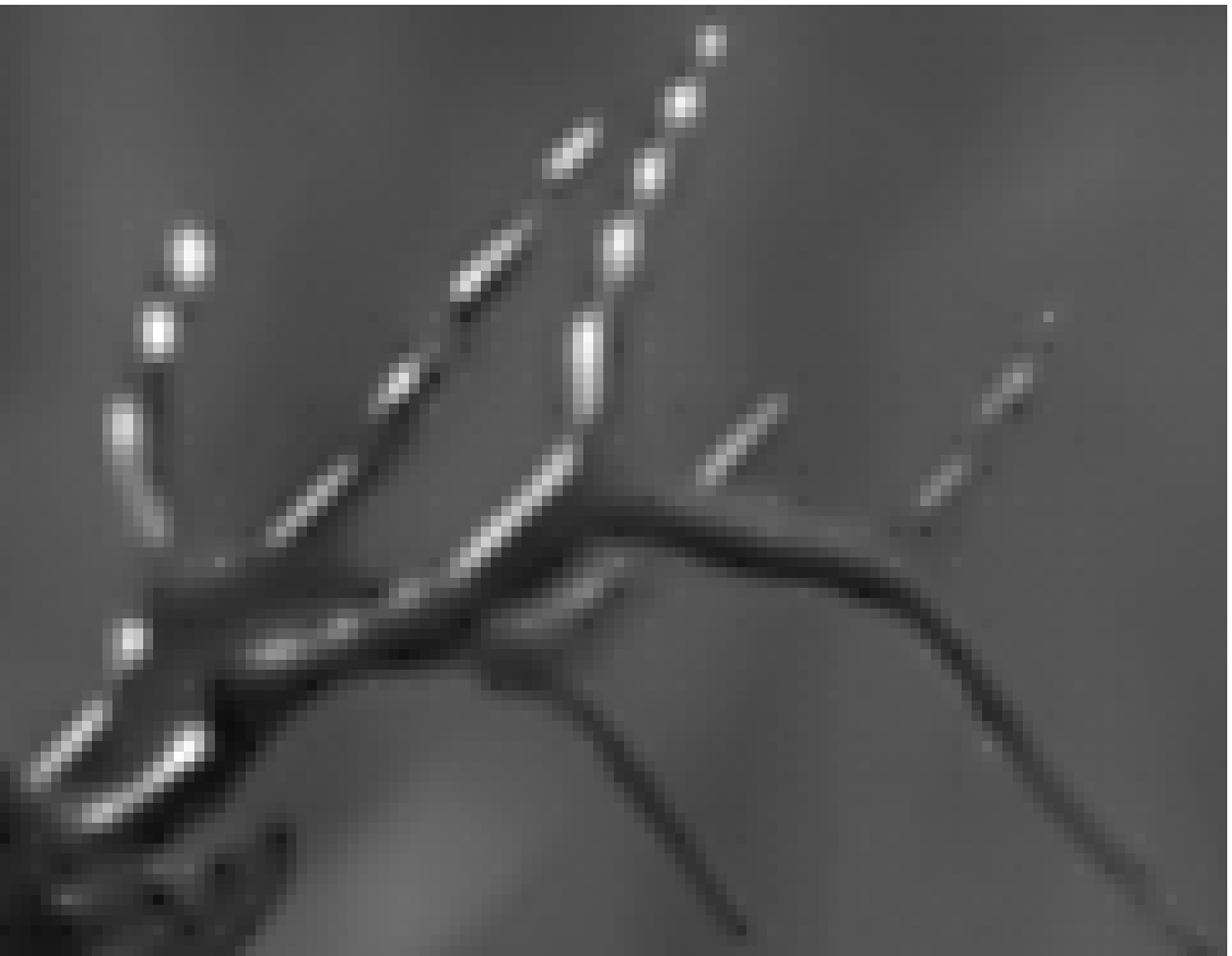}}
\centerline{(c)~NARM}
\end{minipage}
\begin{minipage}[b]{0.21 \linewidth}
\centerline{\includegraphics[width=3.6 cm]{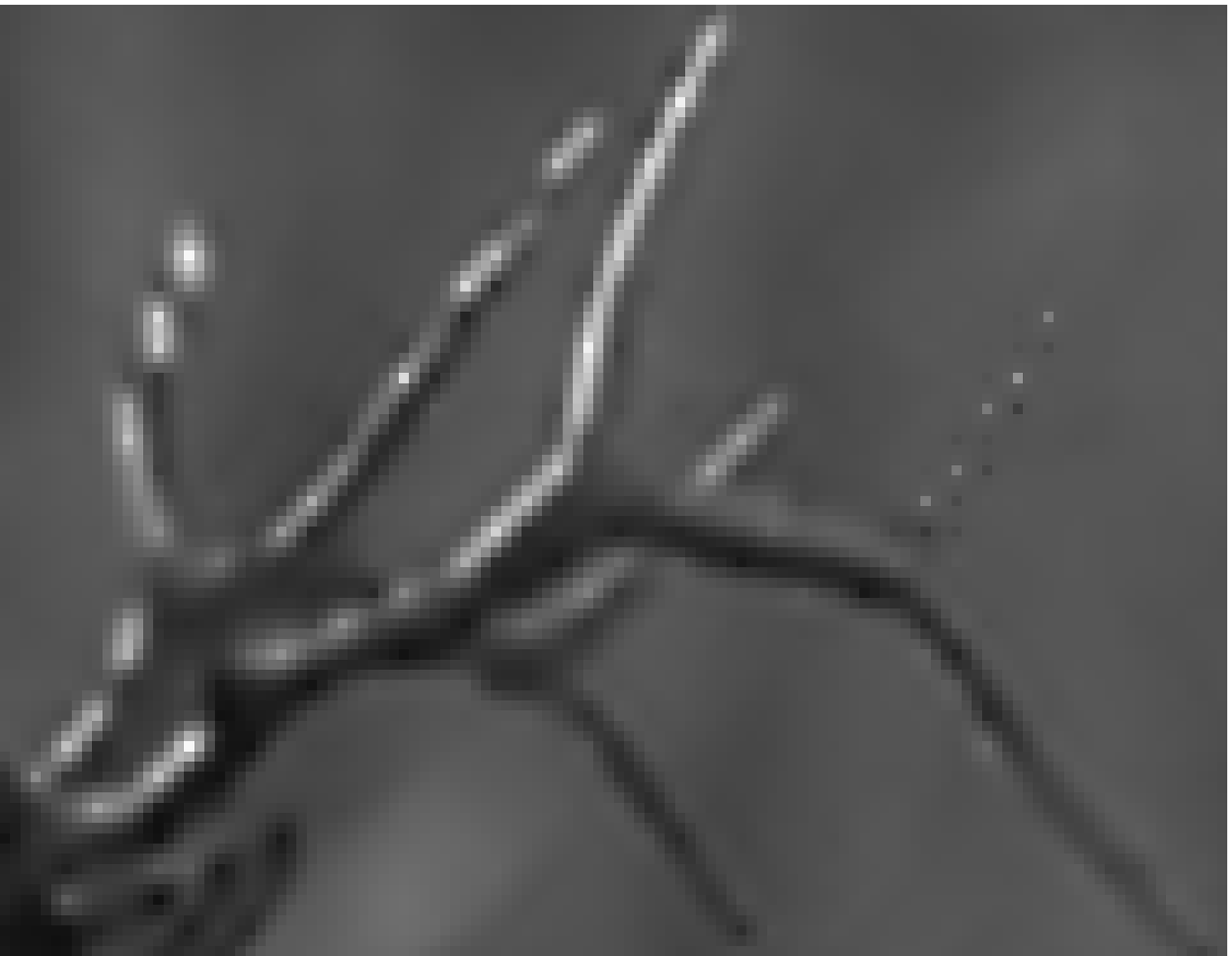}}
\centerline{(d)~ANSM}
\end{minipage}
\begin{minipage}[b]{0.21 \linewidth}
\centerline{\includegraphics[width=3.6 cm]{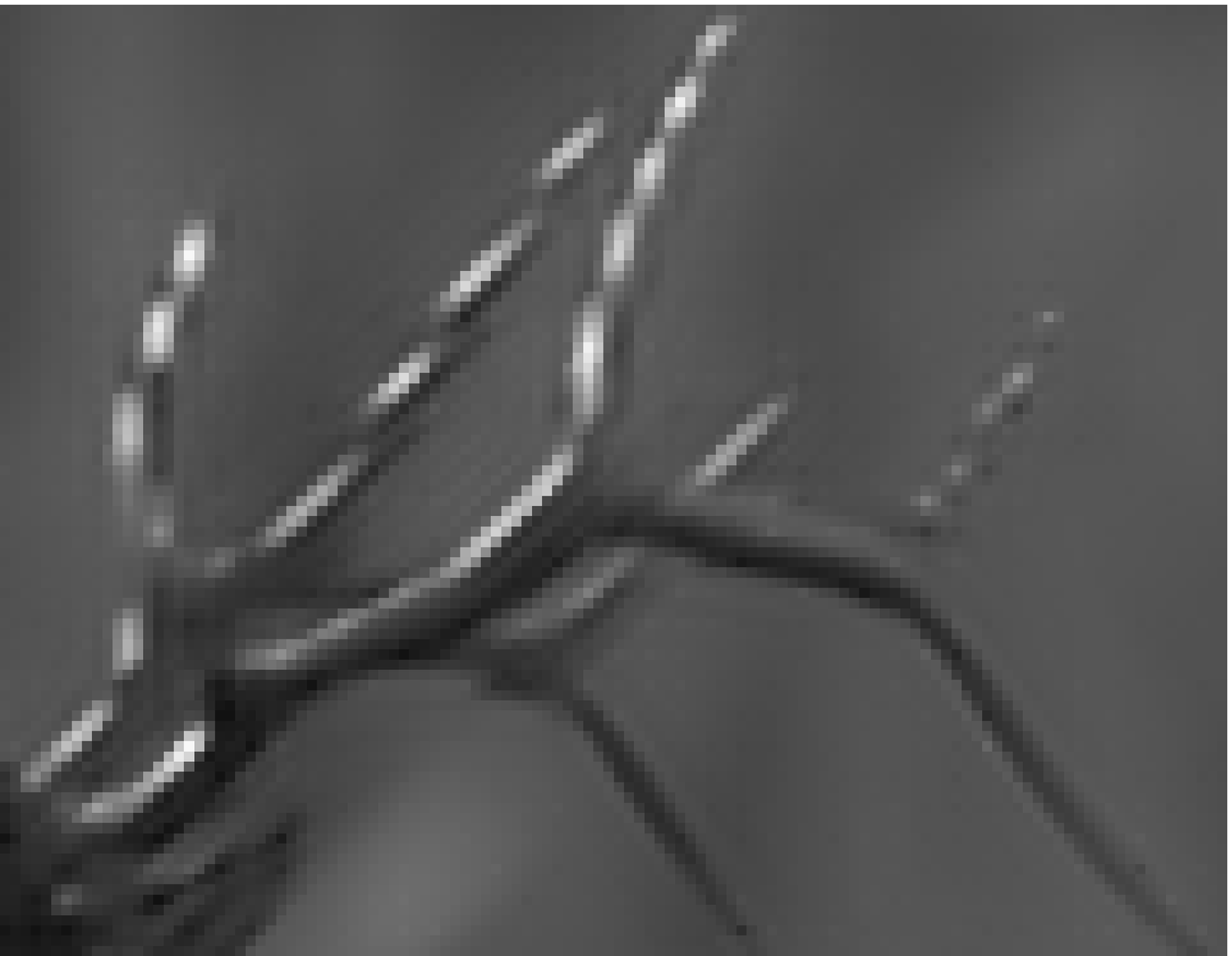}}
\centerline{(e)~NLPC}
\end{minipage} \\
\begin{minipage}[b]{0.21 \linewidth}
\centerline{\includegraphics[width=3.6 cm]{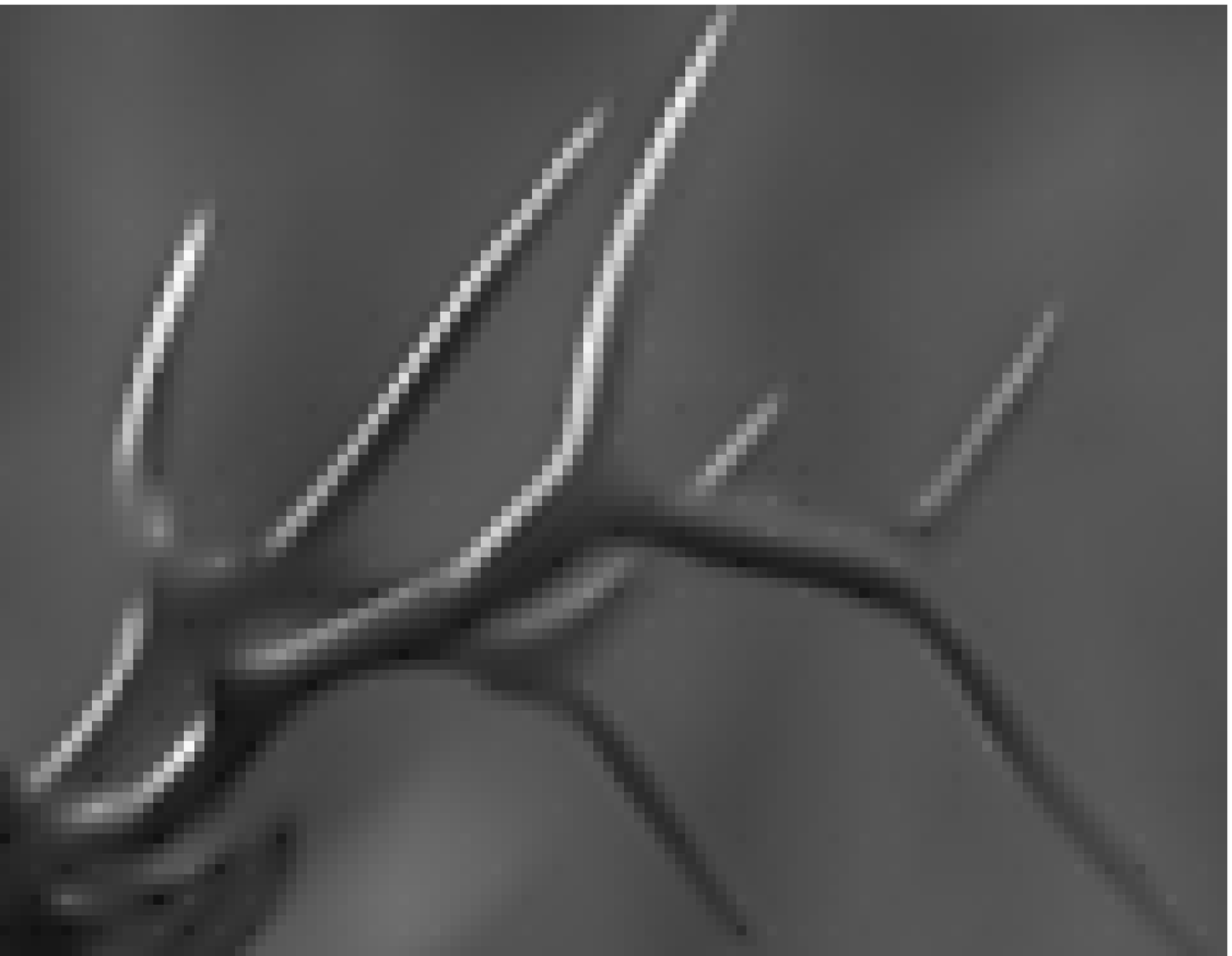}}
\centerline{(f)~MISTER}
\end{minipage}
\begin{minipage}[b]{0.21 \linewidth}
\centerline{\includegraphics[width=3.6 cm]{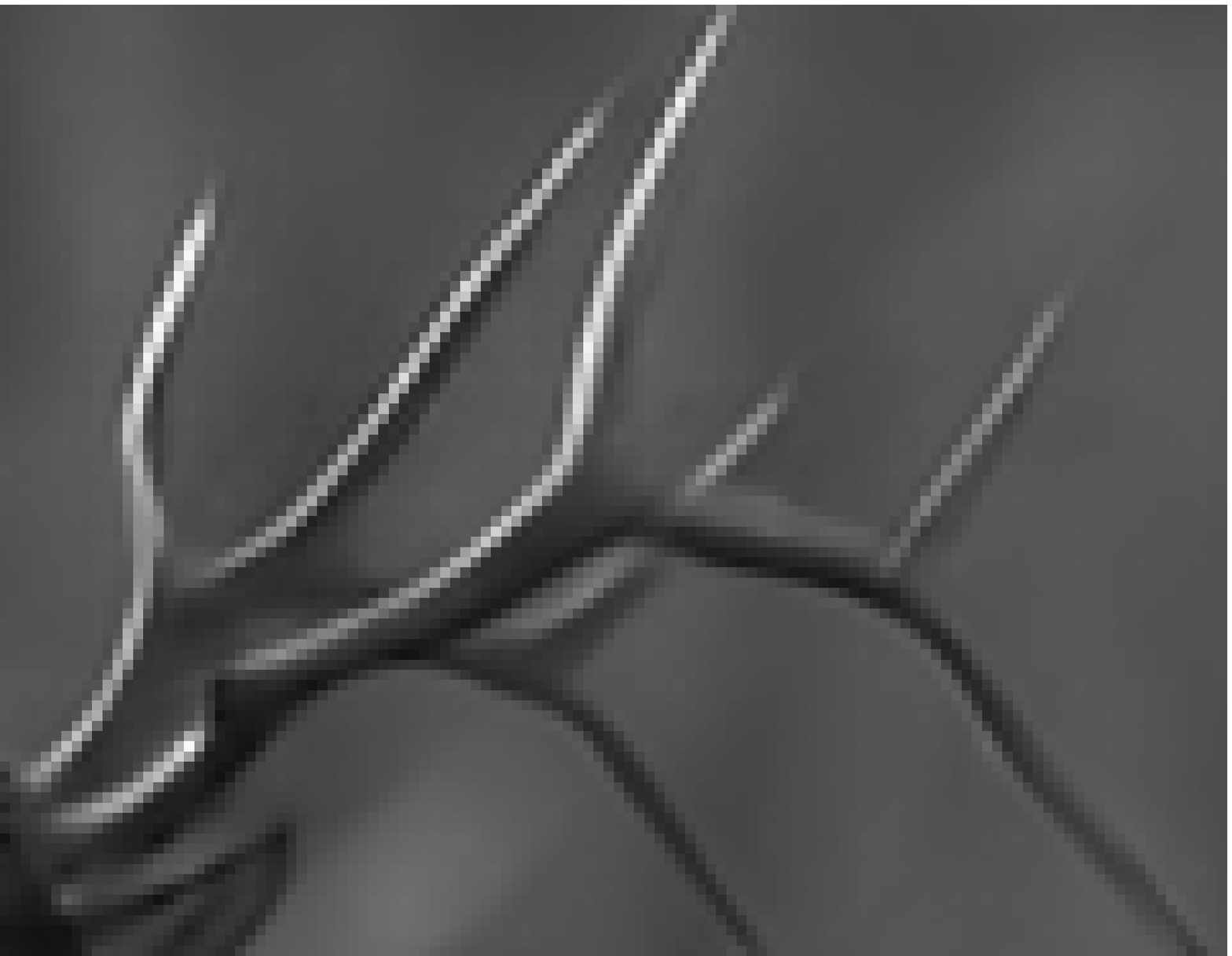}}
\centerline{(g)~MAIN}
\end{minipage}
\begin{minipage}[b]{0.21 \linewidth}
\centerline{\includegraphics[width=3.6 cm]{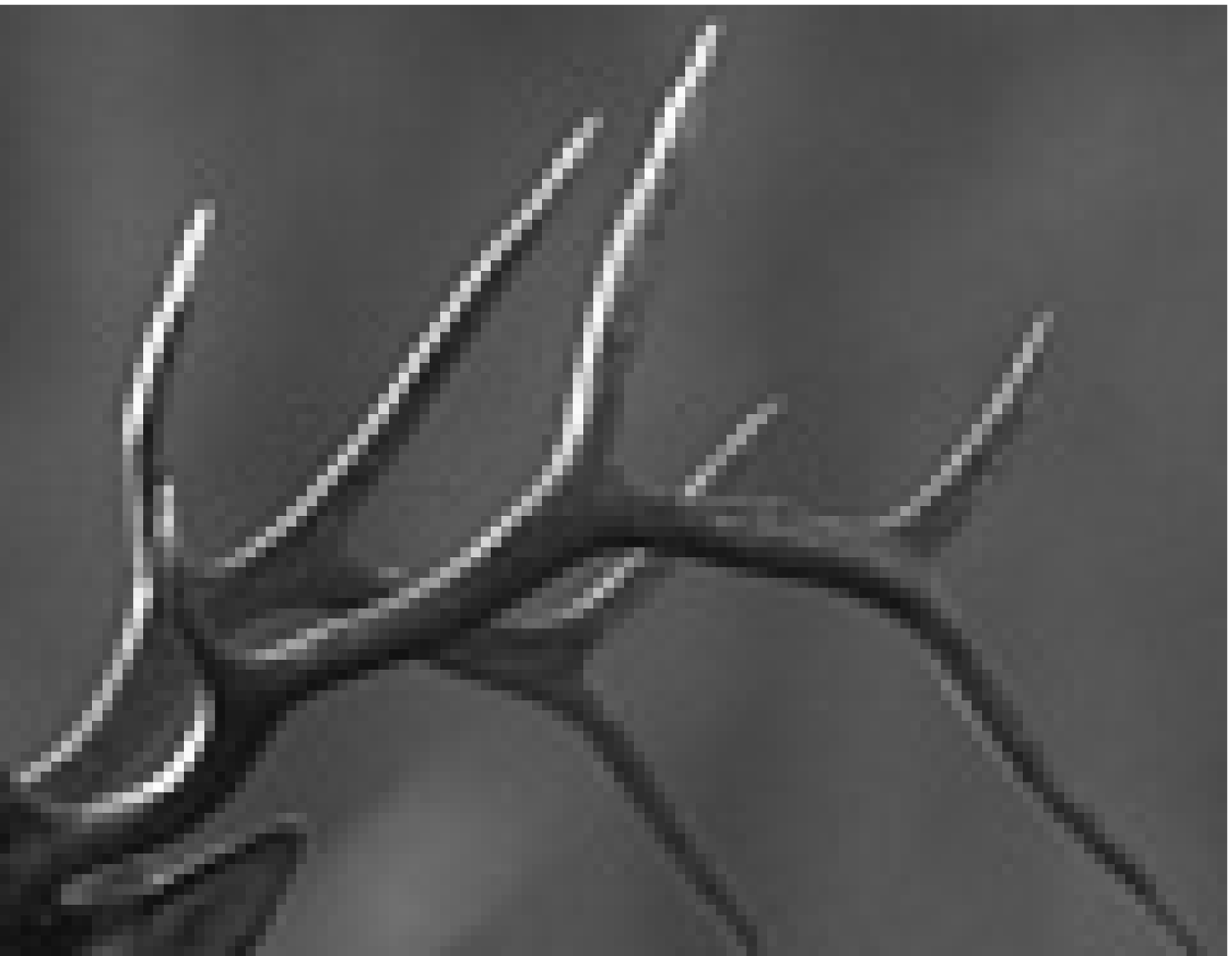}}
\centerline{(h)~Ground Truth}
\end{minipage}
\caption{The zoom-in comparison of crop from~\textit{Elk} in the interpolation task by a factor of $3$.}
\label{fig:edge_elk_3x}
\end{figure*}

\begin{figure*}[tb]
\centering
\begin{minipage}[b]{0.06 \linewidth}
\centerline{\includegraphics[width=1.1 cm]{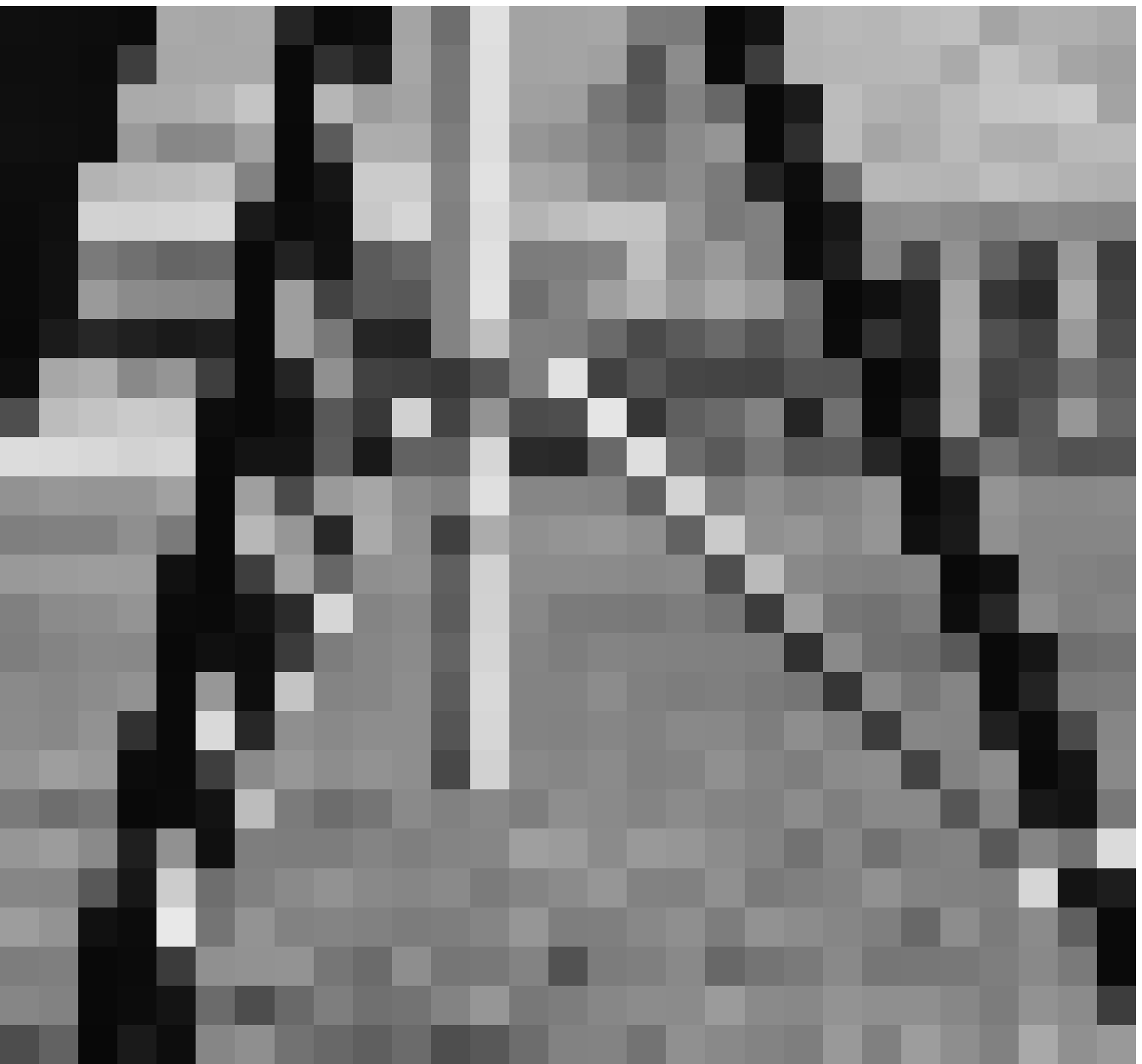}}
\centerline{(a)~LR}
\end{minipage}
\begin{minipage}[b]{0.18 \linewidth}
\centerline{\includegraphics[width=3.3 cm]{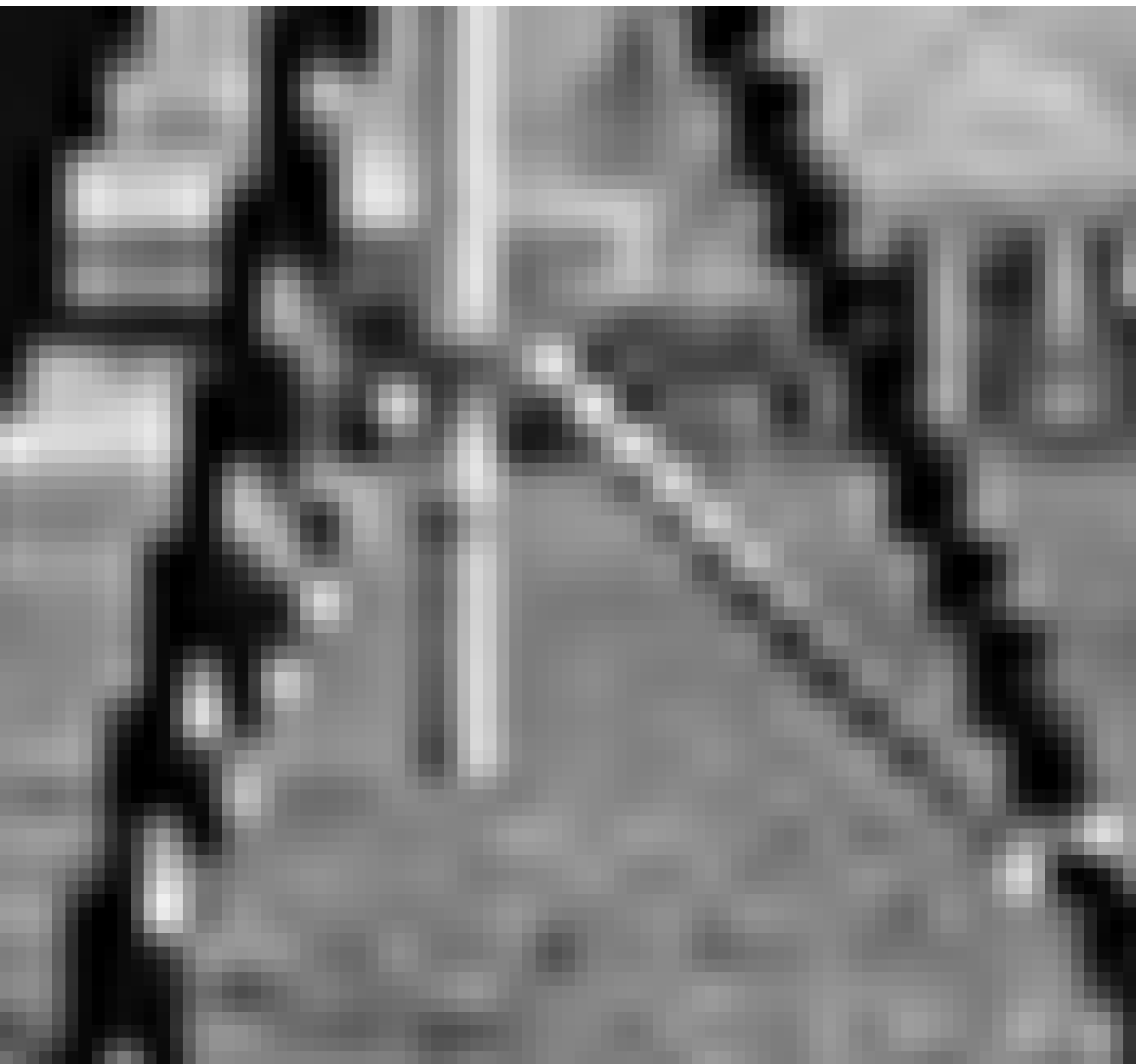}}
\centerline{(b)~Bicubic}
\end{minipage}
\begin{minipage}[b]{0.18 \linewidth}
\centerline{\includegraphics[width=3.3 cm]{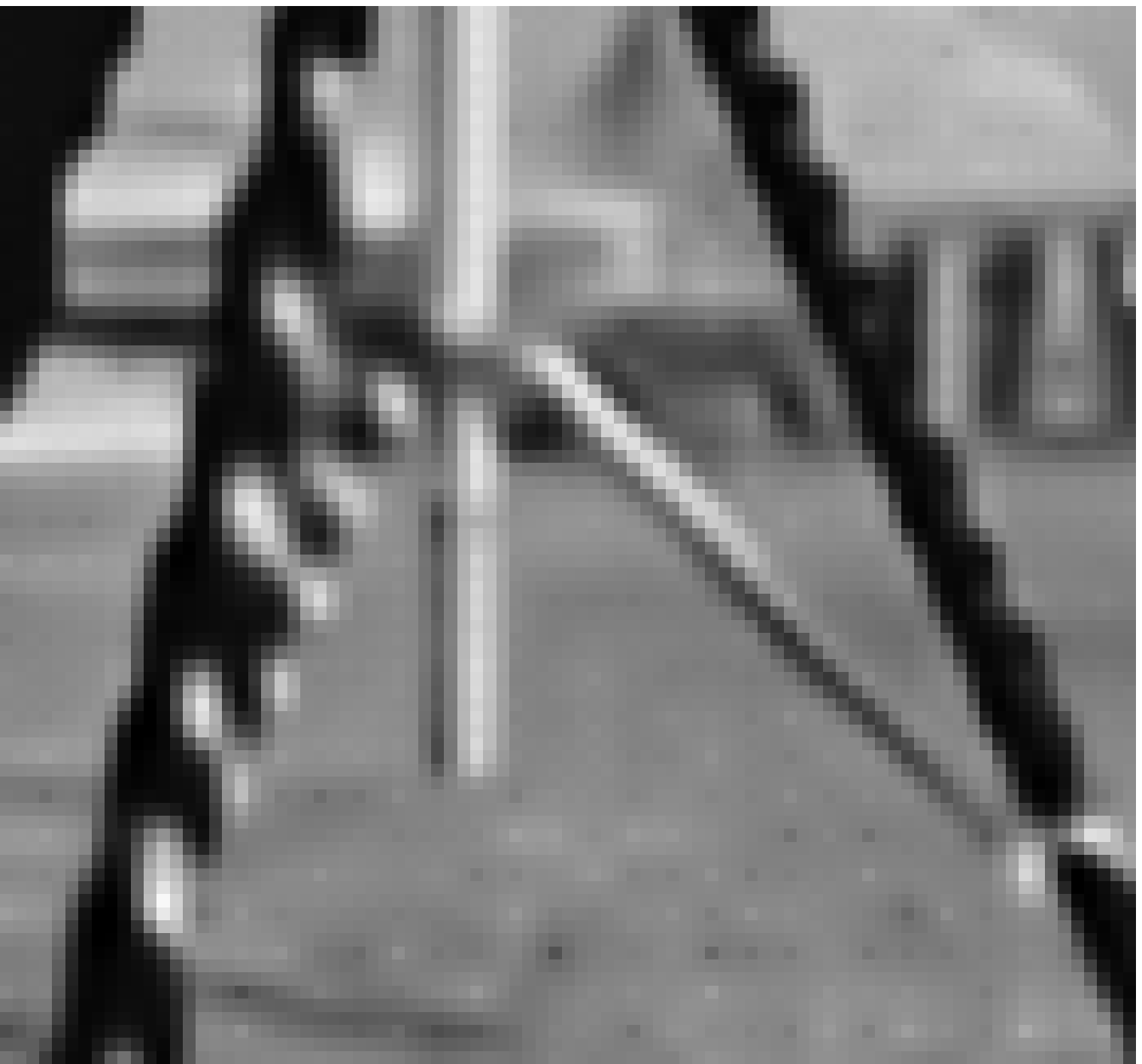}}
\centerline{(c)~NARM}
\end{minipage}
\begin{minipage}[b]{0.18 \linewidth}
\centerline{\includegraphics[width=3.3 cm]{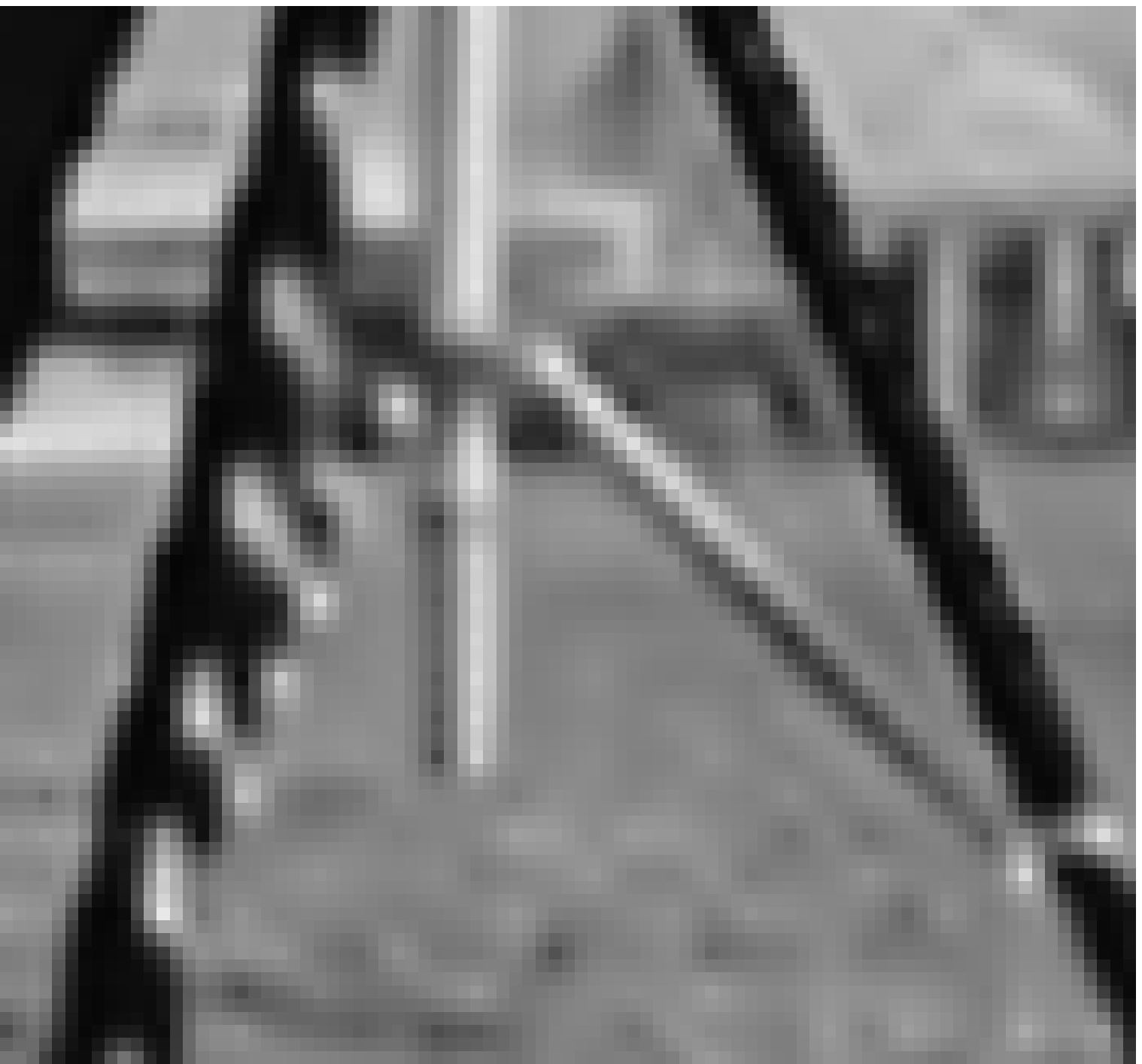}}
\centerline{(d)~ANSM}
\end{minipage}
\begin{minipage}[b]{0.18 \linewidth}
\centerline{\includegraphics[width=3.3 cm]{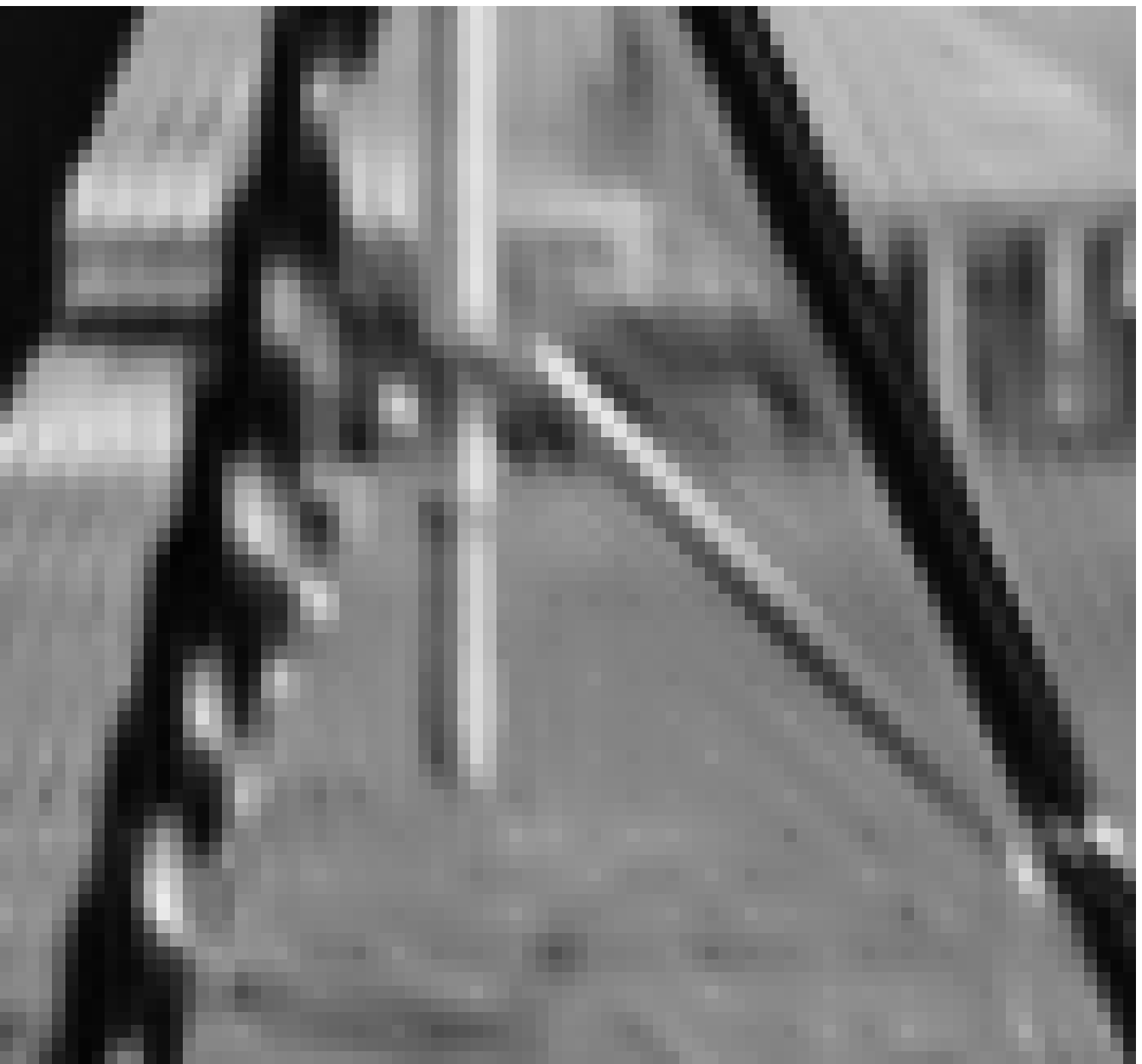}}
\centerline{(e)~NLPC}
\end{minipage} \\
\begin{minipage}[b]{0.18 \linewidth}
\centerline{\includegraphics[width=3.3 cm]{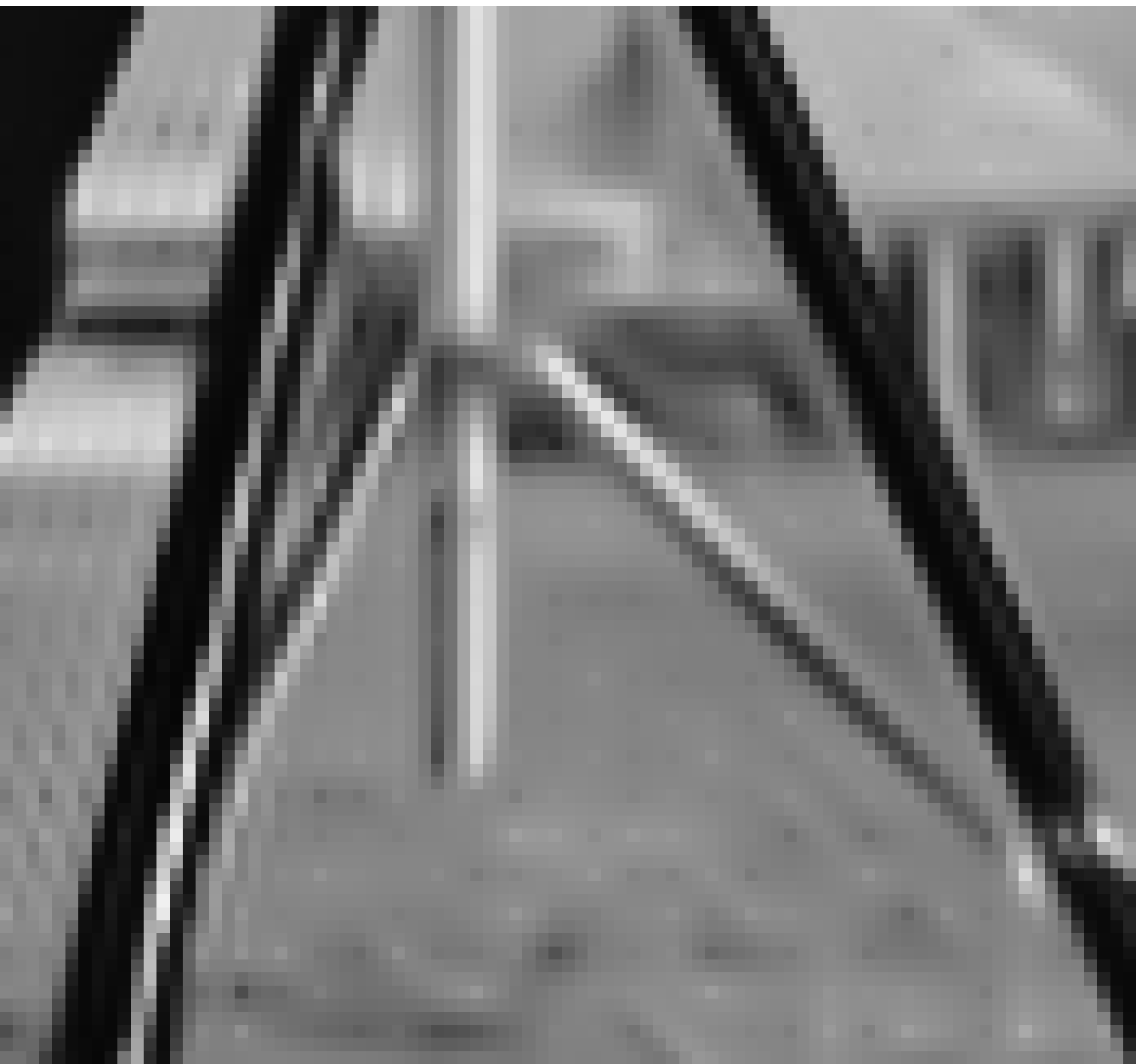}}
\centerline{(f)~MISTER}
\end{minipage}
\begin{minipage}[b]{0.18 \linewidth}
\centerline{\includegraphics[width=3.3 cm]{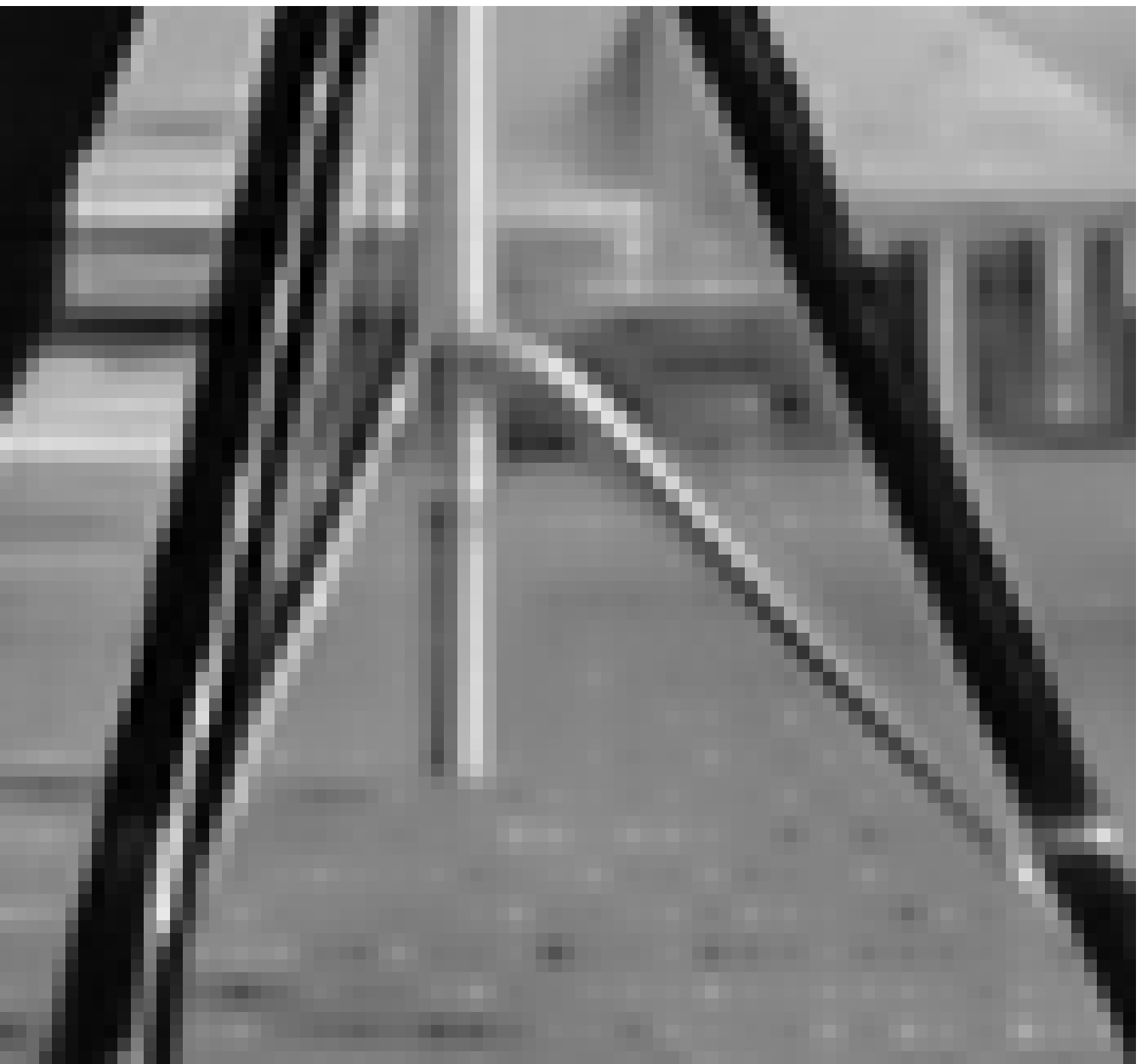}}
\centerline{(g)~MAIN}
\end{minipage}
\begin{minipage}[b]{0.18 \linewidth}
\centerline{\includegraphics[width=3.3 cm]{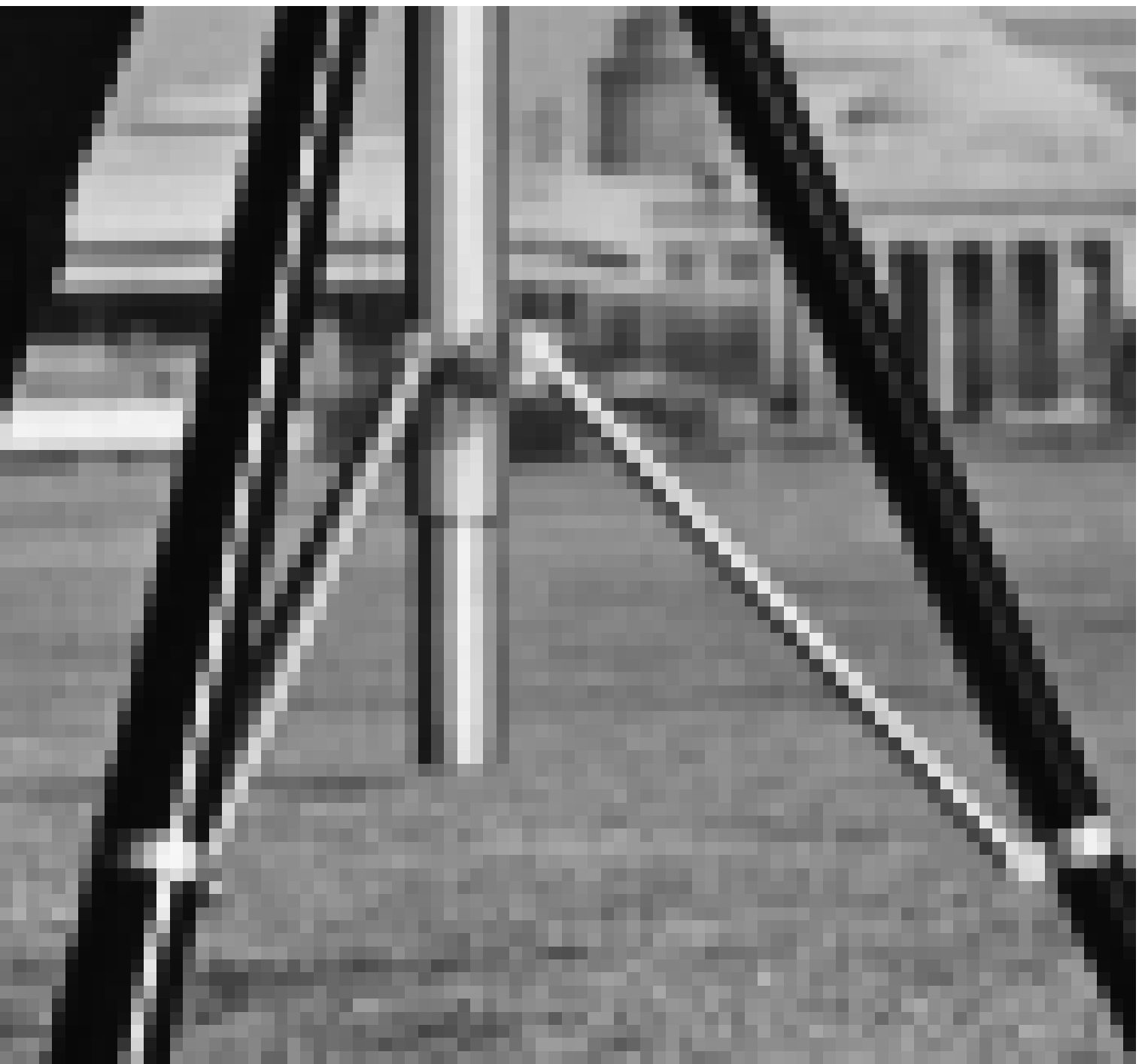}}
\centerline{(h)~Ground Truth}
\end{minipage}
\caption{The zoom-in comparison of crop from \textit{Cameraman} in the interpolation task by a factor of $3$.}
\label{fig:edge_cameraman_3x}
\end{figure*}

\begin{figure*}[tb]
\centering
\begin{minipage}[b]{0.06 \linewidth}
\centerline{\includegraphics[width=1.1 cm]{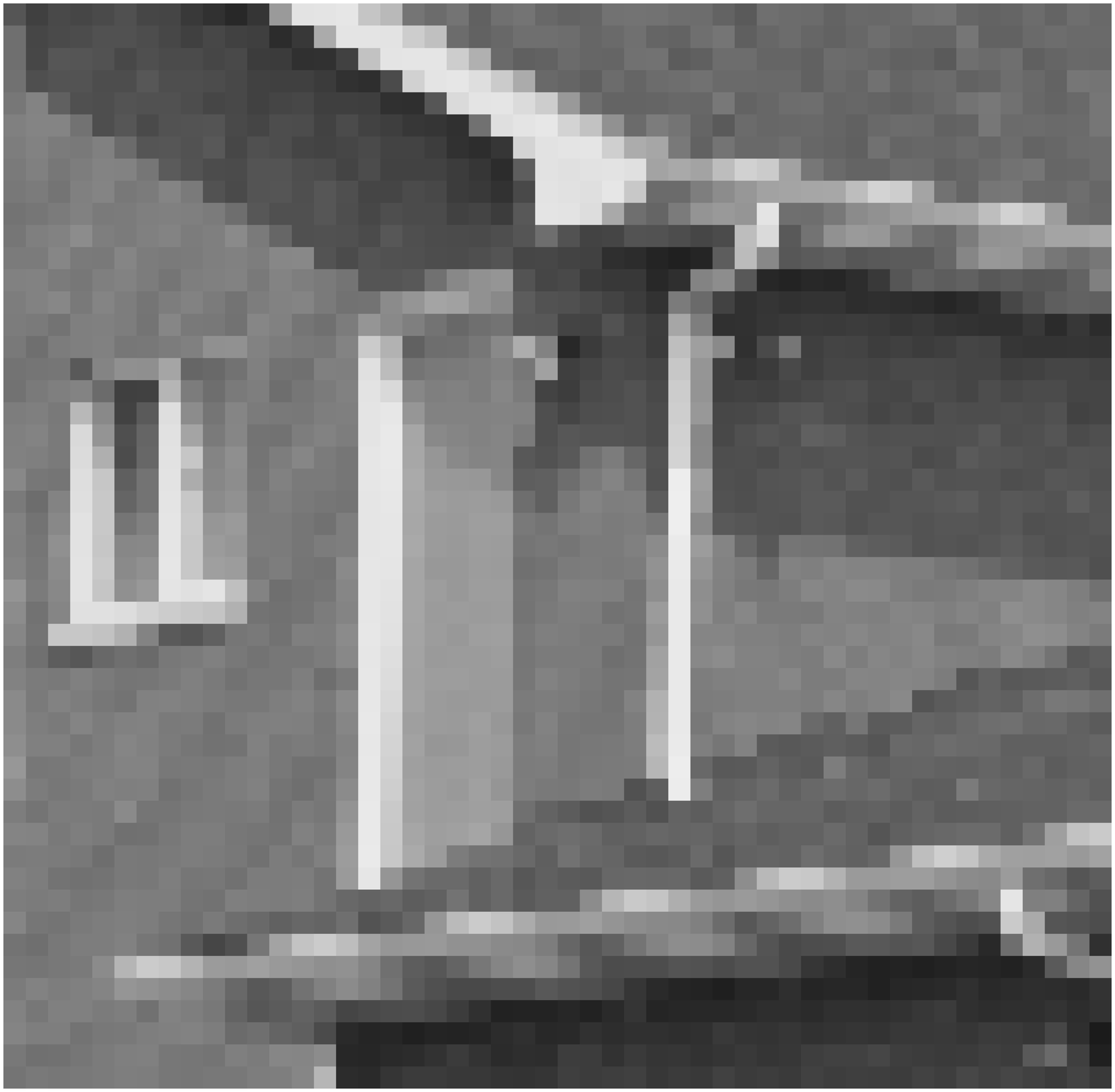}}
\centerline{(a)~LR}
\end{minipage}
\begin{minipage}[b]{0.18 \linewidth}
\centerline{\includegraphics[width=3.3 cm]{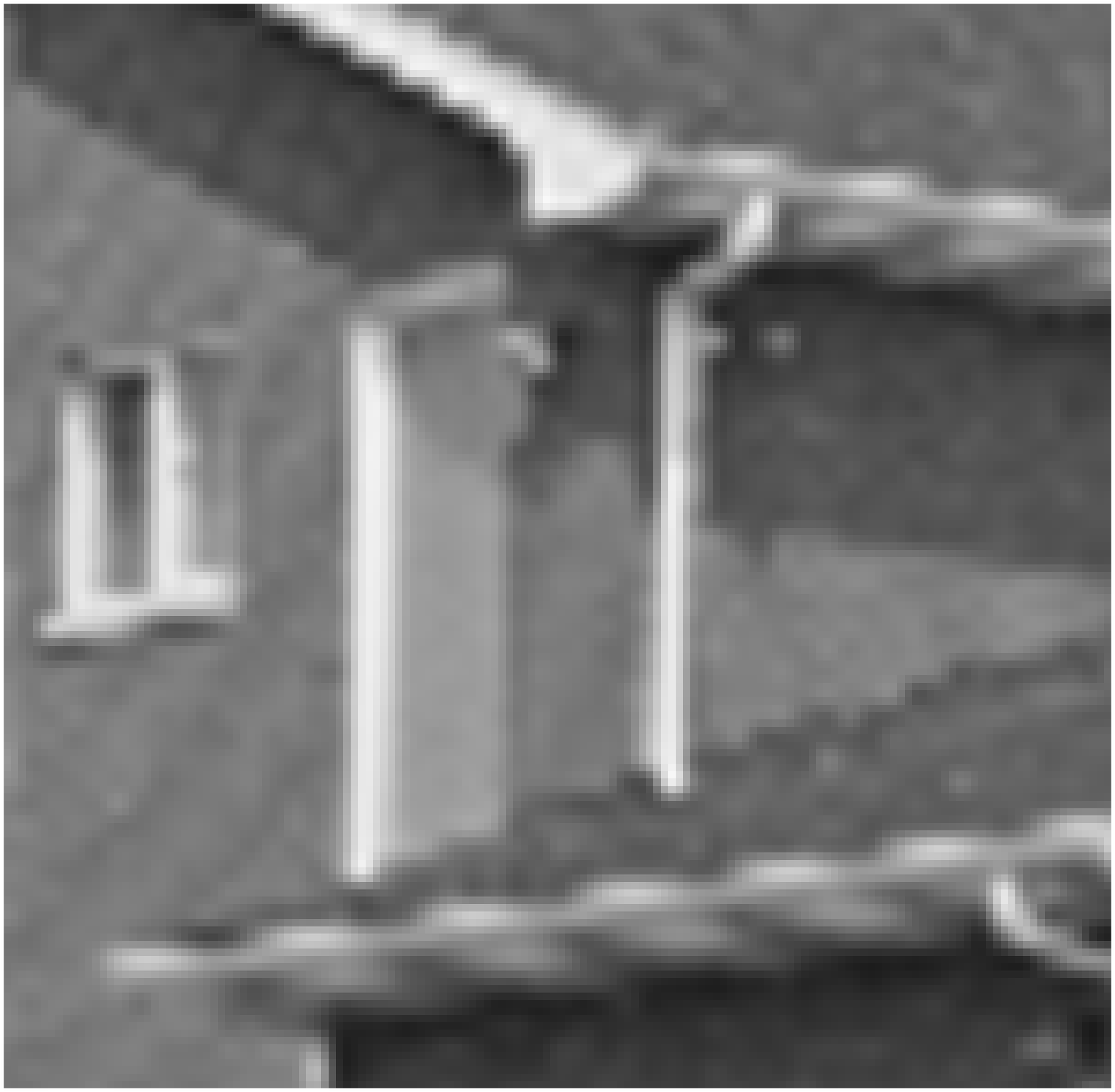}}
\centerline{(b)~Bicubic}
\end{minipage}
\begin{minipage}[b]{0.18 \linewidth}
\centerline{\includegraphics[width=3.3 cm]{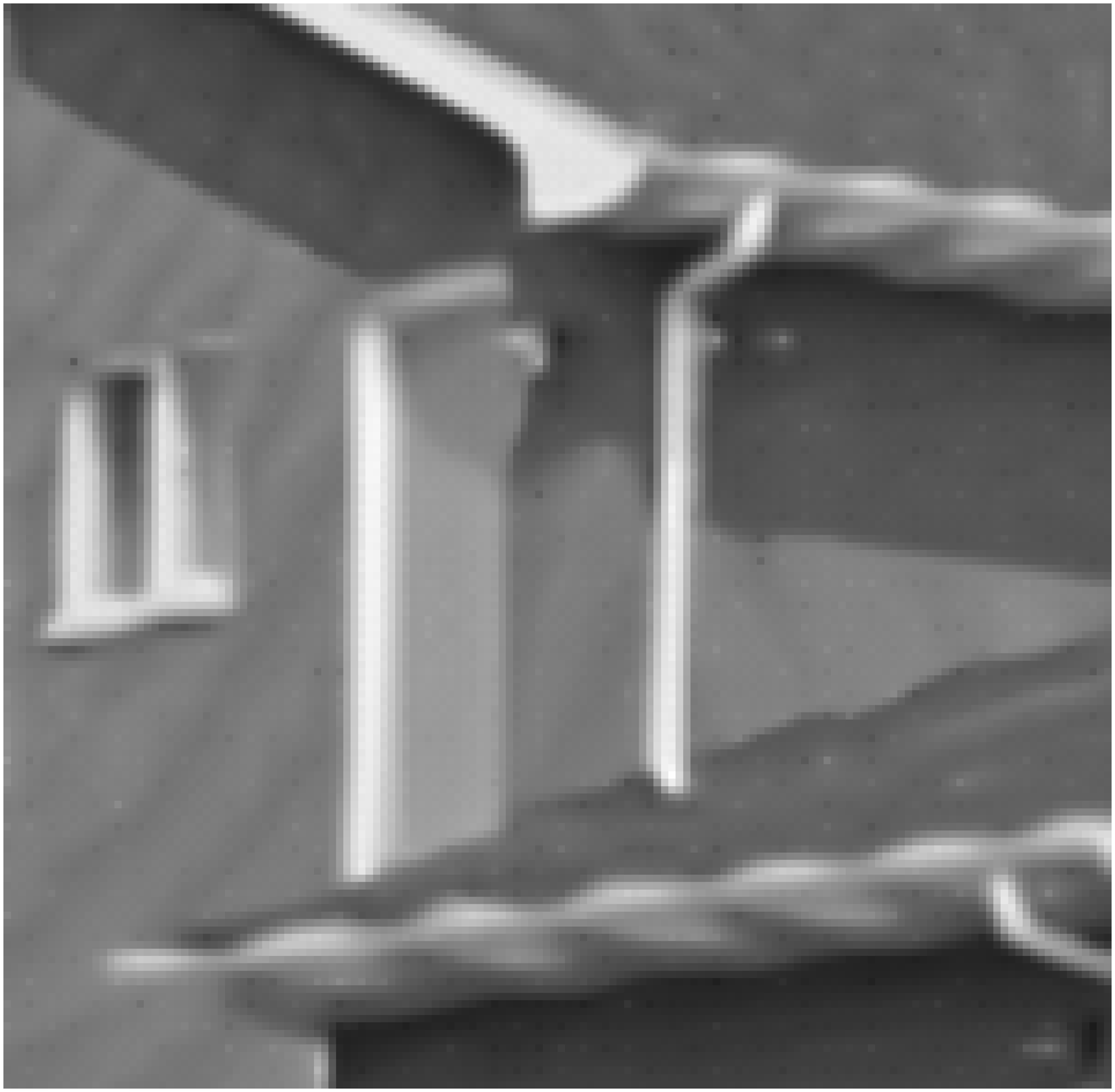}}
\centerline{(c)~NARM}
\end{minipage}
\begin{minipage}[b]{0.18 \linewidth}
\centerline{\includegraphics[width=3.3 cm]{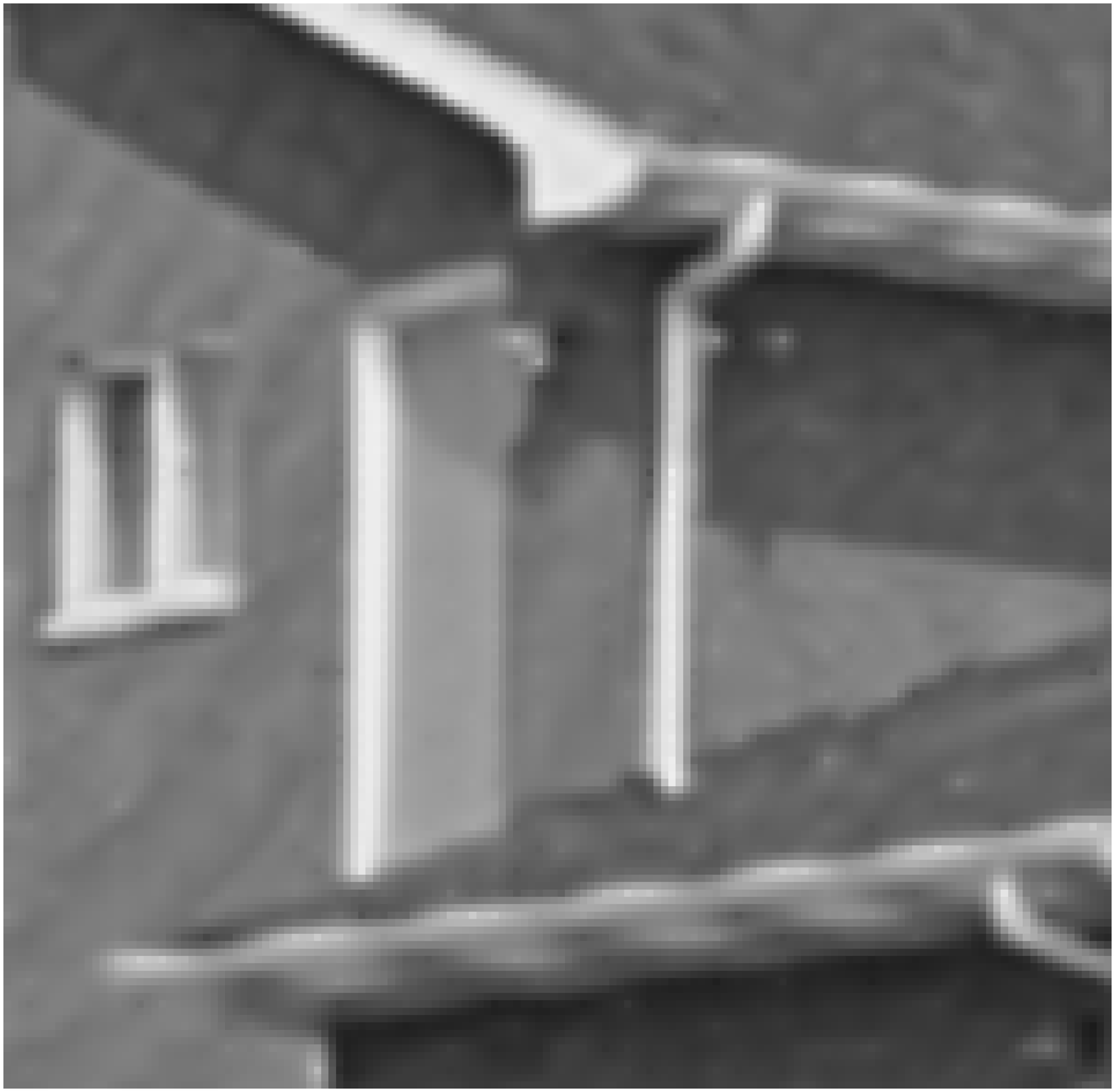}}
\centerline{(d)~ANSM}
\end{minipage}
\begin{minipage}[b]{0.18 \linewidth}
\centerline{\includegraphics[width=3.3 cm]{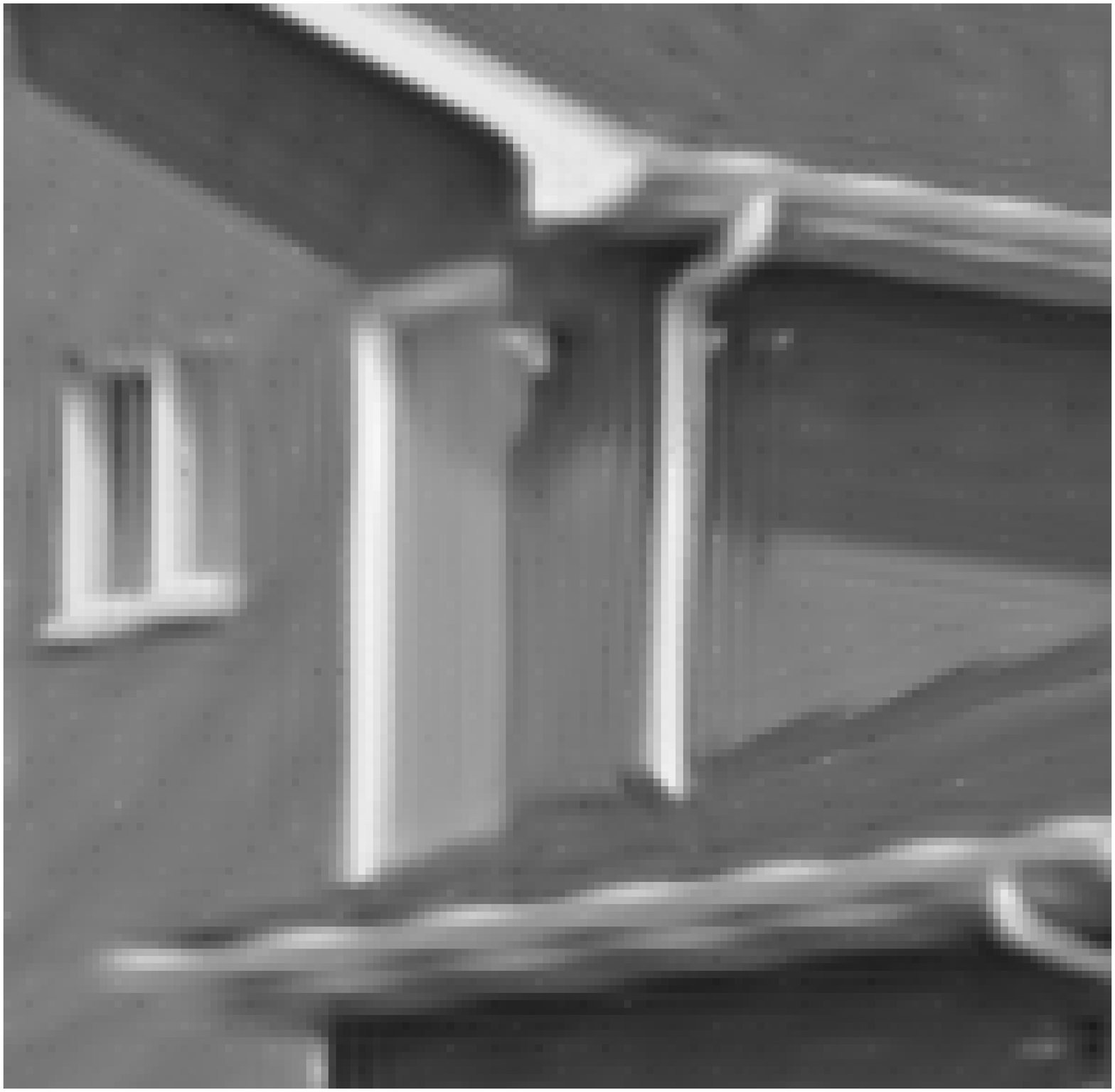}}
\centerline{(e)~NLPC}
\end{minipage} \\
\begin{minipage}[b]{0.18 \linewidth}
\centerline{\includegraphics[width=3.3 cm]{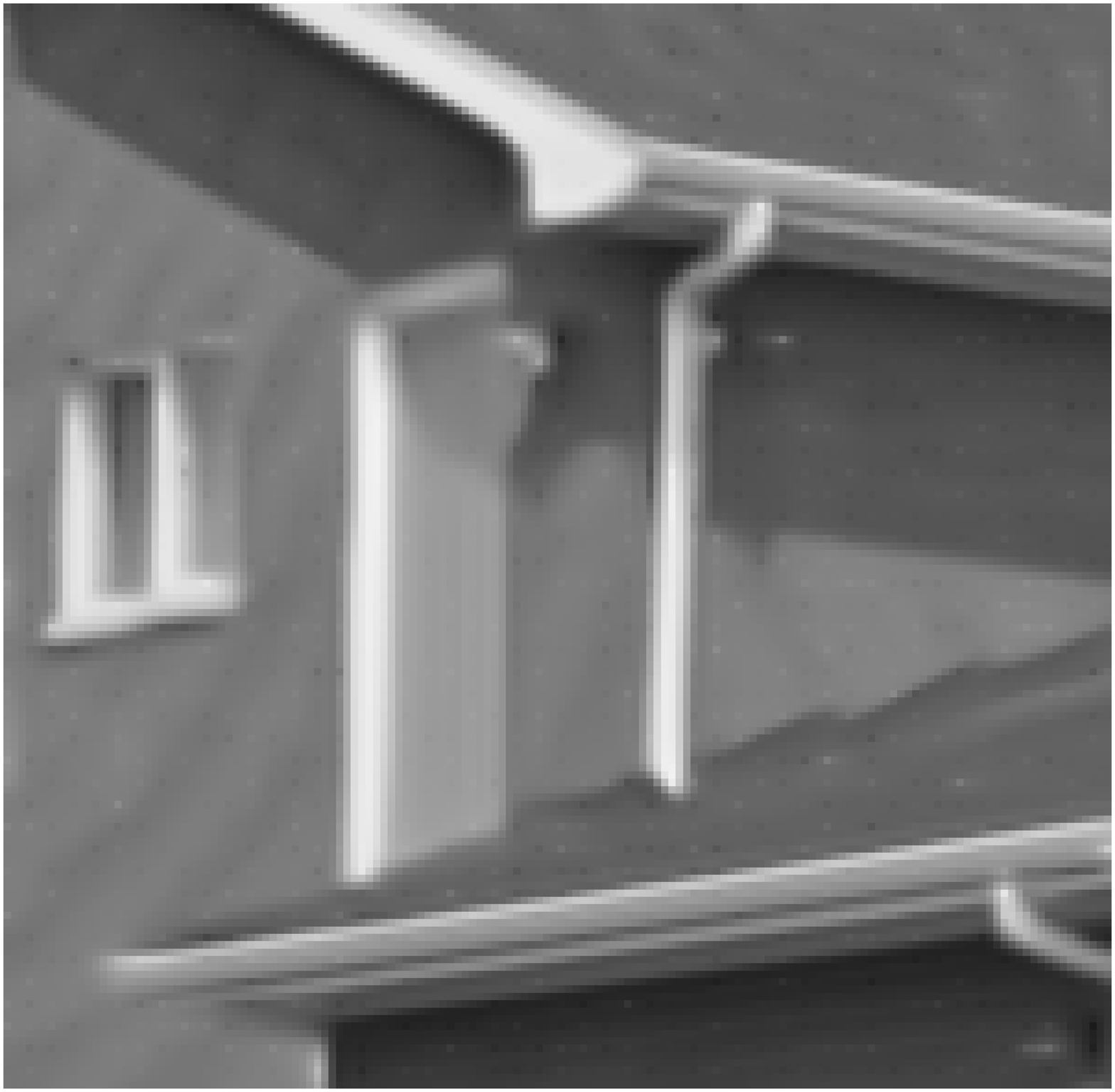}}
\centerline{(f)~MISTER}
\end{minipage}
\begin{minipage}[b]{0.18 \linewidth}
\centerline{\includegraphics[width=3.3 cm]{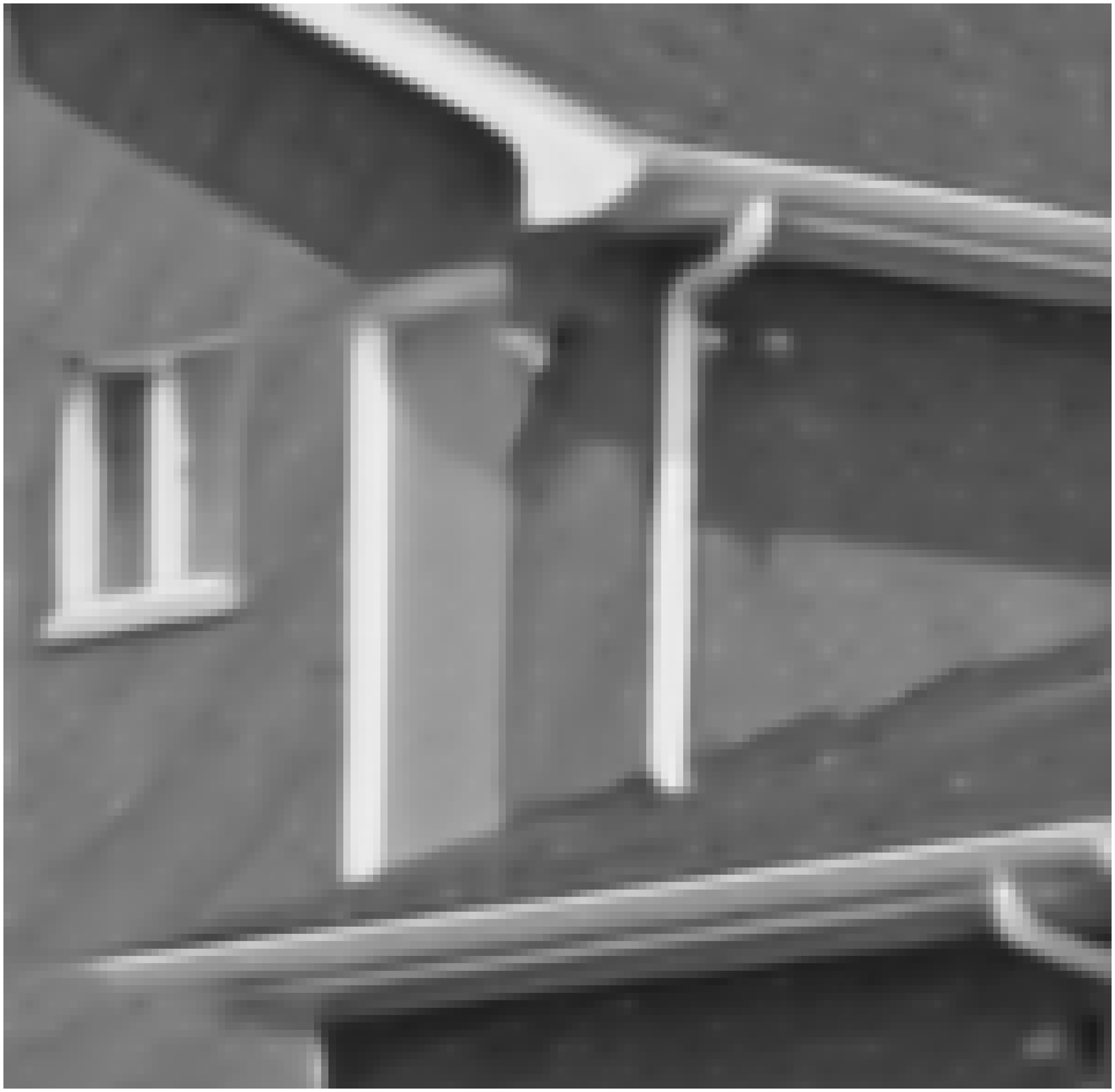}}
\centerline{(g)~MAIN}
\end{minipage}
\begin{minipage}[b]{0.18 \linewidth}
\centerline{\includegraphics[width=3.3 cm]{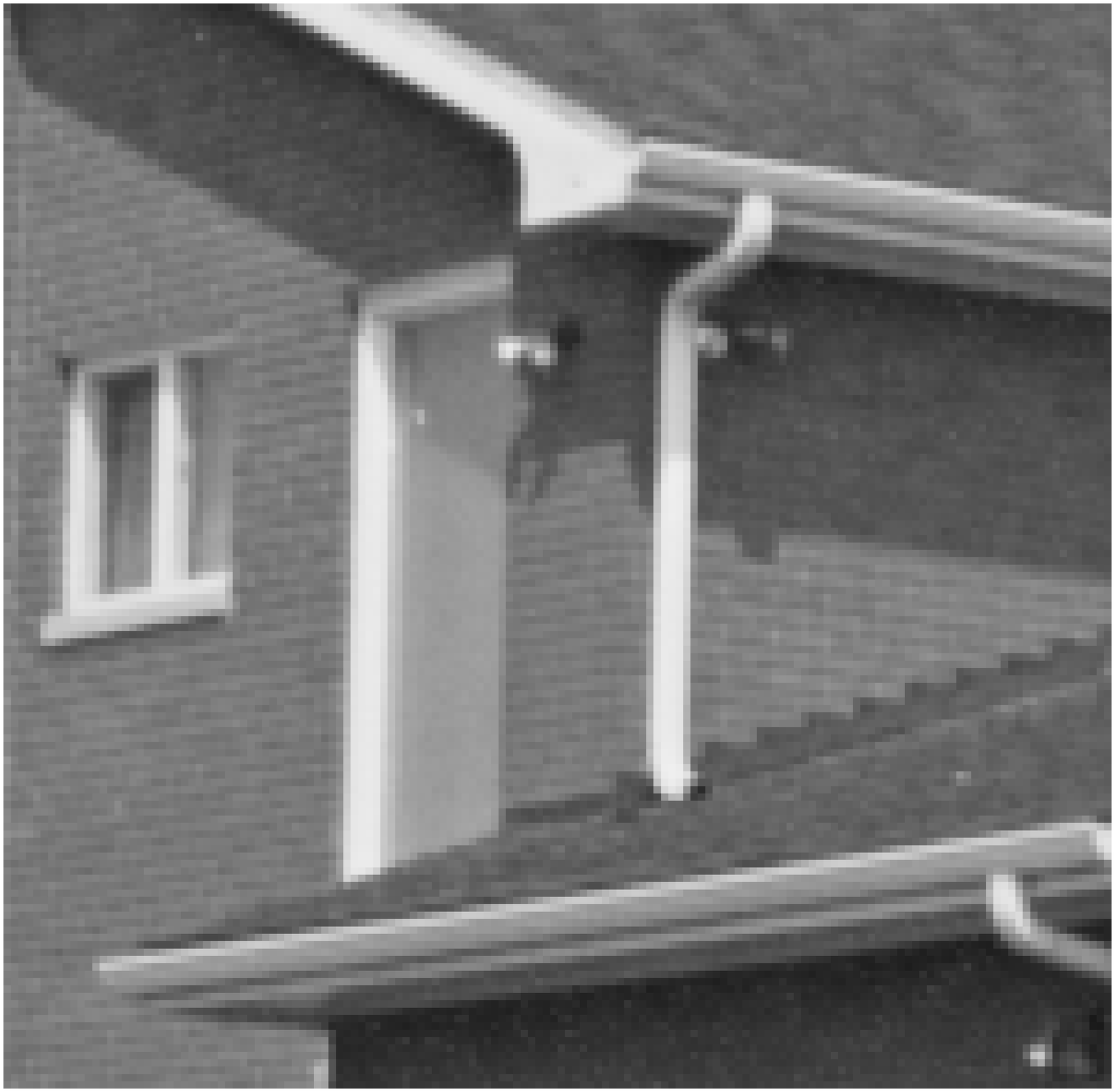}}
\centerline{(h)~Ground Truth}
\end{minipage}
\caption{The zoom-in comparison of crop from~\textit{House} in the interpolation task by a factor of $3$.}
\label{fig:edge_house_3x}
\end{figure*}

We exhibit the visual comparison in Fig.~\ref{fig:edge_elk_3x},~\ref{fig:edge_cameraman_3x},~\ref{fig:edge_house_3x}. MISTER manages to recover edges, such as the elk's horns (Fig.~\ref{fig:edge_elk_3x}(f)), the tripod (Fig.~\ref{fig:edge_cameraman_3x}(f)) and the eaves (Fig.~\ref{fig:edge_house_3x}(f)), even though they appear heavily discontinuous in the original LR image. Unfortunately, the results from compared model-based algorithms often demonstrate discontinuous or distorted contours, thereby sometimes carrying annoying visual perception. For example, the continuity of the contours along the upper and lower eaves in Fig.~\ref{fig:edge_house_3x}(b, c, d, e) are corrupted. The directions of these contours are hardly noticeable.

\section{Contribution}
\label{sec:conclu}

This paper proposes a manifold-inspired approach to modeling and exploiting semi-local similarity for the problem of single image interpolation, leading to the development of an algorithm we call MISTER (Manifold-Inspired Single image inTERpolation).
Our approach explicitly formulates single image interpolation as two mutually exclusive tasks essential to exploiting semi-local similarity: i) the identification of similar patches, and ii) the estimation of the target patches given identified similar patches.  We identify the challenges that strong aliasing poses for each of these two tasks, and propose solutions for each task.

For the task of identifying similar patches, our solution robustly removes aliasing from low-frequency components of the LR image, and identifies reliable similar patches even in the presence of strong aliasing.
For the task of estimating target patches given identified similar patches, our solution i) uses the aliasing-removed image's pixels to estimate the weights of similar patches in the initial iteration, and ii) applies appropriate manifold models to refine the estimated similar patches in later iterations.

MISTER achieves the highest average $\mathrm{PSNR}$ and the best overall visual quality compared with current model-based approaches to single image interpolation.

\bibliographystyle{IEEEtran}
\bibliography{strings2}

\clearpage

\title{Manifold-Inspired Single Image Interpolation \\ Supplemental Materials}

\author{Lantao~Yu,~\IEEEmembership{Member,~IEEE}, Kuida~Liu,
        Michael~T.~Orchard,~\IEEEmembership{Fellow,~IEEE}}

\markboth{IEEE Transactions on Image Processing}%
{Shell \MakeLowercase{\textit{et al.}}: Bare Demo of IEEEtran.cls for IEEE Journals}

\maketitle

\IEEEpeerreviewmaketitle

In this supplemental material, Table IV (see next page) provides details for Table II on Page 11 of the manuscript, with each interpolated image's $\mathrm{PSNR}$ comparison between MISTER and its related approaches in the task of interpolation by a factor of $2$. Table V (see next page) provides details for Table III on Page 11 of the manuscript, with each interpolated image's $\mathrm{PSNR}$ comparison between MISTER and its related approaches in the task of interpolation by a factor of $3$.

\begin{table*}[ht]
\caption{A Detailed Comparison of $\mathrm{PSNR}$s (in decibels) of the results between MISTER and the related approaches in the Task of Interpolation by a Factor of $2$. In each row, the result with the highest $\mathrm{PSNR}$ among model-based approaches is highlighted in dark bold and the result with the highest $\mathrm{PSNR}$ is highlighted in red bold.}
\centering
\setlength\tabcolsep{2 pt}
\begin{tabularx}{\textwidth}{ ||c| *{13}{Y|} | c||}
 \hline
\textbf{Images} &  \textbf{Bicubic} & \textbf{NEDI} & \textbf{SME} & \textbf{SAI} & \textbf{RLLR} & \textbf{MSIA} & \textbf{NGSDG} & \textbf{NARM} & \textbf{ANSM}  & \textbf{NLPC} & \textbf{FIRF} & \textbf{MAIN} & \textbf{MISTER} \\
\hline
\textit{Elk}        & 30.44 & 31.47  & 31.49  & 31.85  & 31.53  & 31.90  & 31.72  & 31.95  & 32.52  & 32.31  & 32.68  & $\textcolor{red}{\bm{33.55}}$ & $\bm{32.86}$ \\      
 \hline
\textit{Birds}      & 33.57 & 33.67  & 34.28  & 34.63  & 34.43  & 33.87  & 34.52  & 35.03  & 34.68  & 35.00  & 34.80  & $\textcolor{red}{\bm{35.61}}$ & $\bm{35.23}$ \\      
 \hline
\textit{Butterfly}  & 25.78 & 26.11  & 26.75  & 26.94  & 27.28  & 26.98  & 27.88  & 28.23  & 27.90  & 27.86  & 28.85 & $\textcolor{red}{\bm{30.09}}$ & $\bm{28.79}$ \\      
 \hline
\textit{Flower}     & 33.21 & 33.21  & 33.76  & 34.09  & 33.88  & 33.36  & 33.67  & 34.44  & 34.16  & 34.22  & 34.62  & $\textcolor{red}{\bm{35.77}}$ & $\bm{34.78}$ \\      
 \hline
\textit{Woman}      & 29.63 & 30.07  & 30.20  & 30.23  & 30.37  & 30.32  & 29.98  & 30.36  & 30.50  & 30.50  & 30.49  &  $\textcolor{red}{\bm{31.41}}$ & $\bm{30.87}$ \\      
 \hline
\textit{Hats}       & 31.75 & 32.15  & 32.33  & 32.51  & 32.61  & 32.21  & 32.27  & 32.56  & 32.80  & 32.62  & 32.97  &  $\textcolor{red}{\bm{34.26}}$ & $\bm{33.34}$ \\      
 \hline
\textit{Leaves}    & 26.64 & 26.88  & 27.97  & 28.23  & 28.02  & 27.33  & 28.76  & 29.36  & 28.86  & 29.23  & 29.51  & $\textcolor{red}{\bm{32.75}}$ & $\bm{30.32}$ \\      
 \hline
\textit{Male}      & 31.76 & 31.64  & 32.16  & 32.16  & 32.20  & 31.61  & 32.20  & 32.41  & 32.41  & 32.43  & 32.43  & $\textcolor{red}{\bm{32.93}}$ & $\bm{32.75}$ \\      
 \hline
\textit{Motorbike}  & 25.85 & 25.99  & 26.71  & 27.00  & 26.63  & 26.43  & 26.62  & 26.83  & 27.11  & 27.16  & 27.74 & $\textcolor{red}{\bm{28.86}}$ & $\bm{27.70}$ \\      
 \hline
\textit{Boat}       & 29.27 & 29.32  & 29.72  & 29.71  & 29.59  & 29.31  & 29.50  & 29.86  & 30.16  & 30.06  & 30.05  & $\textcolor{red}{\bm{31.06}}$ & $\bm{30.41}$ \\      
 \hline
\textit{Cameraman} &  25.51 & 25.68  & 26.24  & 25.99  & 25.90  & 26.52  & 25.65  & 26.04  & 26.62  & 26.31  & 26.68 & 27.09 & $\textcolor{red}{\bm{27.17}}$ \\      
 \hline
\textit{Dragonfly}  & 34.80 & 35.96  & 36.27  & 36.25  & 36.36  & 36.74  & 35.96  & 37.02  & 37.23  & 36.91  & 37.32  &  $\textcolor{red}{\bm{39.24}}$ & $\bm{37.88}$ \\      
 \hline
\textit{Fence}     &  23.45 & 21.11  & 23.47  & 22.68  & 22.98  & 23.43  & 22.99  & 23.64  & 23.74  & 23.75  & 23.51  & $\textcolor{red}{\bm{26.54}}$  & $\bm{23.89}$ \\      
 \hline
\textit{Fighter}   &  29.63 & 30.19  & 30.68  & 30.33  & 30.39  & 30.57  & 30.19  & 30.64  & 31.27  & 30.80  & 31.66  & $\textcolor{red}{\bm{33.26}}$  & $\bm{32.25}$ \\      
 \hline
\textit{Lena}      &  34.00 & 33.95  & 34.64  & 34.74  & 34.51  & 33.86  & 34.68  & 35.08  & 34.86  & 35.08  & 34.83  & $\textcolor{red}{\bm{35.60}}$ & $\bm{35.23}$ \\      
 \hline
\textit{Peppers}   &  32.74 & 33.26  & 33.37  & 33.45  & 33.73  & 33.13  & 33.53  & 34.07  & 33.82  & 34.14  & 33.80  & $\textcolor{red}{\bm{34.26}}$ & $\bm{34.20}$ \\      
 \hline
\textit{Sail}      &  32.17 & 32.68  & 32.79  & 32.79  & 32.95  & 32.88  & 32.47  & 33.13  & 33.39  & 33.33  & 33.39  & $\textcolor{red}{\bm{34.50}}$  & $\bm{33.76}$ \\      
 \hline
\textit{Plane}     &  29.45 & 30.02  & 30.30  & 30.48  & 30.59  & 30.42  & 30.35  & 30.34  & 31.05  & 30.99  & 31.26 & $\textcolor{red}{\bm{32.21}}$ & $\bm{31.37}$ \\      
 \hline
\textit{Vase}      &  34.06 & 34.51  & 34.43  & 34.66  & 34.81  & 34.60  & 34.48  & 34.90  & 34.92  & 34.93  & 34.99  & $\textcolor{red}{\bm{36.55}}$  & $\bm{35.28}$ \\      
 \hline
\textit{House}     &  32.26 & 32.14  & 33.19  & 32.87  & 32.99  & 33.07  & 32.76  & 33.49  & 34.48  & 33.93  & 34.08  & $\textcolor{red}{\bm{35.61}}$  & $\bm{34.87}$ \\      
 \hline
\textit{Parrot}    &  26.53 & 26.45  & 26.85  & 27.34  & 27.16  & 26.81  & 27.34  & 27.19  & 27.46  & 27.59  & 27.69  & $\textcolor{red}{\bm{28.15}}$ & $\bm{27.77}$ \\      
 \hline
\textit{Texture}   &  20.52 & 20.69  & 21.53  & 21.44  & 21.20  & 21.18  & 21.15  & 21.47  & 22.00  & 21.92  & 21.55  & $\textcolor{red}{\bm{23.07}}$ & $\bm{22.42}$ \\      
 \hline
\textit{Foreman}   &  35.27 & 36.58  & 36.78  & 37.00  & 37.24  & 36.84  & 37.28  & 38.41  & 38.26  & 38.03  & 38.24  & $\textcolor{red}{\bm{38.83}}$  & $\bm{38.43}$ \\      
 \hline
\textit{Straws}    &  24.26 & 25.05  & 25.32  & 25.52  & 25.45  & 25.11  & 24.91  & 25.22  & 25.78  & 25.96  & 25.37  &  $\textcolor{red}{\bm{26.64}}$  & $\bm{26.51}$ \\      
 \hline
\textit{Butterfly} &  26.44 & 27.24  & 27.59  & 27.74  & 27.69  & 27.85  & 27.82  & 27.61  & 28.35  & 28.75  & 29.19  &  $\textcolor{red}{\bm{30.18}}$ & $\bm{29.40}$ \\      
 \hline
\textit{Station}   &  24.53 & 24.98  & 25.76  & 25.65  & 25.47  & 25.59  & 25.55  & 26.02  & 25.88  & 25.81  & 27.05  & $\textcolor{red}{\bm{28.93}}$ & $\bm{27.72}$ \\      
 \hline
\textit{Wheel}     &  19.57 & 21.06  & 21.69  & 21.28  & 20.73  & 25.04  & 20.80  & 20.79  & 25.73  & 23.27  & 23.97 & $\textcolor{red}{\bm{30.33}}$ & $\bm{27.20}$ \\       
 \hline
\textbf{AVERAGE}   &  29.00 & 29.37  & 29.86  & 29.91  & 29.88  & 29.89  & 29.82  & 30.23  & 30.59  & 30.48  & 30.69  & $\textcolor{red}{\bm{32.12}}$ & $\bm{31.20}$ \\     
 \hline
\end{tabularx}
\label{tab:test_2xx}
\end{table*}

\begin{table*}[ht]
\caption{A Detailed Comparison of $\mathrm{PSNR}$s (in decibels) of the results between MISTER and the related approaches in the task of interpolation by a factor of $3$. In each row, the result with the highest $\mathrm{PSNR}$ among model-based approaches is highlighted in dark bold and the result with the highest $\mathrm{PSNR}$ is highlighted in red bold.}
\centering
\resizebox{0.5\textwidth}{!}{%
\begin{tabular}{||c|c|c|c|c|c|c||}
 \hline
\textbf{Images}  & \textbf{Bicubic} & \textbf{NARM} & \textbf{ANSM}  & \textbf{NLPC}  & \textbf{MAIN} & \textbf{MISTER} \\
\hline
\textit{Elk}       & 26.82  & 27.69  & 27.99  & 28.11 & $\textcolor{red}{\bm{29.58}}$ & $\bm{28.93}$ \\      
 \hline
\textit{Birds}     & 29.43 & 30.02  & 30.41  & $\bm{30.51}$ & $\textcolor{red}{\bm{31.16}}$ & 30.16 \\      
 \hline
\textit{Butterfly} & 21.82 & 23.46  & 23.54  & 23.54 &$\textcolor{red}{\bm{25.59}}$ & $\bm{24.25}$ \\      
 \hline
\textit{Flower}    & 29.41 & 30.12  & 30.25  & 30.19 &$\textcolor{red}{\bm{30.78}}$ & $\bm{30.46}$ \\      
 \hline
\textit{Woman}     & 27.09 & 27.69  & 27.74  & 28.15 &$\textcolor{red}{\bm{28.73}}$ & $\bm{28.32}$ \\      
 \hline
\textit{Hats}      & 28.40 & 29.16  & 29.28  & 29.61 & $\textcolor{red}{\bm{30.47}}$& $\bm{29.83}$ \\      
 \hline
\textit{Leaves}    & 21.76 & 23.03  & 23.09  & 23.30 & $\textcolor{red}{\bm{25.86}}$& $\bm{23.81}$ \\      
 \hline
\textit{Male}     & 28.30  & 28.78  & 28.92  & 28.98 & $\textcolor{red}{\bm{29.49}}$ & $\bm{29.14}$ \\      
 \hline
\textit{Motorbike} & 22.06 & 22.38  & 22.93  & 23.02 & $\textcolor{red}{\bm{24.25}}$ & $\bm{23.39}$ \\      
 \hline
\textit{Boat}     & 26.06  & 26.57  & 26.80  & 26.72 & $\textcolor{red}{\bm{27.72}}$ & $\bm{27.05}$ \\      
 \hline
\textit{Cameraman} & 22.54  & 22.79  & 23.21  & 23.24 & $\textcolor{red}{\bm{24.51}}$  & $\bm{24.00}$ \\      
 \hline
\textit{Dragonfly} & 31.34 & 32.92  & 33.44  & 33.15 & $\textcolor{red}{\bm{25.60}}$ & $\bm{34.05}$ \\      
 \hline
\textit{Fence}    & 19.58 & 19.43  & 19.55  & $\bm{19.64}$ & $\textcolor{red}{\bm{21.58}}$  & 19.62 \\      
 \hline
\textit{Fighter}  & 26.69  & 27.43  & 27.62  & 27.83 &  $\textcolor{red}{\bm{29.71}}$ & $\bm{28.49}$ \\      
 \hline
\textit{Lena}     & 30.24  & 31.28  & 31.21  & 31.29 &  $\textcolor{red}{\bm{31.99}}$ & $\bm{31.65}$ \\      
 \hline
\textit{Peppers}  & 30.08  & 31.36  & 31.26  & 31.47 & $\textcolor{red}{\bm{32.16}}$ & $\bm{31.72}$ \\      
 \hline
\textit{Sail}    & 29.14   & 29.98  & 29.88  & 30.17 & $\textcolor{red}{\bm{31.32}}$ & $\bm{30.56}$ \\      
 \hline
\textit{Plane}   & 26.08   & 26.74  & 27.36  & 27.46 & $\textcolor{red}{\bm{28.68}}$ & $\bm{27.57}$ \\      
 \hline
\textit{Vase}    & 31.21   & 32.24  & 32.07  & 32.35 & $\textcolor{red}{\bm{33.60}}$ & $\bm{32.53}$ \\      
 \hline
\textit{House}   & 28.78   & 29.84  & 30.10  & 30.23 &  $\textcolor{red}{\bm{31.91}}$ & $\bm{31.45}$ \\      
 \hline
\textit{Parrot}  & 23.14   & 23.59  & 24.03  & 24.03 & $\textcolor{red}{\bm{24.52}}$& $\bm{24.21}$ \\      
 \hline
\textit{Texture} & 16.40   & 16.54  & 17.31  & 17.47 &$\textcolor{red}{\bm{18.92}}$ & $\bm{18.06}$ \\      
 \hline
\textit{Foreman} & 31.86   & 34.18  & 34.51  & 34.99 & $\textcolor{red}{\bm{35.59}}$  & $\bm{35.23}$ \\      
 \hline
\textit{Straws}  & 20.48   & 20.59  & 21.41  & 21.53 & $\textcolor{red}{\bm{22.48}}$ & $\bm{22.12}$ \\      
 \hline
\textit{Butterfly} & 22.98 & 23.78  & 24.19  & 24.40 & $\textcolor{red}{\bm{25.82}}$ & $\bm{24.83}$ \\      
 \hline
\textit{Station}  & 20.81  & 21.38  & 22.05  & 21.90 & $\textcolor{red}{\bm{24.12}}$& $\bm{23.26}$ \\      
 \hline
\textit{Wheel}    & 16.54  & 17.15  & 18.28  & 18.05 & $\textcolor{red}{\bm{24.61}}$ & $\bm{21.22}$ \\      
 \hline
\textbf{AVERAGE}  & 25.52  & 26.30  & 26.61  & 26.72 & $\textcolor{red}{\bm{28.18}}$ & $\bm{27.26}$ \\     
 \hline
\end{tabular}}
\label{tab:test_3xx}
\end{table*}

\end{document}